\pdfoutput=1
\documentclass[10pt,twocolumn,letterpaper]{article}

\usepackage[pagenumbers]{cvpr}

\usepackage{multirow}
\usepackage{placeins}
\definecolor{cvprblue}{rgb}{0.21,0.49,0.74}
\usepackage[pagebackref,breaklinks,colorlinks,allcolors=cvprblue]{hyperref}

\def\confName{CVPR}
\def\confYear{2027}

\title{Believing Is Seeing: How Vision-Language Models Decide to Believe an Image}

\author{Vijay Vankadaru \qquad Vasha DuTell\\
University of California, Berkeley\\
{\tt\small vankadaruvijay@berkeley.edu}
}

\begin{document}
\addtocontents{toc}{\protect\setcounter{tocdepth}{0}}
\maketitle

\begin{abstract}
When responding to a prompt, vision-language models must integrate both text and visual information, along with their priors. When the image contradicts what a model expects to see, the model often decides to answer based on its prior, rather than an answer grounded in the image. We explore how this decision is made in eleven open models across eight different architectures. We measure how strongly the model prefers an image-based answer over a prior-based one, and we intervene on this preference at multiple stages from the input pixels to the output. We find that the decision weighs two independent cues, how clear the image is and how strong the prior is, and that blurring the image lowers a model's image reliance whether the prior is weak or strong. Within the image, the aspect that matters most is the object that contradicts the prior. Degrading the contradicting object can change the decision, degrading an equal area elsewhere does not, and restoring the object alone on a blurred image raises a model's image reliance. Inside the model, we show that the object's evidence moves from the image tokens to the answer position at a characteristic depth, in the second half of the network, where a small set of attention heads write it into the answer. The model uses this pathway whether or not the image conflicts with its prior. Removing the heads hands decisions to the prior, but amplifying them wins back only 12 to 33\% of the overridden images, and no edit of a clean input image that we tried raises the image's win rate. Prior override therefore arises on the model's ordinary pathway for reading images, and our results point to that pathway, not the input image, as the place where a correction must act.
\end{abstract}

\begin{figure}[t]\centering\includegraphics[width=1.0\linewidth]{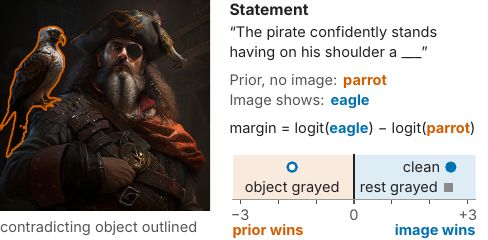}
\caption{\textbf{The conflict and the measurement.} The prior completes the statement one way and the image shows another. The margin is the logit of the image-supported answer minus that of the prior-supported answer, and its sign shows which side the model favors. On this WHOOPS item \cite{bittonguetta2023whoops}, LLaVA-NeXT believes the image until the contradicting object is converted to grayscale.}\label{fig:conflict}\end{figure}

\section{Introduction}\label{sec:intro}
Vision-language models (VLMs) are now used in applications that range from medical imaging and document reading to robotics and autonomous driving. In these settings the cases that matter most are often the unexpected ones, such as an abnormal finding in a scan or an unusual obstacle on the road \cite{li2022coda}, yet an unexpected image is exactly what a VLM is most likely to misreport \cite{bittonguetta2023whoops}. When an image contradicts what a model expects to see, the model often answers from its expectation and ignores the image \cite{goyal2017vqav2,luo2025vilp,vo2026biased}. When asked what sits on the shoulder of the pirate in \cref{fig:conflict}, a model that expects a parrot may answer that it is a parrot even though the image shows an eagle. The answer is fluent either way, so a user cannot tell that the image was set aside. A model that reports what is typical in place of what is present cannot be trusted to notice an anomaly, an error or a deliberate alteration.

This failure, known as prior override, is well documented. We say that the prior \emph{wins} when the model gives the answer its prior supports, and that the image wins when it gives the answer the image supports. Recent work has begun to trace what decides these wins inside the model. Specific attention heads carry the image's answer or the prior's \cite{ortu2026seeing,lietzow2026visiondefault}, and the visual evidence remains decodable from the model's activations even when the prior wins \cite{nooralahzadeh2026arbitration,liao2026decodable}. The model, in other words, often sees the object and overrides it anyway. How it makes that choice remains unclear, and we ask three questions about this phenomenon: What does a model weigh when an image and its prior disagree? Which part of the image supplies the image's side? And where in the network is the choice made, and is that place the same across architectures?

We study this choice through a single quantity, the \emph{margin}, which is the model's preference for the answer the image supports over the answer its prior supports (\cref{fig:conflict}). Following the margin from the input pixels to the output shows what moves it at the input, where in the network it is set, and which components set it. Comparing eleven open models from eight architectures (\cref{tab:models}) shows which of these answers hold in general.

Our earlier paper \cite{vankadaru2026trust} showed, on LLaVA-NeXT, that trust in the image tracks how discernible the conflicting evidence is, and proposed cue combination as an explanation without testing it. We test that explanation across architectures and add the object-level edits, the layer-wise patching, and the head and subspace interventions. We find that: 
\begin{enumerate}
\item \textbf{Two independent cues contribute to prior override in the context of conflict.} An image is more likely to win when it is less degraded, and when the prior is weaker. These two effects add in log-odds with no detectable interaction, and both hold on ten of the eleven models (\cref{sec:rule}).
\item \textbf{An image's evidence is concentrated in the object that contradicts the prior.} Blurring a contradictory object lowers the margin in most models, whereas even graying an equal area elsewhere has little to no effect. When the whole image is blurred, restoring the object alone raises the image's win rate (\cref{sec:what}).
\item \textbf{Evidence follows the same pathway with and without a conflict.} In all models tested, the evidence leaves the image positions and arrives at the answer position at a consistent depth, between 53\% and 74\% of the network. In most models, attention heads located mostly after that depth write it into the answer, and the same pathway is used when the image and the prior agree (\cref{sec:where}).
\item \textbf{Interventions remove the image's influence more readily than they restore it.} Ablating these heads reverses some image-wins, but amplifying them recovers at most a third of the overridden items. Sharpening the object or cropping to it does not raise the image's win rate, and attention steering and direction injection do not correct override on natural photographs (\cref{sec:what,sec:where,sec:photo}).
\end{enumerate}

\section{Related work}\label{sec:related}
\noindent\textbf{Knowledge conflicts in language models.} Previous work has explored contradictions in language models generally. When a context contradicts a fact a language model holds, the model often answers from memory \cite{longpre2021entity}, following the context when it is coherent and convincing \cite{xie2023chameleon} or the prior is weak \cite{wu2024clasheval,yu2023characterizing,du2024persuasion}. This preference is carried by identifiable heads \cite{jin2024cutting,ortu2024competition,yu2023characterizing} and the conflict is detectable in intermediate-layer activations, before the model produces its answer \cite{zhao2024conflictdetect}.

\noindent\textbf{Image versus prior in VLMs.} Benchmarks and studies of language priors \cite{goyal2017vqav2,bittonguetta2023whoops,luo2025vilp,li2026seeingorknowing}, of text--image conflicts \cite{deng2025blindfaith,hua2025conflicting,paraselli2026entities,ghosh2026whichsource} and of conflicts between a model's visual and textual answers \cite{zhu2024crossmodal} establish that VLMs often answer from the prior and that the preference shifts with task and legibility. \citet{barua2026bayesian} find that VLM preferences between conflicting prompt text and images follow a biased Bayesian rule with per-model weights, and name the parametric-prior case as future work. Relatedly, \citet{zhang2025modalitiesconflict} tie modality following to unimodal uncertainty, and \citet{ma2025bayesbench} find Bayes-consistent cue combination in multimodal LLMs on magnitude-estimation tasks. Previous work also explores the opposite failure, following a misleading image \cite{cheng2024typographic,han2024instinctive}.

\noindent\textbf{Mechanisms of the decision.} \citet{ortu2026seeing} identify counterfactual heads on WHOOPS-AHA by logit inspection and steer decision outcomes by reweighting their attention. \citet{lietzow2026visiondefault} locate prior-override heads in the second half of the network (52 to 100\% of depth) on three VLM families that insert image tokens into the stream; ablating them under a prompt contrast flips most prior-grounded predictions. Information-flow studies find that image information moves to the question tokens and, in the middle layers, from there to the last position \cite{zhang2025crossmodal,kaduri2025whats}, or directly from the image tokens to the last position, depending on the task \cite{salazar2026pathways}, and through a single gate token in natively multimodal models \cite{serra2025narrowgate}. Other work reports that visual evidence stays encoded or attended when the prior wins \cite{grob2026bigger,nooralahzadeh2026arbitration,liu2025seeingnotbelieving,zhou2026ignorance,liao2026decodable}, a commitment depth that differs by model \cite{he2026modalityconflict,nooralahzadeh2026arbitration,li2026seeingorknowing,long2026chainofembedding}, a prior-to-visual shift in the mid-to-late layers \cite{golovanevsky2025pixels}, late alignment of visual with textual representations \cite{nikankin2025sametask}, steerable modality preference \cite{zhang2025modalitypreferences}, object information at the object's token positions \cite{neo2024towards,jiang2024devils}, and head subsets that control what the model reports \cite{wang2025vseam,gandikota2026gaze,rudman2026pih,park2026retrievalheads}. These studies build on activation patching and its metrics \cite{zhang2024patching,heimersheim2024patching}, on subspace interventions and their known illusions \cite{geiger2024das,makelov2024illusion}, and on linear concept erasure \cite{belrose2023leace}.

\noindent\textbf{Hallucination detection and training-free intervention.} Hallucination and adversarial benchmarks score answers \cite{rohrbach2018chair,li2023pope,guan2023hallusionbench,li2024naturalbench}. Detectors read the logit lens \cite{nostalgebraist2020logitlens,jiang2024interpretingedit} or probe hidden states \cite{kogilathota2026halp}. Training-free interventions raise the weight of the image by contrastive decoding, an over-trust penalty, attention reweighting or route gating \cite{leng2023vcd,huang2023opera,liu2024pai,kang2025var,cheng2026crg}.

\section{Setup}\label{sec:setup}
\noindent\textbf{Conflicts.} A \emph{conflict} is an image paired with a statement whose most likely completion disagrees with the image (\cref{fig:conflict}). We assemble 2,320 conflicts from four sources: 500 from WHOOPS-AHA \cite{ortu2026seeing,bittonguetta2023whoops}, 600 from ViLP \cite{luo2025vilp}, and 493 color and 727 size items from Visual CounterFact \cite{golovanevsky2025pixels}. A conflict counts for a model only if the prior's answer wins when no image is given, which leaves 391 to 425 valid WHOOPS conflicts per model. Unless another source is named, results are on WHOOPS.

\noindent\textbf{The decision and the margin.} Each answer is a set of accepted strings, which can span several tokens. We score every string by its summed token log-probability, keep the best string on each side, and call the side with the higher score the model's \emph{decision}. A decision for the image is an \emph{image-win} and one for the prior a \emph{prior-win}, and \emph{vision-win} \cite{vankadaru2026trust}, the image's win rate, is the fraction of conflicts that are image-wins. Our main measurement is the \emph{margin}, the logit of the first token of the image-supported string minus that of the prior-supported string at the answer position. The margin is what the interventions move and what we read at intermediate layers, and we count a decision as reversed under an internal intervention when the sign of the margin changes. The \emph{prior} of an item is the same margin measured with no image, in favor of the prior's answer.

\noindent\textbf{Edits and depth.} We edit images in three ways, by converting pixels to grayscale, which we call graying, by Gaussian blur, and by additive Gaussian noise. We call a blur or noise level a \emph{dose}. To localize effects in depth, we divide the decoder's layers into eight equal \emph{bands}. Layer $L$ is the output of decoder block $L$, and relative depth is $L$ over the number of layers. \emph{Arbitration heads} are the counterfactual heads of \citet{ortu2026seeing}, the 20 heads that write most toward the image's answer at the answer position by direct logit attribution (Supp.~P).

\noindent\textbf{Models.} We use eleven open models from eight architectures (\cref{tab:models}). The architectures differ in how the image enters the language model. LLaVA-NeXT-Mistral-7B \cite{liu2024llavanext,jiang2023mistral}, LLaVA-1.5-7B \cite{liu2023llava,liu2024improvedllava} and BakLLaVA-7B \cite{bakllava} share a CLIP ViT-L encoder and an MLP projector, and Qwen2.5-VL-7B \cite{bai2025qwen25vl} has a different encoder with dynamic resolution. We count these four as one architecture, a ViT encoder with an MLP projector. Seven more, one per architecture, are InstructBLIP-Vicuna-7B \cite{dai2023instructblip} (Q-Former), Idefics2-8B \cite{laurencon2024idefics2} (perceiver resampler), LFM2-VL-3B \cite{liquid2025lfm2vl,amini2025lfm2} (a mostly convolutional language model), Fuyu-8B \cite{bavishi2023fuyu} (image patches projected linearly, no vision encoder), Emu3-Chat \cite{wang2024emu3} (discrete image tokens in the text vocabulary), Llama-3.2-11B-Vision \cite{meta2024llama32} (cross-attention, no image tokens in the stream) and Gemma-3-12B \cite{gemmateam2025gemma3} (SigLIP tokens pooled to 256). The behavioral measurements are repeated on twelve further checkpoints and three frontier API models (\cref{sec:rule}). We segment the contradicting object with SAM3 \cite{sam3} from the answer phrase, which yields a mask on 85 to 87\% of valid WHOOPS conflicts. Patching uses the same image, edited, as the counterfactual \cite{zhang2024patching,heimersheim2024patching} and acts on the residual stream \cite{makelov2024illusion}, with hooks verified against the unpatched forward pass (Supp.~C).

\begin{table*}[t]\centering\scriptsize\setlength{\tabcolsep}{2.1pt}
\caption{\textbf{Eleven models, eight architectures} (WHOOPS conflicts). $a$, $b$: precision and prior weights of the additive model with 95\% intervals (\cref{sec:rule}). Dose grid: size in pixels at which blur and noise are applied (full: the original image). Crossing: relative depth at which the image-site and answer-site curves of the band map cross, interpolated on band starts (\cref{sec:where}); single-layer crossings are 0.03 to 0.07 later (Supp.~P). Readout: AUROC of the untrained readout (\cref{sec:read}) at 0.625 / 0.75 of depth. Heads: image-wins reversed by mean-ablating the top-20 heads / range over 20 layer-matched random sets (3 on Qwen2.5-VL). Object $-$ rest: difference in margin shift under gray, in logits. $\dagger$: from the run that fixes the library version (Supp.~M).}\label{tab:models}
\begin{tabular}{@{}llcccccccc@{}}\toprule
Model & Image pathway & Vision-win & Dose grid & $a$ (precision) & $b$ (prior) & Crossing & Readout & Heads / random & Object $-$ rest \\ \midrule
LLaVA-NeXT-7B & CLIP, MLP, multi-crop & 0.736 & full & 0.11 [0.03, 0.18] & 0.19 [0.06, 0.32] & 0.57 & 0.80 / 0.85 & 11.8\% / 0.0--0.7\% & $-$0.55 [$-$0.80, $-$0.32] \\
LLaVA-1.5-7B & CLIP, MLP & 0.734 & 336 & 0.31 [0.23, 0.40] & 0.22 [0.10, 0.35] & 0.58 & 0.83 / 0.88 & 10.5\% / 0.0--1.4\% & $-$0.38 [$-$0.57, $-$0.19] \\
BakLLaVA-7B & CLIP, MLP & 0.743 & 336 & 0.31 [0.24, 0.39] & 0.15 [0.04, 0.26] & 0.59 & 0.80 / 0.86 & 14.7\% / 0.7--3.3\% & $-$0.31 [$-$0.60, $-$0.04] \\
Qwen2.5-VL-7B & ViT, MLP, dynamic res. & 0.798 & full & 0.13 [0.06, 0.20]$^\dagger$ & 0.14 [0.02, 0.26]$^\dagger$ & 0.73 & 0.58 / 0.57 & 7.3\% / 1.8--2.4\% & $-$0.46 [$-$0.79, $-$0.16] \\
InstructBLIP-7B & Q-Former & 0.787 & 224 & 0.33 [0.23, 0.42] & 0.18 [0.04, 0.31] & 0.55 & 0.89 / 0.93 & 8.7\% / 0.0--1.9\% & $-$0.52 [$-$0.71, $-$0.32] \\
Idefics2-8B & Perceiver resampler & 0.816 & full & 0.15 [0.06, 0.24] & 0.19 [0.02, 0.35] & 0.57 & 0.83 / 0.90 & 4.1\% / 0.0--2.4\% & $-$0.40 [$-$0.73, $-$0.11] \\
LFM2-VL-3B & Conv--attention LM & 0.771 & full & 0.17 [0.10, 0.23] & 0.19 [0.04, 0.33] & 0.66 & 0.63 / 0.78 & 21.1\% / 0.0--1.3\% & $-$0.27 [$-$0.50, $-$0.08] \\
Fuyu-8B & Raw patches & 0.675 & full & 0.22 [0.13, 0.31] & 0.24 [0.13, 0.34] & 0.74 & 0.54 / 0.72 & 11.5\% / 0.0--0.8\% & $-$0.57 [$-$0.81, $-$0.36] \\
Emu3-Chat & Discrete VQ tokens & 0.641 & 512$^2$ & 0.31 [0.22, 0.40] & 0.23 [0.11, 0.35] & 0.68 & 0.74 / 0.81 & 25.2\% / 0.7--6.5\% & $-$0.34 [$-$0.63, $-$0.05] \\
Llama-3.2-11B & Cross-attention & 0.745 & 560 & 0.22 [0.14, 0.31] & 0.24 [0.11, 0.38] & 0.53 & 0.82 / 0.90 & 6.5\% / 0.0--1.9\% & $-$0.43 [$-$0.67, $-$0.18] \\
Gemma-3-12B & SigLIP, pooled & 0.806 & full & 0.16 [0.08, 0.25] & 0.09 [$-$0.02, 0.19] & 0.59 & 0.82 / 0.93 & 1.8\% / 0.0\% & $-$0.20 [$-$0.67, 0.27] \\ \bottomrule
\end{tabular}\end{table*}

\begin{figure*}[t]\centering\includegraphics[width=1.0\linewidth]{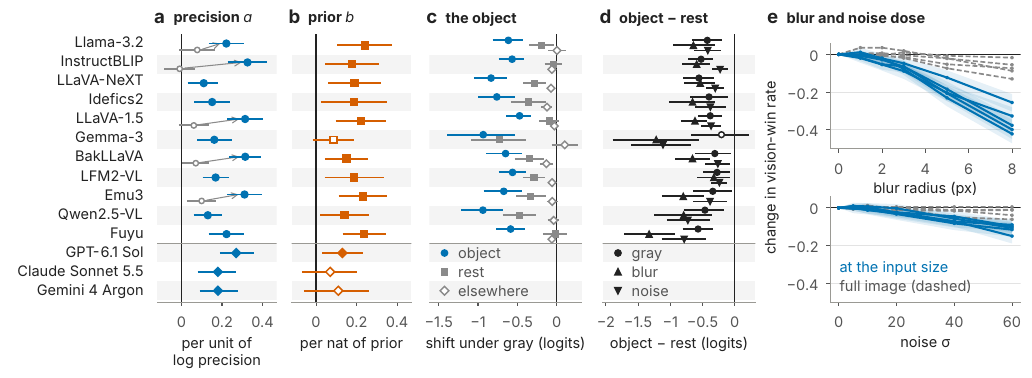}
\caption{\textbf{The two cues and the object, across architectures.} Both cues carry weight on nearly every model (a, b), the object outweighs the rest of the image (c, d), and the clarity effect appears only when the degradation is applied at the model's input size (e). (a, b) Weights of the additive model, $a$ for image precision and $b$ for prior strength, for eleven open and three API models (diamonds; their $b$ uses the open models' prior); gray circles in (a): dose on the full-size image. (c) Margin shift from graying the object, the rest, or a same-size region elsewhere. (d) Object-minus-rest difference under gray, blur and noise. (e) Change in vision-win under blur and noise on the five models whose dose is set at an input size (\cref{tab:models}), with the dose on the full-size image (gray, dashed) and at the input size (blue). Intervals are bootstrap 95\% over items; open markers in (b) and (d) include zero.}\label{fig:cues}\end{figure*}

\section{Clarity versus prior strength}\label{sec:rule}
First, we test how the clarity of the image and the strength of the prior affect the decision. We vary the clarity of the image by blurring it or adding noise in graded doses, and we measure the strength of the prior on each item as the margin the model gives the prior's answer when it sees no image.

\noindent\textbf{Checks on the measurement.} At the final layer the margin predicts the decision almost perfectly on eight of the nine models measured, and less well on Fuyu, whose tokenizer merges words (Supp.~P). On the three LLaVA models the decision agrees with free generation on about nine in ten conflicts, and nearly all image-wins disappear when the image is replaced by a blank gray image, the control of \citet{liao2026decodable} (Supp.~A).

\noindent\textbf{An additive model.} For each VLM we fit a logistic regression with two predictors, $\operatorname{logit} P(\text{image wins}) = a\log\tau - b\,\pi + c$, on WHOOPS conflicts under blur and noise. Here $\tau$ is the nominal precision of the image at each dose ($1/(1+\sigma^2)$ for blur, $1/(1+(\sigma_n/8)^2)$ for noise), which is our measure of its clarity, and $\pi$ is the centred text-only margin. The weights $a$ and $b$ measure how strongly clarity and prior strength move the decision. The fit includes a noise indicator, and intervals are clustered by image. Adapting the split of \citet{barua2026bayesian}, we fit on the boundary cells (clean image at every prior tertile, every dose at the middle tertile) of 70\% of the items and score the interior cells of the rest.

\noindent\textbf{Both cues matter on nearly every model.} The intervals of $a$ and $b$ exclude zero on ten of the eleven models (\cref{tab:models}; \cref{fig:cues}a, b). The exception is the prior weight of Gemma-3, whose interval includes zero on the main split although the weight is positive on every one of 50 item splits. The same signs hold on twelve further checkpoints. These are LLaVA-1.5-13B, InstructBLIP-13B, Gemma-3 at 4B and 27B, Qwen3-VL-8B \cite{bai2025qwen3vl} with and without thinking, MiMo-VL-7B \cite{xiaomi2025mimovl} at its SFT and RL stages, and four training stages of InternVL3.5-8B \cite{wang2025internvl35} (Supp.~P). The weights vary across item splits about as much as across models, so we claim the additive form and the signs of the two effects, not a weight shared by all models.

\noindent\textbf{The two cues act independently.} By independent we mean that the cues add in log-odds, so that the fit needs no dose-by-prior term. Such a term stays inside a pre-specified bound on 11 of 13 fits (all seven with the dose at a fixed input size and four of six at native resolution) and leaves held-out deviance unchanged (Supp.~Q). On the seven fixed-input fits the strongest blur lowers vision-win by 20 to 48 points in every prior tertile, so a clearer image helps whether the prior is weak or strong. The prior's effect is also causal. When we strengthen or weaken the prior within an item by adding a prefix to the statement, the decision moves in the predicted direction on the three LLaVA models, and the fit from unmodified items predicts the modified ones, although the within-item weight is larger than the between-item one (Supp.~H). Both weights are again positive on ViLP, a second source, on all ten open models run (Supp.~Q).

\noindent\textbf{The cues explain little of the variation between items.} The two effects are reliable, but they are small next to the differences between individual conflicts. Together the two terms explain 3 to 10\% of held-out deviance across the eleven models (\cref{tab:size}) and 5 to 10\% on the seven fits at a fixed input size, where the dose term contributes 2 to 5 points (Supp.~Q). The additive model thus describes two tendencies with consistent signs. It does not predict single decisions (AUROC 0.60 to 0.71).

\begin{table}[t]\centering\footnotesize\setlength{\tabcolsep}{2.6pt}
\caption{\textbf{Effect size of the additive model and the enhancement test, per model} (WHOOPS). Held-out deviance explained by the prior alone and by both terms, and the median increment from adding the dose term (Dose), with the dose grid of \cref{tab:models} (50 item splits; Supp.~Q). Enhancement: lowest and highest change in vision-win, in points, over six sharpening and contrast levels on the object; crop: the image cropped to the object ($^*$: 95\% interval excludes zero; Supp.~P). Dash: not run.}\label{tab:size}
\begin{tabular}{@{}lccccc@{}}\toprule
 & \multicolumn{3}{c}{Deviance explained} & \multicolumn{2}{c}{Vision-win change} \\ \cmidrule(lr){2-4}\cmidrule(lr){5-6}
Model & Prior & Both & Dose & Enhancement & Crop \\ \midrule
LLaVA-NeXT & 0.062 & 0.064 & 0.002 & -- & -- \\
LLaVA-1.5 & 0.062 & 0.093 & 0.030 & $-1.4$ to $+0.3$ & $-4.5^*$ \\
BakLLaVA & 0.045 & 0.073 & 0.033 & -- & -- \\
Qwen2.5-VL & 0.052 & 0.062 & 0.008 & $-2.0^*$ to $0.0$ & $-9.7^*$ \\
InstructBLIP & 0.036 & 0.087 & 0.050 & $-2.3^*$ to $+0.6$ & $-2.1$ \\
Idefics2 & 0.051 & 0.059 & 0.008 & $-1.4$ to $+0.8$ & $-6.2^*$ \\
LFM2-VL & 0.049 & 0.054 & 0.005 & $-1.7^*$ to $-0.3$ & $-13.4^*$ \\
Fuyu & 0.067 & 0.082 & 0.014 & $-4.5^*$ to $-0.9$ & $+0.3$ \\
Emu3 & 0.075 & 0.102 & 0.028 & $-1.1$ to $+0.6$ & $-7.8^*$ \\
Llama-3.2 & 0.053 & 0.067 & 0.017 & $-2.2$ to $0.0$ & $-6.9^*$ \\
Gemma-3 & 0.024 & 0.029 & 0.004 & $-1.4$ to $+0.6$ & $-6.9^*$ \\ \bottomrule
\end{tabular}\end{table}

\noindent\textbf{Degradation must be applied at the model's input size.} The clarity effect appears only when the degradation survives the model's own preprocessing. Three of the models receive a fixed 336- or 224-pixel image, and blur or noise applied to the full-size image is mostly removed when the processor downsamples it. On that grid their precision weight is near zero (0.06, 0.07 and $-$0.01). With the same grid applied after resizing to the input size it rises to 0.31, 0.31 and 0.33, with the prior weight unchanged (\cref{fig:cues}a, e; Supp.~P). Llama-3.2, which tiles the image, and Emu3, which caps the number of pixels per image, show the same pattern. Degradation levels should therefore be stated in model-input pixels.

\noindent\textbf{Frontier API models.} We ran the behavioral half of the protocol (text-only answer, clean image, blur and noise, and an object crop against a random crop of equal size) on GPT-6.1 Sol \cite{openai2026gpt61sol}, Claude Sonnet 5.5 \cite{anthropic2026sonnet55} and Gemini 4 Argon \cite{google2026gemini4argon} on WHOOPS (Supp.~R). The precision weight is positive on each model (\cref{fig:cues}a). On the items with a mask, the models keep the image's answer far more often under an object crop (0.83 to 0.91) than under a random crop (0.30 to 0.51).

\section{The contradicting object}\label{sec:what}
A conflict image differs from what the model expects in one place, the object that contradicts the prior (the eagle in \cref{fig:conflict}). We ask whether the model's belief in the image rests on that object or on the image as a whole. If it rests on the object, edits confined to the object should move the decision, and the same edits applied elsewhere should not.

\noindent\textbf{Editing the object moves the decision more than editing the rest.} We apply the same edit to the object alone, to everything except the object, or to a region of the object's area placed elsewhere. On ten of the eleven models, editing the object lowers the margin more than editing the rest of the image under blur, under noise and under gray, and blurring the object alone lowers it by 0.59 to 1.91 logits (\cref{fig:cues}c, d; \cref{tab:models}; Supp.~P). On the eleventh, Gemma-3, the difference holds for blur and noise but not for gray. Graying a region of the same area elsewhere barely moves the margin (at most 0.13 logits), so the effect belongs to the object and not to the amount of image edited.

\begin{figure}[t]\centering\includegraphics[width=1.0\linewidth]{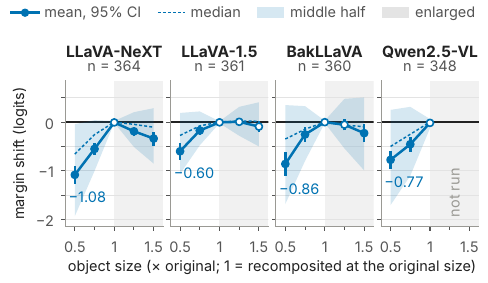}
\caption{\textbf{Object size dose.} Shrinking the object lowers the margin, and enlarging it does not raise it. Margin shift against the original image when the segmented object is recomposited at 0.5 to 1.5 times its size (WHOOPS conflicts with an object mask). Points: means with bootstrap 95\% intervals (open: includes zero); dotted line: median item; band: middle half of items. At 1.0 the recomposite applies every pixel operation except the size change. Qwen2.5-VL was run at 0.5 to 1.0.}\label{fig:sizedose}\end{figure}

\noindent\textbf{Shrinking the object lowers the margin.} We cut the object out, scale it, and paste it back over a blurred fill at 0.5 to 1.5 times its size. The margin falls as the object shrinks, by 0.60 to 1.08 logits at half size on the four models tested (\cref{fig:sizedose}). Recompositing the object at its original size leaves the margin unchanged, so the compositing operations alone do not produce the shift.

\noindent\textbf{Enhancing the object does not help.} Enlarging the object to 1.5 times its size does not raise the margin on the three LLaVA models and lowers it slightly (\cref{fig:sizedose}). We pre-specified the prediction that sharpening the object or raising its contrast would raise vision-win above the clean image, and it fails on all nine models tested (\cref{tab:size}). Cropping the image to the object does not help either. The crop reverses some prior-wins but loses more of the more numerous image-wins, so vision-win falls on eight of the nine models, reliably on seven (\cref{tab:size}; Supp.~P).

\noindent\textbf{Restoring a degraded object helps.} The same kind of edit raises vision-win when there is something to restore. On images blurred at the strongest dose, restoring the object alone raises vision-win by 10 to 35 points on all nine models, while restoring an equal area elsewhere changes it little (Supp.~P). The object's pixels therefore carry most of what the image contributes to the decision, and on a clean image the object edits we tested add nothing on net.

\begin{figure*}[t]\centering\includegraphics[width=1.0\linewidth]{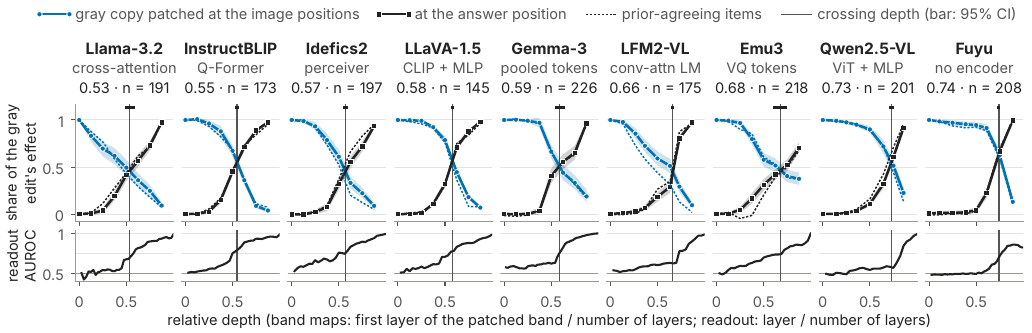}
\caption{\textbf{The image's evidence crosses to the answer position at mid-depth in every architecture.} Top: band map on nine models (LLaVA-NeXT and BakLLaVA are in Supp.~P), ordered by crossing depth (given under each name with the number of items). Each curve is the share of the gray edit's effect reproduced by patching a grayscale copy's activations into the clean run in one band of layers, at the image positions (blue) or the answer position (black). Thin dotted curves: prior-agreeing items. Vertical line: crossing depth, with its 95\% interval as a bar. Bottom: AUROC of the readout by layer (\cref{sec:read}).}\label{fig:path}\end{figure*}

\section{The decision pathway}\label{sec:where}
We next ask where in the network the object's evidence reaches the answer. Converting an image to grayscale removes part of its evidence and lowers the margin. We run the model on the clean image and overwrite its activations with those of the grayscale copy in one of eight bands of layers, at either the image positions or the answer position. If the overwrite reproduces the drop in the margin, the evidence was at that site in that band. The resulting profile is the band map (\cref{fig:path}), computed on the conflicts that the gray edit moves by at least 0.5 logits, and the depth at which the two profiles cross is the crossing depth.

\noindent\textbf{The evidence crosses from the image positions to the answer position.} A crossing has to exist, because patching the image positions in the first band reproduces every later state, and patching the answer position in the last band replaces the state that is read out. The band map shows where it falls. On all eleven models the first band recovers the whole effect from the image positions and none from the answer position, and the late bands do the reverse, less completely on Emu3, Qwen2.5-VL and Gemma-3 (Supp.~O). The curves cross between 0.53 and 0.74 of depth (\cref{tab:models}), earliest on Llama-3.2 and latest on Fuyu, and its interval over resampled items is narrow on most models (bars in \cref{fig:path}). Further checkpoints cross between 0.45 and 0.72, all but one inside this range (Supp.~P).

\noindent\textbf{The route to the answer differs across models.} The crossing depth says when the evidence arrives at the answer position, not how it gets there. With a different image of the same scene as donor (ViLP's prior-consistent image), the share of the effect that passes through the question tokens ranges from almost none on Fuyu to more than half on Gemma-3, and the crossing depth itself shifts with the donor (Supp.~D, P).

\noindent\textbf{Twenty heads write the evidence into the answer.} We next ask which components write the evidence into the answer once it has arrived. We choose the top 20 arbitration heads on one half of the conflicts and ablate them on the other half. Mean-ablating them at the answer position reverses 4 to 25\% of image-wins on seven of eight architectures, against at most 6.5\% for layer-matched random heads (\cref{tab:models}), and on each of these architectures the median top head lies after the crossing. Gemma-3 is the exception. Attribution finds no effective heads there, but heads ranked by the estimated effect of ablating them do reverse more image-wins than random sets (Supp.~P).

\noindent\textbf{Amplifying the heads recovers part of the overridden items.} If the heads write the image's evidence, amplifying them should move decisions toward the image. Scaling each head's deviation from its mean by 3 reverses 12 to 33\% of held-out WHOOPS prior-wins on nine models, more than layer-matched random sets \cite{kadir2026audit} on eight. It is not a general correction, however. The heads were chosen on WHOOPS, and on ViLP they beat random sets on only five of the nine models, while the same scaling also loses more ViLP image-wins than our pre-specified bound allows (Supp.~P).

\noindent\textbf{The pathway is the same without a conflict.} Finally, we ask whether this pathway is specific to conflicts. We repeat the analysis on prior-agreeing items, the WHOOPS items on which the text-only run already prefers the image-supported answer, with the same two candidate answers. There the crossing falls in the same band or the adjacent one on all nine models of \cref{fig:path} (dotted curves), and ablating the top heads lowers the margin by a comparable amount (Supp.~P). The model therefore reads an agreeing image through the same route, and prior override happens on its ordinary pathway for images.

\section{The image-token subspace}\label{sec:levers}
Before the crossing, the band map places the evidence at the image tokens, and we ask what form it takes there.

\begin{figure*}[t]\centering\includegraphics[width=1.0\linewidth]{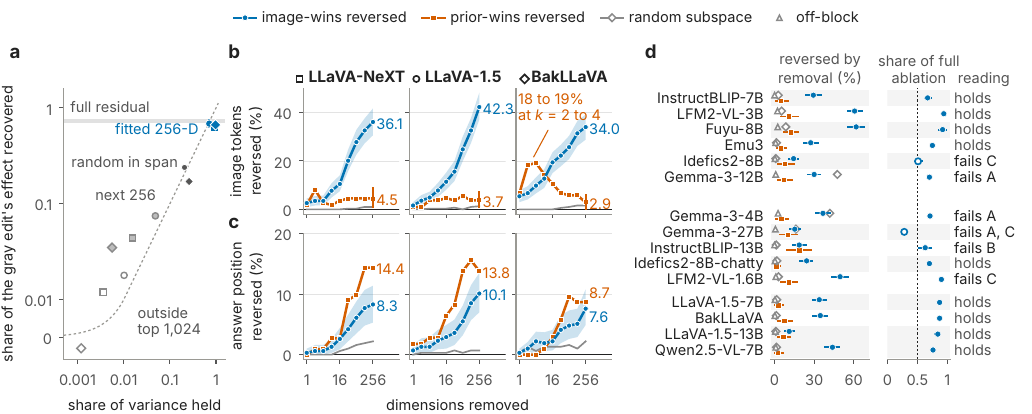}
\caption{\textbf{Subspace recovery and removal at the image tokens.} Removing the subspace at the image tokens reverses image-wins far more than prior-wins (b), and the pattern holds on some other architectures (d). (a) Squares, circles and diamonds are LLaVA-NeXT, LLaVA-1.5 and BakLLaVA. Share of the gray edit's effect recovered by patching through one 256-D subspace at the image tokens, against the share of per-token variance it holds (log axes; dashed: proportionality). Blue: fitted subspace. Gray: a random subspace in the top-1,024 span, the next 256 components, and 256 components outside the top 1,024. Band: full residual. (b) Image-wins and prior-wins reversed when the top $k$ dimensions are projected out at the image tokens; gray: random subspace of equal $k$. (c) The same at the answer position. (d) Other architectures and sizes, with the basis fit over all image positions (so the LLaVA rows differ from b): decisions reversed by removal against two control subspaces; the share of the full ablation's margin change from setting the subspace to its mean (dashed: 0.5; open: interval reaches 0.5); and the test outcome with the criterion that fails (A, B, C as in \cref{sec:levers}). Intervals are cluster-bootstrap 95\%.}\label{fig:subspace}\end{figure*}

\noindent\textbf{A high-variance subspace carries the evidence (three LLaVA models).} We take the top 256 principal components of the image-token residuals on the conflict sets (Supp.~E) and test them in two ways. First, we patch this subspace alone into the clean run at the image tokens of layers 16 to 24. It recovers about two thirds of the gray edit's effect, nine tenths of what the full residual recovers, and no more than its share of the per-token variance (\cref{fig:subspace}a; Supp.~E). Recovery rises with the variance held, and we found no smaller subspace specific to the decision. Second, we project the subspace out. This reverses 34 to 42\% of image-wins and only 3 to 5\% of prior-wins, while random and low-variance subspaces of equal dimension reverse almost none of either (\cref{fig:subspace}b). At the answer position the same removal reverses a share of prior-wins at least as large as the share of image-wins (\cref{fig:subspace}c), so the asymmetry belongs to the image tokens.

\noindent\textbf{The result holds on some other architectures and not on others.} We repeated the analysis, with its controls and with the basis fit over all image positions, on every other architecture that places image tokens in the decoder's sequence. We pre-specified three criteria. Removal must reverse more image-wins than both controls (A) and more image-wins than prior-wins (B), and the subspace set to its mean must give at least half of the full ablation's change in the margin, with the controls under a fifth (C; Supp.~P). The reading holds on four of the six models (InstructBLIP-7B, LFM2-VL, Fuyu and Emu3; \cref{fig:subspace}d). It fails on Gemma-3, where projecting out a random subspace reverses more image-wins than the fitted one, and on Idefics2, where the subspace's share of the full ablation's change falls just short of the criterion. Over all eleven other-architecture models, which add other sizes and a second Idefics2 checkpoint, it holds on five, and a second checkpoint changes the verdict on three architectures. It also holds on LLaVA-1.5-13B and on Qwen2.5-VL, whose encoder differs from the LLaVA models' (Supp.~P).

\section{Reading the decision}\label{sec:read}
If the decision is formed at the answer position by mid-depth, it should be readable there without training. We project the residual at a mid-depth layer onto the unembedding rows of the two candidate answers and use the difference as a score, which we call the \emph{readout}. On WHOOPS the readout at layer 20 predicts the decision at 0.80--0.83 AUROC on the LLaVA models (\cref{tab:models}). We take layer 20 from our earlier work \cite{vankadaru2026trust}, which selected it on LLaVA-NeXT, and do not re-tune it (Supp.~J, M). Across architectures the readout is weaker on the models that cross later, and at three quarters of depth it is lowest on Qwen2.5-VL, which crosses at 0.73 (\cref{fig:path}, bottom; \cref{tab:models}). On most models the readout also adds to what the prior alone predicts (Supp.~S).

\section{Natural photographs}\label{sec:photo}
On NaturalBench \cite{li2024naturalbench} each yes/no question comes with two photographs whose answers differ, so the model's text-only answer is right for one and wrong for the other. We ask which of the findings above carry over to these photographs on the three LLaVA models, which lose 22--37 accuracy points when the image contradicts their text-only answer (\cref{tab:photo}).

\begin{table}[t]\centering\footnotesize\setlength{\tabcolsep}{2.6pt}
\caption{\textbf{What carries over to natural photographs} (NaturalBench, three LLaVA models). Readout: AUROC for the decision. Margin shift: mean absolute change in logits when a region is grayed. Injection: change in accuracy, in points, when the text-only answer direction is injected (Supp.~L).}\label{tab:photo}
\begin{tabular}{@{}lccc@{}}\toprule
 & NeXT & 1.5 & BakLLaVA \\ \midrule
Accuracy, prior agrees / contradicts & 0.72 / 0.43 & 0.69 / 0.47 & 0.77 / 0.40 \\ \midrule
Readout, uncentered & 0.55 & 0.93 & 0.86 \\
Readout, centered & 0.88 & 0.98 & 0.92 \\
Readout, centered on a disjoint half & 0.89 & 0.98 & 0.95 \\ \midrule
Margin shift, object grayed & 0.51 & 0.42 & 0.91 \\
Margin shift, rest grayed & 0.42 & 0.41 & 0.99 \\
Margin shift, random region grayed & 0.23 & 0.19 & 0.37 \\ \midrule
Injection, conflicting items & $+$2.9 & $+$13.6 & $+$13.5 \\
Injection, consistent items & $-$2.9 & $-$11.8 & $-$10.6 \\
Injection, all items & $-$0.6 & $-$1.7 & $-$1.0 \\ \bottomrule
\end{tabular}\end{table}

\noindent\textbf{The readout carries over.} Over all conflicts the readout (read here at layer 19, hidden-state index 20) predicts the decision at 0.88--0.98 AUROC once the yes-minus-no margin is centered on its median before it is oriented toward the image-supported answer, which removes the model's default yes or no bias. The result does not depend on fitting the median to the evaluated items (\cref{tab:photo}), and it holds within the yes items and the no items alone (Supp.~L). The readout also reflects the model's default answer (Supp.~L, M).

\noindent\textbf{The object's advantage and the corrections do not.} The text-only answer on NaturalBench is mostly a default yes or no, a known bias \cite{li2024naturalbench,chan2026yes}, not knowledge about the object. There the object no longer outweighs the rest of the image. Graying it shifts the margin about as much as graying everything else, although more than graying a random region of its size, and its advantage returns only in part on the questions that concern a single object (\cref{tab:photo}; Supp.~L). Attention steering leaves conflicting-item accuracy essentially unchanged, and injecting the text-only answer direction raises accuracy on conflicting items but lowers it on consistent ones, so accuracy over all items falls slightly (\cref{tab:photo}; Supp.~G, K, L).

\section{Discussion}\label{sec:conclusion}
Across eleven models and eight architectures, a vision-language model weighs the clarity of a contradicting image against the strength of its prior, and the two contribute independently. The image's side comes from the contradicting object, reaches the answer position at a characteristic depth, and is written into the answer by a small set of heads. Because the model uses the same route when the image and the prior agree, prior override is a property of how these models read images in general. The additive weighting of clarity and prior strength is reminiscent of cue combination in human perception, where cues are weighted by their reliability \cite{ernst2002humans,knill2004bayesian}, although we do not test whether the model's weighting is optimal.

Intervening against the image is far easier than intervening in its favor. Removing the image's evidence at the object, at the image-token subspace or at the heads hands decisions to the prior, whereas enhancing the object does not help and amplifying the heads recovers only part of the overridden items. On a clean image the object's evidence already reaches the answer, so a clearer object has nothing to add, and the edits help only where the evidence was degraded in the first place. Preprocessing the input is therefore unlikely to correct prior override, and a correction will likely have to act inside the model.

\noindent\textbf{Broader impact.} VLMs are being deployed to read medical images and documents and to perceive the world for robots and vehicles, where the unexpected case is often the one that matters. Our results say when such a case is most likely to be ignored. Override is most likely when the image is degraded at the model's input resolution, as can happen in fog, motion blur or a low-resolution feed, and when the prior is strong, and the strength of the prior can be measured in advance by asking the question without the image. A deployed system can therefore flag the queries on which the image is least likely to be believed, and on most models the decision can be read at mid-depth before the answer is produced. The object is the point of leverage in both directions. Degrading the contradicting object alone can be enough to hand the decision to the prior, so a small local alteration of an image is an attack surface, and an attacker can use the same two cues to find the inputs on which it will succeed. Conversely, restoring the detail of a degraded object recovers many decisions, which makes restoration of degraded regions a promising defense, whereas enhancing an image that is already clear does not help.

\noindent\textbf{Limitations.} The two cues give the direction of each effect but explain only 3 to 10\% of held-out deviance, so they do not predict individual decisions. Three of the eight architectures (Fuyu, Emu3 and Llama-3.2) are represented by a single checkpoint. The steering, injection and photograph analyses cover three LLaVA models only, and on photographs the object's advantage and the corrections do not carry over. Of the 19 pre-specified tests, two were met on every model, seven failed, seven gave mixed outcomes across models and three could not be completed. The failures include the prediction that enhancing the object would raise the image's win rate, and Supp.~O lists each test with its outcome.

\section*{LLM Usage Statement}
\noindent\textbf{Tools.} Claude Opus 5 (Anthropic) was used as a research and coding assistant. GPT-6 (OpenAI) and Gemini 4 Argon (Google) were used to produce critical reviews of drafts. GPT-6.1 Sol, Claude Sonnet 5.5 and Gemini 4 Argon are also subjects of the API experiment of \cref{sec:rule}.

\noindent\textbf{Ideas.} The research question, the hypotheses, the experimental design and the claims of the paper are the authors'. The authors asked Claude which further controls would test their argument and chose which of its suggestions to run.

\noindent\textbf{Code and experiments.} Experiment and analysis code, including the plotting code, was written by the authors together with Claude. Claude's help was largest on the API-model experiment, where it was used to keep the runs within a small compute budget, and on testing. Before a long experiment each notebook runs self-tests that Claude helped write (a patching identity check, and a reproduction of the LLaVA-1.5 band map, readout and head ablation), so that a run does not fail partway. Claude also helped keep the code efficient on the available compute and helped monitor the runs. No data, results or images were generated by an LLM or another generative AI tool.

\noindent\textbf{Writing and literature.} The authors drafted the related work and asked Claude to find further relevant papers, whose metadata was checked against the primary sources. The paper was written by the authors with Claude's assistance, and Claude was used to revise the wording for clarity throughout the paper and the supplement. Claude was also used to audit the submission by comparing the numbers reported in the paper with the saved result files.

\noindent\textbf{Responsibility.} The authors reviewed the text and the results and take responsibility for all content.

{\small
\bibliographystyle{ieeenat_fullname}
\bibliography{refs}

\begin{thebibliography}{131}
\providecommand{\natexlab}[1]{#1}
\providecommand{\url}[1]{\texttt{#1}}
\expandafter\ifx\csname urlstyle\endcsname\relax
  \providecommand{\doi}[1]{doi: #1}\else
  \providecommand{\doi}{doi: \begingroup \urlstyle{rm}\Url}\fi

\bibitem[Agrawal et~al.(2018)Agrawal, Batra, Parikh, and
  Kembhavi]{agrawal2018vqacp}
Aishwarya Agrawal, Dhruv Batra, Devi Parikh, and Aniruddha Kembhavi.
\newblock Don't just assume; look and answer: Overcoming priors for visual
  question answering.
\newblock In \emph{Proceedings of the IEEE/CVF Conference on Computer Vision
  and Pattern Recognition (CVPR)}, 2018.
\newblock arXiv:1712.00377.

\bibitem[Amini et~al.(2025)Amini, Banaszak, Benoit, B{\"o}{\"o}k, Dakhran,
  Duong, Eng, Fernandes, H{\"a}rk{\"o}nen, Harrington, Hasani, Karwa,
  Khrustalev, Labonne, Lechner, Lechner, Lee, Li, Loo, Marks, Mosca, Paech,
  Pak, Parnichkun, Quach, Rogers, Rus, Saxena, Schlager, Seyde, Smith,
  Tadimeti, and Tumma]{amini2025lfm2}
Alexander Amini, Anna Banaszak, Harold Benoit, Arthur B{\"o}{\"o}k, Tarek
  Dakhran, Song Duong, Alfred Eng, Fernando Fernandes, Marc H{\"a}rk{\"o}nen,
  Anne Harrington, Ramin Hasani, Saniya Karwa, Yuri Khrustalev, Maxime Labonne,
  Mathias Lechner, Valentine Lechner, Simon Lee, Zetian Li, Noel Loo, Jacob
  Marks, Edoardo Mosca, Samuel~J. Paech, Paul Pak, Rom~N. Parnichkun, Alex
  Quach, Ryan Rogers, Daniela Rus, Nayan Saxena, Bettina Schlager, Tim Seyde,
  Jimmy T.~H. Smith, Aditya Tadimeti, and Neehal Tumma.
\newblock {LFM2} technical report.
\newblock \emph{arXiv preprint arXiv:2511.23404}, 2025.

\bibitem[{Anthropic}(2026)]{anthropic2026sonnet55}
{Anthropic}.
\newblock {Claude Sonnet 5.5}.
\newblock
  \url{https://platform.claude.com/docs/en/about-claude/models/overview}, 2026.
\newblock API model \texttt{claude-sonnet-5-5}. Accessed 2026-11-12.

\bibitem[Bai et~al.(2025{\natexlab{a}})Bai, Cai, Chen, Chen, Chen, Cheng, Deng,
  Ding, Gao, Ge, Ge, Guo, Huang, Huang, Huang, Hui, Jiang, Li, Li, Li, Li, Lin,
  Lin, Liu, Liu, Liu, Liu, Liu, Liu, Lu, Luo, Lv, Men, Meng, Ren, Ren, Song,
  Sun, Tang, Tu, Wan, Wang, Wang, Wang, Wang, Xie, Xu, Xu, Xu, Yang, Yang,
  Yang, Yang, Yu, Zhang, Zhang, Zhang, Zheng, Zhong, Zhou, Zhou, Zhou, Zhu, and
  Zhu]{bai2025qwen3vl}
Shuai Bai, Yuxuan Cai, Ruizhe Chen, Keqin Chen, Xionghui Chen, Zesen Cheng,
  Lianghao Deng, Wei Ding, Chang Gao, Chunjiang Ge, Wenbin Ge, Zhifang Guo,
  Qidong Huang, Jie Huang, Fei Huang, Binyuan Hui, Shutong Jiang, Zhaohai Li,
  Mingsheng Li, Mei Li, Kaixin Li, Zicheng Lin, Junyang Lin, Xuejing Liu,
  Jiawei Liu, Chenglong Liu, Yang Liu, Dayiheng Liu, Shixuan Liu, Dunjie Lu,
  Ruilin Luo, Chenxu Lv, Rui Men, Lingchen Meng, Xuancheng Ren, Xingzhang Ren,
  Sibo Song, Yuchong Sun, Jun Tang, Jianhong Tu, Jianqiang Wan, Peng Wang,
  Pengfei Wang, Qiuyue Wang, Yuxuan Wang, Tianbao Xie, Yiheng Xu, Haiyang Xu,
  Jin Xu, Zhibo Yang, Mingkun Yang, Jianxin Yang, An Yang, Bowen Yu, Fei Zhang,
  Hang Zhang, Xi Zhang, Bo Zheng, Humen Zhong, Jingren Zhou, Fan Zhou, Jing
  Zhou, Yuanzhi Zhu, and Ke Zhu.
\newblock {Qwen3-VL} technical report.
\newblock \emph{arXiv preprint arXiv:2511.21631}, 2025{\natexlab{a}}.

\bibitem[Bai et~al.(2025{\natexlab{b}})Bai, Chen, Liu, Wang, Ge, Song, Dang,
  Wang, Wang, Tang, Zhong, Zhu, Yang, Li, Wan, Wang, Ding, Fu, Xu, Ye, Zhang,
  Xie, Cheng, Zhang, Yang, Xu, and Lin]{bai2025qwen25vl}
Shuai Bai, Keqin Chen, Xuejing Liu, Jialin Wang, Wenbin Ge, Sibo Song, Kai
  Dang, Peng Wang, Shijie Wang, Jun Tang, Humen Zhong, Yuanzhi Zhu, Mingkun
  Yang, Zhaohai Li, Jianqiang Wan, Pengfei Wang, Wei Ding, Zheren Fu, Yiheng
  Xu, Jiabo Ye, Xi Zhang, Tianbao Xie, Zesen Cheng, Hang Zhang, Zhibo Yang,
  Haiyang Xu, and Junyang Lin.
\newblock {Qwen2.5-VL} technical report.
\newblock \emph{arXiv preprint arXiv:2502.13923}, 2025{\natexlab{b}}.

\bibitem[Barua et~al.(2026)Barua, Mahowald, and Rudman]{barua2026bayesian}
Josh Barua, Kyle Mahowald, and William Rudman.
\newblock Vision-language models integrate conflicting evidence like biased
  {Bayesian} observers.
\newblock \emph{arXiv preprint arXiv:2610.19407}, 2026.

\bibitem[Basu et~al.(2024)Basu, Grayson, Morrison, Nushi, Feizi, and
  Massiceti]{basu2024information}
Samyadeep Basu, Martin Grayson, Cecily Morrison, Besmira Nushi, Soheil Feizi,
  and Daniela Massiceti.
\newblock Understanding information storage and transfer in multi-modal large
  language models.
\newblock In \emph{Advances in Neural Information Processing Systems
  (NeurIPS)}, 2024.
\newblock arXiv:2406.04236.

\bibitem[Bavishi et~al.(2023)Bavishi, Elsen, Hawthorne, Nye, Odena, Somani, and
  Ta{\c{s}}{\i}rlar]{bavishi2023fuyu}
Rohan Bavishi, Erich Elsen, Curtis Hawthorne, Maxwell Nye, Augustus Odena,
  Arushi Somani, and Sa{\u{g}}nak Ta{\c{s}}{\i}rlar.
\newblock {Fuyu-8B}: A multimodal architecture for {AI} agents.
\newblock \url{https://www.adept.ai/blog/fuyu-8b}, 2023.
\newblock Adept AI blog post, October 17, 2023. Accessed 2026-11-11.

\bibitem[Belrose et~al.(2023)Belrose, Schneider-Joseph, Ravfogel, Cotterell,
  Raff, and Biderman]{belrose2023leace}
Nora Belrose, David Schneider-Joseph, Shauli Ravfogel, Ryan Cotterell, Edward
  Raff, and Stella Biderman.
\newblock {LEACE}: Perfect linear concept erasure in closed form.
\newblock In \emph{Advances in Neural Information Processing Systems
  (NeurIPS)}, 2023.
\newblock arXiv:2306.03819.

\bibitem[Bitton-Guetta et~al.(2023)Bitton-Guetta, Bitton, Hessel, Schmidt,
  Elovici, Stanovsky, and Schwartz]{bittonguetta2023whoops}
Nitzan Bitton-Guetta, Yonatan Bitton, Jack Hessel, Ludwig Schmidt, Yuval
  Elovici, Gabriel Stanovsky, and Roy Schwartz.
\newblock Breaking common sense: {WHOOPS}! {A} vision-and-language benchmark of
  synthetic and compositional images.
\newblock In \emph{Proceedings of the IEEE/CVF International Conference on
  Computer Vision (ICCV)}, pages 2616--2627, 2023.
\newblock arXiv:2303.07274.

\bibitem[Carion et~al.(2026)Carion, Gustafson, Hu, Debnath, Hu, Suris, Ryali,
  Alwala, Khedr, Huang, Lei, Ma, Guo, Kalla, Marks, Greer, Wang, Sun,
  R{\"a}dle, Afouras, Mavroudi, Xu, Wu, Zhou, Momeni, Hazra, Ding, Vaze,
  Porcher, Li, Li, Kamath, Cheng, Doll{\'a}r, Ravi, Saenko, Zhang, and
  Feichtenhofer]{sam3}
Nicolas Carion, Laura Gustafson, Yuan-Ting Hu, Shoubhik Debnath, Ronghang Hu,
  Didac Suris, Chaitanya Ryali, Kalyan~Vasudev Alwala, Haitham Khedr, Andrew
  Huang, Jie Lei, Tengyu Ma, Baishan Guo, Arpit Kalla, Markus Marks, Joseph
  Greer, Meng Wang, Peize Sun, Roman R{\"a}dle, Triantafyllos Afouras,
  Effrosyni Mavroudi, Katherine Xu, Tsung-Han Wu, Yu Zhou, Liliane Momeni,
  Rishi Hazra, Shuangrui Ding, Sagar Vaze, Francois Porcher, Feng Li, Siyuan
  Li, Aishwarya Kamath, Ho~Kei Cheng, Piotr Doll{\'a}r, Nikhila Ravi, Kate
  Saenko, Pengchuan Zhang, and Christoph Feichtenhofer.
\newblock {SAM} 3: Segment anything with concepts.
\newblock In \emph{International Conference on Learning Representations
  (ICLR)}, 2026.
\newblock arXiv:2511.16719.

\bibitem[Chan et~al.(2026)Chan, Suresh, Saha, Hahn, and Demberg]{chan2026yes}
Tsan~Tsai Chan, Varsha Suresh, Anisha Saha, Michael Hahn, and Vera Demberg.
\newblock System-mediated attention imbalances make vision-language models say
  yes.
\newblock In \emph{Findings of the Association for Computational Linguistics:
  ACL 2026}, pages 38958--38979, 2026.
\newblock arXiv:2601.12430.

\bibitem[Cheng et~al.(2024)Cheng, Xiao, Gu, Yang, Duan, Zhang, Cao, Xu, and
  Xu]{cheng2024typographic}
Hao Cheng, Erjia Xiao, Jindong Gu, Le Yang, Jinhao Duan, Jize Zhang, Jiahang
  Cao, Kaidi Xu, and Renjing Xu.
\newblock Unveiling typographic deceptions: Insights of the typographic
  vulnerability in large vision-language models.
\newblock In \emph{Proceedings of the European Conference on Computer Vision
  (ECCV)}, pages 179--196, 2024.
\newblock arXiv:2402.19150.

\bibitem[Cheng et~al.(2026)Cheng, Chen, Zhang, and Shen]{cheng2026crg}
Zhe Cheng, Wenyu Chen, Fode Zhang, and Dehuan Shen.
\newblock Mitigating hallucinations in large vision-language models via causal
  route gating.
\newblock In \emph{Proceedings of the International Conference on Machine
  Learning (ICML)}, pages 18828--18860, 2026.
\newblock arXiv:2605.24024.

\bibitem[Dai et~al.(2023)Dai, Li, Li, Tiong, Zhao, Wang, Li, Fung, and
  Hoi]{dai2023instructblip}
Wenliang Dai, Junnan Li, Dongxu Li, Anthony Meng~Huat Tiong, Junqi Zhao,
  Weisheng Wang, Boyang Li, Pascale Fung, and Steven Hoi.
\newblock {InstructBLIP}: Towards general-purpose vision-language models with
  instruction tuning.
\newblock In \emph{Advances in Neural Information Processing Systems
  (NeurIPS)}, 2023.
\newblock arXiv:2305.06500.

\bibitem[Deng et~al.(2025)Deng, Cao, Chen, and Hooi]{deng2025blindfaith}
Ailin Deng, Tri Cao, Zhirui Chen, and Bryan Hooi.
\newblock Words or vision: Do vision-language models have blind faith in text?
\newblock In \emph{Proceedings of the IEEE/CVF Conference on Computer Vision
  and Pattern Recognition (CVPR)}, pages 3867--3876, 2025.
\newblock arXiv:2503.02199.

\bibitem[Du et~al.(2024)Du, Sn{\ae}bjarnarson, Stoehr, White, Schein, and
  Cotterell]{du2024persuasion}
Kevin Du, V{\'e}steinn Sn{\ae}bjarnarson, Niklas Stoehr, Jennifer~C. White,
  Aaron Schein, and Ryan Cotterell.
\newblock Context versus prior knowledge in language models.
\newblock In \emph{Proceedings of the 62nd Annual Meeting of the Association
  for Computational Linguistics (Volume 1: Long Papers)}, pages 13211--13235,
  2024.
\newblock arXiv:2404.04633.

\bibitem[Ernst and Banks(2002)]{ernst2002humans}
Marc~O. Ernst and Martin~S. Banks.
\newblock Humans integrate visual and haptic information in a statistically
  optimal fashion.
\newblock \emph{Nature}, 415\penalty0 (6870):\penalty0 429--433, 2002.

\bibitem[Favero et~al.(2024)Favero, Zancato, Trager, Choudhary, Perera,
  Achille, Swaminathan, and Soatto]{favero2024m3id}
Alessandro Favero, Luca Zancato, Matthew Trager, Siddharth Choudhary,
  Pramuditha Perera, Alessandro Achille, Ashwin Swaminathan, and Stefano
  Soatto.
\newblock {Multi-Modal} hallucination control by visual information grounding.
\newblock In \emph{Proceedings of the IEEE/CVF Conference on Computer Vision
  and Pattern Recognition (CVPR)}, pages 14303--14312, 2024.
\newblock arXiv:2403.14003.

\bibitem[Fu et~al.(2025{\natexlab{a}})Fu, Chen, Shen, Qin, Zhang, Lin, Yang,
  Zheng, Li, Sun, Wu, Ji, Shan, and He]{fu2023mme}
Chaoyou Fu, Peixian Chen, Yunhang Shen, Yulei Qin, Mengdan Zhang, Xu Lin,
  Jinrui Yang, Xiawu Zheng, Ke Li, Xing Sun, Yunsheng Wu, Rongrong Ji, Caifeng
  Shan, and Ran He.
\newblock {MME}: A comprehensive evaluation benchmark for multimodal large
  language models.
\newblock In \emph{Advances in Neural Information Processing Systems (NeurIPS),
  Datasets and Benchmarks Track}, 2025{\natexlab{a}}.
\newblock arXiv:2306.13394.

\bibitem[Fu et~al.(2025{\natexlab{b}})Fu, Bonnen, Guillory, and
  Darrell]{fu2025hidden}
Stephanie Fu, Tyler Bonnen, Devin Guillory, and Trevor Darrell.
\newblock Hidden in plain sight: {VLMs} overlook their visual representations.
\newblock In \emph{Conference on Language Modeling (COLM)}, 2025{\natexlab{b}}.
\newblock arXiv:2506.08008.

\bibitem[Gandikota and Bau(2026)]{gandikota2026gaze}
Rohit Gandikota and David Bau.
\newblock Gaze heads: How {VLMs} look at what they describe.
\newblock \emph{arXiv preprint arXiv:2606.14703}, 2026.

\bibitem[Geiger et~al.(2024)Geiger, Wu, Potts, Icard, and
  Goodman]{geiger2024das}
Atticus Geiger, Zhengxuan Wu, Christopher Potts, Thomas Icard, and Noah~D.
  Goodman.
\newblock Finding alignments between interpretable causal variables and
  distributed neural representations.
\newblock In \emph{Proceedings of the Third Conference on Causal Learning and
  Reasoning (CLeaR)}, pages 160--187, 2024.
\newblock arXiv:2303.02536.

\bibitem[{Gemma Team}(2025)]{gemmateam2025gemma3}
{Gemma Team}.
\newblock {Gemma 3} technical report.
\newblock \emph{arXiv preprint arXiv:2503.19786}, 2025.

\bibitem[Ghosh et~al.(2026)Ghosh, Gupta, and Wang]{ghosh2026whichsource}
Rodela Ghosh, Aviral Gupta, and Guangjing Wang.
\newblock Which source wins? {Task-dependent} reliance in vision-language
  models.
\newblock \emph{arXiv preprint arXiv:2608.17205}, 2026.

\bibitem[Golovanevsky et~al.(2025{\natexlab{a}})Golovanevsky, Rudman, Lepori,
  Bar, Singh, and Eickhoff]{golovanevsky2025pixels}
Michal Golovanevsky, William Rudman, Michael~A. Lepori, Amir Bar, Ritambhara
  Singh, and Carsten Eickhoff.
\newblock Pixels versus priors: Controlling knowledge priors in vision-language
  models through visual counterfacts.
\newblock In \emph{Proceedings of the 2025 Conference on Empirical Methods in
  Natural Language Processing}, pages 24837--24852, 2025{\natexlab{a}}.
\newblock arXiv:2505.17127.

\bibitem[Golovanevsky et~al.(2025{\natexlab{b}})Golovanevsky, Rudman, Palit,
  Eickhoff, and Singh]{golovanevsky2025notice}
Michal Golovanevsky, William Rudman, Vedant Palit, Carsten Eickhoff, and
  Ritambhara Singh.
\newblock What do {VLMs} {NOTICE}? {A} mechanistic interpretability pipeline
  for {Gaussian}-noise-free text-image corruption and evaluation.
\newblock In \emph{Proceedings of the 2025 Conference of the Nations of the
  Americas Chapter of the Association for Computational Linguistics: Human
  Language Technologies (Volume 1: Long Papers)}, pages 11462--11482,
  2025{\natexlab{b}}.
\newblock arXiv:2406.16320.

\bibitem[{Google}(2026)]{google2026gemini4argon}
{Google}.
\newblock {Gemini 4 Argon}.
\newblock \url{https://ai.google.dev/gemini-api/docs/models}, 2026.
\newblock API model \texttt{gemini-4-argon}. Accessed 2026-11-12.

\bibitem[Goyal et~al.(2017)Goyal, Khot, Summers-Stay, Batra, and
  Parikh]{goyal2017vqav2}
Yash Goyal, Tejas Khot, Douglas Summers-Stay, Dhruv Batra, and Devi Parikh.
\newblock Making the {V} in {VQA} matter: Elevating the role of image
  understanding in visual question answering.
\newblock In \emph{Proceedings of the IEEE Conference on Computer Vision and
  Pattern Recognition (CVPR)}, pages 6904--6913, 2017.
\newblock arXiv:1612.00837.

\bibitem[Grob et~al.(2026)Grob, Kim, Schmid, and Akata]{grob2026bigger}
Fabian Grob, Sanghwan Kim, Cordelia Schmid, and Zeynep Akata.
\newblock Bigger is not better: Inverse scaling and arbitration failure in
  counterfactual visual grounding.
\newblock In \emph{ICML 2026 Workshop on Mechanistic Interpretability}, 2026.
\newblock \url{https://openreview.net/forum?id=b7PdSEO2Pm}.

\bibitem[Guan et~al.(2024)Guan, Liu, Wu, Xian, Li, Liu, Wang, Chen, Huang,
  Yacoob, Manocha, and Zhou]{guan2023hallusionbench}
Tianrui Guan, Fuxiao Liu, Xiyang Wu, Ruiqi Xian, Zongxia Li, Xiaoyu Liu, Xijun
  Wang, Lichang Chen, Furong Huang, Yaser Yacoob, Dinesh Manocha, and Tianyi
  Zhou.
\newblock {HallusionBench}: An advanced diagnostic suite for entangled language
  hallucination and visual illusion in large vision-language models.
\newblock In \emph{Proceedings of the IEEE/CVF Conference on Computer Vision
  and Pattern Recognition (CVPR)}, pages 14375--14385, 2024.
\newblock arXiv:2310.14566.

\bibitem[Han et~al.(2024)Han, Lian, Pan, Pi, Zhang, Diao, Lin, and
  Zhang]{han2024instinctive}
Tianyang Han, Qing Lian, Rui Pan, Renjie Pi, Jipeng Zhang, Shizhe Diao, Yong
  Lin, and Tong Zhang.
\newblock The instinctive bias: Spurious images lead to illusion in {MLLMs}.
\newblock In \emph{Proceedings of the 2024 Conference on Empirical Methods in
  Natural Language Processing}, pages 16163--16177, 2024.
\newblock arXiv:2402.03757.

\bibitem[He et~al.(2026)He, AL-Dweikat, Shyutts, AlDahoul, Sonuga, Zhu, Li, and
  Mujumdar]{he2026modalityconflict}
Jiahang He, Dia'a AL-Dweikat, Artur Shyutts, Abdullah AlDahoul, Oyedeji Sonuga,
  Kevin Zhu, Ruizhe Li, and Anusha Mujumdar.
\newblock Mechanistic analysis and inference-time control of modality conflict
  in {VLMs}.
\newblock In \emph{ICML 2026 Workshop on Mechanistic Interpretability}, 2026.
\newblock \url{https://openreview.net/forum?id=1XzMBbB1Rn}.

\bibitem[Hegde et~al.(2026)Hegde, Liu, Patel, and Yang]{hegde2026routing}
Shamanthak Hegde, Xiangrui Liu, Maitreya Patel, and Yezhou Yang.
\newblock Where hallucinations live: A cross-architecture circuit in
  {VQ}-tokenized vision-language models.
\newblock In \emph{Proceedings of the 2026 Conference on Empirical Methods in
  Natural Language Processing}, 2026.
\newblock arXiv:2609.29048.

\bibitem[Heimersheim and Nanda(2024)]{heimersheim2024patching}
Stefan Heimersheim and Neel Nanda.
\newblock How to use and interpret activation patching.
\newblock \emph{arXiv preprint arXiv:2404.15255}, 2024.

\bibitem[Hendrycks and Dietterich(2019)]{hendrycks2019benchmarking}
Dan Hendrycks and Thomas Dietterich.
\newblock Benchmarking neural network robustness to common corruptions and
  perturbations.
\newblock In \emph{International Conference on Learning Representations
  (ICLR)}, 2019.

\bibitem[Hua et~al.(2025)Hua, Yun, and Pavlick]{hua2025conflicting}
Tianze Hua, Tian Yun, and Ellie Pavlick.
\newblock How do vision-language models process conflicting information across
  modalities?
\newblock \emph{arXiv preprint arXiv:2507.01790}, 2025.

\bibitem[Huang et~al.(2024)Huang, Dong, Zhang, Wang, He, Wang, Lin, Zhang, and
  Yu]{huang2023opera}
Qidong Huang, Xiaoyi Dong, Pan Zhang, Bin Wang, Conghui He, Jiaqi Wang, Dahua
  Lin, Weiming Zhang, and Nenghai Yu.
\newblock {OPERA}: Alleviating hallucination in multi-modal large language
  models via over-trust penalty and retrospection-allocation.
\newblock In \emph{Proceedings of the IEEE/CVF Conference on Computer Vision
  and Pattern Recognition (CVPR)}, pages 13418--13427, 2024.
\newblock arXiv:2311.17911.

\bibitem[Jia et~al.(2026)Jia, Du, Jiang, Liang, Ren, Xin, Yang, Feng, Chen, Lu,
  Wang, Qu, Li, and Liu]{jia2026mmkc}
Yifan Jia, Yuntao Du, Kailin Jiang, Yuyang Liang, Qihan Ren, Yi Xin, Rui Yang,
  Fenze Feng, Mingcai Chen, Hengyang Lu, Haozhe Wang, Xiaoye Qu, Qian Li, and
  Dongrui Liu.
\newblock Benchmarking multimodal knowledge conflict for large multimodal
  models.
\newblock In \emph{Proceedings of the AAAI Conference on Artificial
  Intelligence}, pages 22283--22291, 2026.
\newblock arXiv:2505.19509.

\bibitem[Jiang et~al.(2023)Jiang, Sablayrolles, Mensch, Bamford, Chaplot,
  de~las Casas, Bressand, Lengyel, Lample, Saulnier, Lavaud, Lachaux, Stock,
  Le~Scao, Lavril, Wang, Lacroix, and El~Sayed]{jiang2023mistral}
Albert~Q. Jiang, Alexandre Sablayrolles, Arthur Mensch, Chris Bamford,
  Devendra~Singh Chaplot, Diego de~las Casas, Florian Bressand, Gianna Lengyel,
  Guillaume Lample, Lucile Saulnier, L{\'e}lio~Renard Lavaud, Marie-Anne
  Lachaux, Pierre Stock, Teven Le~Scao, Thibaut Lavril, Thomas Wang,
  Timoth{\'e}e Lacroix, and William El~Sayed.
\newblock {Mistral} {7B}.
\newblock \emph{arXiv preprint arXiv:2310.06825}, 2023.

\bibitem[Jiang et~al.(2025{\natexlab{a}})Jiang, Kachinthaya, Petryk, and
  Gandelsman]{jiang2024interpretingedit}
Nick Jiang, Anish Kachinthaya, Suzie Petryk, and Yossi Gandelsman.
\newblock Interpreting and editing vision-language representations to mitigate
  hallucinations.
\newblock In \emph{International Conference on Learning Representations
  (ICLR)}, pages 63582--63605, 2025{\natexlab{a}}.
\newblock arXiv:2410.02762.

\bibitem[Jiang et~al.(2025{\natexlab{b}})Jiang, Chen, Zhu, Luo, Shen, and
  Yang]{jiang2024devils}
Zhangqi Jiang, Junkai Chen, Beier Zhu, Tingjin Luo, Yankun Shen, and Xu Yang.
\newblock Devils in middle layers of large vision-language models:
  Interpreting, detecting and mitigating object hallucinations via attention
  lens.
\newblock In \emph{Proceedings of the IEEE/CVF Conference on Computer Vision
  and Pattern Recognition (CVPR)}, pages 25004--25014, 2025{\natexlab{b}}.
\newblock arXiv:2411.16724.

\bibitem[Jin et~al.(2024)Jin, Cao, Yuan, Chen, Xu, Li, Jiang, Liu, and
  Zhao]{jin2024cutting}
Zhuoran Jin, Pengfei Cao, Hongbang Yuan, Yubo Chen, Jiexin Xu, Huaijun Li,
  Xiaojian Jiang, Kang Liu, and Jun Zhao.
\newblock Cutting off the head ends the conflict: A mechanism for interpreting
  and mitigating knowledge conflicts in language models.
\newblock In \emph{Findings of the Association for Computational Linguistics:
  ACL 2024}, pages 1193--1215, 2024.
\newblock arXiv:2402.18154.

\bibitem[Kadir and Sonntag(2026)]{kadir2026audit}
Md~Abdul Kadir and Daniel Sonntag.
\newblock What do visual-head discovery scores actually find? {Attention}
  source, necessity, and random-control pitfalls in vision-language models.
\newblock \emph{arXiv preprint arXiv:2610.13111}, 2026.

\bibitem[Kaduri et~al.(2025)Kaduri, Bagon, and Dekel]{kaduri2025whats}
Omri Kaduri, Shai Bagon, and Tali Dekel.
\newblock What's in the image? {A} deep-dive into the vision of vision language
  models.
\newblock In \emph{Proceedings of the IEEE/CVF Conference on Computer Vision
  and Pattern Recognition (CVPR)}, pages 14549--14558, 2025.
\newblock arXiv:2411.17491.

\bibitem[Kang et~al.(2025)Kang, Kim, Kim, and Hwang]{kang2025var}
Seil Kang, Jinyeong Kim, Junhyeok Kim, and Seong~Jae Hwang.
\newblock See what you are told: Visual attention sink in large multimodal
  models.
\newblock In \emph{International Conference on Learning Representations
  (ICLR)}, pages 87676--87703, 2025.
\newblock arXiv:2503.03321.

\bibitem[Kim et~al.(2025)Kim, Kang, Park, Kim, and
  Hwang]{kim2025headattribution}
Jinyeong Kim, Seil Kang, Jiwoo Park, Junhyeok Kim, and Seong~Jae Hwang.
\newblock Interpreting attention heads for image-to-text information flow in
  large vision-language models.
\newblock \emph{arXiv preprint arXiv:2509.17588}, 2025.

\bibitem[Knill and Pouget(2004)]{knill2004bayesian}
David~C. Knill and Alexandre Pouget.
\newblock The {Bayesian} brain: the role of uncertainty in neural coding and
  computation.
\newblock \emph{Trends in Neurosciences}, 27\penalty0 (12):\penalty0 712--719,
  2004.

\bibitem[Kogilathota et~al.(2026)Kogilathota, Vallabha E~G, Sun, and
  Zhou]{kogilathota2026halp}
Sai~Akhil Kogilathota, Sripadha Vallabha E~G, Luzhe Sun, and Jiawei Zhou.
\newblock {HALP}: Detecting hallucinations in vision-language models without
  generating a single token.
\newblock In \emph{Proceedings of the 19th Conference of the European Chapter
  of the Association for Computational Linguistics (Volume 1: Long Papers)},
  pages 6067--6085, 2026.
\newblock arXiv:2603.05465.

\bibitem[Lan et~al.(2026)Lan, Sun, Walter, and
  Zhou]{lan2026seeingwithoutlooking}
Zixuan Lan, Luzhe Sun, Matthew~R. Walter, and Jiawei Zhou.
\newblock Seeing without looking: Do vision-language benchmarks really test
  vision?
\newblock In \emph{Proceedings of the IEEE/CVF Conference on Computer Vision
  and Pattern Recognition (CVPR) Workshops}, pages 11260--11273, 2026.
\newblock arXiv:2605.22903.

\bibitem[Lauren{\c{c}}on et~al.(2024)Lauren{\c{c}}on, Tronchon, Cord, and
  Sanh]{laurencon2024idefics2}
Hugo Lauren{\c{c}}on, L{\'e}o Tronchon, Matthieu Cord, and Victor Sanh.
\newblock What matters when building vision-language models?
\newblock In \emph{Advances in Neural Information Processing Systems
  (NeurIPS)}, pages 87874--87907, 2024.
\newblock arXiv:2405.02246.

\bibitem[Lee et~al.(2025)Lee, Kim, Yoon, Kim, Lee, Koh, and Jung]{lee2024vlind}
Kang-il Lee, Minbeom Kim, Seunghyun Yoon, Minsung Kim, Dongryeol Lee, Hyukhun
  Koh, and Kyomin Jung.
\newblock {VLind-Bench}: Measuring language priors in large vision-language
  models.
\newblock In \emph{Findings of the Association for Computational Linguistics:
  NAACL 2025}, 2025.
\newblock arXiv:2406.08702.

\bibitem[Leng et~al.(2024)Leng, Zhang, Chen, Li, Lu, Miao, and
  Bing]{leng2023vcd}
Sicong Leng, Hang Zhang, Guanzheng Chen, Xin Li, Shijian Lu, Chunyan Miao, and
  Lidong Bing.
\newblock Mitigating object hallucinations in large vision-language models
  through visual contrastive decoding.
\newblock In \emph{Proceedings of the IEEE/CVF Conference on Computer Vision
  and Pattern Recognition (CVPR)}, pages 13872--13882, 2024.
\newblock arXiv:2311.16922.

\bibitem[Li et~al.(2024)Li, Lin, Peng, Nyandwi, Jiang, Ma, Khanuja, Krishna,
  Neubig, and Ramanan]{li2024naturalbench}
Baiqi Li, Zhiqiu Lin, Wenxuan Peng, Jean de~Dieu Nyandwi, Daniel Jiang, Zixian
  Ma, Simran Khanuja, Ranjay Krishna, Graham Neubig, and Deva Ramanan.
\newblock {NaturalBench}: Evaluating vision-language models on natural
  adversarial samples.
\newblock In \emph{Advances in Neural Information Processing Systems (NeurIPS),
  Datasets and Benchmarks Track}, pages 17044--17068, 2024.
\newblock arXiv:2410.14669.

\bibitem[Li et~al.(2023{\natexlab{a}})Li, Li, Savarese, and Hoi]{li2023blip2}
Junnan Li, Dongxu Li, Silvio Savarese, and Steven Hoi.
\newblock {BLIP-2}: Bootstrapping language-image pre-training with frozen image
  encoders and large language models.
\newblock In \emph{Proceedings of the International Conference on Machine
  Learning (ICML)}, pages 19730--19742, 2023{\natexlab{a}}.
\newblock arXiv:2301.12597.

\bibitem[Li et~al.(2026)Li, Li, An, Yuan, Liu, Belongie, and
  Sn{\ae}bjarnarson]{li2026seeingorknowing}
Jiaang Li, Chengzu Li, Zhaochong An, Yifei Yuan, Xi Liu, Serge Belongie, and
  V{\'e}steinn Sn{\ae}bjarnarson.
\newblock Seeing or knowing? {Visual} context sensitivity in multimodal large
  language models.
\newblock In \emph{UniWorld Workshop at ECCV 2026}, 2026.
\newblock arXiv:2607.26326.

\bibitem[Li et~al.(2022)Li, Chen, Wang, Hong, Ye, Han, Chen, Zhang, Xu, Yeung,
  Liang, Li, and Xu]{li2022coda}
Kaican Li, Kai Chen, Haoyu Wang, Lanqing Hong, Chaoqiang Ye, Jianhua Han,
  Yukuai Chen, Wei Zhang, Chunjing Xu, Dit-Yan Yeung, Xiaodan Liang, Zhenguo
  Li, and Hang Xu.
\newblock {CODA}: A real-world road corner case dataset for object detection in
  autonomous driving.
\newblock In \emph{Proceedings of the European Conference on Computer Vision
  (ECCV)}, 2022.

\bibitem[Li et~al.(2023{\natexlab{b}})Li, Du, Zhou, Wang, Zhao, and
  Wen]{li2023pope}
Yifan Li, Yifan Du, Kun Zhou, Jinpeng Wang, Wayne~Xin Zhao, and Ji-Rong Wen.
\newblock Evaluating object hallucination in large vision-language models.
\newblock In \emph{Proceedings of the 2023 Conference on Empirical Methods in
  Natural Language Processing}, pages 292--305, 2023{\natexlab{b}}.
\newblock arXiv:2305.10355.

\bibitem[Liang et~al.(2026)Liang, Shi, Zhu, Li, Chen, Liu, Li, and
  Xue]{liang2026enar}
Yuxuan Liang, Fan Shi, Rui Zhu, Xu Li, Xiaolei Chen, Zhe Liu, Bin Li, and
  Xiangyang Xue.
\newblock Envision, attend, then respond: Counterfactual hallucination
  mitigation in large vision-language models.
\newblock In \emph{Proceedings of the IEEE/CVF Conference on Computer Vision
  and Pattern Recognition (CVPR)}, 2026.

\bibitem[Liao et~al.(2026)Liao, Shen, Cao, and Chua]{liao2026decodable}
Chih-Ting Liao, Fei Shen, Xin Cao, and Tat-Seng Chua.
\newblock Decodable is not grounded: A vision-ablation arbiter for {VLM}
  spatial reasoning.
\newblock \emph{arXiv preprint arXiv:2606.31257}, 2026.

\bibitem[Lietzow et~al.(2026)Lietzow, Bitterman, Eickhoff, Rudman, and
  Golovanevsky]{lietzow2026visiondefault}
Niclas Lietzow, Danielle Bitterman, Carsten Eickhoff, William Rudman, and
  Michal Golovanevsky.
\newblock Vision-default, prior-override: Causal mechanisms of
  perception-knowledge conflict in vision-language models.
\newblock In \emph{Findings of the Association for Computational Linguistics:
  EMNLP 2026}, pages 43282--43298, 2026.
\newblock arXiv:2606.28273.

\bibitem[Lin et~al.(2014)Lin, Maire, Belongie, Hays, Perona, Ramanan,
  Doll{\'a}r, and Zitnick]{lin2014coco}
Tsung-Yi Lin, Michael Maire, Serge Belongie, James Hays, Pietro Perona, Deva
  Ramanan, Piotr Doll{\'a}r, and C.~Lawrence Zitnick.
\newblock Microsoft {COCO}: Common objects in context.
\newblock In \emph{Proceedings of the European Conference on Computer Vision
  (ECCV)}, pages 740--755, 2014.
\newblock arXiv:1405.0312.

\bibitem[{Liquid AI}(2025)]{liquid2025lfm2vl}
{Liquid AI}.
\newblock {LFM2-VL-3B}: A new efficient vision-language for the edge.
\newblock
  \url{https://www.liquid.ai/blog/lfm2-vl-3b-a-new-efficient-vision-language-for-the-edge},
  2025.
\newblock Blog post, October 22, 2025. Accessed 2026-11-11.

\bibitem[Liu et~al.(2023)Liu, Li, Wu, and Lee]{liu2023llava}
Haotian Liu, Chunyuan Li, Qingyang Wu, and Yong~Jae Lee.
\newblock Visual instruction tuning.
\newblock In \emph{Advances in Neural Information Processing Systems
  (NeurIPS)}, pages 34892--34916, 2023.
\newblock arXiv:2304.08485.

\bibitem[Liu et~al.(2024{\natexlab{a}})Liu, Li, Li, and
  Lee]{liu2024improvedllava}
Haotian Liu, Chunyuan Li, Yuheng Li, and Yong~Jae Lee.
\newblock Improved baselines with visual instruction tuning.
\newblock In \emph{Proceedings of the IEEE/CVF Conference on Computer Vision
  and Pattern Recognition (CVPR)}, pages 26286--26296, 2024{\natexlab{a}}.
\newblock arXiv:2310.03744.

\bibitem[Liu et~al.(2024{\natexlab{b}})Liu, Li, Li, Li, Zhang, Shen, and
  Lee]{liu2024llavanext}
Haotian Liu, Chunyuan Li, Yuheng Li, Bo Li, Yuanhan Zhang, Sheng Shen, and
  Yong~Jae Lee.
\newblock {LLaVA-NeXT}: Improved reasoning, {OCR}, and world knowledge.
\newblock \url{https://llava-vl.github.io/blog/2024-01-30-llava-next/},
  2024{\natexlab{b}}.
\newblock Accessed 2026-11-04.

\bibitem[Liu et~al.(2024{\natexlab{c}})Liu, Min, Zettlemoyer, Choi, and
  Hajishirzi]{liu2024infinigram}
Jiacheng Liu, Sewon Min, Luke Zettlemoyer, Yejin Choi, and Hannaneh Hajishirzi.
\newblock {Infini-gram}: Scaling unbounded {n-gram} language models to a
  trillion tokens.
\newblock In \emph{Conference on Language Modeling (COLM)}, 2024{\natexlab{c}}.
\newblock arXiv:2401.17377.

\bibitem[Liu et~al.(2026{\natexlab{a}})Liu, Zhong, Wang, Yang, Kang, Dong,
  Zhan, Fang, and Liu]{liu2026dualpathway}
Jiaxin Liu, Ding Zhong, Yue Wang, Zhidong Yang, Zhaolu Kang, Guangyuan Dong,
  Qishi Zhan, Pengcheng Fang, and Aofan Liu.
\newblock Dual-pathway circuits of object hallucination in vision-language
  models.
\newblock \emph{arXiv preprint arXiv:2605.13156}, 2026{\natexlab{a}}.

\bibitem[Liu et~al.(2024{\natexlab{d}})Liu, Zheng, and Chen]{liu2024pai}
Shi Liu, Kecheng Zheng, and Wei Chen.
\newblock Paying more attention to image: A training-free method for
  alleviating hallucination in {LVLMs}.
\newblock In \emph{Proceedings of the European Conference on Computer Vision
  (ECCV)}, pages 125--140, 2024{\natexlab{d}}.
\newblock arXiv:2407.21771.

\bibitem[Liu et~al.(2025{\natexlab{a}})Liu, Wang, Yuan, Huang, Liu, He, and
  Tu]{liu2025insight}
Xiaoyuan Liu, Wenxuan Wang, Youliang Yuan, Jen-tse Huang, Qiuzhi Liu, Pinjia
  He, and Zhaopeng Tu.
\newblock Insight over sight: Exploring the vision-knowledge conflicts in
  multimodal {LLMs}.
\newblock In \emph{Proceedings of the 63rd Annual Meeting of the Association
  for Computational Linguistics (Volume 1: Long Papers)}, pages 17825--17846,
  2025{\natexlab{a}}.
\newblock arXiv:2410.08145.

\bibitem[Liu et~al.(2025{\natexlab{b}})Liu, Liang, Wang, Wu, Tang, He, Li, Liu,
  Yang, Lim, and Zhao]{liu2025ignorance}
Yexin Liu, Zhengyang Liang, Yueze Wang, Xianfeng Wu, Feilong Tang, Muyang He,
  Jian Li, Zheng Liu, Harry Yang, Sernam Lim, and Bo Zhao.
\newblock Unveiling the ignorance of {MLLMs}: Seeing clearly, answering
  incorrectly.
\newblock In \emph{Proceedings of the IEEE/CVF Conference on Computer Vision
  and Pattern Recognition (CVPR)}, pages 9087--9097, 2025{\natexlab{b}}.
\newblock arXiv:2406.10638.

\bibitem[Liu et~al.(2026{\natexlab{b}})Liu, Chen, Liu, Luo, Tang, Wang, Zeng,
  Dai, Shi, Wei, Lu, Dumoulin, and Tong]{liu2025seeingnotbelieving}
Zhining Liu, Ziyi Chen, Hui Liu, Chen Luo, Xianfeng Tang, Suhang Wang, Jingying
  Zeng, Zhenwei Dai, Zhan Shi, Tianxin Wei, Hanqing Lu, Benoit Dumoulin, and
  Hanghang Tong.
\newblock Seeing but not believing: Probing the disconnect between visual
  attention and answer correctness in {VLMs}.
\newblock In \emph{International Conference on Learning Representations
  (ICLR)}, 2026{\natexlab{b}}.
\newblock arXiv:2510.17771.

\bibitem[Long et~al.(2026)Long, Oh, Park, and Li]{long2026chainofembedding}
Lin Long, Changdae Oh, Seongheon Park, and Sharon Li.
\newblock Understanding language prior of {LVLMs} by contrasting
  chain-of-embedding.
\newblock In \emph{International Conference on Learning Representations
  (ICLR)}, 2026.
\newblock arXiv:2509.23050.

\bibitem[Longpre et~al.(2021)Longpre, Perisetla, Chen, Ramesh, DuBois, and
  Singh]{longpre2021entity}
Shayne Longpre, Kartik Perisetla, Anthony Chen, Nikhil Ramesh, Chris DuBois,
  and Sameer Singh.
\newblock Entity-based knowledge conflicts in question answering.
\newblock In \emph{Proceedings of the 2021 Conference on Empirical Methods in
  Natural Language Processing}, pages 7052--7063, 2021.
\newblock arXiv:2109.05052.

\bibitem[L{\"u}ddecke and Ecker(2022)]{luddecke2022clipseg}
Timo L{\"u}ddecke and Alexander Ecker.
\newblock Image segmentation using text and image prompts.
\newblock In \emph{Proceedings of the IEEE/CVF Conference on Computer Vision
  and Pattern Recognition (CVPR)}, pages 7086--7096, 2022.
\newblock CLIPSeg; arXiv:2112.10003.

\bibitem[Luo et~al.(2026)Luo, Fan, Wang, He, Rahman, Abolmaesumi, and
  Sigal]{luo2025sink}
Jiayun Luo, Wan-Cyuan Fan, Lyuyang Wang, Xiangteng He, Tanzila Rahman, Purang
  Abolmaesumi, and Leonid Sigal.
\newblock To sink or not to sink: Visual information pathways in large
  vision-language models.
\newblock In \emph{International Conference on Learning Representations
  (ICLR)}, 2026.
\newblock arXiv:2510.08510.

\bibitem[Luo et~al.(2025)Luo, Cao, Lee, Johnson, and Lee]{luo2025vilp}
Tiange Luo, Ang Cao, Gunhee Lee, Justin Johnson, and Honglak Lee.
\newblock Probing visual language priors in {VLMs}.
\newblock In \emph{Proceedings of the International Conference on Machine
  Learning (ICML)}, pages 41120--41156, 2025.
\newblock arXiv:2501.00569.

\bibitem[Ma et~al.(2025)Ma, Wang, and Fountas]{ma2025bayesbench}
Julian Ma, Jun Wang, and Zafeirios Fountas.
\newblock Emergent {Bayesian} behaviour and optimal cue combination in {LLMs}.
\newblock \emph{arXiv preprint arXiv:2512.02719}, 2025.

\bibitem[Makelov et~al.(2024)Makelov, Lange, Geiger, and
  Nanda]{makelov2024illusion}
Aleksandar Makelov, Georg Lange, Atticus Geiger, and Neel Nanda.
\newblock Is this the subspace you are looking for? {An} interpretability
  illusion for subspace activation patching.
\newblock In \emph{International Conference on Learning Representations
  (ICLR)}, pages 26486--26515, 2024.
\newblock arXiv:2311.17030.

\bibitem[{Meta AI}(2024)]{meta2024llama32}
{Meta AI}.
\newblock {Llama 3.2}: Revolutionizing edge {AI} and vision with open,
  customizable models.
\newblock
  \url{https://ai.meta.com/blog/llama-3-2-connect-2024-vision-edge-mobile-devices/},
  2024.
\newblock Blog post, September 25, 2024. Accessed 2026-11-11.

\bibitem[Minder et~al.(2025)Minder, Du, Stoehr, Monea, Wendler, West, and
  Cotterell]{minder2025knob}
Julian Minder, Kevin Du, Niklas Stoehr, Giovanni Monea, Chris Wendler, Robert
  West, and Ryan Cotterell.
\newblock Controllable context sensitivity and the knob behind it.
\newblock In \emph{International Conference on Learning Representations
  (ICLR)}, 2025.
\newblock arXiv:2411.07404.

\bibitem[Neo et~al.(2025)Neo, Ong, Torr, Geva, Krueger, and
  Barez]{neo2024towards}
Clement Neo, Luke Ong, Philip Torr, Mor Geva, David Krueger, and Fazl Barez.
\newblock Towards interpreting visual information processing in vision-language
  models.
\newblock In \emph{International Conference on Learning Representations
  (ICLR)}, 2025.
\newblock arXiv:2410.07149.

\bibitem[Nguyen et~al.(2025)Nguyen, Michaels, Fiterau, and
  Jensen]{nguyen2025challenges}
Trang Nguyen, Jackson Michaels, Madalina Fiterau, and David Jensen.
\newblock Challenges in understanding modality conflict in vision-language
  models.
\newblock \emph{arXiv preprint arXiv:2509.02805}, 2025.

\bibitem[Nikankin et~al.(2025)Nikankin, Arad, Gandelsman, and
  Belinkov]{nikankin2025sametask}
Yaniv Nikankin, Dana Arad, Yossi Gandelsman, and Yonatan Belinkov.
\newblock Same task, different circuits: Disentangling modality-specific
  mechanisms in {VLMs}.
\newblock In \emph{Advances in Neural Information Processing Systems
  (NeurIPS)}, 2025.
\newblock arXiv:2506.09047.

\bibitem[Niu et~al.(2021)Niu, Tang, Zhang, Lu, Hua, and Wen]{niu2021cfvqa}
Yulei Niu, Kaihua Tang, Hanwang Zhang, Zhiwu Lu, Xian-Sheng Hua, and Ji-Rong
  Wen.
\newblock Counterfactual {VQA}: A cause-effect look at language bias.
\newblock In \emph{Proceedings of the IEEE/CVF Conference on Computer Vision
  and Pattern Recognition (CVPR)}, pages 12700--12710, 2021.

\bibitem[Nooralahzadeh et~al.(2026)Nooralahzadeh, Rohanian, Zhang, F{\"u}rst,
  and Stockinger]{nooralahzadeh2026arbitration}
Farhad Nooralahzadeh, Omid Rohanian, Yi Zhang, Jonathan F{\"u}rst, and Kurt
  Stockinger.
\newblock Arbitration failure, not perceptual blindness: How vision-language
  models resolve visual-linguistic conflicts.
\newblock In \emph{Conference on Language Modeling (COLM)}, 2026.
\newblock arXiv:2604.09364.

\bibitem[{nostalgebraist}(2020)]{nostalgebraist2020logitlens}
{nostalgebraist}.
\newblock interpreting {GPT}: the logit lens.
\newblock
  \url{https://www.lesswrong.com/posts/AcKRB8wDpdaN6v6ru/interpreting-gpt-the-logit-lens},
  2020.
\newblock Accessed 2026-11-04.

\bibitem[{OpenAI}(2026)]{openai2026gpt61sol}
{OpenAI}.
\newblock {GPT-6.1 Sol}.
\newblock \url{https://developers.openai.com/api/docs/models/gpt-6.1-sol},
  2026.
\newblock API model \texttt{gpt-6.1-sol}. Accessed 2026-11-12.

\bibitem[Ortu et~al.(2024)Ortu, Jin, Doimo, Sachan, Cazzaniga, and
  Sch{\"o}lkopf]{ortu2024competition}
Francesco Ortu, Zhijing Jin, Diego Doimo, Mrinmaya Sachan, Alberto Cazzaniga,
  and Bernhard Sch{\"o}lkopf.
\newblock Competition of mechanisms: Tracing how language models handle facts
  and counterfactuals.
\newblock In \emph{Proceedings of the 62nd Annual Meeting of the Association
  for Computational Linguistics (Volume 1: Long Papers)}, pages 8420--8436,
  2024.
\newblock arXiv:2402.11655.

\bibitem[Ortu et~al.(2026)Ortu, Jin, Doimo, and Cazzaniga]{ortu2026seeing}
Francesco Ortu, Zhijing Jin, Diego Doimo, and Alberto Cazzaniga.
\newblock When seeing overrides knowing: Disentangling knowledge conflicts in
  vision-language models.
\newblock In \emph{Proceedings of the 64th Annual Meeting of the Association
  for Computational Linguistics (Volume 1: Long Papers)}, pages 14109--14130,
  2026.
\newblock arXiv:2507.13868.

\bibitem[Palit et~al.(2023)Palit, Pandey, Arora, and
  Liang]{palit2023causaltracing}
Vedant Palit, Rohan Pandey, Aryaman Arora, and Paul~Pu Liang.
\newblock Towards vision-language mechanistic interpretability: A causal
  tracing tool for {BLIP}.
\newblock In \emph{Proceedings of the IEEE/CVF International Conference on
  Computer Vision (ICCV) Workshops}, pages 2856--2861, 2023.

\bibitem[Paraselli et~al.(2026)Paraselli, Hua, and
  Pavlick]{paraselli2026entities}
Athulith Paraselli, Tianze Hua, and Ellie Pavlick.
\newblock Slow to see, slow to suppress: Understanding the effects of modality
  in context-memory conflicts.
\newblock In \emph{Findings of the Association for Computational Linguistics:
  EMNLP 2026}, pages 11751--11778, 2026.
\newblock arXiv:2609.00293.

\bibitem[Park et~al.(2026)Park, Choi, Lee, and Sung]{park2026retrievalheads}
Chanho Park, Daehyeon Choi, Jihyun Lee, and Minhyuk Sung.
\newblock Retrieval heads meet vision: Uncovering how {VLMs} locate and extract
  visual information.
\newblock In \emph{Advances in Neural Information Processing Systems
  (NeurIPS)}, 2026.
\newblock arXiv:2608.27417.

\bibitem[Radford et~al.(2021)Radford, Kim, Hallacy, Ramesh, Goh, Agarwal,
  Sastry, Askell, Mishkin, Clark, Krueger, and Sutskever]{radford2021clip}
Alec Radford, Jong~Wook Kim, Chris Hallacy, Aditya Ramesh, Gabriel Goh,
  Sandhini Agarwal, Girish Sastry, Amanda Askell, Pamela Mishkin, Jack Clark,
  Gretchen Krueger, and Ilya Sutskever.
\newblock Learning transferable visual models from natural language
  supervision.
\newblock In \emph{Proceedings of the International Conference on Machine
  Learning (ICML)}, pages 8748--8763, 2021.
\newblock arXiv:2103.00020.

\bibitem[Rohrbach et~al.(2018)Rohrbach, Hendricks, Burns, Darrell, and
  Saenko]{rohrbach2018chair}
Anna Rohrbach, Lisa~Anne Hendricks, Kaylee Burns, Trevor Darrell, and Kate
  Saenko.
\newblock Object hallucination in image captioning.
\newblock In \emph{Proceedings of the 2018 Conference on Empirical Methods in
  Natural Language Processing}, pages 4035--4045, 2018.
\newblock arXiv:1809.02156.

\bibitem[Rudman et~al.(2026)Rudman, Golovanevsky, Arad, Belinkov, Eickhoff,
  Singh, and Mahowald]{rudman2026pih}
William Rudman, Michal Golovanevsky, Dana Arad, Yonatan Belinkov, Carsten
  Eickhoff, Ritambhara Singh, and Kyle Mahowald.
\newblock Mechanisms of prompt-induced hallucination in vision-language models.
\newblock In \emph{Proceedings of the 64th Annual Meeting of the Association
  for Computational Linguistics (Volume 1: Long Papers)}, pages 41894--41912,
  2026.
\newblock arXiv:2601.05201.

\bibitem[Salazar et~al.(2026)Salazar, Frank, Oneata, Elliott, and
  Fierro]{salazar2026pathways}
Israfel Salazar, Stella Frank, Dan Oneata, Desmond Elliott, and Constanza
  Fierro.
\newblock Pathways of visual information flow in vision-language models.
\newblock In \emph{Advances in Neural Information Processing Systems
  (NeurIPS)}, 2026.
\newblock arXiv:2607.03358.

\bibitem[Schauml{\"o}ffel et~al.(2026)Schauml{\"o}ffel, Vilas, and
  Roig]{schaumloffel2026localization}
Timothy Schauml{\"o}ffel, Martina~G. Vilas, and Gemma Roig.
\newblock Mechanisms of object localization in vision-language models.
\newblock In \emph{Proceedings of the IEEE/CVF Conference on Computer Vision
  and Pattern Recognition (CVPR)}, pages 31356--31365, 2026.

\bibitem[Schild and de~Melo(2026)]{schild2026waldo}
Alexandra Schild and Gerard de Melo.
\newblock Can't find {Waldo}: Evaluating {VLMs}' sensitivity to image
  resolution and detail level.
\newblock In \emph{NeurIPS 2026 Workshop on Grounded and Faithful
  Vision-Language Models for Real-World Deployment}, 2026.
\newblock arXiv:2609.31706.

\bibitem[Serra et~al.(2025)Serra, Ortu, Panizon, Valeriani, Basile, Ansuini,
  Doimo, and Cazzaniga]{serra2025narrowgate}
Alessandro~Pietro Serra, Francesco Ortu, Emanuele Panizon, Lucrezia Valeriani,
  Lorenzo Basile, Alessio Ansuini, Diego Doimo, and Alberto Cazzaniga.
\newblock The narrow gate: Localized image-text communication in native
  multimodal models.
\newblock In \emph{Advances in Neural Information Processing Systems
  (NeurIPS)}, 2025.
\newblock arXiv:2412.06646.

\bibitem[Sim{\'e}oni et~al.(2026)Sim{\'e}oni, Vo, Seitzer, Baldassarre, Oquab,
  Jose, Khalidov, Szafraniec, Yi, Ramamonjisoa, Massa, Haziza, Wehrstedt, Wang,
  Darcet, Moutakanni, Sentana, Roberts, Vedaldi, Tolan, Brandt, Couprie,
  Mairal, J{\'e}gou, Labatut, and Bojanowski]{simeoni2025dinov3}
Oriane Sim{\'e}oni, Huy~V. Vo, Maximilian Seitzer, Federico Baldassarre, Maxime
  Oquab, Cijo Jose, Vasil Khalidov, Marc Szafraniec, Seungeun Yi, Micha{\"e}l
  Ramamonjisoa, Francisco Massa, Daniel Haziza, Luca Wehrstedt, Jianyuan Wang,
  Timoth{\'e}e Darcet, Th{\'e}o Moutakanni, Leonel Sentana, Claire Roberts,
  Andrea Vedaldi, Jamie Tolan, John Brandt, Camille Couprie, Julien Mairal,
  Herv{\'e} J{\'e}gou, Patrick Labatut, and Piotr Bojanowski.
\newblock {DINOv3}.
\newblock \emph{Transactions on Machine Learning Research}, 2026.
\newblock arXiv:2508.10104.

\bibitem[{SkunkworksAI}(2023)]{bakllava}
{SkunkworksAI}.
\newblock {BakLLaVA-1}.
\newblock \url{https://huggingface.co/SkunkworksAI/BakLLaVA-1}, 2023.
\newblock Accessed 2026-11-04.

\bibitem[Soldaini et~al.(2024)Soldaini, Kinney, Bhagia, Schwenk, Atkinson,
  Authur, Bogin, Chandu, Dumas, Elazar, Hofmann, Jha, Kumar, Lucy, Lyu,
  Lambert, Magnusson, Morrison, Muennighoff, Naik, Nam, Peters, Ravichander,
  Richardson, Shen, Strubell, Subramani, Tafjord, Walsh, Zettlemoyer, Smith,
  Hajishirzi, Beltagy, Groeneveld, Dodge, and Lo]{soldaini2024dolma}
Luca Soldaini, Rodney Kinney, Akshita Bhagia, Dustin Schwenk, David Atkinson,
  Russell Authur, Ben Bogin, Khyathi Chandu, Jennifer Dumas, Yanai Elazar,
  Valentin Hofmann, Ananya Jha, Sachin Kumar, Li Lucy, Xinxi Lyu, Nathan
  Lambert, Ian Magnusson, Jacob Morrison, Niklas Muennighoff, Aakanksha Naik,
  Crystal Nam, Matthew Peters, Abhilasha Ravichander, Kyle Richardson, Zejiang
  Shen, Emma Strubell, Nishant Subramani, Oyvind Tafjord, Evan Walsh, Luke
  Zettlemoyer, Noah Smith, Hannaneh Hajishirzi, Iz Beltagy, Dirk Groeneveld,
  Jesse Dodge, and Kyle Lo.
\newblock Dolma: an open corpus of three trillion tokens for language model
  pretraining research.
\newblock In \emph{Proceedings of the 62nd Annual Meeting of the Association
  for Computational Linguistics (Volume 1: Long Papers)}, pages 15725--15788,
  2024.
\newblock arXiv:2402.00159.

\bibitem[Sun et~al.(2026)Sun, Tu, Xue, Jiang, Xu, Yao, Qian, Sun, Zhou, Sun,
  and Lin]{sun2026prive}
Jingyu Sun, Jiachen Tu, Yuyang Xue, Yaoxin Jiang, Guoyi Xu, Zhengtao Yao, Rui
  Qian, Yizheng Sun, Hongpeng Zhou, Jingyuan Sun, and Yan Lin.
\newblock Seeing what is actually there: {PriVE-Bench} and {PriVE-Tools} for
  counterfactual evaluation of agentic visual evidence in {VLMs}.
\newblock In \emph{Proceedings of the 2026 Conference on Empirical Methods in
  Natural Language Processing}, pages 17595--17622, 2026.
\newblock arXiv:2607.16311.

\bibitem[Vankadaru and DuTell(2026)]{vankadaru2026trust}
Vijay Vankadaru and Vasha DuTell.
\newblock Vision-language models trust what they can discern: Image and
  representational properties governing modality arbitration.
\newblock In \emph{Explainable Computer Vision Workshop (eXCV) at ECCV 2026,
  Proceedings Track}, 2026.

\bibitem[Vo et~al.(2026)Vo, Nguyen, Taesiri, Dang, Nguyen, and
  Kim]{vo2026biased}
An Vo, Khai-Nguyen Nguyen, Mohammad~Reza Taesiri, Vy~Tuong Dang, Anh~Totti
  Nguyen, and Daeyoung Kim.
\newblock Vision language models are biased.
\newblock In \emph{International Conference on Learning Representations
  (ICLR)}, 2026.
\newblock arXiv:2505.23941.

\bibitem[Wang et~al.(2025{\natexlab{a}})Wang, Chen, Zhang, Tian, Xu, Deng, and
  Chen]{wang2025deco}
Chenxi Wang, Xiang Chen, Ningyu Zhang, Bozhong Tian, Haoming Xu, Shumin Deng,
  and Huajun Chen.
\newblock {MLLM} can see? {Dynamic} correction decoding for hallucination
  mitigation.
\newblock In \emph{International Conference on Learning Representations
  (ICLR)}, 2025{\natexlab{a}}.

\bibitem[Wang et~al.(2025{\natexlab{b}})Wang, Hu, and Jiang]{wang2025vseam}
Qidong Wang, Junjie Hu, and Ming Jiang.
\newblock {V-SEAM}: Visual semantic editing and attention modulating for causal
  interpretability of vision-language models.
\newblock In \emph{Proceedings of the 2025 Conference on Empirical Methods in
  Natural Language Processing}, pages 17396--17420, 2025{\natexlab{b}}.
\newblock arXiv:2509.14837.

\bibitem[Wang et~al.(2025{\natexlab{c}})Wang, Gao, Gu, Pu, Cui, Wei, Liu, Jing,
  Ye, Shao, Wang, Chen, Zhang, Yang, Wang, Wei, Yin, Li, Cui, Chen, Ding, Tian,
  Wu, Xie, Li, Yang, Duan, Wang, Hou, Hao, Zhang, Li, Zhao, Duan, Deng, Fu, He,
  Wang, He, Shi, He, Xiong, Lv, Wu, Shao, Zhang, Deng, Qi, Ge, Guo, Zhang,
  Zhang, Cao, Lin, Tang, Gao, Huang, Gu, Lyu, Tang, Wang, Lv, Ouyang, Wang,
  Dou, Zhu, Lu, Lin, Dai, Su, Zhou, Chen, Qiao, Wang, and
  Luo]{wang2025internvl35}
Weiyun Wang, Zhangwei Gao, Lixin Gu, Hengjun Pu, Long Cui, Xingguang Wei,
  Zhaoyang Liu, Linglin Jing, Shenglong Ye, Jie Shao, Zhaokai Wang, Zhe Chen,
  Hongjie Zhang, Ganlin Yang, Haomin Wang, Qi Wei, Jinhui Yin, Wenhao Li, Erfei
  Cui, Guanzhou Chen, Zichen Ding, Changyao Tian, Zhenyu Wu, Jingjing Xie,
  Zehao Li, Bowen Yang, Yuchen Duan, Xuehui Wang, Zhi Hou, Haoran Hao, Tianyi
  Zhang, Songze Li, Xiangyu Zhao, Haodong Duan, Nianchen Deng, Bin Fu, Yinan
  He, Yi Wang, Conghui He, Botian Shi, Junjun He, Yingtong Xiong, Han Lv, Lijun
  Wu, Wenqi Shao, Kaipeng Zhang, Huipeng Deng, Biqing Qi, Jiaye Ge, Qipeng Guo,
  Wenwei Zhang, Songyang Zhang, Maosong Cao, Junyao Lin, Kexian Tang, Jianfei
  Gao, Haian Huang, Yuzhe Gu, Chengqi Lyu, Huanze Tang, Rui Wang, Haijun Lv,
  Wanli Ouyang, Limin Wang, Min Dou, Xizhou Zhu, Tong Lu, Dahua Lin, Jifeng
  Dai, Weijie Su, Bowen Zhou, Kai Chen, Yu Qiao, Wenhai Wang, and Gen Luo.
\newblock {InternVL3.5}: Advancing open-source multimodal models in
  versatility, reasoning, and efficiency.
\newblock \emph{arXiv preprint arXiv:2508.18265}, 2025{\natexlab{c}}.

\bibitem[Wang et~al.(2026)Wang, Cui, Wang, Zhang, Wang, Zhang, Luo, Sun, Li,
  Wang, Yu, Zhao, Ao, Min, Men, Wu, Zhao, Zhang, Wang, Liu, He, Yang, Liu, Lin,
  Wang, and Huang]{wang2024emu3}
Xinlong Wang, Yufeng Cui, Jinsheng Wang, Fan Zhang, Yueze Wang, Xiaosong Zhang,
  Zhengxiong Luo, Quan Sun, Zhen Li, Yuqi Wang, Qiying Yu, Yingli Zhao, Yulong
  Ao, Xuebin Min, Chunlei Men, Boya Wu, Bo Zhao, Bowen Zhang, Liangdong Wang,
  Guang Liu, Zheqi He, Xi Yang, Jingjing Liu, Yonghua Lin, Zhongyuan Wang, and
  Tiejun Huang.
\newblock Multimodal learning with next-token prediction for large multimodal
  models.
\newblock \emph{Nature}, 650\penalty0 (8101):\penalty0 327--333, 2026.
\newblock Emu3; preprint arXiv:2409.18869.

\bibitem[Weber et~al.(2024)Weber, Fu, Anthony, Oren, Adams, Alexandrov, Lyu,
  Nguyen, Yao, Adams, Athiwaratkun, Chalamala, Chen, Ryabinin, Dao, Liang,
  R{\'e}, Rish, and Zhang]{weber2024redpajama}
Maurice Weber, Daniel~Y. Fu, Quentin Anthony, Yonatan Oren, Shane Adams, Anton
  Alexandrov, Xiaozhong Lyu, Huu Nguyen, Xiaozhe Yao, Virginia Adams, Ben
  Athiwaratkun, Rahul Chalamala, Kezhen Chen, Max Ryabinin, Tri Dao, Percy
  Liang, Christopher R{\'e}, Irina Rish, and Ce Zhang.
\newblock {RedPajama}: an open dataset for training large language models.
\newblock In \emph{Advances in Neural Information Processing Systems (NeurIPS),
  Datasets and Benchmarks Track}, pages 116462--116492, 2024.
\newblock arXiv:2411.12372.

\bibitem[Wolf et~al.(2020)Wolf, Debut, Sanh, Chaumond, Delangue, Moi, Cistac,
  Rault, Louf, Funtowicz, Davison, Shleifer, von Platen, Ma, Jernite, Plu, Xu,
  Scao, Gugger, Drame, Lhoest, and Rush]{wolf2020transformers}
Thomas Wolf, Lysandre Debut, Victor Sanh, Julien Chaumond, Clement Delangue,
  Anthony Moi, Pierric Cistac, Tim Rault, R{\'e}mi Louf, Morgan Funtowicz, Joe
  Davison, Sam Shleifer, Patrick von Platen, Clara Ma, Yacine Jernite, Julien
  Plu, Canwen Xu, Teven~Le Scao, Sylvain Gugger, Mariama Drame, Quentin Lhoest,
  and Alexander~M. Rush.
\newblock Transformers: State-of-the-art natural language processing.
\newblock In \emph{Proceedings of the 2020 Conference on Empirical Methods in
  Natural Language Processing: System Demonstrations}, pages 38--45, 2020.
\newblock arXiv:1910.03771.

\bibitem[Wu et~al.(2024)Wu, Wu, and Zou]{wu2024clasheval}
Kevin Wu, Eric Wu, and James Zou.
\newblock {ClashEval}: Quantifying the tug-of-war between an {LLM}'s internal
  prior and external evidence.
\newblock In \emph{Advances in Neural Information Processing Systems (NeurIPS),
  Datasets and Benchmarks Track}, pages 33402--33422, 2024.
\newblock arXiv:2404.10198.

\bibitem[Xiao et~al.(2021)Xiao, Engstrom, Ilyas, and Madry]{xiao2021noise}
Kai Xiao, Logan Engstrom, Andrew Ilyas, and Aleksander Madry.
\newblock Noise or signal: The role of image backgrounds in object recognition.
\newblock In \emph{International Conference on Learning Representations
  (ICLR)}, 2021.

\bibitem[{Xiaomi LLM-Core Team}(2025)]{xiaomi2025mimovl}
{Xiaomi LLM-Core Team}.
\newblock {MiMo-VL} technical report.
\newblock \emph{arXiv preprint arXiv:2506.03569}, 2025.

\bibitem[Xie et~al.(2024)Xie, Zhang, Chen, Lou, and Su]{xie2023chameleon}
Jian Xie, Kai Zhang, Jiangjie Chen, Renze Lou, and Yu Su.
\newblock Adaptive chameleon or stubborn sloth: Revealing the behavior of large
  language models in knowledge conflicts.
\newblock In \emph{International Conference on Learning Representations
  (ICLR)}, 2024.
\newblock arXiv:2305.13300.

\bibitem[Yang et~al.(2025{\natexlab{a}})Yang, Zheng, Chen, Zhao, Lin, and
  Shen]{yang2025nullu}
Le Yang, Ziwei Zheng, Boxu Chen, Zhengyu Zhao, Chenhao Lin, and Chao Shen.
\newblock Nullu: Mitigating object hallucinations in large vision-language
  models via {HalluSpace} projection.
\newblock In \emph{Proceedings of the IEEE/CVF Conference on Computer Vision
  and Pattern Recognition (CVPR)}, pages 14635--14645, 2025{\natexlab{a}}.

\bibitem[Yang et~al.(2025{\natexlab{b}})Yang, Li, Cao, and Xu]{yang2025modular}
Tianyun Yang, Ziniu Li, Juan Cao, and Chang Xu.
\newblock Understanding and mitigating hallucination in large vision-language
  models via modular attribution and intervention.
\newblock In \emph{International Conference on Learning Representations
  (ICLR)}, 2025{\natexlab{b}}.

\bibitem[Yu et~al.(2023)Yu, Merullo, and Pavlick]{yu2023characterizing}
Qinan Yu, Jack Merullo, and Ellie Pavlick.
\newblock Characterizing mechanisms for factual recall in language models.
\newblock In \emph{Proceedings of the 2023 Conference on Empirical Methods in
  Natural Language Processing}, pages 9924--9959, 2023.

\bibitem[Zhai et~al.(2023)Zhai, Mustafa, Kolesnikov, and Beyer]{zhai2023siglip}
Xiaohua Zhai, Basil Mustafa, Alexander Kolesnikov, and Lucas Beyer.
\newblock Sigmoid loss for language image pre-training.
\newblock In \emph{Proceedings of the IEEE/CVF International Conference on
  Computer Vision (ICCV)}, pages 11975--11986, 2023.
\newblock arXiv:2303.15343.

\bibitem[Zhang and Nanda(2024)]{zhang2024patching}
Fred Zhang and Neel Nanda.
\newblock Towards best practices of activation patching in language models:
  Metrics and methods.
\newblock In \emph{International Conference on Learning Representations
  (ICLR)}, 2024.
\newblock arXiv:2309.16042.

\bibitem[Zhang et~al.(2025{\natexlab{a}})Zhang, Khayatkhoei, Chhikara, and
  Ilievski]{zhang2025mllmsknow}
Jiarui Zhang, Mahyar Khayatkhoei, Prateek Chhikara, and Filip Ilievski.
\newblock {MLLMs} know where to look: Training-free perception of small visual
  details with multimodal {LLMs}.
\newblock In \emph{International Conference on Learning Representations
  (ICLR)}, 2025{\natexlab{a}}.
\newblock arXiv:2502.17422.

\bibitem[Zhang et~al.(2026{\natexlab{a}})Zhang, Ma, Hou, Bai, Chen, Xiang, Yu,
  and Zhang]{zhang2025modalitypreferences}
Yu Zhang, Jinlong Ma, Yongshuai Hou, Xuefeng Bai, Kehai Chen, Yang Xiang, Jun
  Yu, and Min Zhang.
\newblock Evaluating and steering modality preferences in multi-modal {LLMs}.
\newblock In \emph{Proceedings of the International Conference on Machine
  Learning (ICML)}, 2026{\natexlab{a}}.
\newblock arXiv:2505.20977.

\bibitem[Zhang et~al.(2025{\natexlab{b}})Zhang, Yadav, Han, and
  Shutova]{zhang2025crossmodal}
Zhi Zhang, Srishti Yadav, Fengze Han, and Ekaterina Shutova.
\newblock Cross-modal information flow in multimodal large language models.
\newblock In \emph{Proceedings of the IEEE/CVF Conference on Computer Vision
  and Pattern Recognition (CVPR)}, pages 19781--19791, 2025{\natexlab{b}}.
\newblock arXiv:2411.18620.

\bibitem[Zhang et~al.(2026{\natexlab{b}})Zhang, Wang, Gong, Shi, Wang, Wang,
  and Hu]{zhang2025modalitiesconflict}
Zhuoran Zhang, Tengyue Wang, Xilin Gong, Yang Shi, Haotian Wang, Di Wang, and
  Lijie Hu.
\newblock Beyond fixed biases: Decoding the role of reasoning uncertainty in
  {MLLM} modality conflicts.
\newblock In \emph{Proceedings of the International Conference on Machine
  Learning (ICML)}, 2026{\natexlab{b}}.
\newblock Preprint title: When Modalities Conflict: How Unimodal Reasoning
  Uncertainty Governs Preference Dynamics in {MLLMs}; arXiv:2511.02243.

\bibitem[Zhao et~al.(2024{\natexlab{a}})Zhao, Xu, Gupta, Asthana, Zheng, and
  Gould]{zhao2024firsttoknow}
Qinyu Zhao, Ming Xu, Kartik Gupta, Akshay Asthana, Liang Zheng, and Stephen
  Gould.
\newblock The first to know: How token distributions reveal hidden knowledge in
  large vision-language models?
\newblock In \emph{Proceedings of the European Conference on Computer Vision
  (ECCV)}, pages 127--142, 2024{\natexlab{a}}.

\bibitem[Zhao et~al.(2024{\natexlab{b}})Zhao, Du, Hong, Gema, Devoto, Wang, He,
  Wong, and Minervini]{zhao2024conflictdetect}
Yu Zhao, Xiaotang Du, Giwon Hong, Aryo~Pradipta Gema, Alessio Devoto, Hongru
  Wang, Xuanli He, Kam-Fai Wong, and Pasquale Minervini.
\newblock Analysing the residual stream of language models under knowledge
  conflicts.
\newblock In \emph{NeurIPS 2024 Workshop on Foundation Model Interventions
  (MINT)}, 2024{\natexlab{b}}.
\newblock arXiv:2410.16090.

\bibitem[Zheng et~al.(2023)Zheng, Chiang, Sheng, Zhuang, Wu, Zhuang, Lin, Li,
  Li, Xing, Zhang, Gonzalez, and Stoica]{zheng2023vicuna}
Lianmin Zheng, Wei-Lin Chiang, Ying Sheng, Siyuan Zhuang, Zhanghao Wu, Yonghao
  Zhuang, Zi Lin, Zhuohan Li, Dacheng Li, Eric~P. Xing, Hao Zhang, Joseph~E.
  Gonzalez, and Ion Stoica.
\newblock Judging {LLM-as-a-Judge} with {MT-Bench} and {Chatbot} {Arena}.
\newblock In \emph{Advances in Neural Information Processing Systems (NeurIPS),
  Datasets and Benchmarks Track}, pages 46595--46623, 2023.
\newblock arXiv:2306.05685.

\bibitem[Zhou et~al.(2025)Zhou, Yan, Zou, Wang, Liu, and Hu]{zhou2025causalmm}
Guanyu Zhou, Yibo Yan, Xin Zou, Kun Wang, Aiwei Liu, and Xuming Hu.
\newblock Mitigating modality prior-induced hallucinations in multimodal large
  language models via deciphering attention causality.
\newblock In \emph{International Conference on Learning Representations
  (ICLR)}, 2025.
\newblock arXiv:2410.04780.

\bibitem[Zhou et~al.(2026)Zhou, Zhang, and Wang]{zhou2026ignorance}
Runyu Zhou, Qi Zhang, and Yisen Wang.
\newblock Diagnosing visual ignorance in vision-language models.
\newblock \emph{arXiv preprint arXiv:2606.06890}, 2026.

\bibitem[Zhu et~al.(2024)Zhu, Liu, Wang, Tu, and Chen]{zhu2024crossmodal}
Tinghui Zhu, Qin Liu, Fei Wang, Zhengzhong Tu, and Muhao Chen.
\newblock Unraveling cross-modality knowledge conflicts in large
  vision-language models.
\newblock \emph{arXiv preprint arXiv:2410.03659}, 2024.

\end{thebibliography}
}

\clearpage
\onecolumn
\setcounter{section}{0}
\renewcommand{\thesection}{\Alph{section}}
\renewcommand{\theHsection}{supp.\Alph{section}}
\setcounter{table}{0}
\setcounter{figure}{0}
\renewcommand{\thetable}{S\arabic{table}}
\renewcommand{\thefigure}{S\arabic{figure}}
\renewcommand{\theHtable}{supp.\arabic{table}}
\renewcommand{\theHfigure}{supp.\arabic{figure}}
\addtocontents{toc}{\protect\setcounter{tocdepth}{2}}
\begin{center}
{\Large\bfseries Believing Is Seeing: How Vision-Language Models Decide to Believe an Image}\\[0.5em]
{\large Supplementary Material}
\end{center}
\vspace{0.5em}

\noindent This supplement follows the order of the main paper. Supp.~A to C give the data, the reproduction of earlier results and the checks on our interventions that the setup (main-paper Sec.~3) relies on. For the two cues of main-paper Sec.~4, Supp.~H gives the additive fit of edit effects, Supp.~Q compares the additive model with alternatives, and Supp.~R reports the API models. For the object (main-paper Sec.~5), Supp.~G gives the localized edits and the attention steering, and Supp.~N shows examples. For the pathway inside the model (main-paper Sec.~6), Supp.~D extends the band map, and Supp.~P reports every measurement on the further architectures and checkpoints. For the image-token subspace (main-paper Sec.~7), Supp.~E gives its construction and the removal tests, and Supp.~F and I give the mediation and cross-model analyses. For the readout and the photographs (main-paper Secs.~8 and 9), see Supp.~J to L and S. Supp.~M lists design choices and controls. Supp.~O holds the pre-specified tests, the limitations in full, and the detail behind statements that the main text makes briefly. Supp.~T extends the related work. Per-model heads, subspace bases, the readout and the annotated conflicts will be released.

{\small\setcounter{tocdepth}{2}\makeatletter\renewcommand*\l@section{\@dottedtocline{1}{0em}{1.8em}}\renewcommand*\l@subsection{\@dottedtocline{2}{1.8em}{2.6em}}\makeatother\setlength{\parskip}{0pt}\tableofcontents}
\bigskip

\section{Data and prompts}\label{app:data}
\noindent\textit{Supports main-paper Sec.~3. The conflict sources, the per-model prior filter, the prompts, the checks on the outcome measure and the object masks.}\par\smallskip
\noindent\textbf{Sources.} WHOOPS-AHA: the 500 released conflicts (an earlier study \cite{vankadaru2026trust} retains 435 of them); ViLP: 600 (300 questions $\times$ 2 out-of-distribution images; four rows share a question text with another); Visual CounterFact color: 493 rows, 493 unique objects and images; Visual CounterFact size: 727 rows over 410 unique objects (``dog'' 25$\times$, ``car'' 8$\times$), one image per row. Rows are (object, counterfactual image) pairs as released. Per-model prior filter (text-only answer must be the prior answer, three outcome rules agreeing on $\ge92\%$): valid conflicts across all four sources: LLaVA-NeXT 2,121, LLaVA-1.5 1,950, BakLLaVA 2,024; Qwen2.5-VL 1,133.

{\sloppy \noindent\textbf{Answer conventions.} On the size split the released \texttt{correct\_\allowbreak answer} is the world-knowledge (prior) answer and \texttt{incorrect\_\allowbreak answer} the image-supported one, the reverse of the color split; confirmed by the prior filter's acceptance rates (0.85, 0.64, 0.75).\par}

\noindent\textbf{Prompts.} Statement continuation for WHOOPS and Visual CounterFact; question form for ViLP; model-family templates (Mistral \texttt{[INST]}, Vicuna \texttt{USER/ASSISTANT}, Qwen chat template). Yes/no benchmarks use the in-context continuation scorer; first-token and in-context scoring agree in sign on 100\% of LLaVA-1.5 items.

\noindent\textbf{Outcome validation.} Free-generation agreement with the next-token outcome: 0.87--0.95 across models; gray-image control: 90--99\% of image-wins do not survive image removal (per-model values will be released with the code). Quiet-truth ratios (prior wrong, image right, $n\approx60$): gray 0.56/0.29/0.12, blur 0.44/0.84/0.69, size 0.50/0.85/0.89; saturation boost turns 4.5\%, 7.1\%, 6.5\% of prior-wins into wins for the false answer. The image-right subset used for this loud-versus-quiet comparison has $n\approx60$ against $n\approx360$ image-wrong rows.

\noindent\textbf{Masks.} SAM3 on the image-supported answer phrase; a mask on 85 to 87\% of valid WHOOPS conflicts, 702/727 (size split); median object area 10 to 11\% of pixels on WHOOPS. NaturalBench: yes/no items selected by their answer strings (1,450 of 1,900 rows are yes/no), first 600 pairs; the question noun for masks is the last noun phrase after a determiner (for example ``hat'', ``couch'').

\FloatBarrier
\noindent\textbf{Conventions.} Layer $L$ is the output of decoder block $L$ (hidden-state index $L{+}1$; the NaturalBench readout uses hidden-state index 20, which is layer 19, inside the same 18--24 plateau). The \emph{gap} is the margin of the main paper, the logit difference between the image-supported and the prior-supported answer at the answer position; a \emph{flip} is a reversal of its sign. Edits are grayscale, saturation scaling, Gaussian blur and additive noise. \emph{Head signal} and \emph{discernibility} follow earlier work \cite{vankadaru2026trust}: the summed logit contribution of the arbitration heads to the gap, and how legible, large and attended the conflict evidence is. A \emph{recomposite} is the segmented object pasted back at a new scale over a blurred fill. The Visual CounterFact color release has 493 rows; \citet{lietzow2026visiondefault} describe the color set as 469 examples, two of which they exclude, whereas we filter per model. Valid WHOOPS conflicts number 424, 416 and 419 on LLaVA-NeXT, LLaVA-1.5 and BakLLaVA.

\section{Reproduction of prior results on LLaVA-NeXT}\label{app:repro}
\noindent\textit{Supports main-paper Sec.~3. Our reimplementation of an earlier paper's measurements on LLaVA-NeXT, against the values that paper reports.}\par\smallskip
A reimplementation of the measurements of an earlier paper \cite{vankadaru2026trust} on LLaVA-NeXT, compared with the values that paper reports (reimplementation against reported): attended-region size partial $r$ +0.150 vs +0.174; decodability 0.72 vs 0.71; DINOv3 \cite{simeoni2025dinov3} gap +0.157 vs +0.156; saturation dose 0.718$\to$0.805 vs 0.72$\to$0.81; blur and noise curves within their intervals. The analysis code will be released.

\FloatBarrier
\section{Verification of interventions}\label{app:verify}
\noindent\textit{Supports main-paper Secs.~3 and 6. Checks that the patching hooks and the interventions do what they are meant to do.}\par\smallskip
The steering attention path is registered as a custom attention implementation and verified equal to eager attention with no steering set (max logit difference $<10^{-3}$, and bit-identical outcomes); the default scaled-dot-product attention gives the same result as the explicit implementation on all three LLaVA models (max difference 0.016 in logits, identical outcomes). Interchange and ablation hooks return the unpatched forward exactly when no patch or ablation is specified. Qwen2.5-VL requires bfloat16 (float16 overflows to non-finite logits). 

\FloatBarrier
\section{Band map, extended}\label{app:transport}
\noindent\textit{Supports main-paper Sec.~6. The band map on the three LLaVA models at single-layer resolution, at further token positions, with other donors and on prior-agreeing items.}\par\smallskip

\begin{figure}[h]\centering\includegraphics{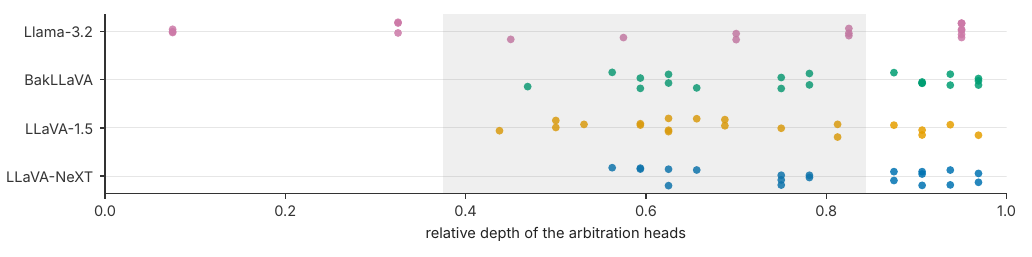}
\caption{Relative depth of the top-20 arbitration heads on four models, one dot per head. Llama-3.2 is the cross-attention model; its heads are at cross-attention layers. Shaded band: relative depth 0.375 to 0.84, layers 12--27 of the three 32-layer LLaVA models.}\end{figure}

Band patches replace the residual stream entering each of the four decoder blocks in the band (forward pre-hooks on the blocks, applied simultaneously) at the chosen positions. The final normalization is never replaced, so even the last band does not reduce to copying the donor's logits (in the three-model run of this section the last band also patches the final hidden state; see below), and recoveries near but not equal to one in that band are expected. \Cref{tab:bandllava} gives the recovered fraction per band with the grayscale copy patched into the clean run.

\begin{table}[h]\centering\footnotesize\setlength{\tabcolsep}{7pt}
\caption{\textbf{Band map on the three LLaVA models and Qwen2.5-VL.} Fraction of the gray edit's effect recovered at the image tokens / at the answer position, per four-layer band. Qwen2.5-VL has 28 layers and seven bands ($n=300$).}\label{tab:bandllava}
\begin{tabular}{lcccc}\toprule
Layers & LLaVA-NeXT & LLaVA-1.5 & BakLLaVA & Qwen2.5-VL \\ \midrule
0--3 & 1.00/0.00 & 1.00/0.00 & 1.00/0.00 & 1.00/0.02 \\
4--7 & 1.00/0.01 & 1.00/0.00 & 0.99/0.01 & 0.97/0.02 \\
8--11 & 0.98/0.03 & 0.98/0.02 & 0.95/0.03 & 0.97/0.04 \\
12--15 & 0.88/0.15 & 0.95/0.12 & 0.89/0.14 & 0.95/0.09 \\
16--19 & 0.69/0.42 & 0.77/0.37 & 0.73/0.39 & 0.84/0.28 \\
20--23 & 0.38/0.53 & 0.28/0.65 & 0.38/0.53 & 0.67/0.62 \\
24--27 & 0.19/0.69 & 0.07/0.86 & 0.11/0.78 & 0.17/1.59 \\
28--31 & 0.11/1.14 & 0.03/0.88 & 0.06/1.06 & -- \\ \bottomrule
\end{tabular}\end{table}

\noindent\emph{Notes on \cref{tab:bandllava}.} Recoveries above 1.0 in the last bands (1.14, 1.06, 1.59) are overshoot on the paired difference: a late-band patch carries the gray copy's entire downstream state at that site, including components unrelated to the edit that happen to push the same way, and the denominator is the mean actual effect. Image-wins lost per band follow the same profile (0.18 at the image tokens, 0.02 at the answer position, crossing at 20--23). Patching layers 16 to 24 at the image tokens recovers 67--77\% of the gray effect and 54--68\% of the blur effect. The 256-D subspace recovers 63, 69 and 67\% of the gray effect (\cref{tab:suffmain}), a random subspace at most 7\% and a single probe direction 0\%, consistent with the band profile.

\noindent\textbf{Band map when the prior agrees.} Agree items are WHOOPS rows whose text-only answer is the image-supported answer (the prior filter fails) and whose clean-image answer is the same. The gray edit moves them less than it moves conflicts, so the per-item statistic keeps fewer of them (\cref{tab:agreellava}). The run that added the prior-agreeing items reproduces the conflict values above within 0.005 in all 16 cells on each model. The crossover band (the first band where the answer position recovers more than the image tokens, per-item mean) is L20--23 in both conditions on all three models. The pre-specified rule (same crossover band and $|$difference$|<0.10$ with intervals including zero at both sites) classifies the crossing as unchanged on all three.

\begin{table}[h]\centering\footnotesize\setlength{\tabcolsep}{5pt}
\caption{\textbf{Conflicts against prior-agreeing items on the three LLaVA models.} Differences are per-item statistics with cluster-bootstrap 95\% intervals.}\label{tab:agreellava}
\begin{tabular}{lccc}\toprule
 & LLaVA-NeXT & LLaVA-1.5 & BakLLaVA \\ \midrule
Prior-agreeing items & 68 & 75 & 69 \\
Mean gray shift, prior-agreeing / conflicts (logits) & $-$0.44 / $-$1.16 & $-$0.21 / $-$0.85 & $-$0.36 / $-$1.31 \\
Items kept, prior-agreeing / conflicts & 31 / 232 & 19 / 206 & 43 / 282 \\
Bootstrap share of L20--23 as crossover, conflict / agree & 1.00/0.98 & 1.00/0.98 & 1.00/0.70 \\ \midrule
Conflict minus agree at L20--23, image tokens & +0.043 [$-$0.065, +0.155] & +0.093 [$-$0.045, +0.212] & +0.007 [$-$0.192, +0.202] \\
Conflict minus agree at L20--23, answer position & +0.026 [$-$0.086, +0.142] & $-$0.063 [$-$0.187, +0.087] & +0.041 [$-$0.131, +0.197] \\ \midrule
Last band, answer position, with the final state & 1.14 & 0.88 & 1.06 \\
Last band without it, conflict / agree & 0.94 / 0.92 & 0.98 / 0.97 & 0.98 / 0.94 \\
Random-index control, image tokens / answer position & 0.94 / 0.74 & 0.94 / 0.82 & 0.91 / 0.81 \\ \bottomrule
\end{tabular}\end{table}

\noindent\emph{Notes on \cref{tab:agreellava}.} The last-band values of \cref{tab:bandllava} include the post-norm final state (index 32) in the patch, which inflates the answer-position recovery above 1 (by ratio of means); the row without it gives per-item means. The control patches four random hidden-state indices per item; its values are under conflict, with the same values within intervals under agreement. On Qwen2.5-VL (306 conflict and 56 agree items after the filter; 13 of 14 cells of the map above reproduced within 0.05; the last-band answer-position cell differs by 0.054) the per-item crossover is at L20--23 in both conditions and the conflict-minus-agree difference there is +0.115 [$-$0.042, +0.268] at the image tokens and $-$0.054 [$-$0.179, +0.076] at the answer position: intermediate by the rule because the image-token point estimate exceeds 0.10, with an interval that includes zero.

\FloatBarrier
\begin{figure}[h]\centering\includegraphics{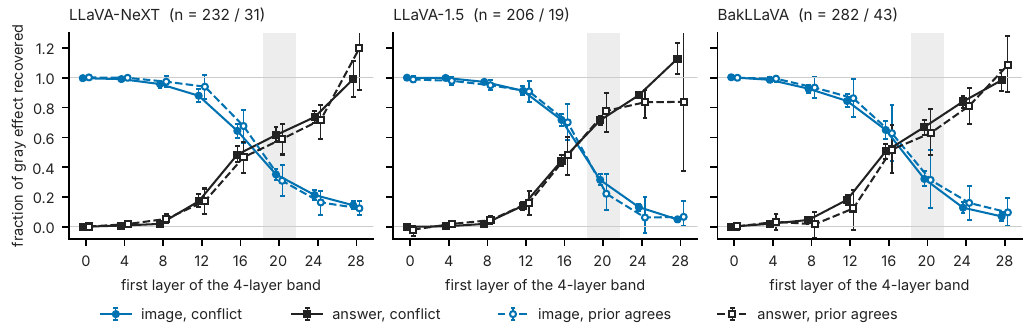}
\caption{\textbf{Band map on the three LLaVA models, conflicts and prior-agreeing items}. Residuals of the grayscale copy of the same image patched into the clean run per 4-layer band, at the image tokens (blue, circles) and at the answer position (black, squares); y = fraction of the actual grayscale effect recovered, per-item mean, cluster-bootstrap 95\% intervals. Solid, filled: WHOOPS conflicts. Dashed, open: WHOOPS images where the text-only answer already agrees with the image. Items are those the gray edit moves by at least 0.5 logits ($n$ in the titles). The crossover band (shaded, 20--23) is the same in both conditions and overlaps the arbitration heads' layers (mean 22 to 25).}\label{fig:transport_llava}\end{figure}

\noindent\textbf{Single-layer map on LLaVA-1.5.} With one decoder layer per band (145 conflicts moved by at least 0.5 logits), the image-site recovery is 0.99 or more through layer 8 and the answer-site recovery 0.01 or less through layer 11; \cref{tab:singlelayer} gives the later layers. The curves cross between layer 20 (0.45 against 0.44) and layer 21 (0.31 against 0.59); the 19 prior-agreeing items cross at layer 20 (0.36 against 0.48). A four-layer band reports the image site at its first layer and the answer site through its last, so the crossing interpolated on band entries (0.58 in the main paper) is earlier than the single-layer crossing by up to one band.

\begin{table}[h]\centering\footnotesize\setlength{\tabcolsep}{7pt}
\caption{\textbf{Single-layer map on LLaVA-1.5}, layers 9 to 31. Fraction of the gray edit's effect recovered when the patch enters at one layer.}\label{tab:singlelayer}
\begin{tabular}{ccc@{\hspace{2.5em}}ccc}\toprule
Layer & Image site & Answer position & Layer & Image site & Answer position \\ \midrule
9 & 0.97 & $\le$ 0.01 & 21 & 0.31 & 0.59 \\
10 & 0.97 & $\le$ 0.01 & 22 & 0.24 & 0.65 \\
11 & 0.95 & $\le$ 0.01 & 23 & 0.19 & 0.71 \\
12 & 0.92 & 0.02 & 24 & 0.18 & 0.72 \\
13 & 0.91 & 0.03 & 25 & 0.12 & 0.79 \\
14 & 0.90 & 0.04 & 26 & 0.12 & 0.80 \\
15 & 0.81 & 0.11 & 27 & 0.07 & 0.88 \\
16 & 0.79 & 0.14 & 28 & 0.07 & 0.88 \\
17 & 0.72 & 0.21 & 29 & 0.05 & 0.91 \\
18 & 0.65 & 0.27 & 30 & 0.01 & 0.97 \\
19 & 0.60 & 0.32 & 31 & 0.00 & 0.98 \\
20 & 0.45 & 0.44 & & & \\ \bottomrule
\end{tabular}\end{table}

\noindent\textbf{Three sites on LLaVA-1.5.} \Cref{tab:threesite} splits the prompt into the image positions, the other prompt positions (the question tokens of the main paper; a median of 20 text tokens) and the answer position, on the same items as the band map. The other prompt positions recover at most 0.13 [0.10, 0.16] of the effect under conflict and 0.14 [0.08, 0.20] on prior-agreeing items, both at layers 16--19. The three sites sum to more than one in the middle bands: the same contribution is recoverable from more than one site while it is being moved.

\begin{table}[h]\centering\footnotesize\setlength{\tabcolsep}{5pt}
\caption{\textbf{Three-site band map, LLaVA-1.5} (144 WHOOPS conflicts moved by at least 0.5 logits; gray copy patched into the clean run). Share of the gray edit's effect recovered at each site, with 95\% intervals.}\label{tab:threesite}
\begin{tabular}{lcccc}\toprule
Layers & Image positions & Other prompt positions & Answer position & Sum \\ \midrule
0--3 & 1.00 [1.00, 1.00] & $-$0.00 [$-$0.01, 0.00] & 0.00 [$-$0.00, 0.00] & 1.00 \\
4--7 & 1.00 [0.99, 1.01] & 0.00 [$-$0.01, 0.01] & 0.00 [$-$0.00, 0.01] & 1.01 \\
8--11 & 0.99 [0.98, 1.01] & 0.04 [0.02, 0.06] & 0.01 [0.00, 0.02] & 1.04 \\
12--15 & 0.93 [0.90, 0.96] & 0.11 [0.08, 0.15] & 0.11 [0.08, 0.14] & 1.15 \\
16--19 & 0.79 [0.75, 0.83] & 0.13 [0.10, 0.16] & 0.32 [0.27, 0.36] & 1.24 \\
20--23 & 0.45 [0.41, 0.50] & 0.09 [0.07, 0.11] & 0.71 [0.67, 0.74] & 1.25 \\
24--27 & 0.19 [0.16, 0.22] & 0.04 [0.03, 0.05] & 0.88 [0.85, 0.90] & 1.11 \\
28--31 & 0.07 [0.05, 0.09] & 0.03 [0.02, 0.04] & 0.98 [0.97, 0.98] & 1.08 \\ \bottomrule
\end{tabular}\end{table}

\noindent\textbf{Three sites on Llama-3.2 and on the other models.} With cross-attention the image site is the vision states read by every cross-attention layer from the band's first layer on. \Cref{tab:threellama} gives the three sites on 191 conflicts over the eight bands of five layers. Every prompt position reads the image directly, so more of the contribution passes through the other prompt positions than on LLaVA-1.5. \Cref{tab:otherpeak} gives the peak of the other prompt positions on eleven models: the gray-copy share that passes through them runs from 0.06 to 0.40.

\begin{table}[h]\centering\footnotesize\setlength{\tabcolsep}{7pt}
\caption{\textbf{Three-site band map, Llama-3.2} (191 conflicts). Share of the gray edit's effect recovered at each site, per band of five layers.}\label{tab:threellama}
\begin{tabular}{lccc}\toprule
Band & Image site & Other prompt positions & Answer position \\ \midrule
1 & 1.00 & 0.14 & 0.00 \\
2 & 0.83 & 0.29 & 0.01 \\
3 & 0.70 & 0.33 & 0.04 \\
4 & 0.62 & 0.36 & 0.19 \\
5 & 0.49 & 0.26 & 0.42 \\
6 & 0.36 & 0.18 & 0.57 \\
7 & 0.25 & 0.13 & 0.73 \\
8 & 0.09 & 0.09 & 0.98 \\ \bottomrule
\end{tabular}\end{table}

\begin{table}[h]\centering\footnotesize\setlength{\tabcolsep}{6pt}
\caption{\textbf{Peak share recovered at the other prompt positions}, gray copy patched into the clean run, with 95\% intervals. Dash: not reported.}\label{tab:otherpeak}
\begin{tabular}{lcccc}\toprule
Model & Conflicts & Peak & Layers of the peak & Peak on prior-agreeing items ($n$) \\ \midrule
LLaVA-1.5 & 144 & 0.13 [0.10, 0.16] & 16--19 & 0.14 [0.08, 0.20] \\
InstructBLIP-7B & 173 & 0.14 [0.11, 0.17] & -- & -- \\
InstructBLIP-13B & -- & 0.16 [0.13, 0.19] & -- & -- \\
Idefics2 & 196 & 0.24 [0.19, 0.30] & -- & -- \\
LFM2-VL & 178 & 0.40 [0.33, 0.47] & 15--18 & 0.52 [0.38, 0.69] (41) \\
Fuyu-8B & 203 & 0.06 [0.04, 0.08] & -- & 0.12 [0.08, 0.17] (42) \\
Emu3 & 227 & 0.16 [0.09, 0.22] & 16--19 & -- \\
Llama-3.2 & 191 & 0.36 [0.28, 0.44] & 15--19 & 0.39 [0.28, 0.50] (35) \\
Gemma-3-4B & -- & 0.23 [0.17, 0.28] & -- & -- \\
Gemma-3-12B & -- & 0.34 [0.29, 0.40] & -- & -- \\
Gemma-3-27B & -- & 0.35 [0.29, 0.40] & -- & -- \\ \bottomrule
\end{tabular}\end{table}

\noindent\emph{Notes on \cref{tab:otherpeak}.} On Gemma-3-12B the three sites sum to 1.41 in layers 24--29, where the image positions still recover 0.66 and the answer position 0.41.

\noindent\textbf{Paired-image band map on ViLP.} The gray-copy map shows where the image's contribution is carried. To see where the evidence that decides is carried, we use ViLP, where each question has a prior-consistent image and two conflict images of the same scene. The donor is the run on the prior-consistent image and the recipient the run on the conflict image; prompt and token layout are the same. All 600 conflict rows match the public release row of the same index on question, expected answer, conflict answer and conflict image. Recovery is the share of the margin difference between the two runs that a patch reproduces; we also count the image-wins whose decision reverses. \Cref{tab:paired} gives the image-wins per band and \cref{tab:pairedsum} the summary per model. Across the eight architectures the share that passes through the question tokens ranges from 0.04 to 0.60, and the curves cross later than with the gray copy on every model.

\begin{table}[h]\centering\footnotesize\setlength{\tabcolsep}{4pt}
\caption{\textbf{Paired-image map on ViLP, per model.} Image-wins mapped; share of items on which the prior-consistent image favors the expected answer; the most the other prompt positions recover, with 95\% interval where computed, and the most image-wins that patching them alone reverses; crossing of the image-site and answer-site curves with the paired-image donor, with the gray copy, and on prior-wins ($n$).}\label{tab:pairedsum}
\begin{tabular}{lccccccc}\toprule
 & & & \multicolumn{2}{c}{Other prompt positions} & \multicolumn{3}{c}{Crossing} \\ \cmidrule(lr){4-5}\cmidrule(lr){6-8}
Model & Image-wins & Favors expected & Recover up to & Reverse up to & Paired & Gray copy & Prior-wins ($n$) \\ \midrule
Gemma-3-12B & 436 & 0.97 & 0.60 & 71\% & 0.69 & 0.59 & 0.80 (155) \\
Idefics2 & 311 & 0.97 & 0.53 & 69\% & 0.70 & 0.57 & 0.73 (266) \\
Llama-3.2 & 461 & 0.98 & 0.43 [0.41, 0.46] & 55\% & 0.72 & 0.53 & 0.76 (136) \\
LFM2-VL & 342 & 0.96 & 0.27 [0.24, 0.30] & 35\% & 0.73 & 0.66 & 0.76 (247) \\
LLaVA-1.5 & 260 & 0.97 & 0.23 [0.20, 0.26] & 38\% & 0.65 & 0.58 & 0.67 (295) \\
Emu3 & 148 & 0.98 & 0.22 [0.17, 0.26] & 35\% & 0.82 & 0.68 & 0.85 (404) \\
InstructBLIP-7B & 246 & 0.97 & 0.21 [0.19, 0.24] & 28\% & 0.59 & 0.55 & 0.64 (309) \\
Fuyu-8B & 281 & 0.92 & 0.04 [0.02, 0.05] & 6\% & 0.79 & 0.74 & 0.78 (284) \\ \bottomrule
\end{tabular}\end{table}

\noindent\emph{Notes on \cref{tab:pairedsum}.} Whether the image tokens are conditioned on the question does not order the models: InstructBLIP's query tokens are (its Q-Former reads the instruction) and Fuyu's patches are not, and both are below Gemma-3 and Idefics2. On Llama-3.2 the image site is the vision states read by the cross-attention layers; in order of crossing depth it is first of the eight architectures with the gray donor and fifth with the paired-image donor. On Emu3 the image positions still recover 0.54 in the last band. In \cref{tab:paired} the bands of LFM2-VL start at 0.13, 0.27, 0.37, 0.63, 0.73 and 0.87 where the row label differs, and Fuyu's bands start at 0.11, 0.39, 0.61 and 0.89 of depth where the row label gives 0.12, 0.38, 0.62 and 0.88. The last column of \cref{tab:paired} is LLaVA-1.5-13B (Supp.~\ref{app:arch}), where the question tokens carry more than at 7B.

\begin{table}[h]\centering\footnotesize\setlength{\tabcolsep}{4pt}
\caption{\textbf{Paired-image band map, image-wins on ViLP.} Share of the margin difference recovered at the image positions / other prompt positions / answer position, and in parentheses the share of image-wins reversed by the patch at each site.}\label{tab:paired}
\begin{tabular}{lccc}\toprule
Band (entry depth) & Gemma-3-12B & Idefics2-8B & Llama-3.2-11B \\ \midrule
1 (0.00) & 1.00 / 0.00 / 0.00 (0.97 / 0.01 / 0.00) & 1.00 / 0.00 / 0.00 (0.96 / 0.01 / 0.00) & 1.00 / 0.05 / 0.00 (0.98 / 0.09 / 0.00) \\
2 (0.12) & 1.00 / 0.01 / 0.00 (0.97 / 0.03 / 0.00) & 1.00 / 0.02 / 0.00 (0.96 / 0.08 / 0.01) & 0.98 / 0.17 / 0.00 (0.97 / 0.21 / 0.00) \\
3 (0.25) & 1.01 / 0.06 / 0.00 (0.97 / 0.09 / 0.00) & 1.01 / 0.13 / 0.00 (0.96 / 0.23 / 0.02) & 0.95 / 0.31 / 0.00 (0.96 / 0.38 / 0.00) \\
4 (0.38) & 0.96 / 0.41 / 0.00 (0.92 / 0.48 / 0.01) & 0.98 / 0.31 / 0.01 (0.92 / 0.46 / 0.03) & 0.90 / 0.43 / 0.06 (0.92 / 0.55 / 0.05) \\
5 (0.50) & 0.80 / 0.60 / 0.19 (0.81 / 0.71 / 0.12) & 0.86 / 0.49 / 0.12 (0.87 / 0.63 / 0.13) & 0.75 / 0.36 / 0.25 (0.85 / 0.41 / 0.25) \\
6 (0.62) & 0.53 / 0.42 / 0.34 (0.56 / 0.42 / 0.29) & 0.46 / 0.53 / 0.29 (0.53 / 0.69 / 0.36) & 0.52 / 0.38 / 0.30 (0.66 / 0.42 / 0.33) \\
7 (0.75) & 0.29 / 0.33 / 0.47 (0.27 / 0.23 / 0.54) & 0.26 / 0.50 / 0.38 (0.33 / 0.64 / 0.50) & 0.35 / 0.43 / 0.43 (0.43 / 0.51 / 0.58) \\
8 (0.88) & 0.11 / 0.28 / 0.98 (0.08 / 0.16 / 0.96) & 0.14 / 0.47 / 0.89 (0.17 / 0.59 / 0.94) & 0.08 / 0.38 / 0.95 (0.07 / 0.46 / 0.97) \\ \bottomrule
\end{tabular}

\vspace{4pt}
\begin{tabular}{lccc}\toprule
Band (entry depth) & LLaVA-1.5-7B & InstructBLIP-7B & Fuyu-8B \\ \midrule
1 (0.00) & 1.00 / 0.00 / 0.00 (0.94 / 0.02 / 0.00) & 1.00 / 0.00 / 0.00 (0.94 / 0.01 / 0.00) & 1.00 / 0.00 / 0.00 (0.86 / 0.01 / 0.00) \\
2 (0.12) & 1.00 / 0.00 / 0.00 (0.93 / 0.03 / 0.01) & 1.00 / 0.02 / 0.00 (0.93 / 0.05 / 0.00) & 1.00 / 0.00 / 0.01 (0.86 / 0.04 / 0.00) \\
3 (0.25) & 1.00 / 0.09 / 0.00 (0.94 / 0.20 / 0.00) & 0.98 / 0.16 / 0.00 (0.92 / 0.23 / 0.01) & 0.99 / 0.00 / 0.00 (0.86 / 0.04 / 0.02) \\
4 (0.38) & 0.92 / 0.23 / 0.07 (0.92 / 0.38 / 0.08) & 0.93 / 0.21 / 0.09 (0.89 / 0.28 / 0.09) & 0.99 / 0.00 / 0.01 (0.86 / 0.05 / 0.01) \\
5 (0.50) & 0.84 / 0.19 / 0.21 (0.85 / 0.33 / 0.29) & 0.79 / 0.17 / 0.33 (0.83 / 0.23 / 0.42) & 0.99 / 0.01 / 0.01 (0.85 / 0.04 / 0.04) \\
6 (0.62) & 0.54 / 0.17 / 0.44 (0.68 / 0.25 / 0.60) & 0.44 / 0.10 / 0.58 (0.56 / 0.10 / 0.76) & 0.97 / 0.02 / 0.09 (0.85 / 0.06 / 0.10) \\
7 (0.75) & 0.22 / 0.14 / 0.63 (0.31 / 0.19 / 0.75) & 0.09 / 0.09 / 0.77 (0.13 / 0.10 / 0.87) & 0.81 / 0.04 / 0.49 (0.76 / 0.06 / 0.54) \\
8 (0.88) & 0.15 / 0.11 / 0.98 (0.23 / 0.14 / 0.93) & 0.06 / 0.08 / 0.97 (0.09 / 0.11 / 0.92) & 0.19 / 0.02 / 1.00 (0.21 / 0.03 / 0.86) \\ \bottomrule
\end{tabular}

\vspace{4pt}
\begin{tabular}{lccc}\toprule
Band (entry depth) & LFM2-VL-3B & Emu3-Chat & LLaVA-1.5-13B \\ \midrule
1 (0.00) & 1.00 / $-$0.01 / 0.00 (0.94 / 0.01 / 0.00) & 1.00 / 0.00 / 0.00 (0.93 / 0.02 / 0.00) & 1.00 / 0.01 / 0.00 (0.94 / 0.03 / 0.00) \\
2 (0.12) & 1.01 / $-$0.01 / 0.00 (0.94 / 0.04 / 0.00) & 1.00 / 0.00 / 0.00 (0.93 / 0.06 / 0.01) & 1.00 / 0.05 / 0.00 (0.95 / 0.11 / 0.00) \\
3 (0.25) & 1.02 / 0.06 / 0.00 (0.93 / 0.14 / 0.01) & 1.01 / 0.06 / 0.00 (0.94 / 0.14 / 0.01) & 0.95 / 0.21 / 0.01 (0.95 / 0.33 / 0.03) \\
4 (0.38) & 1.00 / 0.20 / 0.00 (0.94 / 0.28 / 0.01) & 0.93 / 0.17 / 0.02 (0.90 / 0.30 / 0.10) & 0.86 / 0.35 / 0.12 (0.91 / 0.52 / 0.16) \\
5 (0.50) & 0.92 / 0.27 / 0.03 (0.89 / 0.35 / 0.06) & 0.77 / 0.22 / 0.15 (0.80 / 0.35 / 0.23) & 0.62 / 0.38 / 0.26 (0.74 / 0.58 / 0.40) \\
6 (0.62) & 0.88 / 0.22 / 0.05 (0.85 / 0.28 / 0.09) & 0.73 / 0.20 / 0.29 (0.78 / 0.28 / 0.46) & 0.23 / 0.37 / 0.38 (0.33 / 0.58 / 0.57) \\
7 (0.75) & 0.61 / 0.21 / 0.62 (0.71 / 0.23 / 0.79) & 0.60 / 0.14 / 0.40 (0.70 / 0.18 / 0.57) & 0.11 / 0.34 / 0.45 (0.20 / 0.52 / 0.65) \\
8 (0.88) & 0.20 / 0.13 / 0.99 (0.29 / 0.14 / 0.94) & 0.54 / 0.12 / 0.67 (0.65 / 0.18 / 0.82) & 0.10 / 0.31 / 0.73 (0.20 / 0.46 / 0.85) \\ \bottomrule
\end{tabular}\end{table}

\noindent\textbf{Location by layer band on the LLaVA models.} Through layer 11 the image's contribution is present only in the image tokens: patching them reproduces the whole effect and patching the answer position reproduces none. Across layers 12--27 the two curves cross, and by layer 28 the answer position carries everything. The crossover, at layers 20--23, overlaps the band where the arbitration heads sit (mean layer 22 to 25) and is where the readout becomes predictive. The map measures sufficiency; the layer-0 recovery follows from the paired design. On the WHOOPS images where the text-only answer already agrees with the image and the model answers the same way on the clean image (31, 19 and 43 items moved by at least 0.5 logits by the gray edit), the crossover is at the same band on all three models, and the conflict-minus-agree difference at that band is between $-$0.06 and +0.09 at either site on every model, with intervals that include zero and extend to $\pm$0.2. Crossing depths by the interpolation of Supp.~\ref{app:arch} (first patched layer over the number of layers, linear between bands): 0.58 on LLaVA-NeXT, 0.57 on LLaVA-1.5 (0.58 under the common protocol of Supp.~\ref{app:arch}, on a different item sample), 0.59 on BakLLaVA and 0.72 on Qwen2.5-VL, computed from the per-band values above.

\section{Subspace construction and necessity}\label{app:necessity}
\noindent\textit{Supports main-paper Sec.~7. How the image-token subspace is built, the patching and removal results with their controls, and the check in float32.}\par\smallskip

\noindent \Cref{tab:suffmain} collects the patching and removal results that main-paper Sec.~7 and Fig.~5 report for the three LLaVA models.

\begin{table}[h]\centering\footnotesize\setlength{\tabcolsep}{4pt}
\caption{\textbf{Patching and removal of the image-token subspace} (WHOOPS, $n\approx420$ per model). Recovery: fraction of the gray effect reproduced by patching. Subspace removal: shares of image-wins and of prior-wins reversed by projecting the subspace out of the residual. The basis is fit on the other conflict sources (cross-source). The fitted subspace holds 93\%, 71\% and 98\% of per-token variance. Runs are in float16, except the LLaVA-1.5 cell of the last row (float32; Supp.~E).}\label{tab:suffmain}
\begin{tabular}{p{8.0cm}p{2.5cm}p{2.5cm}p{2.3cm}}\toprule
 & LLaVA-NeXT & LLaVA-1.5 & BakLLaVA \\ \midrule
Patch at the image tokens, layers 16--24, cross-source basis: full / 256-D [95\% CI, clustered by image], \% & 70 / 63 [57, 69] & 77 / 69 [63, 75] & 73 / 67 [60, 73] \\
Same patch: next 256 components / random in top-1,024 span, \% & 4.4 / 17 & 7.5 / 24 & 3.5 / 17 \\
Same patch: 256 components outside the top-1,024 block (at most 2\% of the subspace's variance), \% & 1.2 [0.6, 1.8] & 1.8 [1.0, 2.8] & $-$0.3 [$-$1.3, 0.5] \\
Removing the 256-D subspace at the image tokens: image-wins / prior-wins reversed, \% & 36.1 / 4.5 & 42.3 / 3.7 & 34.0 / 2.9 \\
The same at the answer position: image-wins / prior-wins reversed, \% & 8.3 / 14.4 & 10.1 / 13.8 & 7.6 / 8.7 \\ \bottomrule
\end{tabular}\end{table}

\begin{figure}[h]\centering\includegraphics{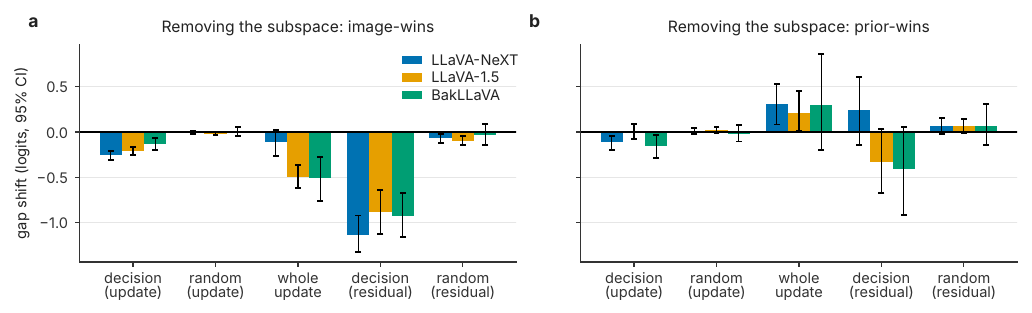}
\caption{\textbf{Removing the fitted 256-D subspace at the answer position.} Gap shift in logits (bootstrap 95\% intervals) on image-wins (a) and on prior-wins (b) when the fitted subspace (labeled ``decision'') or a random 256-D subspace is removed from each layer's own update or from the accumulated residual, and when the whole update is removed. Blue, orange and green: LLaVA-NeXT, LLaVA-1.5 and BakLLaVA.}\end{figure}

\begin{figure}[h]\centering\includegraphics{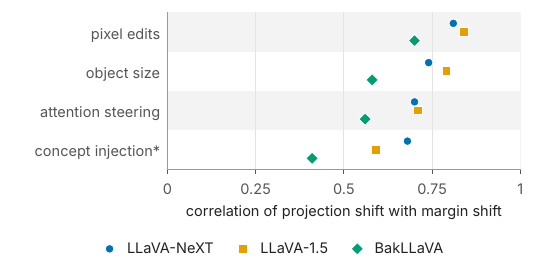}
\caption{\textbf{Answer-concept projection and the decision.} Correlation, over items, between the shift of the layer-20 answer-concept projection at the answer position and the shift of the gap, under pixel edits, a change of object size, attention steering and concept injection, per model. $^*$For concept injection the projection shift is set by the intervention.}\label{fig:levers}\end{figure}

\begin{figure}[h]\centering\includegraphics{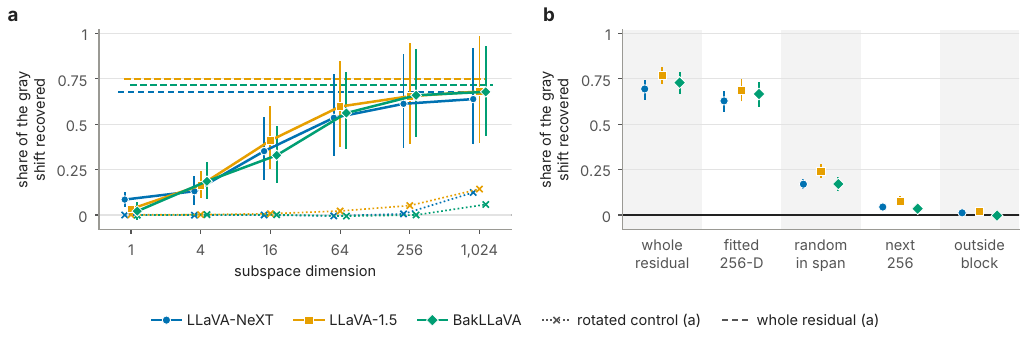}
\caption{\textbf{Subspace sufficiency by dimension, with controls} (share of the gray edit's effect recovered by patching at the image tokens of layers 16 to 24; 95\% intervals). (a) By subspace dimension, with bases fit on half the conflicts and evaluated on the other half (216, 216 and 217 items); crosses: rotated subspaces of the same dimension; dashed: the whole residual patched. (b) With bases fit on the other sources and evaluated on WHOOPS (424, 416 and 419 conflicts; intervals clustered by image): the whole residual, the fitted 256-D subspace, a random subspace within the top-1,024 span, the next 256 principal components, and the 256 highest-variance components outside the top-1,024 block.}\label{fig:suffdim}\end{figure}

\noindent\textbf{Subspace construction.} The 256-dimensional subspace is one basis per layer: the top principal components of the attended image-token residuals captured on the conflict sets, fit once per layer. It is applied at the image tokens for patching the edited copy of the same image and at the answer position for interchange, necessity and injection; the half-split and cross-source held-out designs and the cross-model alignment use the layer-20 basis. Ablation at the answer position is over the band $\{16, 20, 24\}$. \Cref{tab:incabl} gives increment ablation (the subspace removed from each layer's own update) and residual ablation (the subspace removed from the accumulated stream). On LLaVA-NeXT the whole-update flips are symmetric, 5.4\% of image-wins and 6.3\% of prior-wins, so the mean gap shift is near zero while the decision is perturbed. Under pixel edits, object shrinking and attention reallocation, the shift in the layer-20 answer-concept projection correlates with the shift in the decision at $r$ = 0.56--0.84 (\cref{fig:levers}).

\begin{table}[h]\centering\footnotesize\setlength{\tabcolsep}{7pt}
\caption{\textbf{Ablation at the answer position, layers 16, 20 and 24.} Image-wins flipped (\%) and the mean shift of the gap in logits.}\label{tab:incabl}
\begin{tabular}{llccc}\toprule
Ablation & Removed & LLaVA-NeXT & LLaVA-1.5 & BakLLaVA \\ \midrule
Increment & fitted 256-D subspace, flips & 2.6 & 1.6 & 2.6 \\
 & \quad gap & $-0.22$ & $-0.15$ & $-0.14$ \\
 & random subspace, flips & 1.3 & 0.0 & 1.9 \\
 & \quad gap & $+0.01$ & $-0.01$ & $+0.02$ \\
 & whole update, flips & 5.4 & 3.9 & 7.1 \\
 & \quad gap & $-0.006$ & $-0.31$ & $-0.31$ \\ \midrule
Residual & fitted 256-D subspace, flips & 11.8 & 12.1 & 9.3 \\
 & \quad gap over all items & $-0.78$ & $-0.73$ & $-0.79$ \\
 & \quad gap over image-wins only & $-1.14$ & $-0.88$ & $-0.93$ \\
 & random subspace, flips & 2.2 & 1.0 & 2.2 \\ \bottomrule
\end{tabular}\end{table}

\noindent\textbf{Off-block control and recovery per unit variance (main-paper Sec.~7).} The per-token covariance of the image-token residuals at each block output is accumulated over 64 clean WHOOPS forwards (36,864 tokens on LLaVA-1.5) and has full rank 4,096. \Cref{tab:varheld} gives the fraction of that variance held by each 256-D subspace. The 256 components outside the top-1,024 block that hold the most variance reach at most 0.015 of the fitted subspace's variance on LLaVA-1.5 (20 draws weighted by eigenvalue$^{12}$, the cap, are all at the same ratio), so no equal-variance control outside the block exists at this dimension. On BakLLaVA the image-token residual is nearer to low rank than LLaVA-1.5's. \Cref{tab:suffctrl} gives sufficiency for every arm and the recovery per unit of captured variance; \cref{tab:suffmain} reports the same values.

\begin{table}[h]\centering\footnotesize\setlength{\tabcolsep}{7pt}
\caption{\textbf{Per-token variance held by each 256-D subspace} at blocks 16 / 20 / 24. The last two rows are relative to the fitted subspace's variance; the greedy set is chosen after orthogonalising against the fitted basis. Dash: not reported.}\label{tab:varheld}
\begin{tabular}{lccc}\toprule
Subspace & LLaVA-NeXT & LLaVA-1.5 & BakLLaVA \\ \midrule
Top-256 principal components & 0.987/0.972/0.949 & 0.956/0.921/0.879 & 0.993/0.990/0.984 \\
Fitted 256-D subspace (cross-source basis) & 0.968/0.931/0.881 & 0.799/0.714/0.630 & 0.985/0.980/0.968 \\
\quad of which inside the token top block & 0.83/0.75/0.67 & -- & 0.91/0.86/0.79 \\
Next 256 components & 0.008/0.015/0.023 & 0.040/0.048/0.059 & 0.005/0.005/0.007 \\
Random in the top-1,024 span & 0.268/0.245/0.241 & 0.229/0.207/0.188 & 0.273/0.252/0.260 \\
Random 256-D & 0.057/0.062/0.066 & 0.063/0.062/0.063 & 0.058/0.062/0.066 \\ \midrule
Best 256 components outside the top-1,024 block & 0.001/0.002/0.005 & 0.004/0.009/0.015 & 0.000/0.001/0.001 \\
Greedy set outside the block & 0.002/0.004/0.007 & 0.008/0.014/0.022 & 0.001/0.001/0.002 \\ \bottomrule
\end{tabular}\end{table}

\begin{table}[h]\centering\footnotesize\setlength{\tabcolsep}{6pt}
\caption{\textbf{Sufficiency of each subspace} (edit-to-clean patch at the image tokens, all three blocks; share of the gray effect recovered, cluster bootstrap on image), and the recovery per unit of captured variance.}\label{tab:suffctrl}
\begin{tabular}{lccc}\toprule
 & LLaVA-NeXT & LLaVA-1.5 & BakLLaVA \\ \midrule
Full residual & 0.695 [0.636, 0.746] & 0.769 [0.721, 0.814] & 0.730 [0.666, 0.787] \\
Fitted subspace & 0.629 [0.568, 0.686] & 0.687 [0.626, 0.750] & 0.667 [0.598, 0.731] \\
Random in span & 0.170 [0.142, 0.199] & 0.240 [0.206, 0.280] & 0.171 [0.130, 0.211] \\
Next 256 & 0.044 [0.030, 0.062] & 0.075 [0.054, 0.102] & 0.035 [0.013, 0.058] \\
Outside block, greedy & 0.012 [0.006, 0.018] & 0.018 [0.010, 0.028] & $-$0.003 [$-$0.013, 0.005] \\
Outside block, draws & 0.008 to 0.013 & 0.012 to 0.018 & $-$0.004 to 0.001 \\
Outside block, draws pooled & 0.011 [0.005, 0.017] & 0.015 [0.009, 0.024] & $-$0.001 [$-$0.010, 0.007] \\ \midrule
\multicolumn{4}{l}{\emph{Recovery per unit of captured variance}} \\
Fitted subspace & 0.68 & 0.96 & 0.68 \\
Random in span & 0.68 & 1.15 & 0.65 \\
Next 256 & 2.9 & 1.55 & 6.2 \\
Outside block, greedy & 3.2 & 1.72 & undefined \\
Outside block, draws & -- & 1.2 to 1.8 & -- \\ \bottomrule
\end{tabular}\end{table}

\noindent\textbf{Necessity as a cumulative-removal curve.} The top $k$ dimensions of each subspace are projected out of the accumulated residual in all three blocks, and a flip is a sign change of the gap among image-wins. \Cref{tab:neccurve} gives the curves in float16 and \cref{tab:necctrl} the controls, the gap shift and the prior-wins at $k$ = 256. Removal at the image tokens is directional; removal at the answer position reverses decisions in both directions. At the image tokens the prior-win reversals on BakLLaVA are not monotone in $k$, so removing only the top few dimensions is not directional on this model (LLaVA-NeXT: 8.1\% at $k$ = 2 and 1.8 to 4.5\% elsewhere; LLaVA-1.5: 0.9 to 4.6\% throughout).

\begin{table}[h]\centering\footnotesize\setlength{\tabcolsep}{5.5pt}
\caption{\textbf{Cumulative removal of the fitted subspace, float16.} Image-wins flipped (\%) when its top $k$ dimensions are projected out; last row, prior-wins. $^\dagger$: non-finite gaps, counted as reversals (\cref{tab:f32curve} gives float32).}\label{tab:neccurve}
\begin{tabular}{llccccccccc}\toprule
Model & Site & $k$ = 1 & 2 & 4 & 8 & 16 & 32 & 64 & 128 & 256 \\ \midrule
LLaVA-NeXT & image tokens & 2.6 & 3.5 & 3.5 & 7.7 & 10.5 & 20.1 & 27.8 & 32.6 & 36.1 \\
 & answer position & 0.3 & 0.6 & 0.6 & 1.3 & 2.6 & 4.2 & 6.1 & 7.7 & 8.3 \\
LLaVA-1.5 & image tokens & 7.2$^\dagger$ & 13.0$^\dagger$ & 4.9 & 7.5 & 12.1 & 15.3 & 25.7 & 31.6 & 42.3 \\
 & answer position & 0.7 & 1.6 & 1.6 & 2.3 & 2.9 & 3.6 & 5.2 & 8.5 & 10.1 \\
BakLLaVA & image tokens & 5.4 & 6.7 & 9.8 & 12.7 & 19.4 & 22.2 & 25.7 & 31.4 & 34.0 \\
 & answer position & 0.3 & 1.0 & 2.2 & 1.9 & 2.2 & 4.1 & 4.8 & 5.1 & 7.6 \\ \midrule
BakLLaVA, prior-wins & image tokens & 6.7 & 18.3 & 19.2 & 14.4 & 10.6 & 6.7 & 5.8 & 5.8 & 2.9 \\ \bottomrule
\end{tabular}\end{table}

\begin{table}[h]\centering\footnotesize\setlength{\tabcolsep}{5pt}
\caption{\textbf{Controls for the removal curve, and prior-wins at $k$ = 256.} Range of image-wins flipped (\%) over $k$ for the random and off-block subspaces; shift of the gap in logits at $k$ = 256; prior-wins reversed (\%) at $k$ = 256 by the fitted / random / off-block subspace.}\label{tab:necctrl}
\begin{tabular}{llccccc}\toprule
 & & \multicolumn{2}{c}{Image-wins flipped over $k$} & Gap shift, & Prior- & Prior-wins reversed, \\
Model & Site & Random & Off-block & fitted & wins & fitted / random / off-block \\ \midrule
LLaVA-NeXT & image tokens & 0.0 to 1.0 & 0.0 to 0.6 & $-$3.5 & 111 & 4.5 / 1.8 / 0.9 \\
 & answer position & 0.0 to 2.2 & 0.3 to 1.6 & $-$0.89 [$-$1.10, $-$0.69] & & 14.4 / 4.5 / 6.3 \\
LLaVA-1.5 & image tokens & 0.0 to 1.3 & 0.0 to 0.3 & $-$3.3 & 109 & 3.7 / 0.0 / 0.9 \\
 & answer position & 0.0 to 1.0 & 0.0 to 1.3 & $-$0.71 & & 14.7 / 3.7 / 5.5 \\
BakLLaVA & image tokens & 0.0 to 1.6 & 0.0 to 1.3 & $-$4.2 & 104 & 2.9 / 1.9 / 1.0 \\
 & answer position & 0.3 to 2.2 & 0.3 to 3.5 & $-$0.76 & & 8.7 / 4.8 / 10.6 \\ \bottomrule
\end{tabular}\end{table}

\noindent\emph{Notes on \cref{tab:neccurve,tab:necctrl}.} At the answer position at $k$ = 256 the gap shift of the random subspace on LLaVA-NeXT is $-$0.11, and that of the off-block subspace is $+$0.07 [$+$0.02, $+$0.12] on LLaVA-NeXT, a rise, and $+$0.06 [$-$0.04, $+$0.16] on BakLLaVA. \emph{Runs.} LLaVA-1.5: 416 conflicts. BakLLaVA: the same design on 419 conflicts with 1,061 fitting examples. LLaVA-NeXT: the same design on 424 conflicts, with bases fit on 1,081 fitting examples from the other sources and necessity on all 424 conflicts (313 image-wins, 111 prior-wins). On BakLLaVA and LLaVA-NeXT all five implementation checks pass (hook identity, full and subspace self-patch, ablation identity, bootstrap). On the first 200 conflicts of LLaVA-NeXT alone (147 image-wins, 53 prior-wins) the rates are 34.7\% and 1.9\% at the image tokens and 10.2\% and 11.3\% at the answer position. Under the other basis convention on BakLLaVA the fitted subspace recovers 0.659, the random subspace in the span 0.148 and the next 256 components 0.029.

\noindent\textbf{Precision.} The runs of this section are in float16. On LLaVA-1.5 the float16 removal curve at the image tokens has non-finite gaps at 1 and 2 dimensions, which are counted as reversals. We repeated the removal and patching in float32 with the same bases (\cref{tab:f32curve,tab:f32sum}). No forward pass in float32 gives a non-finite gap (23,712 passes on LLaVA-1.5, 23,883 on BakLLaVA, 24,168 on LLaVA-NeXT). On LLaVA-NeXT and BakLLaVA the float32 curves differ from the float16 ones by at most one item; on LLaVA-1.5 by at most three items (1.0 point of image-wins) wherever the float16 gap was finite. Main-paper Fig.~5b and c show the float32 curve for LLaVA-1.5.

\begin{table}[h]\centering\footnotesize\setlength{\tabcolsep}{4.5pt}
\caption{\textbf{Removal curve in float32 and float16.} Decisions reversed (\%) when the top $k$ dimensions of the subspace are projected out, at the image tokens unless noted. $^\dagger$: non-finite gaps in float16, counted as reversals. Dash: not reported.}\label{tab:f32curve}
\begin{tabular}{lllccccccccc}\toprule
 & & & \multicolumn{9}{c}{Dimensions removed} \\ \cmidrule(lr){4-12}
Model & Reversed & Precision & 1 & 2 & 4 & 8 & 16 & 32 & 64 & 128 & 256 \\ \midrule
LLaVA-NeXT & image-wins & float32 & 2.6 & 3.5 & 3.5 & 7.7 & 10.5 & 20.1 & 27.8 & 32.9 & 36.1 \\
 & & float16 & 2.6 & 3.5 & 3.5 & 7.7 & 10.5 & 20.1 & 27.8 & 32.6 & 36.1 \\
 & prior-wins & float32 & 1.8 & 8.1 & 2.7 & 1.8 & 3.6 & 4.5 & 4.5 & 4.5 & 4.5 \\
 & image-wins, answer position & float32 & 0.3 & 0.6 & 0.6 & 1.6 & 2.6 & 4.2 & 6.1 & 7.7 & 8.0 \\
 & & float16 & 0.3 & 0.6 & 0.6 & 1.3 & 2.6 & 4.2 & 6.1 & 7.7 & 8.3 \\ \midrule
LLaVA-1.5 & image-wins & float32 & 1.6 & 3.6 & 4.9 & 7.8 & 11.4 & 15.6 & 24.8 & 32.6 & 42.3 \\
 & & float16 & 7.2$^\dagger$ & 13.0$^\dagger$ & 4.9 & 7.5 & 12.1 & 15.3 & 25.7 & 31.6 & 42.3 \\
 & prior-wins & float32 & 0.9 & 0.9 & 3.7 & 3.7 & 4.6 & 3.7 & 5.5 & 3.7 & 3.7 \\
 & & float16 & 0.9 & 0.9 & 2.8 & 3.7 & 4.6 & 3.7 & 4.6 & 3.7 & 3.7 \\ \midrule
BakLLaVA & image-wins & float32 & 5.4 & 6.7 & 9.8 & 12.4 & 19.0 & 21.9 & 25.4 & 31.4 & 34.3 \\
 & & float16 & 5.4 & 6.7 & 9.8 & 12.7 & 19.4 & 22.2 & 25.7 & 31.4 & 34.0 \\
 & prior-wins & float32 & 6.7 & 19.2 & 18.3 & 14.4 & 10.6 & 6.7 & 5.8 & 5.8 & 2.9 \\
 & & float16 & 6.7 & 18.3 & 19.2 & 14.4 & 10.6 & 6.7 & 5.8 & 5.8 & 2.9 \\ \bottomrule
\end{tabular}\end{table}

\begin{table}[h]\centering\footnotesize\setlength{\tabcolsep}{5pt}
\caption{\textbf{Removal and sufficiency at 256 dimensions, float32 (float16 in parentheses or on the next row).} Top: decisions reversed (\%) and the change in the gap, in logits. Middle: image-wins reversed (\%) under mean ablation, and the share of the full ablation's change in the gap that the subspace at its mean gives. Bottom: share of the gray edit's effect recovered by patching. Intervals are 95\%. Dash: not reported.}\label{tab:f32sum}
\begin{tabular}{lccc}\toprule
 & LLaVA-NeXT & LLaVA-1.5 & BakLLaVA \\ \midrule
Image-wins reversed & 36.1 [31.0, 41.4] (36.1) & 42.3 [37.1, 48.2] (42.3) & 34.3 (34.0) \\
Prior-wins reversed & 4.5 (4.5) & 3.7 (3.7) & 2.9 (2.9) \\
Gap change & $-3.50$ ($-3.51$) & $-3.28$ ($-3.28$) & -- \\
Answer position: image-wins reversed & 8.0 (8.3) & 10.1 (10.1) & 7.9 (7.6) \\
Answer position: prior-wins reversed & 14.4 (14.4) & 13.8 (14.7) & 8.7 (8.7) \\ \midrule
Subspace at its mean & 43.1 (43.5) & 47.2 (47.9) & 35.9 (36.2) \\
Share of the full ablation & 0.91 [0.89, 0.93] & -- & 0.89 [0.87, 0.92] \\
\quad in float16 & 0.91 & -- & 0.89 [0.87, 0.92] \\
Every image position at the mean & 46.3 (46.6) & 55.4 (54.7) & 40.0 (40.3) \\
Only the subspace kept & 8.3 (8.3) & 13.4 (13.4) & 6.7 (6.7) \\ \midrule
Fitted subspace & 0.629 [0.568, 0.685] & 0.686 [0.625, 0.751] & 0.668 [0.599, 0.732] \\
\quad in float16 & 0.629 [0.568, 0.686] & 0.687 [0.626, 0.750] & 0.667 [0.598, 0.731] \\
Whole residual & 0.694 (0.695) & 0.767 (0.769) & 0.729 (0.730) \\
Random subspace in the span & 0.170 (0.170) & 0.241 (0.240) & 0.168 (0.171) \\
Next 256 components & 0.043 (0.044) & 0.076 (0.075) & 0.035 (0.035) \\
Set outside the block & 0.011 (0.012) & 0.016 (0.018) & -- \\ \bottomrule
\end{tabular}\end{table}

\noindent\emph{Note on \cref{tab:f32curve}.} On LLaVA-NeXT a random subspace reverses at most 1.0\% of image-wins and the set outside the block at most 0.6\%.

\FloatBarrier
\noindent\textbf{Notes.} The answer-concept direction is candidate-dependent, a family with one construction rule at one site, which is the sense in which the representation is shared. Part of the correlation between the projection shift and the decision shift ($r$ = 0.56--0.84) is structural, since the projection is a component of the margin; the injection contrast is the stronger evidence. On the variance control, recovery per unit of variance on LLaVA-1.5 is 0.96 for the fitted 256-D subspace, 1.15 for a random 256-D subspace in the top-1,024 span, 1.55 for the next 256 principal components and 1.7 for the 256 components outside the top-1,024 block. BakLLaVA's residual is nearer to low rank (fitted 256-D subspace 98\% of the variance and 67\% recovered; next 256 components 0.6\% and 3.5\%, its one arm above the line; off-block 0.1\% and nothing) and LLaVA-NeXT lies between (93\% and 63\%). The equal-variance control we pre-specified does not exist, because the fitted subspace holds most of the variance: a 256-dimensional block of the image-token residual carries an edit's effect in proportion to how much of the residual it is. Injecting the answer-concept direction split into its components inside and outside the subspace, each at its natural norm, the part inside carries 28--38\% of the effect at $\alpha$ = 0.5 from 26--29\% of the norm and the complement carries the rest. The four BakLLaVA control values come from a separate run (Supp.~\ref{app:faq}).

\noindent\textbf{The subspace against removing the image at the image tokens.} Projecting 256 dimensions out of the image tokens could act simply by damaging the image there. On the three LLaVA models we compared, at the image tokens of blocks 16, 20 and 24 in one forward: the projection used above; setting the subspace component to its mean over 36,864 image tokens of 64 clean forwards (184,272 on LLaVA-NeXT); the same for the off-block and the random 256-D control; setting every image token to that mean; and keeping only the subspace, with the other 3,840 dimensions set to their mean. \Cref{tab:imgablate} gives the decisions reversed under each arm and \cref{tab:imgshare} the share of the full ablation's change in the gap that the subspace produces.

\begin{table}[h]\centering\footnotesize\setlength{\tabcolsep}{6pt}
\caption{\textbf{Share of the full ablation's change in the gap} (every image token set to the mean), with 95\% intervals. Dash: not reported.}\label{tab:imgshare}
\begin{tabular}{lcccc}\toprule
 & \multicolumn{2}{c}{Conflicts} & \multicolumn{2}{c}{Prior-agreeing items} \\ \cmidrule(lr){2-3}\cmidrule(lr){4-5}
Model & Subspace at its mean & Only the subspace kept & Subspace at its mean & Only the subspace kept \\ \midrule
LLaVA-NeXT & 0.91 [0.89, 0.93] & 0.23 [0.20, 0.26] & -- & -- \\
LLaVA-1.5 & 0.86 [0.84, 0.88] & 0.33 [0.30, 0.36] & 0.92 [0.81, 1.02] & 0.38 [0.27, 0.50] \\
BakLLaVA & 0.89 [0.87, 0.92] & 0.11 [0.09, 0.14] & 0.92 [0.81, 1.02] & -- \\ \bottomrule
\end{tabular}\end{table}

\noindent\emph{BakLLaVA.} Setting every image token to the mean reverses 40.3\% of 315 image-wins, setting only the subspace to its mean 36.2\%, and keeping only the subspace 6.7\%, where keeping the off-block or the random subspace reverses 41.0\% and 39.0\%. The 69 prior-agreeing items show the same pattern at a lower rate: 8.7\% reversed with every image token at the mean, 7.2\% with the subspace at its mean, 1.4\% with only the subspace kept. At these blocks the subspace accounts for about nine tenths of what the image tokens contribute to the answer (0.89 on conflicts, 0.92 on prior-agreeing items).

\noindent\emph{LLaVA-1.5} gives the same ordering with a larger remainder (416 conflicts, 307 image-wins; 75 prior-agreeing items). Setting every image token to the mean reverses 54.7\% of image-wins, setting only the subspace to its mean 47.9\%, the off-block and the random control at their mean 0.3\% and 1.0\%, and keeping only the subspace 13.4\%, against 55.0\% when either control is kept. The 256 dimensions carry most of the image tokens' contribution at these blocks, and the other 3,840 carry about a third of it that the subspace alone does not restore.

\noindent\emph{LLaVA-NeXT,} with about 2,900 image tokens per image, falls between the two (424 conflicts, 313 image-wins; 68 prior-agreeing items): every image token at the mean reverses 46.6\%, the subspace at its mean 43.5\%, the controls at their mean 0.6\% and 1.9\%, and the subspace kept alone 8.3\%, against 46.3\% and 47.0\% when a control is kept.

\begin{table}[h]\centering\footnotesize\setlength{\tabcolsep}{4pt}
\caption{\textbf{Subspace against image-token ablation} (image tokens of blocks 16, 20, 24; cluster-bootstrap 95\% intervals over images). BakLLaVA: 419 WHOOPS conflicts (315 image-wins, 104 prior-wins) and 69 prior-agreeing items. LLaVA-1.5: 416 conflicts (307 image-wins, 109 prior-wins) and 75 prior-agreeing items. LLaVA-NeXT: 424 conflicts (313 image-wins, 111 prior-wins) and 68 prior-agreeing items.}\label{tab:imgablate}
\begin{tabular}{lccc}\toprule
Arm & Image-wins reversed & Prior-wins reversed & Agreeing items reversed \\ \midrule
\multicolumn{4}{l}{\emph{BakLLaVA}} \\
Subspace projected out & 34.0\% [28.8, 39.2] & 2.9\% [0.0, 6.0] & 8.7\% [2.9, 15.9] \\
Subspace set to its mean & 36.2\% [30.7, 41.6] & 2.9\% [0.0, 6.0] & 7.2\% [1.4, 14.5] \\
Off-block control set to its mean & 1.3\% [0.3, 2.5] & 1.9\% [0.0, 4.9] & 1.4\% [0.0, 4.3] \\
Random control set to its mean & 1.3\% [0.3, 2.5] & 1.9\% [0.0, 4.8] & 0.0\% \\
Every image token set to the mean & 40.3\% [34.7, 45.8] & 3.8\% [0.9, 7.6] & 8.7\% [2.9, 15.9] \\
Only the subspace kept & 6.7\% [4.1, 9.6] & 3.8\% [0.8, 7.9] & 1.4\% [0.0, 4.3] \\
Only the off-block control kept & 41.0\% [35.4, 46.7] & 3.8\% [0.9, 7.6] & 8.7\% [2.9, 15.9] \\
Only the random control kept & 39.0\% [33.7, 44.5] & 2.9\% [0.0, 6.0] & 8.7\% [2.9, 15.9] \\ \midrule
\multicolumn{4}{l}{\emph{LLaVA-1.5}} \\
Subspace projected out & 42.3\% [37.1, 48.2] & 3.7\% [0.9, 7.4] & 5.3\% [1.3, 10.7] \\
Subspace set to its mean & 47.9\% [42.5, 53.5] & 2.8\% [0.0, 6.5] & 6.7\% [1.3, 13.3] \\
Off-block control set to its mean & 0.3\% [0.0, 1.0] & 0.0\% & 0.0\% \\
Random control set to its mean & 1.0\% [0.0, 2.2] & 0.9\% [0.0, 3.1] & 0.0\% \\
Every image token set to the mean & 54.7\% [49.2, 60.3] & 2.8\% [0.0, 6.2] & 6.7\% [1.3, 13.3] \\
Only the subspace kept & 13.4\% [9.7, 17.4] & 0.9\% [0.0, 3.1] & 4.0\% [0.0, 9.3] \\
Only the off-block control kept & 55.0\% [49.7, 60.6] & 2.8\% [0.0, 6.2] & 6.7\% [1.3, 13.3] \\
Only the random control kept & 55.0\% [49.7, 60.6] & 2.8\% [0.0, 6.2] & 6.7\% [1.3, 13.3] \\ \midrule
\multicolumn{4}{l}{\emph{LLaVA-NeXT}} \\
Subspace projected out & 36.1\% [31.0, 41.4] & 4.5\% [1.0, 8.8] & 5.9\% [1.5, 11.8] \\
Subspace set to its mean & 43.5\% [38.2, 49.4] & 1.8\% [0.0, 4.5] & 5.9\% [1.5, 11.8] \\
Off-block control set to its mean & 0.6\% [0.0, 1.6] & 0.9\% [0.0, 2.9] & 0.0\% \\
Random control set to its mean & 1.9\% [0.6, 3.6] & 0.9\% [0.0, 3.1] & 1.5\% [0.0, 4.4] \\
Every image token set to the mean & 46.6\% [41.4, 52.5] & 1.8\% [0.0, 4.5] & 7.4\% [1.5, 14.7] \\
Only the subspace kept & 8.3\% [5.3, 11.4] & 4.5\% [0.9, 8.9] & 2.9\% [0.0, 7.4] \\
Only the off-block control kept & 46.3\% [41.2, 52.4] & 1.8\% [0.0, 4.5] & 7.4\% [1.5, 14.7] \\
Only the random control kept & 47.0\% [41.7, 52.9] & 1.8\% [0.0, 4.5] & 7.4\% [1.5, 14.7] \\ \bottomrule
\end{tabular}\end{table}

\noindent\textbf{A language-prior subspace at the answer position (pre-specified test).} The image-token subspace is fit on image runs. We pre-specified the corresponding test for the prior on the three LLaVA models: at the answer position of blocks 16, 20 and 24, take the top 256 principal components of the residual over text-only runs of the cross-source items (the \emph{prior basis}), and compare it with the top 256 components over image runs (the \emph{image-run basis}), the next 256 text-run components, 256 random directions inside the top image-run components, the image-token subspace of the main text, and a random basis. The hypothesis was that the prior basis is specific to the prior: that setting it to its image-run mean reverses more prior-wins than image-wins and more than the random and image-run bases, and that its share of the block-20 readout tracks the text-only prior and does not move with blur. It fails on all three. \Cref{tab:priorsub} gives the decisions reversed by each basis and \cref{tab:priorextra} the overlap of the two bases, the readout tests and the patching of the text-only run.

\begin{table}[h]\centering\footnotesize\setlength{\tabcolsep}{6pt}
\caption{\textbf{The prior basis against the image-run basis.} Overlap at block 20 (mean squared cosine of the principal angles; 0.06 by chance); the prior basis's share of the readout against the remainder; image-wins reversed (\%) when the text-only run's answer-position residual is patched into the image run, whole or through one basis. 95\% intervals in brackets. Dash: not reported.}\label{tab:priorextra}
\begin{tabular}{lccc}\toprule
 & LLaVA-NeXT & LLaVA-1.5 & BakLLaVA \\ \midrule
Fit items & 1,075 & 1,059 & 1,055 \\
Mean squared cosine of the two bases & 0.46 & 0.53 & 0.42 \\
Image-run variance held by the prior basis & 0.70 & 0.71 & 0.62 \\ \midrule
Correlation with the text-only prior: prior basis / remainder & $-0.21$ / $-0.20$ & $-0.22$ / $-0.17$ & $-0.18$ / $-0.17$ \\
Fall under blur at radius 8, prior basis (logits) & 0.40 [0.32, 0.50] & 1.84 [1.59, 2.10] & 0.84 [0.73, 0.96] \\
Fall under blur at radius 8, remainder (logits) & 0.62 [0.50, 0.73] & 2.65 [2.33, 2.98] & 2.08 [1.80, 2.35] \\ \midrule
Text-only residual patched whole & 70.3 & 82.7 & 70.2 \\
\quad through the prior basis & 25.6 & 32.2 & 18.1 \\
\quad through the image-run basis & 34.2 & 45.6 & 19.4 \\
\quad through the image-token subspace & -- & 30.6 & -- \\
\quad through the random basis & 3.2 & 3.9 & 4.1 \\ \bottomrule
\end{tabular}\end{table}

\noindent\emph{LLaVA-1.5.} Setting the prior basis to its mean reverses 23.9\% of 109 prior-wins and 5.2\% of 307 image-wins, above the random basis (4.6\% and 0.7\%), and the image-run basis reverses the same 23.9\% of prior-wins. The two bases share most of their directions: 58 principal angles have a cosine above 0.9, and the prior basis holds 0.95 of text-run variance at block 20. The prior basis's share of the readout correlates with the text-only prior no more than the remainder does, and it falls from clean to blur. A single direction fit to the text-only gap ($R^2$ 0.70, five-fold 0.54) reverses no decision when set to its mean.

\noindent\emph{BakLLaVA} agrees (419 conflicts). The prior basis reverses 21.2\% of 104 prior-wins and 5.7\% of 315 image-wins, the image-run basis 19.2\% and 6.3\%, and the random basis 4.8\% and 2.9\%; the two intervals on prior-wins overlap almost entirely.

\noindent\emph{LLaVA-NeXT} (424 conflicts). The removal half of the criterion is met by point estimate and the readout half is not. The prior basis reverses 33.3\% [24.7, 42.1] of 111 prior-wins and 3.8\% of 313 image-wins, the image-run basis 25.2\% [17.5, 34.5] and 7.0\%, and the random basis 7.2\% and 2.2\%.

\noindent\emph{Reading.} On all three models every basis that spans the answer-position residual, set to its mean over image runs, moves prior-wins toward the image's answer and image-wins away from it; the prior basis does this to prior-wins no more than the image-run basis does within the intervals. At rank 256 the answer position does not hold the prior apart from the image's evidence: the directions that carry one carry the other, which is what a single decision variable formed there predicts, and why removal at the answer position reverses decisions in both directions (Main-paper Fig.~5c).

\noindent\emph{The subspace energy as a predictor.} We refit the additive model on the clean, blur and noise grid with the energy of the image-token subspace at the image tokens added as a predictor (\cref{tab:energyfit}). On the two 336-pixel models that energy stands in for all or half of the precision term and for none of the prior. On LLaVA-NeXT it stands in for neither; there the energy does not rise with blur.

\begin{table}[h]\centering\footnotesize\setlength{\tabcolsep}{7pt}
\caption{\textbf{The additive model with the subspace energy added as a predictor.} Weights with 95\% intervals. Dash: not reported.}\label{tab:energyfit}
\begin{tabular}{lccc}\toprule
 & LLaVA-NeXT & LLaVA-1.5 & BakLLaVA \\ \midrule
Precision weight, without the energy & 0.08 [0.05, 0.11] & 0.29 [0.24, 0.34] & 0.29 [0.24, 0.34] \\
Precision weight, with the energy & 0.08 [0.05, 0.11] & 0.06 [$-0.04$, 0.16] & 0.16 [0.10, 0.23] \\
Prior weight, with the energy & -- & 0.21 [0.16, 0.28] & 0.14 [0.08, 0.19] \\ \bottomrule
\end{tabular}\end{table}

\begin{table}[h]\centering\footnotesize\setlength{\tabcolsep}{4pt}
\caption{\textbf{Bases at the answer position} (blocks 16, 20, 24; cluster-bootstrap 95\% intervals over images). LLaVA-1.5: 416 WHOOPS conflicts (307 image-wins, 109 prior-wins). BakLLaVA: 419 conflicts (315 image-wins, 104 prior-wins). LLaVA-NeXT: 424 conflicts (313 image-wins, 111 prior-wins). Each 256-D basis is set to its mean over image runs. Gap change is in logits toward the image's answer.}\label{tab:priorsub}
\begin{tabular}{lcccc}\toprule
Basis & Prior-wins reversed & Image-wins reversed & Gap change, prior-wins & Gap change, image-wins \\ \midrule
\multicolumn{5}{l}{\emph{LLaVA-1.5}} \\
Prior basis (text-only runs) & 23.9\% [16.3, 32.4] & 5.2\% [2.8, 7.8] & $+$0.94 [0.56, 1.29] & $-$0.38 [$-$0.58, $-$0.17] \\
Image-run basis & 23.9\% [16.4, 31.6] & 11.4\% [7.9, 15.1] & $+$0.76 [0.32, 1.18] & $-$1.00 [$-$1.25, $-$0.75] \\
Image-token subspace & 16.5\% [9.6, 23.6] & 5.9\% [3.4, 8.7] & $+$0.53 [0.20, 0.86] & $-$0.53 [$-$0.72, $-$0.35] \\
Next 256 text-run components & 8.3\% [3.0, 13.6] & 2.6\% [1.0, 4.3] & $+$0.27 [0.16, 0.38] & $-$0.11 [$-$0.17, $-$0.04] \\
Random inside image-run components & 8.3\% [3.1, 13.7] & 3.6\% [1.7, 5.6] & $+$0.21 [0.02, 0.40] & $-$0.40 [$-$0.51, $-$0.30] \\
Random basis & 4.6\% [0.9, 9.2] & 0.7\% [0.0, 1.6] & $+$0.30 [0.20, 0.39] & $-$0.08 [$-$0.14, $-$0.03] \\
Prior direction (1-D) & 0.0\% & 0.0\% & $+$0.01 [0.00, 0.02] & $-$0.02 [$-$0.03, $-$0.01] \\ \midrule
\multicolumn{5}{l}{\emph{BakLLaVA}} \\
Prior basis (text-only runs) & 21.2\% [13.9, 29.5] & 5.7\% [3.3, 8.5] & $+$1.20 [0.74, 1.67] & $-$0.52 [$-$0.80, $-$0.28] \\
Image-run basis & 19.2\% [12.4, 26.7] & 6.3\% [3.8, 9.3] & $+$1.00 [0.47, 1.49] & $-$0.80 [$-$1.05, $-$0.54] \\
Image-token subspace & 14.4\% [7.6, 21.2] & 7.0\% [4.3, 9.9] & $+$0.19 [$-$0.29, 0.69] & $-$0.54 [$-$0.78, $-$0.29] \\
Next 256 text-run components & 12.5\% [6.7, 19.3] & 6.0\% [3.7, 8.8] & $+$0.14 [$-$0.17, 0.43] & $-$0.15 [$-$0.30, $-$0.01] \\
Random inside image-run components & 16.3\% [9.8, 23.6] & 4.1\% [2.0, 6.5] & $+$0.49 [0.13, 0.82] & $-$0.28 [$-$0.44, $-$0.12] \\
Random basis & 4.8\% [1.0, 9.3] & 2.9\% [1.3, 4.9] & $+$0.18 [$-$0.03, 0.37] & $+$0.06 [$-$0.06, 0.18] \\
Prior direction (1-D) & 0.0\% & 0.6\% [0.0, 1.6] & $-$0.02 [$-$0.06, 0.01] & $+$0.03 [0.01, 0.04] \\ \midrule
\multicolumn{5}{l}{\emph{LLaVA-NeXT}} \\
Prior basis (text-only runs) & 33.3\% [24.7, 42.1] & 3.8\% [1.9, 6.1] & $+$1.70 [1.35, 2.07] & $-$0.09 [$-$0.29, 0.09] \\
Image-run basis & 25.2\% [17.5, 34.5] & 7.0\% [4.3, 9.9] & $+$1.24 [0.84, 1.62] & $-$0.76 [$-$0.96, $-$0.55] \\
Image-token subspace & 20.7\% [13.9, 28.7] & 7.3\% [4.5, 10.2] & $+$1.17 [0.84, 1.52] & $-$0.73 [$-$0.92, $-$0.56] \\
Next 256 text-run components & 8.1\% [3.5, 13.3] & 3.8\% [1.9, 6.1] & $+$0.37 [0.20, 0.54] & $-$0.18 [$-$0.26, $-$0.10] \\
Random inside image-run components & 11.7\% [6.2, 18.0] & 2.9\% [1.0, 5.0] & $+$0.65 [0.45, 0.88] & $-$0.22 [$-$0.33, $-$0.11] \\
Random basis & 7.2\% [2.7, 12.4] & 2.2\% [0.7, 4.0] & $+$0.27 [0.17, 0.37] & $-$0.10 [$-$0.16, $-$0.04] \\
Prior direction (1-D) & 0.0\% & 0.3\% [0.0, 1.0] & $+$0.02 [0.01, 0.04] & $-$0.01 [$-$0.02, 0.00] \\ \bottomrule
\end{tabular}\end{table}

\section{Mediation and encoder analysis}\label{app:mediation}
\noindent\textit{Background for main-paper Sec.~5. How much of each edit's effect passes through measured properties of the object.}\par\smallskip

\begin{figure}[h]\centering\includegraphics{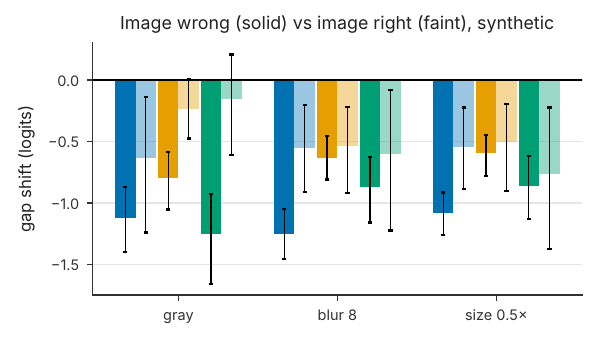}
\caption{Gap shift under the gray edit, blur at radius 8 and the object at half size, on WHOOPS rows where the image-supported answer is wrong (solid, $n\approx360$) and where it is right (faint, $n\approx60$), with bootstrap intervals. Blue, orange and green: LLaVA-NeXT, LLaVA-1.5 and BakLLaVA.}\end{figure}

Product-of-coefficients mediation with a cluster bootstrap over images: saturation acts $\approx20\%$ through concept overlap on all three models (and 12\% and 26\% through probe decodability on LLaVA-NeXT and BakLLaVA), blur 45\%, 8\% and 27\% through attention mass on the object (LLaVA-NeXT, LLaVA-1.5, BakLLaVA) (concept overlap, probe decodability and attention mass as defined in earlier work \cite{vankadaru2026trust}); direct effects remain. The CLIP encoder's generic representational change of the object (cosine and L2 between clean and edited pooled object features) does not predict the gap shift across seven edits (grayscale changes the encoder less than blur 8 yet moves the gap more); concept-specific change correlates positively but not significantly (Spearman 0.72, 0.36, 0.46, $n=7$).

\FloatBarrier
\noindent\textbf{Summary.} Discernibility changes the decision when it is manipulated but does not predict it across items: the object's area adds nothing to the model's own uncertainty in predicting which conflicts it will follow, and the mediation analysis above attributes at most half of each edit's effect to the object's concept overlap and attention mass. On the relational conflicts, steering attention onto the resized object lowered the margin, and steering onto its surroundings, which contain the comparison object, raised it on two of three models.

\section{Attention steering and localized edits}\label{app:steer}
\noindent\textit{Supports main-paper Secs.~5 and 9. Localized edits, the size dose, attention steering, head ablation under the earlier protocol and direction injection, on the LLaVA models.}\par\smallskip

\begin{figure}[h]\centering\includegraphics{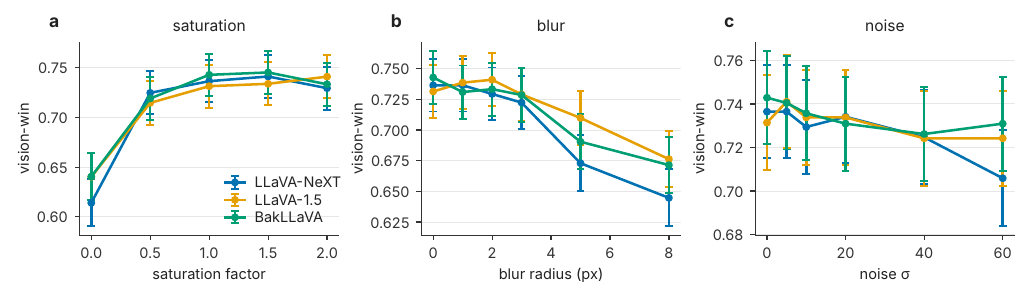}
\caption{Vision-win under saturation (a), blur (b) and noise (c) applied to the full-size image, on the three LLaVA models, with binomial intervals. Qwen2.5-VL is not plotted; its blur and noise rows are fit in Supp.~\ref{app:equation}.}\end{figure}

\begin{figure}[h]\centering\includegraphics{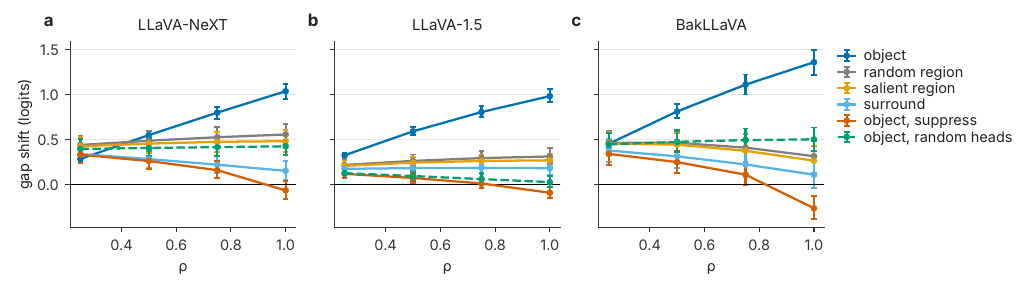}
\caption{Six steering arms by reallocated attention fraction $\rho$ on LLaVA-NeXT (a), LLaVA-1.5 (b) and BakLLaVA (c), with bootstrap intervals. Dashed: the object mask with random heads. The all-heads and head-scaling arms are not drawn. The unsteered margins of this run differ from those of the other runs (see text), so it is not used in the main paper.}\label{fig:steerarms}\end{figure}

\begin{table}[h]\centering\small
\caption{Attention steering on WHOOPS (same masks and conflicts as the object-level edits): gap shift toward the image by arm. Dash: not run.}\label{tab:steer}
\begin{tabular}{lcccc}\toprule
Model & Object, 20 heads & Object, 10 heads & Object, 5 heads & Random region \\ \midrule
LLaVA-NeXT & $+1.07$ & $+0.64$ & $+0.26$ & $+0.15$ \\
LLaVA-1.5 & $+0.82$ & $+0.45$ & $+0.20$ & $-0.08$ \\
BakLLaVA & $+1.45$ & $+0.85$ & $+0.40$ & $-0.07$ \\
Qwen2.5-VL & $+1.61$ & -- & -- & -- \\ \bottomrule
\end{tabular}\end{table}

\begin{figure}[h]\centering\includegraphics{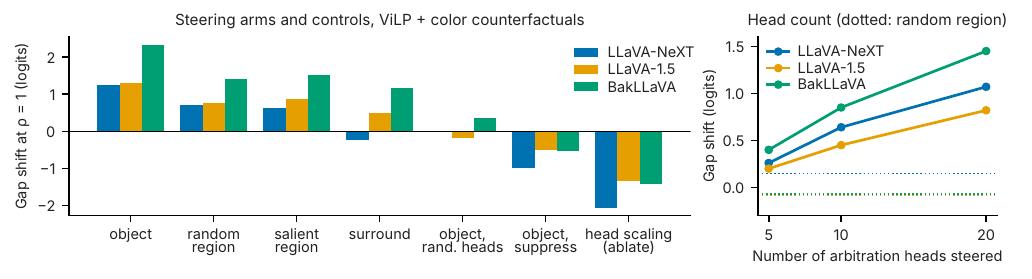}
\caption{Left: steering arms and controls at $\rho=1$ on ViLP and the color counterfactuals. Right: gap shift on WHOOPS when 5, 10 or 20 arbitration heads are steered onto the object; dotted lines: the random-region arm.}\label{fig:controls}\end{figure}

\begin{figure}[h]\centering\includegraphics{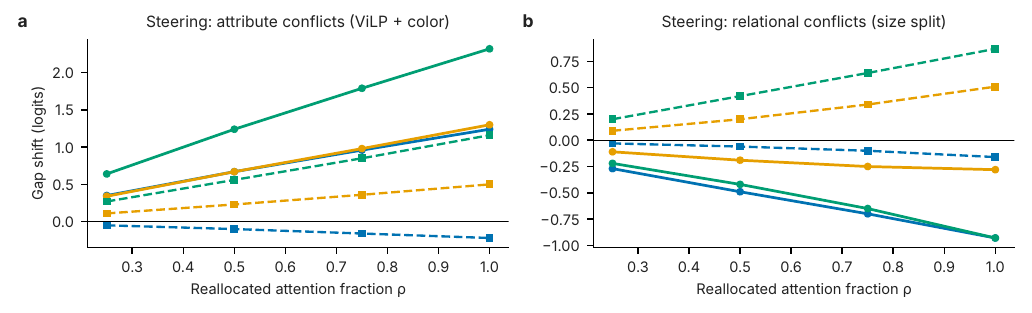}
\caption{Attention steering onto the object (solid) vs the surround (dashed) by dose, on attribute conflicts (a) and on the relational size split (b), where the object arm inverts. Blue, orange and green: LLaVA-NeXT, LLaVA-1.5 and BakLLaVA.}\label{fig:steerrel}\end{figure}

\noindent\textbf{Localized edits and the size dose (three LLaVA models and Qwen2.5-VL).} \Cref{tab:loc} gives the margin shifts under object-only and rest-only edits on WHOOPS conflicts with an object mask, and the size dose on the same items.

\begin{table}[h]\centering\footnotesize\setlength{\tabcolsep}{5pt}
\caption{\textbf{Localized edits and the size dose.} Margin shifts in logits against the original image, with 95\% intervals. Top: the object-only edit, the rest-only edit and their paired difference. Bottom: the object recomposited at another size, per item. Dash: not run.}\label{tab:loc}
\begin{tabular}{lcccc}\toprule
 & LLaVA-NeXT & LLaVA-1.5 & BakLLaVA & Qwen2.5-VL \\
Conflicts with a mask & 364 & 361 & 360 & 348 \\ \midrule
Gray, object & $-0.83$ & $-0.47$ & $-0.64$ & -- \\
Gray, rest & $-0.28$ & $-0.09$ & $-0.34$ & -- \\
Gray, object minus rest & $-0.56$ [$-0.80$, $-0.33$] & $-0.38$ [$-0.58$, $-0.18$] & $-0.30$ [$-0.59$, $-0.03$] & -- \\
Blur 8, object minus rest & $-0.53$ [$-0.78$, $-0.30$] & $-0.62$ [$-0.83$, $-0.42$] & $-0.69$ [$-0.96$, $-0.44$] & -- \\
Noise 60, object minus rest & $-0.30$ [$-0.43$, $-0.16$] & $-0.30$ [$-0.45$, $-0.17$] & $-0.30$ [$-0.49$, $-0.11$] & -- \\ \midrule
Half size: items whose margin falls (\%) & 77 & 63 & 62 & 65 \\
Half size: median shift & $-0.66$ & $-0.27$ & $-0.35$ & $-0.50$ \\
Half size: mean shift & $-1.08$ & $-0.60$ & $-0.86$ & $-0.77$ \\
1.25 times the size: mean shift & $-0.19$ [$-0.30$, $-0.10$] & $+0.01$ [$-0.07$, 0.08] & $-0.05$ [$-0.17$, 0.07] & -- \\
1.5 times the size: mean shift & $-0.34$ [$-0.48$, $-0.20$] & $-0.09$ [$-0.20$, 0.02] & $-0.23$ [$-0.41$, $-0.05$] & -- \\ \bottomrule
\end{tabular}\end{table}

\noindent\emph{Notes on \cref{tab:loc}.} Holm-corrected paired inside-against-outside contrasts over all levels of the localized edits are significant at corrected $p<0.05$ for blur 5/8 and noise 60 on all three models, for noise 40 on LLaVA-NeXT and LLaVA-1.5 (BakLLaVA $p=0.13$), and for grayscale on LLaVA-NeXT ($p<0.001$) and LLaVA-1.5 ($p=0.001$). They are not significant for grayscale on BakLLaVA ($p=0.14$) or for saturation 1.5/2.0 on any model, and noise 20 is significant only on LLaVA-1.5 ($p=0.04$). With a matched-area random region on LLaVA-1.5 (gray, WHOOPS, $n$ = 361) the shifts are $-0.47$ [$-0.64$, $-0.32$] for the object, $-0.09$ [$-0.22$, $+0.03$] for the rest and $-0.10$ [$-0.18$, $-0.04$] for a random region of the object's area. This region is a square of the object's area that may overlap the object by up to a quarter of that area, and the items are all WHOOPS items with a mask; the common protocol (Supp.~\ref{app:arch}) moves the object's own mask with under a tenth of overlap, on prior-valid conflicts, and its region elsewhere shifts the margin by at most 0.12.

\noindent\textbf{Attention steering, head ablation and injection.} The steering arms are the object (SAM3 mask), a random region (matched token count), a salient region (spectral-residual saliency), the surround (annulus), the object with random heads, object suppression, all heads, and head-output scaling (\cref{fig:controls} shows all but the all-heads arm). \Cref{tab:steerinj} gives the margin shifts.

\begin{table}[h]\centering\footnotesize\setlength{\tabcolsep}{6pt}
\caption{\textbf{Steering, head ablation and injection on the LLaVA models.} Margin shifts in logits unless a percentage is given. Steering is at full dose ($\rho=1$) with the same masks as \cref{tab:loc}. Head ablation is mean ablation of the counterfactual heads at the answer position, with decisions counted by the scoring rule of Supp.~\ref{app:data}. Injection is at $\alpha=0.5$. Dash: not reported here.}\label{tab:steerinj}
\begin{tabular}{lccc}\toprule
 & LLaVA-NeXT & LLaVA-1.5 & BakLLaVA \\ \midrule
\multicolumn{4}{l}{\emph{Attention steering, WHOOPS}} \\
Onto the object & $+1.07$ & $+0.82$ & $+1.45$ \\
Onto the object, 10 heads & $+0.64$ & $+0.45$ & $+0.85$ \\
Onto the object, 5 heads & $+0.26$ & $+0.20$ & $+0.40$ \\
Onto a random region & $+0.15$ & $-0.08$ & $-0.07$ \\
Onto the object, by mask confidence (two halves) & $+1.00$ vs $+1.13$ & $+0.70$ vs $+0.93$ & $+1.46$ vs $+1.44$ \\ \midrule
\multicolumn{4}{l}{\emph{Attention steering, other sources}} \\
ViLP and Visual CounterFact color, object & $+1.24$ & $+1.30$ & $+2.32$ \\
ViLP and Visual CounterFact color, random region & $+0.70$ & $+0.75$ & $+1.40$ \\
Visual CounterFact size, object & $-0.93$ & $-0.28$ & $-0.93$ \\ \midrule
\multicolumn{4}{l}{\emph{Head ablation: image-wins reversed (\%) and margin shift}} \\
Top 5 heads & 0.0\% ($-0.32$) & 0.3\% ($-0.16$) & -- \\
Top 10 heads & 4.2\% ($-0.88$) & 0.7\% ($-0.31$) & -- \\
Top 20 heads & 7.0\% ($-1.43$ [$-1.56$, $-1.30$]) & 1.3\% ($-0.95$ [$-1.07$, $-0.83$]) & -- \\ \midrule
\multicolumn{4}{l}{\emph{Injected direction}} \\
Answer-concept direction & $+2.61$ & $+1.98$ & $+2.30$ \\
Discriminant direction for the decision & $-0.01$ & $-0.13$ & $-0.07$ \\
Unrelated concept direction & $-0.05$ & $-0.11$ & $+0.04$ \\
Answer-concept, full / inside / outside the subspace & $+2.60$ / $+0.93$ / $+1.88$ & $+1.96$ / $+0.54$ / $+1.55$ & -- \\
Random-subspace split, inside / outside & $+0.19$ / $+2.47$ & $+0.16$ / $+1.85$ & -- \\ \bottomrule
\end{tabular}\end{table}

\noindent\emph{Notes on \cref{tab:steerinj}.} The steering values are from the factorial analysis of Supp.~\ref{app:equation}, whose unsteered margins equal those of the localized-edit analysis item by item. A separate run over doses gives the specificity contrasts (object against each control, $p<10^{-4}$ on the three LLaVA models), a monotone negative suppression arm, the all-heads arm, which breaks the model, and \cref{fig:steerarms}. The unsteered margins of that run differ from those of the other runs on the same items (LLaVA-1.5: mean $+1.03$ against $+2.81$, correlation 0.70; LLaVA-NeXT and BakLLaVA: correlation 0.97 and 0.96), so it is shown for completeness and is not used in the main paper. On the other sources the random regions are also positive because the object fills the frame, and grayscale baselines lower the gain (LLaVA-NeXT $+0.66$ at saturation 0). On the size split the surround is positive on LLaVA-1.5 and BakLLaVA, and the random region is positive on LLaVA-1.5 and BakLLaVA and null on LLaVA-NeXT ($-0.03$). For head ablation, layer-matched random heads reverse 0.3--0.7\% of image-wins (margin shift $-0.01$ to $-0.05$), and prior-wins flip on 0--2.7\% with a shift of $-0.07$ to $-0.68$. The common protocol of Supp.~\ref{app:arch}, with its own head selection and item sample, gives 10.5\% and a gap change of $-0.88$ on LLaVA-1.5, and \cref{tab:textid} counts reversals by the sign of the margin on the text-premise item sample and gives 13.3\%, 8.6\% and 16.3\%. The heads' write projection is 0.318/0.305 of the answer-position output norm inside the fitted 256-D subspace (LLaVA-NeXT / LLaVA-1.5), against 0.249/0.249 for a random 256-D subspace. The discriminant direction, fit to classify the decision on held-out conflicts, is inert. BakLLaVA's values for the injected directions come from the separate run described in Supp.~\ref{app:faq}. Qwen2.5-VL: object steering gives $+1.61$ at full dose, object steering with random heads $+0.30$ (19\% of the effect) and a random region $+0.18$, with no interaction with the discernibility measures of earlier work \cite{vankadaru2026trust}. Llama-3.2: measurement only (heads at cross-attention layers 3 to 38, most of them at 23 to 38; $r=0.45$; attention mass to the image correlates with the gap on ViLP, +0.29).

\noindent\textbf{Text-premise conflicts.} The image of each prior-valid WHOOPS conflict was replaced by one sentence asserting the image's answer, placed before the statement (``USER: \{stem\} \{image answer\}. ASSISTANT: \{stem\}''), with the same answer tokens. Per-head scores are the logit-lens contribution of each head's output to the image-answer-minus-expected-answer margin at the answer position, under the image condition and the text condition, by the direct logit attribution of main-paper Sec.~3; the top-20 heads of each condition are compared by Jaccard index against 1,000 layer-matched random draws and by the Spearman correlation of the full 32$\times$32 score matrices against within-layer permutations. Each set, and 10 layer-matched random sets, is mean-ablated at the answer position on the items of both conditions. The readout direction is the difference of class means of the residual at hidden states 16, 20 and 24 between items the premise or image wins and items the prior wins (testable only with at least 30 prior-wins per condition). Emergence is the first hidden state at which the logit-lens margin reaches half its final value. The pre-specified criterion for one shared circuit was a Jaccard index of at least 0.40 and a cross-ablation ratio of at least 0.5. \Cref{tab:textid} gives the results. The sets overlap five to ten times more than layer-matched chance, their score profiles correlate at 0.51 to 0.60, and each set's ablation lowers the other condition's margin more than every random set, asymmetrically: the text heads matter for image conflicts 1.6 to 5.7 times more than the image heads matter for text conflicts. Both measures fall below the pre-specified bar. BakLLaVA is classed as disjoint only by the direction criterion, which its 16 text prior-wins leave untested; without it the outcome is partial on all three models. The image heads recomputed here match the heads used in the main paper (Jaccard 0.90 to 1.00).

\begin{table}[h]\centering\footnotesize
\caption{\textbf{Text-premise conflicts against image conflicts} (WHOOPS; 416, 408, 412 items). Text-win: fraction of conflicts the premise (or image) answer wins. Jaccard: top-20 heads by score, with the layer-matched null mean and 99th percentile. Ablation rows: fraction of wins flipped and gap shift when a set is mean-ablated at the answer position; random: mean over 10 layer-matched sets (flips) or the most negative of them (gap). Cross ratio: flips by the other condition's heads over flips by the condition's own heads. Emergence: hidden-state index.}\label{tab:textid}
\setlength{\tabcolsep}{3pt}
\begin{tabular}{p{5.4cm}ccc}
\toprule
 & LLaVA-NeXT & LLaVA-1.5 & BakLLaVA \\
\midrule
Text-win: premise / no premise / image & 0.952 / 0.067 / 0.743 & 0.983 / 0.064 / 0.740 & 0.961 / 0.131 / 0.757 \\
Image heads recomputed against the main analysis (Jaccard) & 0.90 & 1.00 & 1.00 \\
Top-20 image-answer heads, Jaccard (null mean / p99) & 0.18 (0.04 / 0.11) & 0.33 (0.03 / 0.11) & 0.21 (0.04 / 0.11) \\
Top-20 expected-answer heads, Jaccard & 0.38 & 0.33 & 0.25 \\
Head-score Spearman (null p99) & 0.51 (0.10) & 0.60 (0.11) & 0.51 (0.09) \\
Text conflicts, flips: image heads / text heads / random & 0.005 / 0.083 / 0.001 & 0.007 / 0.062 / 0.002 & 0.028 / 0.149 / 0.008 \\
Text conflicts, gap: image heads / text heads / random & $-$0.29 / $-$2.79 / $-$0.25 & $-$0.50 / $-$2.59 / $-$0.14 & $-$0.49 / $-$3.25 / $-$0.31 \\
Image conflicts, flips: text heads / image heads / random & 0.045 / 0.133 / 0.011 & 0.033 / 0.086 / 0.011 & 0.051 / 0.163 / 0.018 \\
Image conflicts, gap: text heads / image heads / random & $-$0.52 / $-$1.33 / $-$0.10 & $-$0.41 / $-$0.83 / $-$0.20 & $-$0.54 / $-$1.75 / $-$0.03 \\
Cross ratio: image heads on text / text heads on image & 0.06 / 0.34 & 0.12 / 0.38 & 0.19 / 0.31 \\
Readout direction cosine, hidden states 16 / 20 / 24 (null p99 0.12--0.21) & 0.25 / 0.28 / 0.34 & 0.28 / 0.30 / 0.33 & 0.01 / 0.06 / 0.06 \\
Text prior-wins (direction needs 30) & 20 & 7 & 16 \\
Logit-lens emergence: text / image & 20 / 22 & 18 / 20 & 21 / 22 \\
Outcome (pre-specified rule) & partial & partial & disjoint (direction only) \\
\bottomrule
\end{tabular}
\end{table}

\FloatBarrier
\begin{figure}[h]\centering\includegraphics{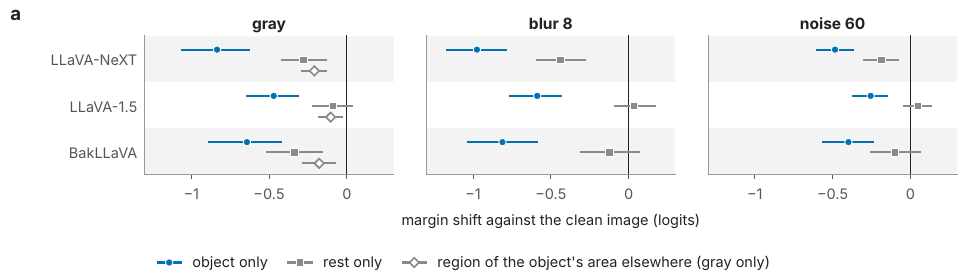}\\[2pt]
\includegraphics{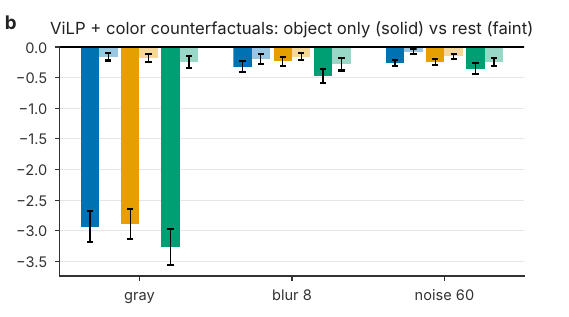}
\caption{\textbf{Object-only versus rest-only edits on the LLaVA models}. Margin shift against the clean image for gray, blur and noise edits at their strongest level (bootstrap 95\% intervals). (a) WHOOPS conflicts with an object mask ($n$ = 364/361/360); diamonds: a gray region of the object's area placed elsewhere (\cref{tab:matched}). (b) ViLP plus color counterfactuals; solid bars: object only, faint bars: rest only; blue, orange and green: LLaVA-NeXT, LLaVA-1.5 and BakLLaVA.}\label{fig:object_llava}\end{figure}

\begin{table}[h]\centering\footnotesize\setlength{\tabcolsep}{4pt}
\caption{\textbf{Matched-area control} (WHOOPS conflicts with an object mask; $n$ = 364/361/360). Margin shift in logits under the gray edit. The region elsewhere is a square of the object's area that may overlap the object by up to a quarter of that area. BakLLaVA: object-clustered bootstrap 95\% interval for its object row [$-0.87$, $-0.40$].}\label{tab:matched}
\begin{tabular}{p{5.2cm}ccc}\toprule
 & LLaVA-NeXT & LLaVA-1.5 & BakLLaVA \\ \midrule
Gray object & $-0.83$ & $-0.47$ & $-0.64$ \\
Gray rest & $-0.28$ & $-0.09$ & $-0.34$ \\
Gray region of the object's area elsewhere & $-0.21$ & $-0.10$ & $-0.18$ \\ \bottomrule
\end{tabular}\end{table}

\begin{table}[h]\centering\footnotesize\setlength{\tabcolsep}{6pt}
\caption{\textbf{Object-level effects and injections on the LLaVA models} (WHOOPS; edit rows on the $n$ = 364/361/360 conflicts with an object mask). Margin shift in logits against the clean image, except the head-signal row. The $1.0\times$ recomposite applies every pixel operation except the size change. Steering: 20 arbitration heads at $\rho=1$ with the SAM3 object mask or a random region of matched token count. Injection: a direction added at the image tokens at strength $\alpha$ relative to the local residual norm. Bracketed values are bootstrap 95\% CIs. Per-item spread of the size rows is in the text above.}\label{tab:llavaobj}
\begin{tabular}{p{7.4cm}ccc}\toprule
 & LLaVA-NeXT & LLaVA-1.5 & BakLLaVA \\ \midrule
Gray object & $-0.83$ & $-0.47$ & $-0.64$ \\
Gray rest of the image & $-0.28$ & $-0.09$ & $-0.34$ \\
Object at $0.5\times$ [CI] & $-1.08$ [$-1.26$, $-0.91$] & $-0.60$ [$-0.78$, $-0.43$] & $-0.86$ [$-1.11$, $-0.62$] \\
Object at $0.75\times$ & $-0.55$ & $-0.17$ & $-0.25$ \\
Recomposite at $1.0\times$ (no size change) & $-0.01$ & $-0.01$ & $-0.00$ \\
Attention steering onto the object & $+1.07$ & $+0.82$ & $+1.45$ \\
Attention steering onto a random region & $+0.15$ & $-0.08$ & $-0.07$ \\
Head signal $r$ with vision-win (all sources) & $+0.61$ & $+0.58$ & $+0.65$ \\ \midrule
Answer-concept direction injected, $\alpha{=}0.5$ & $+2.61$ & $+1.98$ & $+2.30$ \\
Decoding direction injected, $\alpha{=}0.5$ & $-0.01$ & $-0.13$ & $-0.07$ \\
Concept injection $\alpha{=}1$ & $+4.0$ & $+2.4$ & $+1.4$ \\ \bottomrule
\end{tabular}\end{table}

\begin{table}[h]\centering\footnotesize\setlength{\tabcolsep}{4pt}
\caption{\textbf{ViLP plus color counterfactuals} ($n\approx970$). Gap shift in logits under the gray edit and under attention steering at full dose. Random regions are also positive under steering because the object fills the frame.}\label{tab:vilp}
\begin{tabular}{p{5.2cm}ccc}\toprule
 & LLaVA-NeXT & LLaVA-1.5 & BakLLaVA \\ \midrule
Gray object & $-2.9$ & $-2.9$ & $-3.3$ \\
Gray rest of the image & $-0.2$ & $-0.2$ & $-0.2$ \\
Steering onto the object & $+1.24$ & $+1.30$ & $+2.32$ \\
Steering onto a random region & $+0.70$ & $+0.75$ & $+1.40$ \\ \bottomrule
\end{tabular}\end{table}

\noindent\textbf{Summary.} On the mask-matched subset, where a random region of the object's own area can be drawn, graying the object moved the margin 1.9 to 5.2 times as much as graying the rest (\cref{tab:matched}). On the color conflicts, where graying the object removes the conflicting color itself, the object edit was 14 to 18 times larger than the rest edit (\cref{tab:vilp}). The head-ablation rates (7.0\% and 1.3\% of image-wins on LLaVA-NeXT and LLaVA-1.5) are below the 68--96\% that \citet{lietzow2026visiondefault} report for prior-grounded predictions under a prompt contrast, a different measurement on a complementary set of heads; fewer heads flip fewer, and prior-wins flip at 0 to 2.7\%. In the text-premise test, the text heads flip 31--38\% as many image-wins as the image heads, and the image heads 6--19\% as many text-wins as the text heads.

\FloatBarrier
\section{Additive fit of edit effects and prediction across sources}\label{app:equation}
\noindent\textit{Supports main-paper Sec.~4. The first-stage additive fit on the three LLaVA models and Qwen2.5-VL, and the test that changes the prior within an item.}\par\smallskip
The additive fit $\Delta\text{gap} = c_1\,\text{gray} + c_2\,\text{blur} + c_3\,\text{small} + c_4\,\rho_{\text{obj}} + c_5\,\alpha + c_0$ is estimated on half the WHOOPS examples and scored by $R^2$ on the other half (\cref{tab:addcoef}). Interactions ($p<0.01$, all models): edit $\times$ injection is positive or zero.

\begin{table}[h]\centering\footnotesize\setlength{\tabcolsep}{7pt}
\caption{\textbf{Additive fit of edit effects.} Coefficients of the fit on half the WHOOPS examples, $R^2$ on the other half, and the attention $\times$ injection interaction. Dash: not reported.}\label{tab:addcoef}
\begin{tabular}{lcccccccc}\toprule
Model & gray & blur & small & $\rho_{\text{obj}}$ & $\alpha$ & intercept & $R^2$ & attention $\times$ injection \\ \midrule
LLaVA-NeXT & $-0.64$ & $-0.86$ & $-0.66$ & $+1.20$ & $+5.21$ & $-0.41$ & 0.40 & $+2.7$ \\
LLaVA-1.5 & $-0.36$ & $-0.38$ & $-0.23$ & $+0.94$ & $+3.67$ & -- & 0.27 & $+2.9$ \\
BakLLaVA & $-0.02$ & $-0.32$ & $-0.21$ & $+2.17$ & $+4.86$ & -- & 0.19 & $+7.7$ \\ \bottomrule
\end{tabular}\end{table}

\noindent The equation does not predict effect magnitudes on other sources (Spearman 0.12): grayscale on color counterfactuals is $-2.9$ against a predicted $-0.6$, and the object term reverses on the size split.

\noindent\textbf{Cue-combination fit on the full-size grid (three LLaVA models and Qwen2.5-VL).} \emph{Items and grid.} Items: WHOOPS conflicts after the prior filter (425, 417, 420 for LLaVA-NeXT, LLaVA-1.5, BakLLaVA) (clean vision-win on LLaVA-NeXT: 0.736, 313 of 425), blur $\sigma\in\{0,1,2,3,5,8\}$ and noise $\sigma_n\in\{0,5,10,20,40,60\}$, one clean row per item; outcome: an image-win (the image-supported answer scores above the prior-supported answer); precision proxy $\tau = 1/(1+\sigma^2)$ for blur and $1/(1+(\sigma_n/8)^2)$ for noise; prior strength $\pi$ = the negative text-only margin, centred at the pooled median (4.74 nats), tertile edges per model.

\noindent\emph{Fit and protocol.} Fit: logistic regression with cluster-robust covariance (cluster = image). Protocol: items split 70/30 with seed 42, shared across models; boundary cells (clean image in every tertile, every dose in the middle tertile) of the training items fit the model and interior cells (dosed levels in the outer tertiles) of the held-out items score it; no interior test cell has fewer than 37 items. Null: 200 label permutations. Bootstrap: 2,000 resamples of training images, shared draws across models.

\noindent\emph{Pooled fit.} \Cref{tab:cuefit} gives the per-model fits; the pooled fit with shared slopes and per-model intercepts has $a$ = 0.081 [0.033, 0.129], $b$ = 0.210 [0.120, 0.300], and the likelihood-ratio test of per-model slopes against shared slopes gives 7.99 on 4 degrees of freedom ($p$ = 0.092). Pairwise equivalence tests at a 20\% relative margin (90\% intervals) are inconclusive for every pair and parameter; the half-widths are 0.4 to 1.5 relative units against the 0.2 margin.

\noindent\emph{Alternative proxies and attribute type.} Alternative proxies move $a$ little (within-family standardised $\log\tau$: 0.185, 0.087, 0.090; all five families with family terms: 0.116, 0.096, 0.085; $\sigma_0$ = 4 or 16 in the noise proxy: $\pm$0.01). Attribute type (color, count, size, other), added to the prior-only model on the clean rows of all four sources, improves the fit on every model: likelihood ratio $p<10^{-4}$ with deviance gains of 0.061, 0.148, 0.028 and 0.078 (Qwen2.5-VL) and at least two category coefficients of 0.30 logits or more with intervals excluding zero; adding the source on top of attribute type gives $p$ = 0.30, 0.003 and 0.008 (deviance gains under 0.015); source is confounded with attribute type and we do not interpret it.

\noindent\emph{Qwen2.5-VL} is fit on the run of Supp.~\ref{app:faq} that fixes the library version (407 prior-valid WHOOPS conflicts, 4,070 dose rows), with the same specification and centring but its own 70/30 split (seed 42 over its items; prior tertile edges 2.84 and 5.05 nats, median 3.94) and a permutation null that refits on permuted training labels (95th percentile 0.045, wider than the null of the other rows). Over 50 splits its $a$ and $b$ have medians 0.165 and 0.181 (ranges 0.11 to 0.22 and 0.11 to 0.25), both positive on every split, and held-out deviance explained ranges from 0.01 to 0.15 (median 0.09).

\noindent\emph{Scope.} The pooled fit and the pairwise tests above cover the three LLaVA models. The band-readout regression below uses the per-model linear predictor $\eta$ of these fits. The second stage of the pre-specified analysis (a per-item CLIP evidence margin on the object crop in place of the level proxy, five prior paraphrases, and the prior manipulation by prefix) is reported below under measured inputs and changing the prior within an item.

\begin{table}[h]\centering\footnotesize
\caption{\textbf{Cue-combination fit per model} (boundary-to-interior protocol). $a$ per unit $\log\tau$, $b$ per nat of prior strength, $c$ the clean-image log-odds at the median prior; 95\% cluster-robust intervals. $n$ = fitted rows / image clusters. Held out: deviance explained on the interior cells of the held-out items against the permutation null (95th percentile); in-distribution: the item-disjoint split over all cells. $\Delta$int.: change in held-out deviance explained when a dose-by-prior interaction is added. $^*$Own item split and permutation null (see text).}\label{tab:cuefit}
\setlength{\tabcolsep}{3pt}
\begin{tabular}{lccccccccc}
\toprule
Model & $a$ & $b$ & $c$ & $n$ & held out & null & in-dist. & AUROC & $\Delta$int. \\
\midrule
LLaVA-NeXT & 0.127 [0.049, 0.206] & 0.163 [0.031, 0.295] & 1.24 [0.91, 1.57] & 1206/296 & 0.126 & 0.002 & 0.070 & 0.71 & $-$0.222 \\
LLaVA-1.5 & 0.060 [$-$0.012, 0.132] & 0.310 [0.150, 0.470] & 1.15 [0.80, 1.50] & 1266/286 & $-$0.003 & 0.001 & 0.046 & 0.62 & $-$0.284 \\
BakLLaVA & 0.062 [$-$0.015, 0.139] & 0.181 [0.069, 0.292] & 1.29 [0.94, 1.64] & 1171/291 & 0.098 & 0.003 & 0.057 & 0.67 & $-$0.279 \\
Qwen2.5-VL$^*$ & 0.126 [0.056, 0.196] & 0.140 [0.018, 0.262] & 1.27 [0.94, 1.61] & 1215/285 & 0.091 & 0.045 & 0.076 & 0.70 & $-$0.087 \\
\bottomrule
\end{tabular}
\end{table}

\noindent\textbf{Precision-proxy check.} Per family, the level effects of a model with level dummies (prior slope fixed) are compared with $\log\tau$ by Spearman correlation: LLaVA-NeXT blur 0.94, noise 0.94; BakLLaVA 0.94, 0.94; Qwen2.5-VL 0.94, 1.00; LLaVA-1.5 blur 0.77, noise 0.60. On LLaVA-1.5 the vision-win rate rises by 0.01 at blur $\sigma$ = 1 and 2 (0.731 to 0.739 and 0.741; 0.04 to 0.05 in log-odds) and by 0.00 to 0.01 at noise 5 to 20 before it falls, which is why the additive-model fitted on its boundary cells does not predict its interior cells.

\noindent\textbf{Band readout against the fit.} Per model, the centred logit-lens margin at blocks 20 to 23 (mean over the band) is regressed on $\eta$ with five item folds. On the clean image the dose term is constant, so $\eta$ is affine in the text-only margin and the regression measures how much of the readout the prior term explains: held-out $R^2$ = 0.223, 0.244, 0.250 (LLaVA-NeXT, LLaVA-1.5, BakLLaVA), peaking at 0.267 to 0.313 (block 28 on LLaVA-NeXT and BakLLaVA, block 31 on LLaVA-1.5), against a label-shuffle null of 0.003. On the four conditions that vary the dose (clean, grayscale, blur 8, half size), the block-20 concept readout gives held-out $R^2$ = 0.099, 0.099, 0.067 on $\eta$ from the fit with all five edit families, with a partial $R^2$ beyond the prior term of 0.002 or less. On NaturalBench natural conflicts, with the WHOOPS slopes transferred, $R^2$ = 0.01 to 0.03.

\noindent\textbf{Measured inputs.} Two inputs of the fit were then measured on the WHOOPS conflicts. Prior: the text-only margin under five fixed phrasings (the original and the prefixes ``Common knowledge:'', ``Fact:'', ``It is well known that'', ``In general,''); the original reproduces the prior of the main fit (Pearson 1.000), a single phrasing has ICC 0.93, 0.95, 0.91 (LLaVA-NeXT, LLaVA-1.5, BakLLaVA), and a slope fitted on one phrasing is therefore attenuated by a factor of 0.93, 0.95, 0.91. Image cue: the margin of CLIP ViT-L/14-336, the models' own encoder, between ``a photo of a'' plus each candidate answer on the object's SAM3 bounding box padded by 10\%, at every blur, noise and size level, entered as $\log\max(|\text{margin}|, 0.05)$. On the clean image the crop margin favors the image answer on 81\% of 418 conflicts (CLIP ViT-B/32: 80\%); it falls monotonically with blur and noise (Spearman of the level means $-1.00$, $-0.94$) but far less than the closed-form $\tau$: at blur 8 it keeps 0.75 of its clean mean (closed form 0.015), at noise 60 0.86 (0.017). \cref{tab:cuefit2} refits the additive model with each input replaced in turn, on the WHOOPS conflicts that have both measurements. The prior weight excludes zero in every specification but one (LLaVA-1.5 with the five-phrasing prior and the closed-form $\tau$). The precision weight excludes zero on BakLLaVA under all three, on LLaVA-NeXT under all three, and on LLaVA-1.5 under none. With the measured image cue, held-out deviance explained falls to $-0.08$ to 0.07, below the level-based fits, and per-model slopes differ ($p$ = 0.005 with both inputs measured, driven by LLaVA-1.5's flat precision weight; $p$ = 0.09 with the closed-form $\tau$). No specification gains from a dose-by-prior interaction (gain at most 0.006). On the clean image the band readout gains at most 0.015 in held-out $R^2$ from the measured image cue beyond the prior.

\begin{table}[h]\centering\footnotesize
\caption{\textbf{Cue-combination fit with measured inputs} (boundary-to-interior protocol, WHOOPS). Prior: the five-phrasing mean (P5) or the single phrasing of \cref{tab:cuefit}. Image cue: the closed-form $\tau$ of the main text, or the measured CLIP crop margin, whose $a$ is per unit of $\log|\text{margin}|$ and not comparable in size with the closed-form $a$. Held out: deviance explained on the interior cells of the held-out items (permutation-null 95th percentile 0.002 to 0.004). $\Delta$int. as in \cref{tab:cuefit}.}\label{tab:cuefit2}
\setlength{\tabcolsep}{3pt}
\begin{tabular}{llccccc}
\toprule
Inputs & Model & $a$ & $b$ & $n$ & held out & $\Delta$int. \\
\midrule
P5, closed-form $\tau$ & LLaVA-NeXT & 0.128 [0.043, 0.212] & 0.244 [0.088, 0.401] & 1286/296 & 0.145 & $-$0.021 \\
 & LLaVA-1.5 & 0.034 [$-$0.034, 0.101] & 0.095 [$-$0.027, 0.217] & 1186/286 & 0.057 & $-$1.044 \\
 & BakLLaVA & 0.078 [0.007, 0.149] & 0.162 [0.050, 0.274] & 1191/291 & 0.088 & $-$0.298 \\
Single, measured cue & LLaVA-NeXT & 0.409 [0.148, 0.671] & 0.238 [0.084, 0.393] & 1049/249 & 0.044 & $+$0.002 \\
 & LLaVA-1.5 & 0.047 [$-$0.235, 0.329] & 0.392 [0.205, 0.580] & 1074/244 & $-$0.079 & $-$0.008 \\
 & BakLLaVA & 0.390 [0.151, 0.629] & 0.245 [0.132, 0.358] & 966/246 & 0.048 & $+$0.001 \\
P5, measured cue & LLaVA-NeXT & 0.369 [0.131, 0.607] & 0.312 [0.137, 0.487] & 1119/249 & 0.074 & $-$0.017 \\
 & LLaVA-1.5 & 0.066 [$-$0.191, 0.323] & 0.194 [0.046, 0.342] & 1004/244 & 0.015 & $+$0.006 \\
 & BakLLaVA & 0.309 [0.074, 0.544] & 0.225 [0.097, 0.353] & 986/246 & 0.018 & $+$0.000 \\
\bottomrule
\end{tabular}
\end{table}

\noindent\textbf{Corpus counts and the prior.} For each WHOOPS conflict we counted, with the infini-gram engine \cite{liu2024infinigram} on Dolma v1.7 \cite{soldaini2024dolma} (2.6T Llama-2 tokens in the infini-gram index), the exact occurrences of the longest suffix of the statement seen at least 50 times followed by the expected answer, and of the same suffix followed by the image's answer (median suffix four words). The predictor is the log ratio of the two counts. For the color conflicts of Visual CounterFact the predictor is the log ratio, over the two answers, of the counts of ``\{color\} \{object\}'' plus ``\{object\} is \{color\}''. The prior is the negative text-only margin of each of the five models. The controlled correlation is a partial Spearman correlation with the log ratio of the two answers' unigram frequencies and, on WHOOPS, attribute type held fixed. \Cref{tab:h6} gives the results. With a two-word context the uncontrolled WHOOPS correlation is 0.43 to 0.48; against the five-phrasing prior of the measured-inputs fit it is 0.39 to 0.42 (controlled 0.33 to 0.37). On the size conflicts, with an adjective proxy, the controlled correlation is 0.06 to 0.14 (exploratory). On RedPajama \cite{weber2024redpajama} (1.4T tokens in the infini-gram index) the controlled WHOOPS correlation is 0.34 to 0.42 (lowest on BakLLaVA, 0.34 [0.26, 0.42]) and the color correlation 0.07 to 0.18. The last column of \cref{tab:h6} regresses the override (prior-win) indicator on the standardized corpus ratio without and with the prior: the corpus ratio's effect on the override falls to a direct effect whose interval includes zero on four of five models in Dolma (Llama~3.2-Vision 0.29 [0.01, 0.57]) and on all five in RedPajama. Corpus counts of the statement's context predict the prior on all five models, and most of their effect on the override passes through the prior. The pre-specified color test (controlled correlation of at least 0.3) fails. Neither corpus is the training data of these models' language models; both stand in for web text of the same kind.

\begin{table}[h]\centering\footnotesize
\caption{\textbf{Corpus counts and the prior} (Dolma v1.7, exact counts). $\rho$: Spearman correlation of the corpus log ratio with the prior; controlled: partial correlation with answer frequency (and attribute type on WHOOPS) held fixed, 95\% bootstrap intervals. Override: logistic coefficient of the prior-win indicator per standard deviation of the corpus ratio, without (total) and with (direct) the prior in the model. RedPajama values are in the text.}\label{tab:h6}
\setlength{\tabcolsep}{4pt}
\begin{tabular}{lcccc}
\toprule
Model & WHOOPS $\rho$ & WHOOPS, controlled & Color, controlled & Override total / direct \\
\midrule
LLaVA-NeXT & 0.53 & 0.47 [0.40, 0.54] & 0.16 [0.07, 0.26] & 0.40 / $-$0.00 \\
LLaVA-1.5 & 0.54 & 0.47 [0.39, 0.53] & 0.12 [0.03, 0.21] & 0.45 / 0.14 \\
BakLLaVA & 0.51 & 0.42 [0.35, 0.50] & 0.16 [0.07, 0.25] & 0.37 / 0.15 \\
Qwen2.5-VL & 0.51 & 0.44 [0.37, 0.50] & 0.17 [0.08, 0.26] & 0.46 / 0.21 \\
Llama~3.2-Vision & 0.50 & 0.44 [0.36, 0.51] & 0.17 [0.09, 0.24] & 0.55 / 0.29 \\
\bottomrule
\end{tabular}
\end{table}

\noindent\textbf{Changing the prior within an item.} Three prefixes were placed before the statement, identically in the text-only and the image prompt: strengthen, ``Normally, \{stem\} \{expected answer\}.''; generic strengthen, ``This image shows an ordinary, typical scene.''; weaken, ``This image shows something unusual and unexpected.''. The text-only margin was measured under each prefix with the same scorer as the unprefixed prior; without a prefix it reproduces the prior of the main fit (Pearson 1.0000 on all three models), and the image scorer reproduces its dose grid ($r$ = 1.0000). The strengthen and weaken prefixes moved the prior in their intended direction on all three models (\cref{tab:c5}); the generic prefix met the validity rule only on LLaVA-NeXT (median shift $+$0.75 nats, 72\% of items in the intended direction; $+$0.13 and $+$0.41 nats on LLaVA-1.5 and BakLLaVA) and is used only there.  Model: on every row (each item under no prefix and each valid prefix, on the clean image and the blur and noise doses), a Mundlak logistic regression $\operatorname{logit} P(\text{vision win}) = a\log\tau - b_w(\pi - \bar\pi_i) - b_b\,\bar\pi_i + c + \text{family}$, with $\pi$ the prior measured under the row's prefix and $\bar\pi_i$ its mean over the item's prefixes; $b_w$ is the within-item weight. The pre-specified test compares $b_w$ on the strengthen and weaken rows with the stage-1 $b$ of \cref{tab:cuefit} by two one-sided tests at a 20\% margin on the relative difference (difference over the mean of the two, 90\% interval): EQUIVALENT if the interval lies inside $\pm$0.2, DIFFERENT if it lies outside, otherwise INCONCLUSIVE. An item bootstrap refits both slopes on the same resampled items. Secondary analyses add one indicator per prefix (its direct effect beyond the prior shift it causes), restrict to the prefixes that do not name an answer, compare the observed change in vision-win with the change the stage-1 parameters predict from the measured prior shift, test the $\log\tau \times (\pi-\bar\pi_i)$ interaction, and score the stage-1 boundary-to-interior protocol on the prefixed rows.

\Cref{tab:c5} gives the results. The pre-specified test fails: $b_w$ is 1.7 to 3.0 times $b$, DIFFERENT on LLaVA-NeXT and INCONCLUSIVE on the other two, and DIFFERENT on all three by the item bootstrap. The strengthen prefix, which states the expected answer, lowers the vision-win log-odds by 1.8 to 2.0 beyond its effect through the prior on all three models; the weaken prefix has a direct effect of $-$0.39 on LLaVA-1.5 and none on the others. With these terms $b_w$ is 0.23, 0.32 and 0.14 against $b$ = 0.16, 0.31 and 0.18. On the prefixes that do not name an answer alone, $b_w$ is 0.09 to 0.13, below $b$ (DIFFERENT on LLaVA-1.5, INCONCLUSIVE on the others), and the observed changes in vision-win are close to the predicted ones on LLaVA-NeXT and BakLLaVA and below them on LLaVA-1.5. The strengthen prefix lowers vision-win 2.5 to 4.3 times as much as its prior shift predicts. The interaction is $-$0.009 [$-$0.017, $-$0.002] on LLaVA-NeXT, about a fiftieth of its $b_w$, and indistinguishable from zero on the others. The stage-1 parameters, fit with no prefix, predict the prefixed interior cells (pseudo-$R^2$ 0.14 to 0.27), and the same protocol with the unshifted prior fed in scores below zero; on the non-naming prefixes the calibration gap is at most 0.04. Manipulating the prior moves the decision in the predicted direction, and the first-stage fit predicts the manipulated items; the within-item weight is not shown to equal the between-item weight, and an answer stated in the prompt acts on the decision beyond the prior it induces, as the text-premise conflicts of Supp.~\ref{app:steer} also show.

\begin{table}[h]\centering\footnotesize
\caption{\textbf{Changing the prior within an item} (WHOOPS; 425, 417, 420 items). Prior shift: median change of the text-only margin in nats (fraction of items in the intended direction). $b_w$: within-item weight with its standard error. Relative difference: $(b_w - b)/\text{mean}(b_w, b)$ with its 90\% interval. Pseudo-$R^2$ on the interior cells of the held-out items (boundary-to-interior protocol): refit on the prefixed rows / unshifted prior fed in / stage-1 parameters.}\label{tab:c5}
\setlength{\tabcolsep}{3pt}
\begin{tabular}{p{5.4cm}ccc}
\toprule
 & LLaVA-NeXT & LLaVA-1.5 & BakLLaVA \\
\midrule
Prior shift, strengthen & $+$3.65 (0.90) & $+$3.29 (0.91) & $+$5.84 (0.92) \\
Prior shift, generic strengthen & $+$0.75 (0.72) & $+$0.13 (0.56), not valid & $+$0.41 (0.62), not valid \\
Prior shift, weaken & $-$1.31 (0.80) & $-$1.67 (0.85) & $-$1.81 (0.78) \\
Vision-win on the clean image: none / strengthen / weaken & 0.74 / 0.19 / 0.79 & 0.73 / 0.16 / 0.76 & 0.74 / 0.17 / 0.79 \\
Stage-1 $b$ & 0.163 (0.067) & 0.310 (0.082) & 0.181 (0.057) \\
$b_w$, strengthen and weaken & 0.481 (0.030) & 0.539 (0.033) & 0.329 (0.021) \\
Relative difference, pre-specified test & $+$0.99 [0.47, 1.51] & $+$0.54 [0.13, 0.95] & $+$0.58 [0.10, 1.07] \\
Outcome: two one-sided tests / item bootstrap & DIFFERENT / DIFFERENT & INCONCLUSIVE / DIFFERENT & INCONCLUSIVE / DIFFERENT \\
Strengthen prefix, direct effect (logits) & $-$1.82 [$-$2.14, $-$1.50] & $-$1.92 [$-$2.25, $-$1.59] & $-$2.00 [$-$2.34, $-$1.65] \\
$b_w$ with one indicator per prefix & 0.231 (0.035) & 0.315 (0.040) & 0.136 (0.026) \\
$b_w$, prefixes that do not name an answer & 0.131 (0.021) & 0.115 (0.027) & 0.088 (0.021) \\
Relative difference, non-naming prefixes & $-$0.22 [$-$0.94, 0.50] & $-$0.92 [$-$1.37, $-$0.46] & $-$0.69 [$-$1.27, $-$0.11] \\
Change in vision-win, weaken: observed / predicted & $+$0.058 / $+$0.045 & $+$0.034 / $+$0.091 & $+$0.047 / $+$0.061 \\
Change in vision-win, generic: observed / predicted & $-$0.029 / $-$0.027 & -- & -- \\
Change in vision-win, strengthen: observed / predicted & $-$0.54 / $-$0.13 & $-$0.57 / $-$0.22 & $-$0.56 / $-$0.23 \\
$\log\tau \times$ within-item prior interaction & $-$0.009 [$-$0.017, $-$0.002] & $-$0.002 [$-$0.010, 0.006] & $-$0.001 [$-$0.006, 0.005] \\
Pseudo-$R^2$, strengthen and weaken & 0.23 / $-$0.04 / 0.14 & 0.25 / $-$0.10 / 0.27 & 0.25 / $-$0.04 / 0.23 \\
Pseudo-$R^2$, non-naming prefixes (calibration gap) & 0.13 / 0.07 / 0.12 (0.04) & 0.14 / 0.01 / 0.14 (0.03) & 0.12 / 0.06 / 0.14 (0.04) \\
\bottomrule
\end{tabular}
\end{table}

\FloatBarrier
\begin{figure}[h]\centering\includegraphics{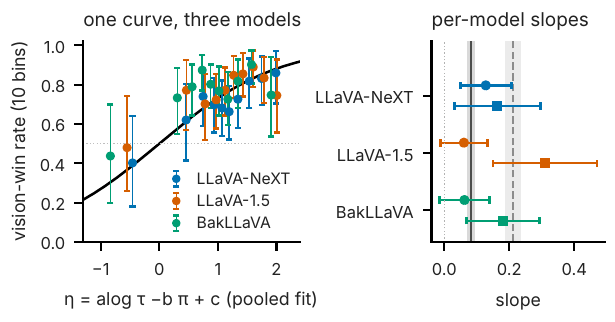}
\caption{\textbf{The additive model on the three LLaVA models, full-size dose grid}. Left: observed vision-win rate in ten quantile bins of the pooled linear predictor $\eta = a\log\tau - b\,\pi + c$ on the WHOOPS blur and noise doses, with the logistic curve; cluster-bootstrap 95\% intervals. Right: per-model slopes $a$ (precision, circles) and $b$ (prior, squares) with 95\% intervals; the solid and the dashed vertical line are the pooled $a$ and $b$, each with a grey band of $\pm$10\%.}\label{fig:rule_llava}\end{figure}

\noindent\textbf{Summary: full-size grid.} With blur and noise applied to the full-size image, the prior slope is positive on all three LLaVA models ($b$ = 0.16 [0.03, 0.30], 0.31 [0.15, 0.47], 0.18 [0.07, 0.29] per nat on LLaVA-NeXT, LLaVA-1.5, BakLLaVA); the precision slope excludes zero on LLaVA-NeXT ($a$ = 0.13 [0.05, 0.21]) and not on the two 336-pixel models (LLaVA-1.5 and BakLLaVA) (0.06 [$-$0.01, 0.13], 0.06 [$-$0.02, 0.14]). Held-out deviance explained is 0.13 and 0.10 on LLaVA-NeXT and BakLLaVA against a permutation null of 0.002, and zero on LLaVA-1.5 in that split. Qwen2.5-VL gives $a$ = 0.13 [0.06, 0.20], $b$ = 0.14 [0.02, 0.26] and 0.09 held out. Per-model and shared slopes (pooled $a$ = 0.08, $b$ = 0.21) are not separated ($p$ = 0.09) nor shown equivalent. Attribute type adds to the prior on all four models. The layer-20 to 23 readout tracks the prior term of the fit ($R^2$ = 0.22 to 0.25 held out) and nothing beyond it. These values are from one 70/30 item split. Under the common protocol of Supp.~\ref{app:arch}, 50 splits of the same LLaVA-1.5 data give $b$ from 0.17 to 0.39 (median 0.25) and held-out deviance explained from $-$0.08 to 0.19, so the single-split values above lie inside the split-to-split range; the main text reports the common-protocol fits. The weak precision weight on the two 336-pixel models (LLaVA-1.5 and BakLLaVA) is a property of this grid: applied after resizing to the 336-pixel input, the same levels give $a$ = 0.31 and 0.31 (\cref{tab:archfit}).

\noindent\textbf{Changing the prior within an item.} We changed each WHOOPS conflict's prior with a prefix that states the expected answer (``Normally, the \emph{object} is \emph{expected}.'') or calls the image unusual, and measured the margin under the same prefix. Vision-win moved in the predicted direction on all three models (from 0.73--0.74 to 0.16--0.19 and 0.76--0.79), and the fit made without prefixes predicts the prefixed items (pseudo-$R^2$ 0.14 to 0.27). The pre-specified test that the within-item weight equals $b$ within 20\% fails (0.48, 0.54, 0.33): the stated answer lowers the decision by a further 1.8 to 2.0 logits beyond its effect on the margin; with that term the weight is 0.23, 0.32, 0.14, and the prefixes that name no answer give 0.09 to 0.13. The prior moves the decision in the direction the additive model predicts; its within-item weight depends on how the prior is changed.

\section{Cross-model alignment, transferred directions and the projector-only checkpoint}\label{app:portable}
\noindent\textit{Background for main-paper Sec.~7. Whether subspaces and directions transfer between the LLaVA models, and a checkpoint trained with the projector only.}\par\smallskip

\begin{figure}[h]\centering\includegraphics{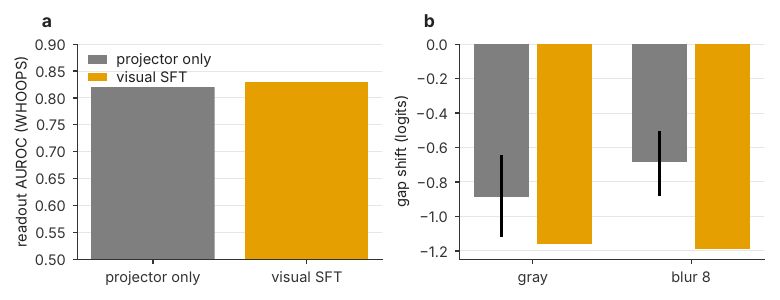}
\caption{\textbf{Projector-only checkpoint.} The projector-only LLaVA-1.5 (gray) against the tuned model (orange, labeled ``visual SFT''): readout AUROC on WHOOPS (a) and the gap shift under grayscale and blur at radius 8 (b; bootstrap intervals shown for the projector-only bars). It followed the image on 0.755 of WHOOPS conflicts against 0.733 for the tuned model, had a readout AUROC of 0.82 against 0.83, and lost 58--76\% as much margin to grayscale and blur.}\label{fig:stage1}\end{figure}

\begin{figure}[h]\centering\includegraphics{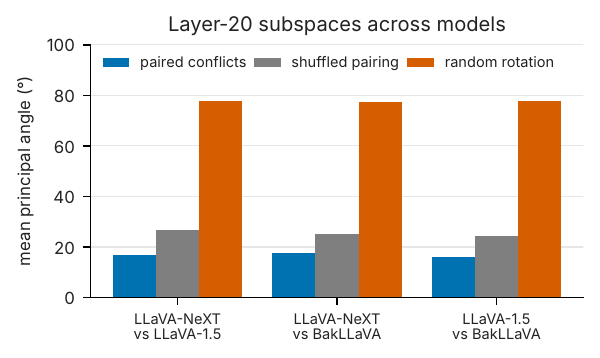}
\caption{Mean principal angle between two models' layer-20 256-D subspaces after orthogonal Procrustes on paired conflicts (blue), against a shuffled pairing (gray) and a random rotation (orange).}\end{figure}

Principal angles between models' layer-20 256-D principal subspaces after orthogonal Procrustes on paired conflicts (NeXT--1.5, NeXT--BakLLaVA, 1.5--BakLLaVA): 16.7$^\circ$, 17.6$^\circ$, 16.2$^\circ$; shuffled pairing 26.8$^\circ$, 25.1$^\circ$, 24.3$^\circ$; random rotation 77.6$^\circ$. Per-example unembedding-difference directions ported through the same map have cosine 0.01--0.04 with the target's own direction (random 0.016) and move the target by $+0.52$ (into LLaVA-NeXT), $+0.43$ (into LLaVA-1.5), $+0.21$ (into BakLLaVA) at $\alpha=0.5$ (the target's own direction in that run: $+2.63$, $+1.97$, $+2.30$; random $\approx0$). The three $\alpha=0.5$ full-injection values in this supplement ($+2.60$/$+1.96$ at natural norms, $+2.61$/$+1.98$ renormalized, $+2.63$/$+1.97$ here) come from three runs on overlapping item subsets; each comparison is within its run.

\FloatBarrier
\section{Detector, calibration and early exit}\label{app:detector}
\noindent\textit{Supports main-paper Sec.~8. The readout used as a detector, its calibration, and early exit.}\par\smallskip

\begin{figure}[h]\centering\includegraphics{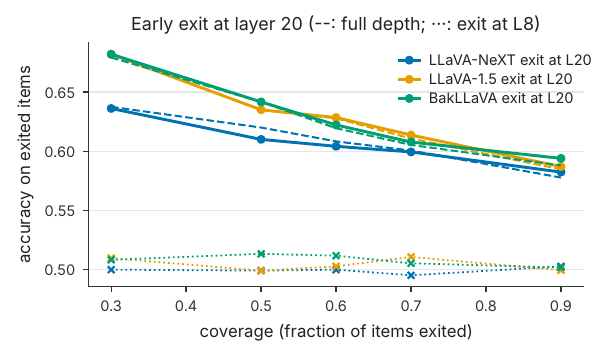}
\caption{\textbf{Early exit at layer 20.} Accuracy of the exited answer on the items the centered margin is most confident about (solid) against full depth on the same items (dashed) and an exit at layer 8 (dotted), by coverage.}\label{fig:exit}\end{figure}

\begin{figure}[h]\centering\includegraphics{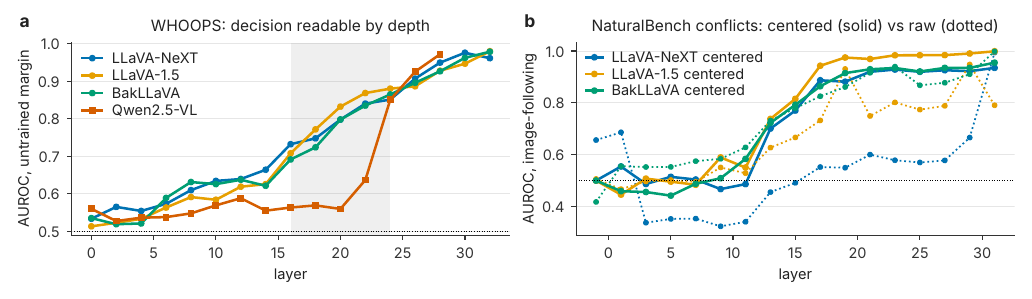}
\caption{Untrained margin AUROC by layer. (a) WHOOPS, four models; shaded band: layers 16--24. (b) NaturalBench conflicts, centered (solid) and raw (dotted).}\label{fig:depth}\end{figure}

\begin{table}[h]\centering\footnotesize
\caption{Early exit at layer 20 on NaturalBench: exited accuracy / full-depth accuracy on the same items, and a random-layer exit at matched coverage.}\label{tab:exit}
\begin{tabular}{lccc}\toprule
Coverage & LLaVA-NeXT & LLaVA-1.5 & BakLLaVA \\ \midrule
30\% & 0.636 / 0.637 (L26: 0.654) & 0.682 / 0.682 (L26: 0.675) & 0.682 / 0.679 (L26: 0.686) \\
50\% & 0.610 / 0.620 (L18: 0.619) & 0.635 / 0.635 (L18: 0.636) & 0.642 / 0.642 (L18: 0.631) \\
60\% & 0.604 / 0.608 (L16: 0.569) & 0.628 / 0.628 (L16: 0.582) & 0.622 / 0.619 (L16: 0.588) \\
70\% & 0.599 / 0.601 (L8: 0.495) & 0.614 / 0.611 (L8: 0.511) & 0.608 / 0.605 (L8: 0.505) \\
90\% & 0.582 / 0.578 (L10: 0.504) & 0.587 / 0.585 (L10: 0.499) & 0.594 / 0.588 (L10: 0.509) \\ \bottomrule
\end{tabular}\end{table}

The untrained margin's AUROC on WHOOPS rises with the hidden-state index on LLaVA-NeXT: 0.64 at 12, 0.73 at 16, 0.80 at 20, 0.84 at 22, 0.95 at 28. Early exit on NaturalBench uses the centered layer-20 margin and exits when its magnitude exceeds the coverage quantile; \cref{tab:exit} gives the exited accuracy against full depth on the same items, with a random-layer control at matched coverage (one draw per row, the layer in parentheses). \Cref{tab:detsum} collects the other detector results.

\begin{table}[h]\centering\footnotesize\setlength{\tabcolsep}{6pt}
\caption{\textbf{Detector, early exit and probes on the three LLaVA models.} AURC: lower is better. Fixed-layer exits on NaturalBench use the sign of the centered margin as the exit answer and its magnitude for coverage; differences are in accuracy points at 90\% coverage with a paired cluster bootstrap on the item. Dash: not reported.}\label{tab:detsum}
\begin{tabular}{lccc}\toprule
 & LLaVA-NeXT & LLaVA-1.5 & BakLLaVA \\ \midrule
\multicolumn{4}{l}{\emph{Correctness-AURC on NaturalBench}} \\
Centered layer-20 margin & 0.368 & 0.348 & 0.350 \\
Final margin & 0.357 & 0.335 & 0.341 \\
Random ordering & 0.436 & 0.418 & 0.434 \\ \midrule
\multicolumn{4}{l}{\emph{Fixed-layer exit minus full depth}} \\
Index 18 & $+$0.004 [$-$0.011, $+$0.019] & $-$0.009 [$-$0.019, $+$0.002] & $-$0.011 [$-$0.025, $+$0.002] \\
Index 16 & $-$0.018 & $-$0.033 & $-$0.022 \\
Index 20, conflict items: image-following (points) & $+$6.8 & $+$1.6 & $+$7.0 \\ \midrule
\multicolumn{4}{l}{\emph{Trained last-position probes at layer 20, AUROC}} \\
ViLP & 0.86 & 0.85 & 0.79 \\
Visual CounterFact & -- & 0.86 & 0.85 \\
WHOOPS & 0.73 & 0.60 & 0.70 \\ \midrule
Mid-depth yes/no margin, error prediction on POPE & 0.62 & 0.57 & 0.74 \\ \bottomrule
\end{tabular}\end{table}

\noindent\emph{Notes on \cref{tab:detsum}.} Fixed-layer exits were run at hidden-state indices 16, 18, 20, 22 and 26. At 90\% coverage the index-18 exit already matches full depth on every model and index 16 does not (intervals at or below zero). On the conflict items alone the mid-depth exit sides with the image more often than the final answer, so the parity on all items is a parity of totals: relative to the layer-20 reading, the final layers move conflicts toward the prior. The Visual CounterFact probes reach 0.88 at the best layer. Selective prediction on LLaVA-NeXT: the trained probe calibrates better (AURC 0.09 vs 0.23 on ViLP; 0.14 vs 0.19 on WHOOPS). Benchmark error prediction with the mid-depth yes/no margin is 0.5--0.6 on MME and 0.45--0.48 on HallusionBench (below chance); the post-hoc final margin gives 0.75--0.84 on POPE and MME and 0.51--0.57 on HallusionBench.

\FloatBarrier
\section{Benchmarks}\label{app:bench}
\noindent\textit{Supports main-paper Sec.~9. Steering and injection on standard benchmarks, and how often those benchmarks are answered without the image.}\par\smallskip

\begin{figure}[h]\centering\includegraphics{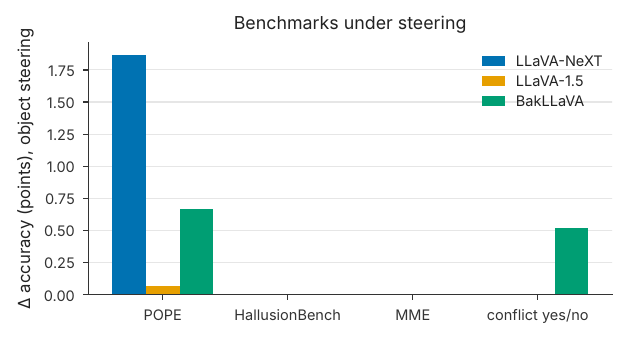}
\caption{Change in accuracy, in points, under object steering at full dose, per benchmark and model.}\end{figure}

\begin{table}[h]\centering\small
\caption{Benchmark accuracy, baseline $\to$ object steering at full dose (answer position); conflict yes/no under concept injection at $\alpha{=}1$ in the last column. Dash: not reported (the run failed an implementation check).}\label{tab:bench}
\begin{tabular}{lccccc}\toprule
 & POPE & HallusionBench & MME & Conflict yes/no & Conflict yes/no, injection \\ \midrule
LLaVA-NeXT & 0.697 $\to$ 0.716 & 0.485 $\to$ 0.485 & 0.617 $\to$ 0.617 & 0.744 $\to$ 0.744 & -- \\
LLaVA-1.5 & 0.741 $\to$ 0.742 & 0.484 $\to$ 0.484 & 0.652 $\to$ 0.652 & 0.786 $\to$ 0.786 & 0.787 $\to$ 0.787 \\
BakLLaVA & 0.743 $\to$ 0.749 & 0.463 $\to$ 0.463 & 0.677 $\to$ 0.677 & 0.727 $\to$ 0.733 & 0.728 $\to$ 0.791 \\ \bottomrule
\end{tabular}\end{table}

POPE, HallusionBench, MME with object steering at $\rho\in\{0.5, 1\}$ and concept injection at $\alpha\in\{0.5, 1\}$; CHAIR \cite{rohrbach2018chair} on 500 COCO captions at $\rho\in\{0.25, 0.5\}$. Object steering at full dose changes POPE by $+1.9$, $+0.1$, $+0.6$ points and leaves HallusionBench and MME unchanged (\cref{tab:bench}); concept injection improves conflict yes/no accuracy by 6.3 points on BakLLaVA (prior-override splits +11--12) and 0 on LLaVA-1.5; the LLaVA-NeXT run failed its implementation check and is not reported; CHAIR$_s$ 0.110$\to$0.102 on NeXT, higher on LLaVA-1.5, mixed on BakLLaVA. Injection is undefined on benchmark items without a candidate answer pair; those items are excluded.

\noindent\textbf{How much of a yes/no benchmark the prior answers.} For each item we compared the model's yes/no answer with and without the image. The share whose answer does not change is 0.77, 0.51, 0.45 on POPE, 0.73, 0.77, 0.80 on HallusionBench and 0.64, 0.43, 0.55 on MME (item-bootstrap half-widths about 0.03); text-only accuracy on the same benchmarks is 0.44 to 0.56, so these items are answered by the model's polarity default rather than by knowledge, and a benchmark cannot measure visual grounding on them for that model. The three models agree only partly on which items they are (pairwise Jaccard 0.36 to 0.49 on POPE, 0.62 to 0.68 on HallusionBench, 0.36 to 0.44 on MME). The magnitude of the centered layer-20 margin separates image-dependent from prior-answerable items on LLaVA-NeXT's POPE (AUROC 0.82 [0.79, 0.85]) and nowhere else (0.39 to 0.55), so the readout does not separate these items on the other benchmarks.

\FloatBarrier
\section{The readout on NaturalBench photographs}\label{app:real}
\noindent\textit{Supports main-paper Sec.~9. Scoring, centering and the object edits on NaturalBench photographs.}\par\smallskip

\begin{figure}[h]\centering\includegraphics{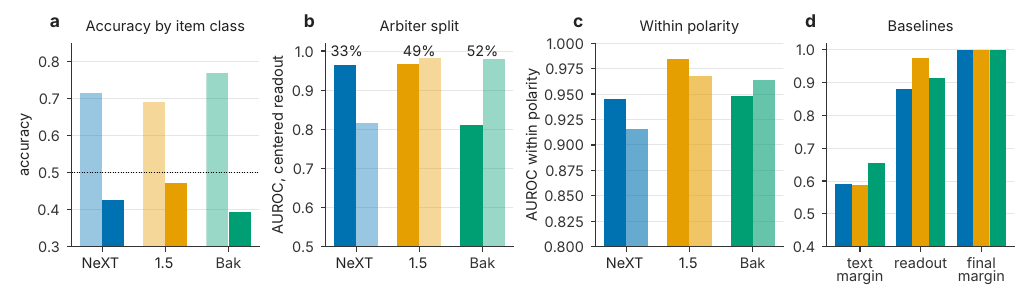}
\caption{\textbf{Real photographs.} NaturalBench yes/no pairs, 1,200 natural conflicts per model. (a) Accuracy when the prior agrees (faint) against when the image contradicts it (solid). (b) The centered readout on the items where a gray image changes the answer (solid) against the default items (faint), with the image-dependent share above each pair. (c) The readout within each answer polarity (solid: yes items; faint: no items). (d) The readout against the text-only margin and the final-layer margin on the same items; bars are the three models in the colors of (a). NeXT, 1.5 and Bak: LLaVA-NeXT, LLaVA-1.5 and BakLLaVA.}\label{fig:real}\end{figure}

\begin{figure}[h]\centering\includegraphics{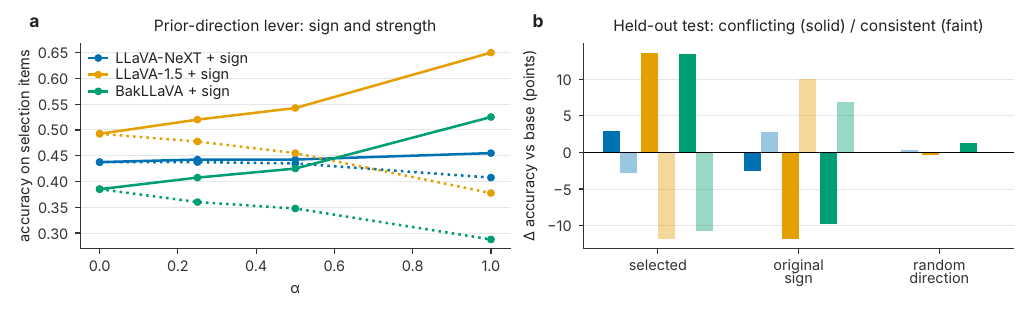}
\caption{Injection of the text-only answer direction. (a) Accuracy on the 400 conflicting selection items by strength $\alpha$; solid: the selected sign; dotted: the original sign. (b) Held-out change in accuracy, in points, on conflicting (solid) and consistent (faint) items for the selected sign at $\alpha=1$, the original sign and a random direction. Blue, orange and green: LLaVA-NeXT, LLaVA-1.5 and BakLLaVA.}\label{fig:lever}\end{figure}

\begin{figure}[h]\centering\includegraphics{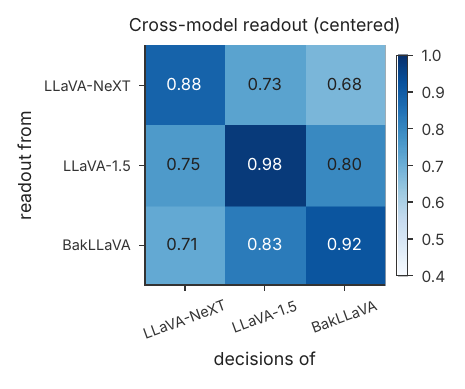}
\caption{One model's centered readout predicting another model's decisions on the same photographs (rows: readout from; columns: decisions of). Diagonal: each model's own readout.}\end{figure}

\noindent\textbf{Scoring.} For each item we compute the layer-20 logit-lens margin $m_i = z_{\mathrm{yes},i} - z_{\mathrm{no},i}$ at the answer position. The raw readout is $s_i = p_i m_i$ with $p_i = +1$ if the image-supported (correct) answer is yes and $-1$ otherwise, so that a high score predicts that the model follows the image; the target is whether the model's final answer equals the image-supported answer. The centered readout is $s_i = p_i (m_i - c)$ with $c = \mathrm{median}_i\, m_i$ over the evaluation items; because $p_i$ is item-dependent, subtracting $c$ before orienting changes the ranking across yes and no items, which is why centering raises AUROC on LLaVA-NeXT, whose $m_i$ carries a large negative offset. The orientation $p_i$ is used only to define the evaluation target; the early exit and the gated interventions use $m_i - c$ alone and predict yes when it is positive. The trained probes use the same target and the same folds. \Cref{tab:nbscore} gives the readout and its controls, \cref{tab:nbcross} the transfer between models and \cref{tab:nbinject} the interventions.

\begin{table}[h]\centering\footnotesize\setlength{\tabcolsep}{7pt}
\caption{\textbf{The readout on NaturalBench, three LLaVA models.} AUROC unless noted. Dash: not reported.}\label{tab:nbscore}
\begin{tabular}{lccc}\toprule
 & LLaVA-NeXT & LLaVA-1.5 & BakLLaVA \\ \midrule
\multicolumn{4}{l}{\emph{Scorer validation, agreement with the greedy generated answer (200 items)}} \\
In-context yes/no scorer & 99.5\% & 100\% & 98\% \\
Statement-continuation scorer & 62\% & 69\% & 66.5\% \\ \midrule
\multicolumn{4}{l}{\emph{With the validated scorer}} \\
Yes-rate & 0.33 & 0.52 & 0.66 \\
Accuracy, prior agrees / contradicts & 0.72/0.43 & 0.69/0.47 & 0.77/0.40 \\
Raw layer-20 readout & 0.55 & 0.93 & 0.86 \\
Centered readout & 0.88 & 0.98 & 0.92 \\
Text-only-subtracted readout & 0.82 & 0.95 & 0.93 \\
Within-polarity & 0.95/0.92 & 0.99/0.97 & 0.95/0.96 \\
Trained 5-fold probe & 0.64 & 0.66 & 0.75 \\ \midrule
\multicolumn{4}{l}{\emph{Checks of the centering}} \\
Held-out centering, median from a disjoint half ($n$) & 0.886 (615) & 0.976 (587) & 0.947 (579) \\
Non-default-prior subset & 0.916 & 0.985 & 0.948 \\
Text-only control & 0.847 & 0.907 & 0.935 \\
1-D margin probe, same folds & 0.478 & 0.930 & 0.861 \\
2-D probe (margin, text-only margin) & 0.850 & 0.967 & 0.953 \\ \midrule
\multicolumn{4}{l}{\emph{Gray-blank arbiter}} \\
Answers that change under a gray image, all items & 0.320 & 0.495 & 0.530 \\
Answers that change, natural conflicts & 0.332 & 0.487 & 0.517 \\
Centered readout, image-dependent conflicts ($n$) & 0.963 (398) & 0.967 (584) & 0.812 (621) \\
\quad follows-image rate & 0.588 & 0.452 & 0.343 \\
Centered readout, image-independent conflicts ($n$) & 0.816 (802) & 0.983 (616) & 0.980 (579) \\ \midrule
\multicolumn{4}{l}{\emph{Object and prior direction}} \\
Object-area correlation $r$ (n.s.) & $+0.04$ & $-0.00$ & $-0.05$ \\
Size recomposite at $0.5\times$, conflicting items & $-0.13$ & $-0.07$ & $-0.09$ \\
Prior-direction sign test, base & 0.438 & 0.492 & 0.385 \\
\quad original sign & 0.435 & 0.455 & 0.347 \\
\quad opposite sign & 0.443 & 0.542 & 0.425 \\ \bottomrule
\end{tabular}\end{table}

\noindent\emph{Notes on \cref{tab:nbscore}.} The statement-continuation scorer used for the four conflict sources agrees with generation on too few items here and is therefore not used for yes/no benchmarks. The final-layer margin has AUROC 1.000. Depth curves by hidden-state index: LLaVA-NeXT 0.66, 0.34, 0.35, 0.46, 0.55, 0.60, 0.57, 0.67, 0.99 at 0, 4, 8, 14, 20, 22, 26, 30, 32; LLaVA-1.5 0.51 to 0.93 at 20 and 0.95 at 30; BakLLaVA 0.42 to 0.86 at 20, 0.94 at 24, 1.00 at 32. The prior-direction sign test uses 400 conflicting items per model.

\begin{table}[h]\centering\footnotesize\setlength{\tabcolsep}{8pt}
\caption{\textbf{The readout across models.} AUROC of the centered readout of the row model, centered with its own median, on the decisions of the column model. Diagonal: the model's own readout. Uncentered, the off-diagonal values are 0.46--0.83.}\label{tab:nbcross}
\begin{tabular}{lccc}\toprule
Readout from & LLaVA-NeXT & LLaVA-1.5 & BakLLaVA \\ \midrule
LLaVA-NeXT & 0.88 & 0.73 & 0.68 \\
LLaVA-1.5 & 0.75 & 0.98 & 0.80 \\
BakLLaVA & 0.71 & 0.83 & 0.92 \\ \bottomrule
\end{tabular}\end{table}

\begin{table}[h]\centering\footnotesize\setlength{\tabcolsep}{7pt}
\caption{\textbf{Held-out test of the injection on NaturalBench} (sign $+$ and $\alpha=1$, selected on 400 conflicting items). Share of items on which the model follows the image: base / always-on / gated. Bottom: masks from dependency-parsed nouns (63\% coverage), change in points at $\alpha=0.5$.}\label{tab:nbinject}
\begin{tabular}{lccc}\toprule
 & LLaVA-NeXT & LLaVA-1.5 & BakLLaVA \\ \midrule
Conflicting items ($n=800$) & 0.421/0.450/0.450 & 0.464/0.600/0.551 & 0.400/0.535/0.446 \\
Consistent items ($n=1{,}200$) & 0.717/0.688/0.685 & 0.690/0.572/0.612 & 0.768/0.662/0.723 \\
All items & 0.599/0.593/0.591 & 0.600/0.583/0.588 & 0.621/0.611/0.613 \\
Random direction, always-on, conflicting (base) & 0.421 (0.421) & 0.460 (0.464) & 0.412 (0.400) \\ \midrule
Parsed-noun masks: object-area $r$ (n.s.; $n$) & $+0.07$ (797) & $-0.01$ (746) & $-0.07$ (706) \\
Injection, conflicting, always-on & $+0.5$ & $+5.1$ & $+4.0$ \\
Injection, conflicting, gated & $+0.2$ & $+2.4$ & $+1.1$ \\
Injection, consistent & $+0.1$ & $-4.4$ & $-3.5$ \\ \bottomrule
\end{tabular}\end{table}

\noindent\emph{Notes on \cref{tab:nbinject}.} Always-on applies the intervention to every item; gated applies it only where the centered margin favors the prior. The answer-agnostic interventions ($\alpha=0.5$; object steering and prior suppression) change the rate by $-4.7$ to $+4.7$ points always-on and by $-1.5$ to $+1.8$ gated. With the parsed-noun masks object steering changes it by $+0.4$ to $+0.6$ always-on and by $-0.6$ to $+0.7$ gated. Early-exit values are in Supp.~\ref{app:detector}; head-ablation, matched-region and injection-component values in Supp.~\ref{app:steer}.

\begin{table}[h]\centering\footnotesize\setlength{\tabcolsep}{4pt}
\caption{Object-only against rest-only edits on NaturalBench photographs, 300 items per model, SAM3 masks. Ratios are object/rest of mean $|\Delta m|$ with cluster-bootstrap 95\% intervals; $|\Delta|$ rows are in logits.}\label{tab:photoobj}
\begin{tabular}{lccc}\toprule
 & LLaVA-NeXT & LLaVA-1.5 & BakLLaVA \\ \midrule
Pooled ratio, final margin $m$ & 0.92 [0.78, 1.11] & 0.80 [0.66, 0.97] & 0.76 [0.65, 0.90] \\
Pooled ratio, hidden-state-20 margin & 0.89 [0.76, 1.06] & 0.84 [0.68, 1.03] & 0.76 [0.65, 0.90] \\
Gray / blur / noise ratio, $m$ & 1.22 / 0.79 / 0.87 & 1.02 / 0.65 / 0.85 & 0.92 / 0.62 / 0.82 \\
Gray $|\Delta m|$: object / rest & 0.51 / 0.42 & 0.42 / 0.41 & 0.91 / 0.99 \\
Gray $|\Delta m|$: random region / salient region of matched area & 0.23 / 0.19 & 0.19 / 0.18 & 0.37 / 0.40 \\
Recomposite at scale 1.0, signed $\Delta m$ & $-$0.01 [$-$0.03, $-$0.00] & $-$0.00 [$-$0.01, 0.01] & $-$0.00 [$-$0.02, 0.02] \\
Scale 1.5 minus 0.5, signed $\Delta m$ & 0.13 [0.03, 0.22] & 0.08 [$-$0.01, 0.17] & 0.19 [$-$0.01, 0.39] \\
Answer flips under gray: object-only / rest-only & 9.7 / 6.0\% & 6.7 / 5.7\% & 8.0 / 9.0\% \\
Rule & fail & fail & fail \\ \bottomrule
\end{tabular}\end{table}

\noindent\textbf{Object-only versus rest-only edits on photographs (main-paper Sec.~9).} For each model, 300 NaturalBench items balanced over truth (yes/no) and prior agreement (the model's clean answer equal to its text-only answer or not), 75 per cell, drawn from the 2,400-item yes/no pool used for the readout. The question noun is extracted with the same heuristic as the attention-steering experiment and segmented with SAM3 (facebook/sam3, instance threshold 0.5, with transformers 5.22.1; the LLaVA models run with 4.57.6); an item enters when the mask score is at least 0.5 and the mask covers 1 to 50\% of the image (667 of the 2,400 pool items qualify; median score 0.86, median area 0.09). Thirteen forwards per item: clean; gray, blur (radius 8) and noise ($\sigma$ = 40) applied to the object only or to everything except the object; gray and blur on a random square of the object's pixel area; gray on the most salient non-object pixels of matched area; and the recomposite at scale 1.0 (artifact control), 0.5 and 1.5. Outcome: $|\Delta|$ of the yes/no margin $m$ against the clean image, signed toward the true answer, and the same for the hidden-state-20 margin; cluster bootstrap over photographs, 2,000 replicates. Pre-specified rule: pass if the pooled object/rest ratio of mean $|\Delta|$ is at least 2 with the interval excluding 1 on both margins and the scale-1.0 control covers zero.

\noindent\emph{Result.} All three models fail the rule (\cref{tab:photoobj}), with every pooled ratio below 1 (the interval excludes 1 on LLaVA-1.5 and BakLLaVA). Gray is the edit under which the object counts most (1.22, 1.02, 0.92); blur and noise of the larger remainder move the margin more than the same edits on the object. Per unit area the WHOOPS pattern holds: graying the object moves the margin 2.2, 2.2 and 2.5 times as much as graying a random square of its area, and 2.7, 2.3 and 2.3 times as much as graying the most salient non-object pixels of that area. The size dose keeps its sign on every model (scale 1.5 minus 0.5 positive; interval above zero on LLaVA-NeXT). Splits by prior agreement, truth and prior consistency give pooled ratios of 0.67 to 1.01.

\noindent\emph{By question type and mask.} An exploratory split by question type (a regular expression for a second object or a spatial or interaction relation; 225 to 228 of the 300 items are relational) gives gray ratios of 2.03 [1.23, 3.11], 1.61 [0.94, 2.52] and 1.59 [1.07, 2.26] on the 72 to 75 one-object questions (pooled over the three edits 1.30, 1.12, 0.94) and 0.99, 0.84 and 0.75 on the relational ones (pooled 0.77 [0.64, 0.93], 0.68 [0.56, 0.81], 0.69 [0.58, 0.82]): where the question names one object, graying that object moves the answer more than graying the rest; where it names a relation, the rest of the photograph, which holds the second object, carries more. With CLIPSeg masks (median area 0.14; median IoU with SAM3 0.59 on the 593 items both segmented; 79\% of the selected items in common) gave pooled ratios of 1.20, 1.07 and 0.99 and object-to-random ratios of 1.9 to 2.1: tighter masks lower the object-to-rest ratio, since the object region shrinks and the rest grows, and leave the per-area contrast in place.
\FloatBarrier
\noindent\textbf{Readout, early exit and bounds.} The text-only answer on NaturalBench is mostly a default yes or no rather than knowledge about the object (yes-rates 0.33, 0.52, 0.66), so what is overridden is a polarity default. The readout score is $s_i = p_i\,(m_i - c)$: $m_i$ the yes-minus-no margin at layer 20, $c$ its median over the evaluation items, and $p_i=\pm1$ the orientation toward the image-supported answer. Centering removes LLaVA-NeXT's polarity offset; over all conflicts the raw readout is 0.55--0.93 and the centered one 0.88--0.98. On BakLLaVA the readout is higher on default items (0.98) than on image-dependent ones (0.81), so there it mostly reflects the default answer. The early exit saves 12 of 32 decoder layers on the exited items (19\% of decoder compute at 50\% coverage, not end-to-end latency); an exit anywhere in layers 18--26 does as well, and one at layers 8--16 does worse. \Cref{tab:readfull} lists every readout variant.

\begin{table}[h]\centering\footnotesize\setlength{\tabcolsep}{6pt}
\caption{\textbf{Readout on the LLaVA models, all variants} (main-paper Secs.~8 and 9). AUROC of the untrained layer-20 margin for the decision. NB: NaturalBench, 1,200 natural conflicts per model (hidden-state index 20), centered on the median margin over all evaluation items unless raw; image-dependent: items a gray image changes ($n$ = 398/584/621); text-only control: the language model's own text-only answers. WHOOPS $n$ = 424/416/419. The last two rows give the scale of the prior override.}\label{tab:readfull}
\begin{tabular}{lccc}\toprule
 & LLaVA-NeXT & LLaVA-1.5 & BakLLaVA \\ \midrule
WHOOPS, within source & 0.80 & 0.83 & 0.80 \\
NB, image-dependent & 0.96 & 0.97 & 0.81 \\
NB, default items & 0.82 & 0.98 & 0.98 \\
NB, all, centered & 0.88 & 0.98 & 0.92 \\
NB, all, raw & 0.55 & 0.93 & 0.86 \\
NB, median from other half & 0.89 & 0.98 & 0.95 \\
NB, within yes / no & 0.95/0.92 & 0.99/0.97 & 0.95/0.96 \\
Text-only control & 0.85 & 0.91 & 0.94 \\
NB, trained linear probe (Supp.~M) & 0.64 & 0.66 & 0.75 \\
All curated sources, untrained margin & 0.85 & 0.87 & 0.83 \\
All curated sources, tuned probe (Supp.~M) & 0.85 & 0.82 & 0.85 \\ \midrule
Acc., agrees / contradicts & 0.72/0.43 & 0.69/0.47 & 0.77/0.40 \\
Prior override (points) & 29 & 22 & 37 \\ \bottomrule
\end{tabular}\end{table}

Two things that hold on curated conflicts weaken on photographs. Object locality: on 300 items per model with SAM3 masks of the question noun, graying the object shifts the margin 0.9--1.2 times as much as graying the other 91\%, and 0.8--0.9 times pooled over gray, blur and noise, against 2--5 times on WHOOPS; per unit area it still counts 2.2--2.5 times as much as a random region of its size. For the quarter of questions about one object the gray ratio is 1.6--2.0; for relational ones, whose second object lies in the rest, 0.8--1.0. Correction fails there: noun-mask attention steering changes conflicting-item accuracy by at most $+0.7$ points, and injecting the model's own text-only answer direction trades 2.9--13.6 points of held-out conflicting accuracy for 2.9--11.8 points of consistent-item accuracy (a random direction changes nothing), a polarity-bias vector; steering leaves POPE \cite{li2023pope}, HallusionBench \cite{guan2023hallusionbench} and MME within two points.

\section{Architecture choices and controls}\label{app:faq}
\noindent\textit{Supports the whole paper. The choices behind each analysis and the controls for them.}\par\smallskip
\noindent\textbf{Edited-to-clean patching.} The same image, edited, is the counterfactual; patching the edited copy's residuals into the clean run is a noising patch, so a recovered fraction says the edited activation at that site suffices to reproduce the edit, equivalently that the clean activation there is needed for the clean outcome. Gaussian-noise corruptions were avoided on the recommendation of \citet{zhang2024patching}, who recommend an in-distribution counterfactual input; here that input is the edited copy of the same image. Cross-example interchange is reported separately at the answer position.

\noindent\textbf{The margin and its scale.} The gap is the logit difference between the two candidate answers at the answer position; its sign is the decision. Its scale is given by the accompanying changes in vision-win (grayscale moves vision-win from 0.78 to 0.69 on LLaVA-NeXT). The gap is not a sufficient statistic for every ablation: the whole-update ablation on LLaVA-NeXT flips 5.4\% of image-wins and 6.3\% of prior-wins with a mean shift near zero, which is why flip rates are reported alongside shifts.

\noindent\textbf{Layer 20 and the hidden-state offset.} Layer 20 is the layer at which earlier work \cite{vankadaru2026trust} read representations on LLaVA-NeXT; we do not re-tune it. The NaturalBench analysis indexes hidden states from the embeddings, so its index 20 is decoder layer 19 in this paper's convention. The centered curve plateaus across 18--24 on all three models, so the result is insensitive to the choice within that band.

\noindent\textbf{Median centering.} The margin carries a per-model answer-polarity offset on NaturalBench. Subtracting the median over the evaluation items is label-free; estimating the median on a disjoint half changes the AUROCs by at most 0.03; and the within-polarity AUROCs (0.92--0.99) show the ranking exists before centering.

\noindent\textbf{Trained probe against the untrained direction.} The unembedding difference is one direction in the 4,096-dimensional residual, so a probe could in principle recover it. Two things stop the 5-fold probe: with 1,200 items and L2 regularization at $C=0.01$ it is strongly shrunk, and the class-conditional distributions of the margin differ by answer polarity, so a single threshold fits neither polarity (a 1-D probe on the margin itself with the same folds reaches only 0.48, 0.93, 0.86). Supplying the polarity as a second feature (the text-only margin) lifts the 2-D probe to 0.85, 0.97, 0.95, matching the centered readout. The readout's advantage is that it fixes the direction and needs only an offset; the probe has to learn both from 1,200 items in 4,096 dimensions. On the curated conflicts the same holds with the regularization chosen properly: with $C$ selected by nested cross-validation grouped by object ($C$ = 0.001 to 0.003), the trained layer-20 probe reaches 0.85 [0.82, 0.87], 0.82 [0.80, 0.85] and 0.85 [0.83, 0.88] AUROC over the three sources without the size split and 0.73, 0.58, 0.66 on WHOOPS alone, against the untrained margin's 0.85, 0.87, 0.83 and 0.80, 0.83, 0.80 on the same items (with $C$ = 1 and ungrouped folds: 0.83, 0.82, 0.81 and 0.73, 0.53, 0.61).

\noindent\textbf{Text-only control.} The same lens, run without an image, predicts the language model's own text-only answer at 0.85--0.94; those are AUROCs for a different target than the image-run decision. The comparison shows the lens is a general property of the language model; it does not by itself say what the image run adds. The gray-image control split does: 0.96 and 0.97 on the image-dependent items of LLaVA-NeXT and LLaVA-1.5, 0.81 on BakLLaVA.

\noindent\textbf{Sign of the injected direction and its selection.} Suppressing the model's text-only answer direction at image tokens lowered conflicting-item accuracy; the opposite sign raised it. The direction reaches the answer through attention value projections whose sign need not match the unembedding; we have no further account. Sign and strength were selected on 400 items and tested on the remaining 800 with random-direction and text-site controls; the gain held out on two models and was offset by consistent-item losses. One hypothesis for the inversion: the unembedding difference points along the direction that \emph{writes} the answer at the final position, while at image tokens the direction that the arbitration heads' value projections carry forward need not align with it, and the reported steering effects say that pulling attention onto the image raises the margin; adding the prior direction at image tokens may make those tokens more similar to the prior for the heads that read them.

\noindent\textbf{BakLLaVA.} The control analyses of this section begin with a check on three items that patches layer 0 alone at the image positions. On BakLLaVA it recovered 0.24--0.80 of the gray shift, although the identity checks of the band map passed and its L0--3 band recovers 1.00 on average. The BakLLaVA values below come from a separate run with a stricter check, which patches the gray copy's block-0 output at every position (recovery 1.000 on all 25 of the first 30 items whose gray shift exceeds 0.2), and keeps the image-position check (median 1.00, interquartile range 0.98 to 1.03 on the same items); both pass on transformers 4.57.6. The four controls are reported for all three models. Matched area on 360 items: object $-0.64$ [$-0.87$, $-0.40$], rest $-0.34$, random region of the object's area $-0.18$. Top-20 head ablation at the answer position: 7.7\% of image-wins flipped (gap $-1.98$), layer-matched random heads 2.2\% ($-0.07$). The arbitration heads write 0.31 of their output norm into the fitted 256-D subspace against 0.25 for a random 256-dimensional subspace (0.32 and 0.31 on LLaVA-NeXT and LLaVA-1.5). Injection at natural component norms: the in-subspace part carries 38\% of the $\alpha{=}0.5$ effect from 29\% of the norm. BakLLaVA's grayscale contrast fails Holm correction ($p=0.14$) and its injection dose is non-monotone (gap shift $+2.23$ at $\alpha{=}0.5$, $+1.33$ at $\alpha{=}1$).

\noindent\textbf{Qwen2.5-VL version sensitivity.} Three runs of the outcome measurement under two minor versions of the transformers library gave valid counts of 356/457/345 and 477/574/407 and vision-wins of 0.79/0.34/0.48 and 0.98/0.57/0.79 by source. The object-level battery, band map and readout by layer are reported on the later run and replicate the LLaVA ordering. A repeat with the library fixed at version 4.57.6 (bfloat16, eager attention) reproduces the later run exactly: image and text margins correlate at Pearson 1.0000 on all 1,593 items, every win and prior-validity flag agrees, and a repeated forward pass on 40 items differs by 0 logits. The additive-model fit of Supp.~\ref{app:equation} and the model-specific object, band-map and readout numbers come from the later run; the common-protocol numbers of Supp.~\ref{app:arch} and main-paper Table~1 come from a third run under the common protocol (410 prior-valid WHOOPS conflicts, vision-win 0.798); the outcome statistics of this model depend on the library version and should be read under it.

\noindent\textbf{Clustering and multiplicity.} Intervals for the additive model are cluster-robust by image, and the other intervals in the main tables resample items, one row per item; only the mediation analysis clusters on prompts. The size split has 727 rows over 410 objects and ViLP has 300 questions over 600 images. We therefore recomputed the headline quantities with a cluster bootstrap (2,000 replicates) on the object and, separately, on the question (object clusters: ViLP 210 over 600 rows, the size split 410 over 727, WHOOPS 383 over 500, Visual CounterFact one row per object). \Cref{tab:cluster} gives the object-clustered intervals.

\begin{table}[h]\centering\footnotesize\setlength{\tabcolsep}{6pt}
\caption{\textbf{Headline quantities with intervals clustered on the object} (cluster bootstrap, 2,000 replicates).}\label{tab:cluster}
\begin{tabular}{lccc}\toprule
 & LLaVA-NeXT & LLaVA-1.5 & BakLLaVA \\ \midrule
Gray inside-minus-outside on WHOOPS & $-$0.56 [$-$0.87, $-$0.29] & $-$0.38 [$-$0.59, $-$0.17] & $-$0.31 [$-$0.61, $-$0.03] \\
Half-size recomposite against the scale-1.0 control & $-$1.08 [$-$1.27, $-$0.89] & $-$0.59 [$-$0.79, $-$0.41] & $-$0.86 [$-$1.12, $-$0.60] \\
Vision-win drop at full gray & $-$0.12 [$-$0.16, $-$0.08] & $-$0.09 [$-$0.14, $-$0.05] & $-$0.10 [$-$0.15, $-$0.06] \\
Centered layer-20 readout on NaturalBench conflicts & 0.88 [0.86, 0.91] & 0.98 [0.97, 0.98] & 0.92 [0.90, 0.93] \\ \bottomrule
\end{tabular}\end{table}

\noindent Question clustering moves interval ends by at most 0.1 and changes no conclusion among these. Under clustering the following intervals include their null value: the arms designed as null controls (random heads, random subspaces, the scale-1.0 recomposite, the contrast, luminance, sharpening and resolution doses), the enlargement arms (scale 1.25 and 1.5) on LLaVA-1.5 and BakLLaVA, and the blur inside-over-outside ratio on the color-transfer set (1.6 [1.0, 3.1] and 1.5 [0.9, 2.5]). Inside-versus-outside contrasts are Holm-corrected across the edit grid.

\noindent\textbf{Two held-out designs.} The half-split design fits the subspace on one half of WHOOPS and evaluates on the other; the cross-source design fits on the non-WHOOPS conflicts with pooled examples and evaluates on WHOOPS, and is the one used for the variance-aware comparison.

\FloatBarrier
\noindent\textbf{Discarded run.} One early run on LLaVA-1.5 and LLaVA-NeXT (subspace sufficiency, head ablation, natural-norm injection, head-write projection, concept projection and the matched-region control) produced clean margins that differ from every other run on the same items (means $-0.40$ against $+2.79$ on LLaVA-1.5 and $-1.83$ against $+3.34$ on LLaVA-NeXT; correlations 0.71 and 0.60, against 0.96 or more between other runs), so its forward pass was altered. Every number from it was replaced by a rerun that first checks its clean margins against the localized-edit run (mean absolute difference under 0.15 on 40 items), and none is reported here. The attention-steering run on LLaVA-1.5 shows a smaller mismatch of the same kind (Supp.~\ref{app:steer}); its values are listed there and are not used in the main paper.

\section{Qualitative examples}\label{app:qual}
\noindent\textit{Examples of the conflicts, the edits and the photograph pairs.}\par\smallskip

\begin{figure}[h]\centering\includegraphics{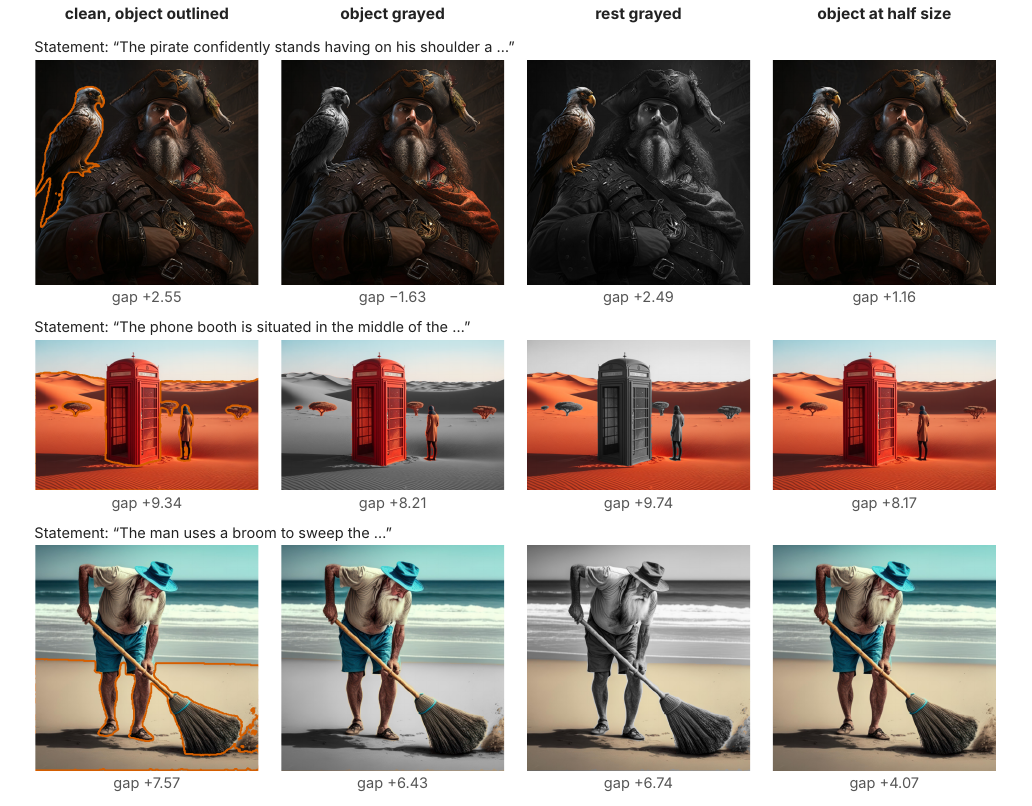}
\caption{\textbf{The edits on three WHOOPS conflicts (LLaVA-NeXT).} For three WHOOPS conflicts (images from \citet{bittonguetta2023whoops}): the image with the SAM3 object outlined and the statement, then the object-only gray edit, the rest-only gray edit and the $0.5\times$ recomposite, with the gap under each. Graying the object alone reverses the margin in the first example and lowers it by 1.1 logits in the other two; graying everything else changes it by $-0.06$, $+0.40$ and $-0.83$; halving the object lowers it in all three.}\label{fig:concept}\end{figure}

\begin{figure}[h]\centering\includegraphics{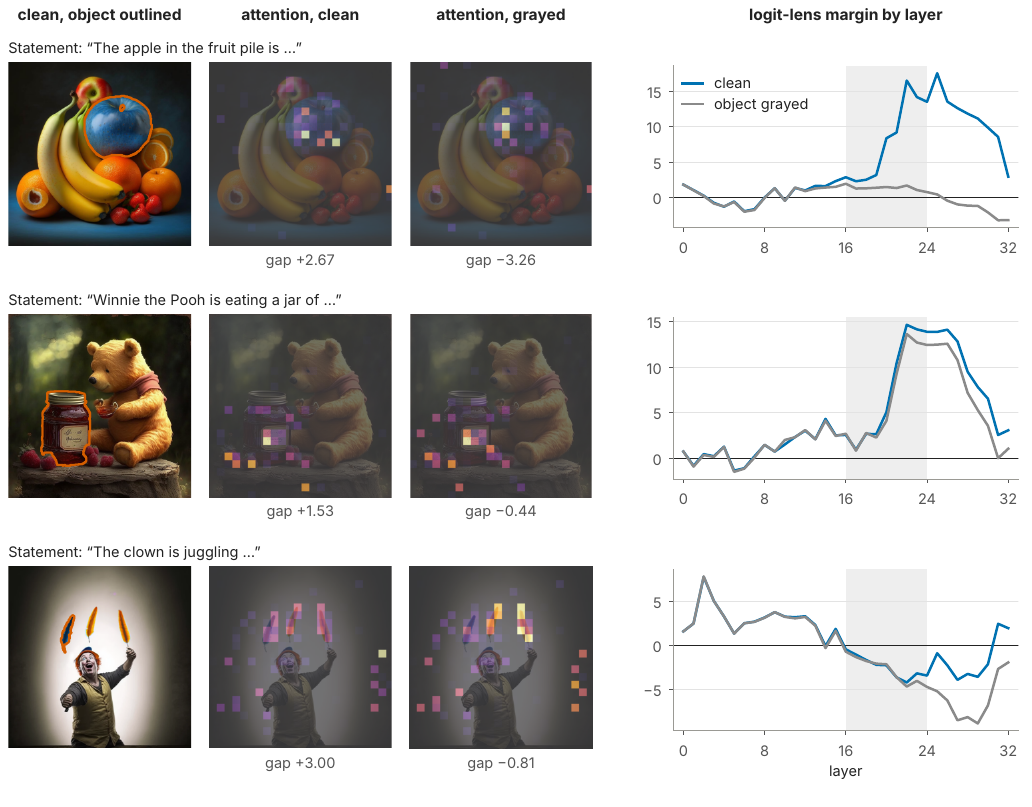}
\caption{The arbitration heads on three WHOOPS examples (LLaVA-1.5; images from \citet{bittonguetta2023whoops}): the clean image with the SAM3 object outlined and the statement; the heads' mean attention from the answer position over image tokens on the clean image and on the object-grayed image, with the gap under each; and the logit-lens margin by layer for both conditions (band 16--24 shaded). The heads attend to the contradicting object, and graying it reverses the decision in all three examples (gap under each image). The two logit-lens margins separate inside the band in the first example and at or after its end in the other two. The gap is the logit difference between the two candidate answers at the output; the last point of a logit-lens curve differs from it by up to 1.6 logits in these examples. Examples are chosen to illustrate the mechanism, not to estimate its size.}\end{figure}

\begin{figure}[h]\centering\includegraphics{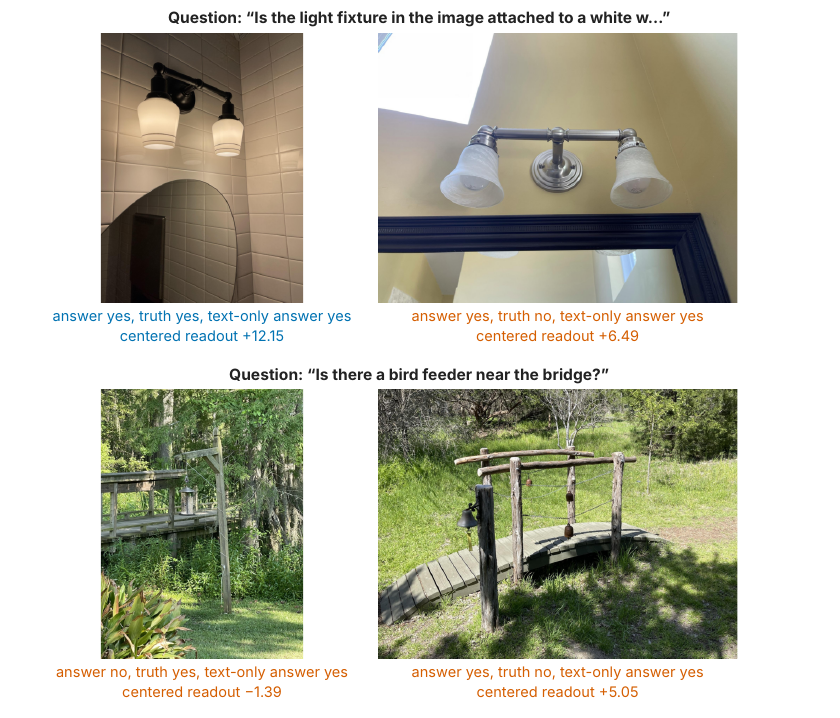}
\caption{NaturalBench pairs (LLaVA-1.5; photographs from \citet{li2024naturalbench}): the same question on two photographs; under each, the model's answer, the truth, its text-only answer, and the centered layer-20 readout (blue when the answer is correct, vermillion when it is wrong).}\end{figure}

\FloatBarrier
\section{Limitations, extended}\label{app:limits}
\noindent\textit{The limitations in full, the detail behind statements that the main text makes briefly, and every pre-specified test with its outcome.}\par\smallskip
Intervals for the additive model are cluster-robust by image, and the other intervals in the main tables resample items, one row per item; the object- and question-clustered recomputation in Supp.~\ref{app:faq} leaves every main interval of the three LLaVA models excluding its null value and widens the enlargement and color-transfer blur contrasts past it. Typical margin magnitudes per model and source are not tabulated; the accompanying vision-win changes in Supp.~\ref{app:faq} give the scale of the main shifts. Not run: steering, the band map and subspace injection on the projector-only checkpoint, and the union-mask test on the relational split. On BakLLaVA one implementation check failed in one analysis (0.24--0.80; Supp.~\ref{app:faq}); its head-ablation, matched-region and natural-norm injection values come from a separate run with a stricter check, and its subspace controls pass all five checks (Supp.~\ref{app:necessity}). Its other numbers in \cref{tab:suffmain} and Fig.~3 and in Supp.~\ref{app:transport} come from the gray-copy patching and band-map analyses, whose hooks passed their identity checks (band L0--3 recovers 1.00 there). The prior manipulation does not establish that the within-item prior weight equals the between-item weight, and the answer-naming prefix acts on the decision beyond the prior it induces (Supp.~\ref{app:equation}); the text-premise heads overlap the image heads below the pre-specified bar for one circuit (Supp.~\ref{app:steer}); the corpus counts come from public corpora, not the models' own training data. A read-space versus nullspace decomposition of the injected direction is not reported, because the next layers' input projections are full rank. The band map is measured with gray, blur and noise donors on WHOOPS and with a paired image on ViLP; the off-block subspace control and the necessity curve are reported on all three models (Supp.~\ref{app:necessity}); the prior-agreeing band map rests on 19 to 43 items per model (Supp.~\ref{app:transport}); the relational (size) split's union-mask prediction was pre-specified and not run; enlargement is confounded with legibility; the projector-only comparison uses one assembled checkpoint of one architecture; NaturalBench scorers are polarity-biased and the noun heuristic for masks is coarse; cross-attention steering is not implemented.

\noindent\textbf{Limitations in detail.} By direct attribution the head result holds on seven of eight architectures; on Gemma-3 it needs heads ranked by estimated ablation effect, and at 27B those reverse only 1.9\% of image-wins. For the additive model and the crossing, three architectures are one model each, which confounds architecture with size and training data. Qwen2.5-VL's outcomes depend on the library version (Supp.~M). The injection, steering and photograph analyses are on three LLaVA models that share an encoder. On LLaVA-1.5 the share of image-wins the head ablation reverses depends on the protocol (10.5\% against 1.3\%; Supp.~G, P). Within an item the prior weight is 1.7 to 3.0 times the between-item $b$ (Supp.~H). The prior-agreeing band map rests on as few as 19 items. Of the 19 pre-specified tests in Supp.~O (ten criteria from the analysis plan, nine written during the study), two were met, seven failed, seven gave mixed outcomes, one lost its reference run, one could not be built and one was not run; one of the two met is post hoc for three models; four further predictions of the experiment plan did not hold (Supp.~O).

\noindent\textbf{Detail behind statements of the main text.} The main text states the following results briefly; the full statements are given here. \emph{Sec. 4, eleven models.} On four of the twelve (Gemma-3-4B, both MiMo-VL stages and the pretrained InternVL3.5 stage) the held-out fit on the main split does not exceed its permutation null, but does at the median of the 50 splits. \emph{Sec. 4, how much the additive model explains.} We measure the interaction as the change in $a$ per standard deviation of the prior, as a share of $a$. Its 95\% interval lies inside $\pm$0.32 on all seven fixed-input fits and inside $\pm$0.41 on four of the six others, against a pre-specified bound of $\pm$0.5; on LFM2-VL and Gemma-3-4B it is wider than the bound. Over all 28 grids refit on every cell the interaction excludes zero on two (Supp.~Q); both MiMo-VL stages, not among the 28, show one in the boundary cells. On ViLP, a second source, both weights are positive with intervals that exclude zero on all ten open models run (eight architectures; every-cell refits: $a$ = 0.07 to 0.28, $b$ = 0.10 to 0.24). The dose adds 0.02 to 0.03 on the five models with the dose at an input size and at most 0.012 on the others, and the interaction is inside $\pm$0.42 on nine and present on Gemma-3 ($-$0.31 [$-$0.59, $-$0.04]). The held-out test passes on nine; on Idefics2 it fails on the main split and passes at the median split (Supp.~Q). \emph{Sec. 4, frontier models.} With the prior taken from the open models the prior weight is positive on all three and its interval excludes zero on one; with Gemini's own log-odds as the prior it is $-0.05$ [$-0.17$, 0.08]. \emph{Sec. 6, route through the question tokens.} With this donor the crossing is at 0.59 to 0.82 of depth, 0.04 to 0.19 later than with the gray copy on the eight models measured, one per architecture. With a blurred copy as donor (nine models) or a noised one (four) it is 0.01 to 0.09 earlier; the pre-specified equivalence with the gray donor is met on four of the nine and unresolved on five (Supp.~D, P). \emph{Sec. 6, prior-agreeing items.} on all nine models of main-paper Fig.~4; over four further sizes the largest difference is 0.07, on Gemma-3-27B. Against a pre-specified bound of 0.10 of depth, the 95\% interval of the difference lies inside it on six of the nine; it is unresolved on Llama-3.2 and Emu3, and on LFM2-VL the crossing is 0.05 earlier without a conflict (Supp.~P). \emph{Sec. 6, no separate prior subspace.} On the three LLaVA models we pre-specified a test for a separate prior subspace at the answer position, and it fails: at rank 256 the directions that carry the prior also carry the image's evidence (Supp.~E). \emph{Sec. 6, scaling the heads up.} Applied only on items where the mid-depth readout favors the prior, the scaling changes vision-win by 0 to $+$7.0 points on WHOOPS and $-$2.0 to $+$6.0 on ViLP; on Fuyu its interval excludes zero on both sources. \emph{Sec. 8, photographs.} With no image the same margin predicts the model's text-only answer at 0.85--0.94, so the readout also carries the default answer; on BakLLaVA it is 0.98 on the items a gray image does not change (Supp.~L, M). \emph{Sec. 6, band map.} except that the last band still recovers 0.38 from the image positions on Emu3, 0.23 on Qwen2.5-VL and 0.19 on Gemma-3. \emph{Sec. 6, heads.} When the heads are ranked by estimated ablation effect instead of attribution, 14 to 19 of the 20 lie after the crossing on six models and about half on LLaVA-1.5 and InstructBLIP-7B (median layer 18 of 32; Supp.~P). \emph{Sec. 6, amplifying the heads.} and the mean WHOOPS margin is above its unscaled value at every scale above 1 on all nine. \emph{Sec. 7, mean ablation of the image tokens (LLaVA-NeXT, LLaVA-1.5, BakLLaVA).} Setting every image token to its mean reverses 47\%, 55\% and 40\% of image-wins on the three models, with the same asymmetry. Setting only the subspace to its mean gives 0.91, 0.86 and 0.89 of that change in the margin, and keeping only the subspace reverses 8\%, 13\% and 7\%. \emph{Sec. 4, variation explained.} On six fits with the dose on the full-size image and no smaller fixed input (Fuyu, Idefics2, LFM2-VL, Gemma-3 at three sizes) the two terms explain 0.03 to 0.08 of held-out deviance and the dose adds 0.004 to 0.014. \emph{Sec. 4, dose at the input size.} On the four models with a fixed input size the dose adds 0.03 to 0.05 of held-out deviance at the input size. \emph{Sec. 3, depth convention.} The NaturalBench readout reads hidden-state index 20, the output of block 19 (called layer 20). \emph{Sec. 6, donor dependence of the crossing.} The crossing depth also depends moderately on the donor. It is 0.04 to 0.19 later with this donor and 0.01 to 0.09 earlier with a blurred or noised copy (Supp.~D, P). \emph{Sec. 5, size-dose control (the control at the original size does not expose the blurred fill that surrounds a shrunken object).} so the compositing operations alone do not produce the shift, although this control does not expose the blurred fill that surrounds a shrunken object. \emph{Sec. 8, photographs.} On the quarter of questions about one object the gray ratio is 1.6 to 2.0 (Supp.~L). \emph{Sec. 4, frontier API models.} On the items with a mask, the prior overrides the clean image on only 2 to 4 per model.

\noindent\textbf{Values behind the summaries of the main text.} The main text states several results in words and refers to the supplement for the values. They are collected here by section of the main paper.

\noindent\emph{Main-paper Sec.~4.} At the final layer the margin predicts the decision at 0.98 to 1.00 AUROC on eight of nine models and at 0.82 on Fuyu. On the three LLaVA models the decision agrees with free generation on 87--95\% of conflicts, and 90--99\% of image-wins disappear when the image is replaced by gray. The prior weight of Gemma-3 is 0.09 [$-$0.02, 0.19] on the main split and positive on all 50 item splits (median 0.14); on the twelve further checkpoints both weights are positive on all 50 item splits with intervals that exclude zero. The dose-by-prior term is measured as the change in $a$ per standard deviation of the prior, as a share of $a$. Its 95\% interval lies inside the pre-specified bound of $\pm$0.5 on 11 of 13 fits (all seven with the dose at a fixed input size and four of six at native resolution), and adding the term changes held-out deviance by at most 0.001. The fit from unmodified items predicts the prefix-modified ones at pseudo-$R^2$ 0.14 to 0.27. On the seven fixed-input fits the two terms together explain a median 0.05 to 0.10 of held-out deviance and the prior alone 0.01 to 0.08; the dose adds a median 0.02 to 0.05, on 98 to 100\% of splits. With the dose on the full-size image the precision weight is 0.06 [$-$0.01, 0.13], 0.07 [0.00, 0.14] and $-$0.01 [$-$0.09, 0.07] on the three models with a fixed 336- or 224-pixel input, and 0.31, 0.31 and 0.33 with the dose at the input size. Llama-3.2 tiles the image at up to 1,120 pixels, and $a$ = 0.08 [$-$0.01, 0.17] on the full-size grid against 0.22 [0.14, 0.31] with the image resized to one 560-pixel tile. For Emu3, with a budget of 512$^2$ pixels per image, $a$ is 0.10 and 0.31. On these five models the dose adds at most 0.004 of held-out deviance on the full-size grid and 0.02 to 0.05 at the input size. On the API models the precision weight is 0.27 [0.19, 0.36] on GPT-6.1 Sol, 0.18 [0.08, 0.27] on Claude Sonnet 5.5 and 0.18 [0.09, 0.28] on Gemini 4 Argon. On the items with a mask ($n$ = 58, 53, 41), 0.83, 0.91 and 0.90 keep the image's answer under an object crop and 0.31, 0.30 and 0.51 under a random crop.

\noindent\emph{Main-paper Sec.~5.} On ten of the eleven models blurring the object lowers the margin by 0.59 to 1.91 logits, and the paired difference between the object edit and the rest edit excludes zero for blur, for noise and for gray. On Gemma-3 the difference excludes zero for blur and noise but not for gray, at 4B, 12B and 27B. At half size the margin falls by 0.60 to 1.08 logits on the three LLaVA models and by 0.77 on Qwen2.5-VL, and at three-quarter size by 0.17 to 0.55. Recompositing the object at its original size moves the margin by at most 0.02. Enlarged to 1.5 times its size, the object changes the margin by $-0.34$, $-0.09$ and $-0.23$ logits on the three LLaVA models. Over six enhancement levels vision-win changes by $-4.5$ to $+0.8$ points, with no interval above zero and four below it. Cropping to the object changes vision-win by $-13.4$ to $+0.3$ points, with seven of nine intervals below zero; the crop reverses 13 to 29\% of prior-wins (a random crop reverses 2 to 16\%) and loses 7 to 19\% of image-wins. On images blurred at the strongest dose, restoring an equal area elsewhere changes vision-win by $-$2 to $+$6 points.

\noindent\emph{Main-paper Sec.~6.} The band map uses 145 to 282 conflicts per model. Over 2,000 resamples of items the 95\% interval of the crossing is narrower than 0.06 of depth on six of the nine models and at most 0.15 (Emu3). Further checkpoints cross between 0.45 and 0.72 (Supp.~P). With ViLP's prior-consistent image as donor, the question tokens recover 0.04 of the effect at mid-depth on Fuyu and 0.60 on Gemma-3, patching them alone reverses 6\% and 71\% of image-wins, and the crossing depth shifts by up to 0.19. On Gemma-3 the heads ranked by attribution reverse 1.8\% of image-wins against 0.0\% for random sets, and ten of the top 20 sit in layer 0; 20 heads ranked by the estimated effect of ablating them reverse 7.4\% [3.7, 11.7] against at most 2.5\%. Scaling the heads by 3 is evaluated on 31 to 72 held-out prior-wins per model against five layer-matched random sets. Applied to every item it changes WHOOPS vision-win by $+$0.5 to $+$7.0 points, and on ViLP it loses 3 to 12\% of image-wins, above the pre-specified 2\% bound. On prior-agreeing items the crossing is within 0.06 of depth of the conflict crossing on the nine models, and ablating the top heads lowers the margin by 0.42 to 1.58 logits on these items and by 0.71 to 2.01 on conflicts, on the eight models tested.

\noindent\emph{Main-paper Sec.~7.} The subspace recovers 63\%, 69\% and 67\% of the gray edit's effect on LLaVA-NeXT, LLaVA-1.5 and BakLLaVA and holds 93\%, 71\% and 98\% of the per-token variance. Random and low-variance subspaces of equal dimension reverse at most 1.9\% of image-wins or of prior-wins. On the four other-architecture models where the reading holds, removal reverses 27 to 62\% of image-wins and 5 to 13\% of prior-wins, against at most 8.5\% for a control. On Gemma-3, projecting out a random subspace reverses 48\% of image-wins against 30\% for the fitted one. On Idefics2 the share of the full ablation's change is 0.51 [0.47, 0.57], and the criterion requires the lower end of the interval at 0.5 or above. On Qwen2.5-VL removal reverses 44\% of image-wins against 1.5\% for a control.

\noindent\emph{Main-paper Secs.~8 and 9.} At three quarters of depth the readout predicts the decision at 0.85 to 0.93 AUROC on the models that cross by 0.59, and at 0.57 to 0.81 on those that cross later. The text-only margin predicts the WHOOPS decision at 0.56--0.68 AUROC, and adding the mid-depth readout raises a cross-validated predictor by 0.18--0.26 on eight of the twelve models measured, by 0.08--0.09 on Emu3 and Gemma-3-4B, and not on LFM2-VL or Fuyu. On NaturalBench, on the 33 to 52\% of items whose answer a gray image changes, the centered readout predicts the decision at 0.96 and 0.97 AUROC on LLaVA-NeXT and LLaVA-1.5 and at 0.81 on BakLLaVA. Over all conflicts it is 0.88--0.98 centered and 0.55--0.93 uncentered, 0.89--0.98 with the median taken from a disjoint half, and 0.92--0.99 within the yes items and the no items alone. Graying the object shifts the margin 0.9--1.2 times as much as graying the rest (about nine tenths of the image) and 2.2--2.5 times as much as graying a random region of the object's size; on the quarter of questions that concern a single object the ratio to the rest is 1.6 to 2.0. Attention steering changes conflicting-item accuracy by at most 0.7 points. Injecting the text-only answer direction raises conflicting-item accuracy by 2.9 to 13.6 points and lowers consistent-item accuracy by 2.9 to 11.8, so accuracy over all items falls by 0.6 to 1.7 points.

\noindent\textbf{Pre-specified tests and their outcomes.} \Cref{tab:pre-specified,tab:pre-specified2} list the tests whose criterion was fixed before the run and whose outcome this paper reports. The criteria of \cref{tab:pre-specified} were written in the analysis plan before any experiment. Those of its last two rows and of \cref{tab:pre-specified2} were written during the study, each before the run it applies to. Criterion A$'$ was added after the results of three models were known and is marked as post hoc for them. Each row is labeled with its outcome. Of the nineteen tests, two were met on every model, seven failed, seven gave mixed outcomes across models, one lost its reference run, one could not be built and one was not run. The experiment plan made four further predictions that are not rows of the tables, and none held: a crossing between 0.55 and 0.70 of depth on every model (measured 0.45 to 0.74; outside the range on five of fifteen models, \cref{tab:crossci} and Supp.~\ref{app:transport}); no effect of attribute type once precision and prior are in the additive model (it improves the fit on every model, Supp.~\ref{app:equation}); one function across models, with slopes within 20\% (inconclusive for every pair, Supp.~\ref{app:equation}); and a monotone shift of the weights across training stages (not seen, \cref{tab:stage}). The plan treated a crossing in the same band on prior-agreeing items as a result against a band specific to conflict; the first table reports it as the transport test.

\begin{table}[h]\centering\footnotesize\setlength{\tabcolsep}{4pt}
\caption{\textbf{Pre-specified tests.} Criterion as fixed before the run; outcome as observed. Three LLaVA models unless stated. Recorded: when the criterion was written down (day of November 2026 and time, UTC); `plan': fixed in the experiment plan before any experiment.}\label{tab:pre-specified}
\begin{tabular}{p{3.1cm}p{1.2cm}p{4.9cm}p{6.2cm}}\toprule
Test & Recorded & Pre-specified criterion & Outcome \\ \midrule
Rule, held-out fit [mixed] & plan & Held-out deviance explained above the 95th percentile of a permutation null, fit on boundary cells and scored on interior cells & Met on the eleven models of main-paper Table~1 and on eight of the twelve further checkpoints; on Gemma-3-4B, both MiMo-VL stages and the pretrained InternVL3.5 stage it fails on the main split, and on LLaVA-1.5 it fails with the dose on the full-size image (\cref{tab:cuefit}, Supp.~\ref{app:equation}) \\
Within-item prior weight [failed] & plan & Equal to the between-item $b$ within 20\% (two one-sided tests) & Fails: 1.7 to 3.0 times $b$; DIFFERENT on LLaVA-NeXT, INCONCLUSIVE on the other two, DIFFERENT on all three by item bootstrap (Supp.~\ref{app:equation}) \\
Corpus counts, color conflicts [failed] & plan & Controlled correlation with the prior of at least 0.3 & Fails: 0.12 to 0.17 on Dolma, 0.07 to 0.18 on RedPajama (\cref{tab:h6}) \\
Band map on prior-agreeing items [met] & plan & Same crossover band and $|$difference$|<0.10$ with intervals including zero at both sites classifies the handoff as transport & Transport on all three models (Supp.~\ref{app:transport}) \\
Equal-variance subspace control [not built] & plan & 256 components outside the top-1,024 block matched to the fitted subspace's variance within 5\% & Cannot be built: the remaining spectrum reaches at most 2\% of that variance; recovery per unit variance is reported instead (Supp.~\ref{app:necessity}) \\
Reproduction of the first subspace run [reference discarded] & plan & Each repeated value within 0.05 of the first run & Sufficiency: not testable; the first run on LLaVA-1.5 and LLaVA-NeXT was discarded (altered forward pass, Supp.~\ref{app:faq}). Necessity: within 0.05 wherever a first value exists \\
Text premises, one shared circuit [failed] & plan & Head-set Jaccard of at least 0.40 and cross-ablation ratio of at least 0.5 & Below the bar on all three models; outcome partial (\cref{tab:textid}) \\
Relational conflicts, union mask [not run] & plan & Steering onto the union of the two objects raises the margin & Not run \\
Object against rest on photographs (NaturalBench) [failed] & plan & Pooled object/rest ratio of the mean absolute margin shift at least 2, with its interval excluding 1, on the final and the mid-depth margin, and the scale-1.0 recomposite covering zero & Fails on all three models: pooled ratios 0.76 to 0.92, every one below 1 (\cref{tab:photoobj}) \\
Object-only enhancement (LLaVA-1.5, Idefics2, InstructBLIP-7B, LFM2-VL, Llama-3.2, Emu3, Gemma-3-12B, Qwen2.5-VL, Fuyu-8B) [failed] & plan & Sharpening or contrast applied to the object alone raises vision-win above the clean image; falsified if it stays at the clean baseline & Fails on all nine: $-4.5$ to $+0.8$ points at six levels; no interval lies above zero and four lie below it (Supp.~\ref{app:arch}) \\
Image-token subspace on other architectures [mixed] & 10th, 20:25 & At the image positions, 256 dimensions: removal reverses more image-wins than a random and an off-block control (A), more image-wins than prior-wins (B), and the subspace set to its mean gives at least half of the gap change of removing every image position, controls under 0.2 (C) & Holds on InstructBLIP-7B, LFM2-VL-3B, Fuyu-8B and Emu3. Fails on Gemma-3-12B at A: a random subspace projected out reverses 47.6\% of image-wins against 30.0\% for the fitted one (41.9\% against 36.7\% at 4B). Fails on Idefics2 at C: the share is 0.51 [0.47, 0.57], its lower end under 0.5. Other sizes: Gemma-3-27B fails A and C (share 0.28 [0.24, 0.33]); InstructBLIP-13B fails B (18.7\% of image-wins against 18.6\% of prior-wins); LLaVA-1.5-13B and Qwen2.5-VL hold; LFM2-VL-1.6B fails C (a control exceeds 0.2). A second Idefics2 checkpoint (Idefics2-8B-chatty) holds, with a share of 0.70 [0.68, 0.73] (\cref{tab:archsub,tab:archsubmean}) \\
Language-prior subspace at the answer position [failed] & 10th, 17:00 & Setting the top 256 principal components of text-only runs to their mean at the answer position reverses more prior-wins than image-wins and more than a random and an image-run basis of equal rank; its share of the readout tracks the text-only prior and does not move with blur & Fails on all three: 21 to 33\% of prior-wins against 4 to 6\% of image-wins and 5 to 7\% for the random basis, but 19 to 25\% for the image-run basis, with overlapping intervals; its share of the readout falls by 0.40 to 1.84 logits under blur and tracks the prior no better than the remainder (\cref{tab:priorsub}) \\ \bottomrule
\end{tabular}\end{table}

\begin{table}[h]\centering\footnotesize\setlength{\tabcolsep}{4pt}
\caption{\textbf{Pre-specified tests, continued.} Columns as in \cref{tab:pre-specified}.}\label{tab:pre-specified2}
\begin{tabular}{p{3.1cm}p{1.2cm}p{4.9cm}p{6.2cm}}\toprule
Test & Recorded & Pre-specified criterion & Outcome \\ \midrule
Heads on Gemma-3, ranked by estimated ablation effect [mixed] & 10th, 14:38 & The 20 selected heads reverse more image-wins than every one of 20 layer-matched random sets, and their median layer is after the crossing & Met at 4B (14.1\% against at most 2.5\%) and at 12B (7.4\% against at most 2.5\%). Formally met at 27B, where the set reverses 1.9\% (3 items) against at most 0.6\%; we do not count 27B as showing the heads (\cref{tab:headgrad}) \\
Subspace criterion A$'$ [met] & 10th, 23:15 & The fitted subspace set to its mean reverses more image-wins than a random and an off-block control set to their means & Added after the results of InstructBLIP-7B, LFM2-VL-3B and Gemma-3-12B were known, so post hoc for those three, which meet it. Holds on the six models analyzed afterwards (Idefics2, Fuyu, Emu3, Gemma-3-4B, Gemma-3-27B, InstructBLIP-13B) and on LLaVA-1.5-13B (Supp.~\ref{app:arch}) \\
Rule on a second source (ViLP) [mixed] & 11th, 18:55 & Both weights positive with intervals that exclude zero, and held-out deviance above the 95th percentile of the permutation null & Met on nine of ten models. On Idefics2 both weights are positive and the held-out half fails on the main split ($-$0.070 against 0.034) and passes at the median of 50 splits (Supp.~\ref{app:rulecmp}) \\
Head scaling as a correction [failed] & 11th, 19:20--19:40 & (1) At a scale of 2 or 3 the selected heads reverse more prior-wins than every random set, on WHOOPS and on ViLP; (2) the margin rises with the scale; (3) at the smallest scale meeting (1), at most 2\% of image-wins and of prior-agreeing answers are lost. Closed loop: a correction if its paired interval excludes zero on both sources & Not met in full on any of nine models (eight architectures): (1) holds on both sources on five and (2) on WHOOPS on all nine; (3) holds on none, since where (1) holds 2.7 to 12.3\% of the ViLP image-wins are lost. The closed loop is a correction on Fuyu only (\cref{tab:dial}) \\
No dose-by-prior interaction [mixed] & 11th, 20:45--21:00 & Absent on a fit if the 95\% interval of the change in $a$ per standard deviation of the prior, as a share of $a$, lies inside $\pm$0.5; present if the interval excludes zero & Inside the bound on all seven WHOOPS fits with the dose at a fixed input size, on four of the six native-resolution fits and on nine of the ten ViLP fits. Wider than the bound on LFM2-VL-3B and Gemma-3-4B and on full-size grids where $a$ is near zero. Present on two of 28 fits (Gemma-3-12B on ViLP and the full-size grid of Llama-3.2, both negative) (Supp.~\ref{app:rulecmp}) \\
Handoff without a conflict [mixed] & 11th, 21:10--21:30 & The same on a model if the 95\% interval of the crossing on prior-agreeing items minus the crossing on conflicts lies inside $\pm$0.10 of depth & Eight bands: inside on six of the nine models of main-paper Fig.~4, not resolved on Llama-3.2 and Emu3, earlier without a conflict on LFM2-VL-3B. One layer per band: inside on six of nine (Supp.~\ref{app:arch}) \\
Handoff with another edit as donor [mixed] & 11th, 22:37 & Independent of the edit if the donor's crossing is in the same or the adjacent band as the gray crossing and the 95\% interval of the difference lies inside $\pm$0.10 of depth; dependent if the interval lies outside & Blur: met on four of nine models, not resolved on five, outside on none. Noise: same band on all four; interval inside on three and 0.004 past the margin on Idefics2 (\cref{tab:donor}) \\ \bottomrule
\end{tabular}\end{table}

\FloatBarrier
\section{Further architectures}\label{app:arch}
\noindent\textit{Supports main-paper Secs.~4 to 7. Every measurement on the architectures beyond the LLaVA family, under one common protocol.}\par\smallskip
\subsection{Protocol and calibration}
\noindent\textbf{Protocol.} One protocol measures every model; only the prompt format, the image positions, the decoder layers, the final norm and the unembedding are model-specific. Per model it scores every conflict with and without the image by the protocol of Supp.~\ref{app:data}, computes the logit-lens margin of the first answer token at every layer, runs the band map on up to 300 WHOOPS conflicts and all prior-agreeing items that the gray edit moves by at least 0.5 logits, runs the blur and noise grid on the prior-valid WHOOPS conflicts with the fit of Supp.~\ref{app:equation}, attributes and ablates heads, and applies the object edits. A repeated forward pass on 40 items reproduces every margin exactly on every model for which the check was recorded (all but Emu3; Qwen2.5-VL was checked in its model-specific run), the final-layer lens equals the gap, and the band-map identity check (clean activations patched into the clean run) returns the clean gap with zero deviation. For cross-attention (Llama-3.2) the image site of the band map is the vision states read by every cross-attention layer from the band onward, which is what patching image tokens does in a decoder-only model. Fuyu's tokenizer has multi-word pieces, so its answers are scored from the common prefix of the candidate strings. Emu3 receives 512 $\times$ 512 images (4,096 codes).

\noindent\textbf{Calibration on LLaVA-1.5.} The common protocol reproduces the model-specific analyses: 417 prior-valid WHOOPS conflicts (416), vision-win 0.734 (0.733), crossover band 20--23 in both conditions, readout at 0.625 of depth 0.83 (0.83), precision weight on the full-size grid 0.06 [$-$0.01, 0.13] (the same), gray-object and gray-rest shifts $-0.47$ and $-0.09$ (the same). Per-band recoveries differ by up to 0.17 in bands 20 to 31, on a 145-item sample with a different answer-token rule. The head ablation reverses 10.5\% of image-wins with a gap change of $-0.88$; the model-specific run of Supp.~\ref{app:steer}, with its own head selection and decision rule, gives 1.3\% and $-0.95$.

\begin{table}[h]\centering\footnotesize\setlength{\tabcolsep}{4pt}
\caption{\textbf{Outcomes and readout by depth} (common protocol). Prior-valid WHOOPS conflicts, vision-win by source (ViLP and Visual CounterFact color where run), and AUROC of the untrained logit-lens margin for the WHOOPS decision at five relative depths; the last column is the text-only control at 0.625.}\label{tab:archout}
\begin{tabular}{lcccccccccc}\toprule
Model & $n$ & WHOOPS & ViLP & VCF & 0.25 & 0.50 & 0.625 & 0.75 & 1.0 & text \\ \midrule
LLaVA-1.5 & 417 & 0.734 & -- & -- & 0.59 & 0.71 & 0.83 & 0.88 & 0.98 & 0.79 \\
BakLLaVA & 421 & 0.743 & -- & -- & 0.63 & 0.69 & 0.80 & 0.86 & 0.98 & 0.80 \\
InstructBLIP & 403 & 0.787 & 0.397 & 0.978 & 0.56 & 0.75 & 0.89 & 0.93 & 0.98 & 0.78 \\
Idefics2 & 418 & 0.816 & 0.499 & 0.956 & 0.67 & 0.76 & 0.83 & 0.90 & 0.99 & 0.84 \\
LFM2-VL & 411 & 0.771 & 0.551 & 0.974 & 0.55 & 0.61 & 0.63 & 0.78 & 0.98 & 0.72 \\
Fuyu & 391 & 0.675 & 0.473 & 0.960 & 0.49 & 0.50 & 0.54 & 0.72 & 0.82 & 0.68 \\
Emu3 & 407 & 0.641 & 0.241 & 0.963 & 0.60 & 0.70 & 0.74 & 0.81 & 1.00 & 0.82 \\
Llama-3.2 & 423 & 0.745 & 0.756 & 0.983 & 0.55 & 0.68 & 0.82 & 0.90 & 0.99 & 0.81 \\
Gemma-3 & 423 & 0.806 & 0.716 & 0.981 & 0.59 & 0.62 & 0.82 & 0.93 & 1.00 & 0.75 \\ \bottomrule
\end{tabular}\end{table}

\begin{table}[h]\centering\footnotesize\setlength{\tabcolsep}{3.2pt}
\caption{\textbf{Band maps} (WHOOPS conflicts moved by at least 0.5 logits by the gray edit). Share of the gray edit's effect recovered by patching the gray copy's activations into the clean run in each of eight bands of equal relative depth, at the image site / at the answer position. $n$: conflicts (prior-agreeing items). Crossing depths are in \cref{tab:crossci}. In the last band the image site still recovers 0.13 on Fuyu and 0.38 on Emu3.}\label{tab:archband}
\begin{tabular}{lccccccccc}\toprule
Model & $n$ & band 1 & 2 & 3 & 4 & 5 & 6 & 7 & 8 \\ \midrule
LLaVA-1.5 & 145 (19) & 1.00/0.00 & 1.00/0.00 & 0.99/0.01 & 0.92/0.11 & 0.79/0.32 & 0.45/0.71 & 0.18/0.88 & 0.07/0.98 \\
InstructBLIP & 173 (46) & 1.00/0.00 & 1.01/0.00 & 0.96/0.03 & 0.88/0.15 & 0.67/0.45 & 0.36/0.71 & 0.09/0.89 & 0.04/0.98 \\
Idefics2 & 197 (42) & 1.00/0.00 & 0.99/0.00 & 0.93/0.02 & 0.79/0.08 & 0.61/0.32 & 0.33/0.56 & 0.22/0.72 & 0.09/0.93 \\
LFM2-VL & 175 (39) & 1.00/0.00 & 0.98/0.01 & 0.85/0.03 & 0.73/0.06 & 0.59/0.18 & 0.51/0.29 & 0.29/0.80 & 0.10/0.98 \\
Fuyu & 208 (44) & 1.00/0.01 & 0.98/0.01 & 0.97/0.01 & 0.96/0.02 & 0.95/0.04 & 0.91/0.22 & 0.62/0.67 & 0.13/1.00 \\
Emu3 & 218 (49) & 1.00/0.00 & 0.99/0.01 & 0.97/0.05 & 0.80/0.17 & 0.58/0.31 & 0.52/0.42 & 0.40/0.52 & 0.38/0.71 \\
Llama-3.2 & 191 (35) & 1.00/0.00 & 0.83/0.01 & 0.70/0.04 & 0.62/0.19 & 0.49/0.42 & 0.36/0.57 & 0.25/0.73 & 0.09/0.98 \\
Gemma-3 & 226 (51) & 1.00/0.00 & 1.00/0.00 & 0.99/0.00 & 0.97/0.04 & 0.66/0.41 & 0.44/0.56 & 0.33/0.65 & 0.19/0.97 \\ \bottomrule
\end{tabular}\end{table}

\begin{table}[h]\centering\footnotesize\setlength{\tabcolsep}{5pt}
\caption{\textbf{Crossing depth with intervals} (main-paper Fig.~4). Relative depth at which the image-site and answer-site curves of the gray-copy band map cross, interpolated on the first layer of each band, on WHOOPS conflicts and on prior-agreeing items. Brackets are 95\% intervals over 2,000 bootstrap resamples of items (seed 0). Upper block: one model per architecture and Qwen2.5-VL, ordered by crossing depth; lower block: other sizes. The last column is prior-agreeing minus conflict.}\label{tab:crossci}
\begin{tabular}{lccccc}\toprule
Model & $n$ & Conflicts & $n$ & Prior-agreeing & Difference \\ \midrule
Llama-3.2-11B-Vision & 191 & 0.53 [0.48, 0.59] & 35 & 0.50 [0.40, 0.59] & $-$0.04 \\
InstructBLIP-7B & 173 & 0.55 [0.53, 0.57] & 46 & 0.55 [0.52, 0.57] & 0.00 \\
Idefics2-8B & 197 & 0.57 [0.54, 0.59] & 42 & 0.53 [0.49, 0.56] & $-$0.04 \\
LLaVA-1.5-7B & 145 & 0.58 [0.57, 0.59] & 19 & 0.57 [0.54, 0.60] & $-$0.01 \\
Gemma-3-12B & 226 & 0.59 [0.54, 0.64] & 51 & 0.59 [0.53, 0.67] & 0.00 \\
LFM2-VL-3B & 175 & 0.66 [0.65, 0.68] & 39 & 0.61 [0.49, 0.66] & $-$0.05 \\
Emu3-Chat & 218 & 0.68 [0.60, 0.75] & 49 & 0.69 [0.55, 0.83] & $+$0.01 \\
Qwen2.5-VL-7B & 201 & 0.73 [0.71, 0.75] & 50 & 0.70 [0.68, 0.72] & $-$0.03 \\
Fuyu-8B & 208 & 0.74 [0.73, 0.76] & 44 & 0.73 [0.70, 0.76] & $-$0.01 \\
\midrule
LLaVA-1.5-13B & 146 & 0.53 [0.51, 0.55] & 24 & 0.48 [0.45, 0.52] & $-$0.04 \\
InstructBLIP-13B & 160 & 0.45 [0.44, 0.47] & 66 & 0.45 [0.42, 0.47] & 0.00 \\
Gemma-3-4B & 238 & 0.61 [0.59, 0.64] & 48 & 0.60 [0.57, 0.66] & $-$0.01 \\
Gemma-3-27B & 235 & 0.59 [0.56, 0.62] & 38 & 0.52 [0.48, 0.57] & $-$0.07 \\ \bottomrule
\end{tabular}\end{table}

\subsection{The crossing: prior-agreeing items and other donors}
\noindent\textbf{The crossing without a conflict.} We pre-specified a bound: the crossing counts as the same on a model if the 95\% interval of the crossing on prior-agreeing items minus the crossing on conflicts lies inside $\pm$0.10 of depth (\cref{tab:pre-specified2}; the same 0.10 margin as the band-map test of \cref{tab:pre-specified}). Items are resampled independently in the two sets, 2,000 times. \Cref{tab:agreecross} gives the difference at eight bands and at one layer per band. Where the two crossings differ, the handoff is slightly earlier when the prior agrees.

\begin{table}[h]\centering\footnotesize\setlength{\tabcolsep}{5pt}
\caption{\textbf{Crossing on prior-agreeing items minus crossing on conflicts}, in relative depth with 95\% intervals. Reading against the $\pm$0.10 bound: inside, not resolved (the interval crosses the bound), or earlier (the interval excludes zero). Right: the same difference, and the conflict crossing, with one layer per band on the nine models of main-paper Fig.~4. Dash: not run.}\label{tab:agreecross}
\begin{tabular}{lcccc}\toprule
 & \multicolumn{2}{c}{Eight bands} & \multicolumn{2}{c}{One layer per band} \\ \cmidrule(lr){2-3}\cmidrule(lr){4-5}
Model & Difference & Reading & Difference & Conflict crossing \\ \midrule
LLaVA-1.5 & $-$0.01 [$-$0.04, 0.02] & inside & $-$0.01 [$-$0.03, 0.01] & 0.63 [0.62, 0.63] \\
InstructBLIP-7B & 0.00 [$-$0.04, 0.03] & inside & $-$0.02 [$-$0.06, 0.02] & 0.60 [0.58, 0.61] \\
Idefics2 & $-$0.04 [$-$0.09, 0.00] & inside & $-$0.02 [$-$0.05, 0.01] & 0.61 [0.59, 0.63] \\
Gemma-3-12B & 0.00 [$-$0.08, 0.09] & inside & $-$0.02 [$-$0.09, 0.08] & 0.65 [0.62, 0.68] \\
Qwen2.5-VL & $-$0.03 [$-$0.06, 0.00] & inside & $-$0.04 [$-$0.07, $-$0.00] & 0.77 [0.74, 0.79] \\
Fuyu & $-$0.01 [$-$0.04, 0.02] & inside & $-$0.04 [$-$0.06, $-$0.01] & 0.78 [0.77, 0.81] \\
Llama-3.2 & $-$0.04 [$-$0.15, 0.08] & not resolved & $-$0.01 [$-$0.18, 0.11] & 0.59 [0.58, 0.70] \\
Emu3 & 0.01 [$-$0.15, 0.17] & not resolved & $+$0.02 [$-$0.04, 0.20] & 0.71 [0.62, 0.75] \\
LFM2-VL-3B & $-$0.05 [$-$0.17, $-$0.00] & earlier & $-$0.04 [$-$0.13, $-$0.00] & 0.72 [0.71, 0.72] \\ \midrule
LLaVA-1.5-13B & $-$0.04 [$-$0.09, 0.00] & inside & -- & -- \\
InstructBLIP-13B & 0.00 [$-$0.04, 0.03] & inside & -- & -- \\
Gemma-3-4B & $-$0.01 [$-$0.05, 0.06] & inside & -- & -- \\
Gemma-3-27B & $-$0.07 [$-$0.12, $-$0.02] & earlier & -- & -- \\ \bottomrule
\end{tabular}\end{table}

\noindent\emph{Notes on \cref{tab:agreecross}.} At eight bands the point estimate is below zero on eleven of the thirteen models and above it on two (Gemma-3-12B, $+$0.001, and Emu3, $+$0.013). At one layer per band the difference is inside the margin on six of the nine, with LFM2-VL, Llama-3.2 and Emu3 as at eight bands, and below zero on eight of the nine. The single-layer conflict crossings are 0.03 to 0.07 of depth later than the eight-band estimates of \cref{tab:crossci}, which interpolate between band starts, and they keep the order of the models, except that Emu3 and LFM2-VL, 0.02 apart at eight bands, change places. On Emu3 the handoff is spread out: at single-layer resolution the image site stays at 0.47 to 0.49 and the answer site at 0.33 to 0.39 from layer 18 to layer 22 of 32, and they cross at layer 23, which is why its crossing has the widest interval.

\noindent\textbf{The handoff with another edit as donor.} The band map patches a grayscale copy. To ask whether the crossing depth belongs to the model or to that edit, we repeated the patching, on the same clean runs, with a blurred copy as donor (Gaussian blur of radius 8; in model-input pixels on the fixed-input models LLaVA-1.5 and InstructBLIP-7B, in full-size pixels on the native-resolution models Idefics2, LFM2-VL-3B, Fuyu and Qwen2.5-VL, and scaled to a 560-pixel short side on Llama-3.2, an 896-pixel one on Gemma-3-12B and a 512-pixel one on Emu3, the size of one input tile or of the input budget). The pre-specified criterion: the depth does not depend on the edit if the blur crossing falls in the same band as the gray crossing or the adjacent one and the 95\% interval of the difference lies inside $\pm$0.10 of depth, and it depends on the edit if that interval lies outside. Items are those the donor moves by at least 0.5 logits; every identity check is exact. On all nine models the early bands recover the whole effect from the image positions and none from the answer position; the late bands do the reverse, except that the last band still recovers 0.33 from the image positions on Emu3 and 0.19 on Qwen2.5-VL, as with gray.

\Cref{tab:donor} gives the crossings. The blur crossing is in the same band on LLaVA-1.5, InstructBLIP-7B, Fuyu, Qwen2.5-VL, Gemma-3-12B and Emu3 and in the adjacent, earlier band on Idefics2, LFM2-VL-3B and Llama-3.2, so the band part of the criterion holds on all nine. The crossing is earlier with blur on every model, by 0.01 to 0.09 of depth, and the difference excludes zero on five. The interval of the difference is inside the margin on LLaVA-1.5, InstructBLIP-7B, Fuyu and Qwen2.5-VL; on LFM2-VL-3B it is inside on the items both maps kept and reaches 0.01 past it over all items; on Idefics2 the point estimate is at the margin ($-$0.09 to $-$0.10) and the interval crosses it; on Llama-3.2, Gemma-3-12B and Emu3, whose gray crossings have the widest intervals of the nine, the point estimate is $-$0.06, $-$0.06 and $-$0.03 and the interval reaches $-$0.13, $-$0.12 and $-$0.11. The criterion is therefore met on four models and unresolved on five, and on no model does the interval lie outside the margin. On prior-agreeing items the difference is $-$0.05 to $-$0.06 on LLaVA-1.5, InstructBLIP-7B, LFM2-VL-3B and Llama-3.2, $-$0.03 on Qwen2.5-VL, $-$0.02 on Gemma-3-12B, $-$0.01 on Fuyu and 0.00 on Idefics2; on Emu3 it is $-$0.08 with an interval from $-$0.33 to 0.16. Together with the paired-image donor of Supp.~D, where the crossing is 0.04 to 0.19 later than with gray, the crossing shifts by at most 0.19 of depth relative to the gray donor across the three other donors: earlier when detail is degraded, later when the donor is another image of the scene.

\noindent\textbf{Noise as a further donor.} To see whether the earlier crossing is specific to blur we repeated the run on the four native-resolution models with Gaussian pixel noise of standard deviation 60 as donor (the strongest level of the dose grid; full-size noise does not reach the fixed-input models, Supp.~\ref{app:arch}). The crossing is in the same band as with gray on all four and again earlier (lower block of \cref{tab:donor}): by 0.02 on Qwen2.5-VL, by 0.03 on LFM2-VL-3B and Fuyu and by 0.07 on Idefics2. The interval is inside the margin on Fuyu and Qwen2.5-VL and, over all items, on LFM2-VL-3B; on Idefics2 it reaches 0.004 past it. On prior-agreeing items the difference is 0.00 to $-$0.04 with intervals that include zero. Blur and noise both degrade detail and leave color; gray removes color and leaves detail. The two detail edits cross earlier than the color edit on every model measured, by 0.01 to 0.09 of depth, which is the order one would expect if color evidence reaches the answer position slightly later than shape evidence; we have not tested this explanation directly.

\begin{table}[h]\centering\footnotesize\setlength{\tabcolsep}{3pt}
\caption{\textbf{Crossing depth with a blurred or a noised donor against the gray donor} (WHOOPS conflicts; 95\% intervals over 2,000 resamples of items). Band: index of the first band in which the answer site recovers at least as much as the image site. Common: items kept by both maps. Intervals of the gray crossing are resampled for each comparison and can differ from \cref{tab:crossci} by 0.01.}\label{tab:donor}
\begin{tabular}{@{}lcccc@{}}\toprule
Model & Gray ($n$, band) & Donor ($n$, band) & Donor $-$ gray, all items & Common ($n$) \\ \midrule
\multicolumn{5}{@{}l}{\emph{Blur 8 as donor}} \\
LLaVA-1.5 & 0.58 [0.57, 0.59] (145, 5) & 0.56 [0.55, 0.58] (267, 5) & $-$0.017 [$-$0.037, 0.002] & $-$0.017 [$-$0.044, 0.006] (134) \\
InstructBLIP-7B & 0.55 [0.53, 0.57] (173, 5) & 0.51 [0.49, 0.53] (269, 5) & $-$0.035 [$-$0.062, $-$0.009] & $-$0.039 [$-$0.070, $-$0.006] (153) \\
Idefics2 & 0.57 [0.54, 0.59] (197, 5) & 0.48 [0.45, 0.51] (266, 4) & $-$0.094 [$-$0.129, $-$0.053] & $-$0.100 [$-$0.144, $-$0.049] (172) \\
LFM2-VL-3B & 0.66 [0.65, 0.68] (175, 6) & 0.60 [0.56, 0.64] (240, 5) & $-$0.064 [$-$0.109, $-$0.020] & $-$0.036 [$-$0.098, $-$0.002] (147) \\
Fuyu & 0.74 [0.73, 0.76] (208, 6) & 0.73 [0.71, 0.75] (246, 6) & $-$0.010 [$-$0.035, 0.015] & $-$0.009 [$-$0.034, 0.015] (172) \\
Qwen2.5-VL & 0.73 [0.71, 0.75] (201, 6) & 0.71 [0.69, 0.72] (267, 6) & $-$0.025 [$-$0.055, 0.006] & $-$0.011 [$-$0.043, 0.022] (185) \\
Llama-3.2 & 0.53 [0.48, 0.59] (191, 5) & 0.47 [0.44, 0.50] (273, 4) & $-$0.064 [$-$0.128, $-$0.004] & $-$0.056 [$-$0.129, 0.018] (173) \\
Gemma-3-12B & 0.59 [0.54, 0.65] (226, 5) & 0.53 [0.49, 0.56] (265, 5) & $-$0.057 [$-$0.123, $-$0.003] & $-$0.066 [$-$0.135, $-$0.005] (203) \\
Emu3 & 0.68 [0.60, 0.75] (218, 6) & 0.65 [0.60, 0.70] (279, 6) & $-$0.027 [$-$0.110, 0.058] & $-$0.032 [$-$0.124, 0.050] (203) \\ \midrule
\multicolumn{5}{@{}l}{\emph{Noise 60 as donor}} \\
Idefics2 & 0.57 [0.55, 0.60] (197, 5) & 0.50 [0.48, 0.52] (231, 5) & $-$0.070 [$-$0.104, $-$0.037] & $-$0.072 [$-$0.110, $-$0.033] (157) \\
LFM2-VL-3B & 0.66 [0.65, 0.68] (175, 6) & 0.63 [0.59, 0.65] (173, 6) & $-$0.030 [$-$0.077, $-$0.008] & $-$0.047 [$-$0.107, $-$0.014] (108) \\
Fuyu & 0.74 [0.73, 0.76] (208, 6) & 0.71 [0.68, 0.74] (254, 6) & $-$0.030 [$-$0.064, 0.004] & $-$0.040 [$-$0.077, $-$0.002] (180) \\
Qwen2.5-VL & 0.73 [0.71, 0.75] (201, 6) & 0.71 [0.69, 0.74] (244, 6) & $-$0.018 [$-$0.051, 0.017] & $-$0.013 [$-$0.045, 0.024] (173) \\ \bottomrule
\end{tabular}\end{table}

\subsection{Qwen2.5-VL, and larger and smaller models of an architecture}
\noindent\textbf{Qwen2.5-VL and four models of another size.} Qwen2.5-VL was run under the common protocol, and four architectures were run at a second size: LLaVA-1.5-13B (same encoder, projector type and training recipe as LLaVA-1.5-7B, with a 40-layer Vicuna-13B), Gemma-3-4B and Gemma-3-27B (same architecture as Gemma-3-12B, SigLIP tokens pooled to 256, which attend to one another bidirectionally in the released implementation, with a 34-layer and a 62-layer language model), and InstructBLIP-Vicuna-13B (same Q-Former architecture as InstructBLIP-7B with a 40-layer Vicuna-13B). The four other sizes use all 1,593 items. \Cref{tab:fivefit} gives the outcomes and the additive model, \cref{tab:fiveband} the band maps, \cref{tab:fivepath} the pathway and the heads, and \cref{tab:fiveobj} the object edits.

\begin{table}[h]\centering\footnotesize\setlength{\tabcolsep}{4pt}
\caption{\textbf{Outcomes and the additive model on five further models.} Vision-win on prior-valid conflicts; $a$, $b$ with 95\% intervals; held-out deviance explained against the 95th percentile of its permutation null; median weights over 50 item splits (both weights are positive on 50 of 50 splits on each model listed). $^\ast$: weights from the per-model run of Supp.~\ref{app:equation} (407 conflicts, vision-win 0.789), reported in main-paper Table~1. Dash: not reported here.}\label{tab:fivefit}
\begin{tabular}{lccccc}\toprule
 & Qwen2.5-VL & LLaVA-1.5-13B & Gemma-3-4B & Gemma-3-27B & InstructBLIP-13B \\ \midrule
Dose grid & full & 336 & full & full & 224 \\
Prior-valid WHOOPS conflicts & 410 & 425 & 421 & 428 & 324 \\
Vision-win, WHOOPS & 0.798 & 0.744 & 0.803 & 0.811 & 0.784 \\
Vision-win, ViLP / color & -- & 0.369 / 0.975 & 0.701 / 0.976 & 0.734 / 0.975 & 0.413 / 0.970 \\
$a$ & $^\ast$ & 0.25 [0.17, 0.34] & 0.13 [0.04, 0.22] & 0.18 [0.10, 0.26] & 0.29 [0.19, 0.39] \\
$b$ & $^\ast$ & 0.22 [0.10, 0.35] & 0.15 [0.06, 0.25] & 0.13 [0.03, 0.23] & 0.05 [0.00, 0.09] \\
Held out / null 95th & -- & 0.132 / 0.041 & 0.008 / 0.017 & 0.050 / 0.033 & 0.049 / 0.025 \\
Median $a$ / $b$ over splits & -- & 0.27 / 0.20 & 0.14 / 0.13 & 0.19 / 0.14 & 0.32 / 0.05 \\
Vision-win at blur 8 & -- & 0.362 & 0.701 & 0.659 & 0.407 \\
Vision-win at noise 60 & -- & -- & -- & 0.736 & 0.682 \\ \bottomrule
\end{tabular}\end{table}

\noindent\emph{Notes on \cref{tab:fivefit}.} On Gemma-3-4B the held-out deviance on the main split is below the null, and it is 0.073 at the median split. On Gemma-3-27B the interval of the prior weight excludes zero, where it includes zero at 12B. On InstructBLIP-13B the text-only answer is valid on fewer items than at 7B (1,033 prior-valid conflicts over all sources), the lower bound of $b$ is 0.004, and the prior weight is smaller than at 7B (0.18).

\begin{table}[h]\centering\footnotesize\setlength{\tabcolsep}{6pt}
\caption{\textbf{Band maps on five further models.} Share of the gray edit's effect recovered at the image site / at the answer position, per band of layers (band 1 is the earliest), with the number of conflicts and of prior-agreeing items mapped.}\label{tab:fiveband}
\begin{tabular}{lccccc}\toprule
Band & Qwen2.5-VL & LLaVA-1.5-13B & Gemma-3-4B & Gemma-3-27B & InstructBLIP-13B \\ \midrule
Conflicts / prior-agreeing & 201 / 50 & 146 / 24 & 238 / 48 & 235 / 38 & 160 / 66 \\ \midrule
1 & 1.00/0.00 & 1.00/0.00 & 1.00/0.00 & 1.00/0.00 & 1.00/$-$0.01 \\
2 & 0.99/0.00 & 0.99/0.01 & 1.00/0.01 & 1.00/0.00 & 0.99/0.00 \\
3 & 0.97/0.01 & 0.93/0.06 & 1.01/0.01 & 1.00/0.00 & 0.91/0.12 \\
4 & 0.95/0.03 & 0.86/0.23 & 0.93/0.06 & 0.98/0.08 & 0.73/0.34 \\
5 & 0.90/0.11 & 0.62/0.50 & 0.77/0.19 & 0.66/0.35 & 0.47/0.70 \\
6 & 0.72/0.32 & 0.32/0.73 & 0.54/0.58 & 0.37/0.51 & 0.16/0.87 \\
7 & 0.51/0.61 & 0.13/0.79 & 0.27/0.74 & 0.26/0.60 & 0.04/0.90 \\
8 & 0.23/0.92 & 0.10/0.92 & 0.10/0.90 & 0.17/0.95 & 0.03/0.98 \\ \bottomrule
\end{tabular}\end{table}

\begin{table}[h]\centering\footnotesize\setlength{\tabcolsep}{4pt}
\caption{\textbf{Pathway and heads on five further models.} Crossing in relative depth; readout AUROC at 0.625 / 0.75 of depth; decisions reversed by mean-ablating the top 20 heads, with the range over random sets; mean gap change in logits. Dash: not reported.}\label{tab:fivepath}
\begin{tabular}{lccccc}\toprule
 & Qwen2.5-VL & LLaVA-1.5-13B & Gemma-3-4B & Gemma-3-27B & InstructBLIP-13B \\ \midrule
Crossing & 0.73 & 0.53 & 0.61 & 0.59 & 0.45 \\
Crossing, prior-agreeing items & -- & 0.48 & same band & -- & 0.45 \\
Other positions recover at most & -- & -- & 0.23 [0.17, 0.28] & 0.35 [0.29, 0.40] & 0.16 [0.13, 0.19] \\
Readout at 0.625 / 0.75 & 0.58 / 0.57 & 0.84 / 0.91 & 0.69 / 0.93 & 0.89 / 0.95 & 0.86 / 0.89 \\
Text-only control at 0.625 & -- & 0.85 & 0.76 & 0.84 & 0.61 to 0.63 \\ \midrule
Median layer of the top 20 heads & 26 of 28 & 27 of 40 & -- & 0 & 26 of 40 \\
Image-wins in the evaluation half & 165 & 149 & 162 & 160 & 129 \\
Image-wins reversed (\%) & 7.3 [3.6, 11.5] & 9.4 [5.4, 14.1] & 23.5 & 0.0 & 7.8 \\
Random sets, image-wins (\%) & 1.8 to 2.4 & at most 1.3 & 1.9--3.1 & 0.0 & 0.0 \\
Prior-wins reversed (\%) & -- & 3.9 & 21.1 & -- & 3.1 \\
Mean gap change & -- & $-0.87$ [$-1.01$, $-0.73$] & $-$5.56 & -- & $-0.89$ \\ \bottomrule
\end{tabular}\end{table}

\begin{table}[h]\centering\footnotesize\setlength{\tabcolsep}{4pt}
\caption{\textbf{Object edits on five further models.} Margin shift in logits with 95\% intervals, on the conflicts with a mask. Left: the gray edit on the object, on the rest of the image, and on a region of the object's size elsewhere. Right: object minus rest under each edit. Dash: not reported.}\label{tab:fiveobj}
\begin{tabular}{lccccccc}\toprule
 & & \multicolumn{3}{c}{Gray edit on} & \multicolumn{3}{c}{Object minus rest} \\ \cmidrule(lr){3-5}\cmidrule(lr){6-8}
Model & $n$ & object & rest & elsewhere & gray & blur & noise \\ \midrule
Qwen2.5-VL & 351 & $-0.94$ & $-0.48$ & $-0.04$ & $-0.46$ [$-0.79$, $-0.16$] & $-0.79$ [$-1.25$, $-0.36$] & $-0.73$ [$-1.05$, $-0.38$] \\
LLaVA-1.5-13B & 360 & $-0.48$ & $-0.15$ & $-0.04$ & $-0.33$ [$-0.52$, $-0.15$] & $-0.56$ [$-0.76$, $-0.37$] & $-0.30$ [$-0.43$, $-0.16$] \\
Gemma-3-4B & 362 & -- & -- & -- & $-0.37$ [$-0.94$, $+0.17$] & $-1.39$ [$-2.21$, $-0.61$] & $-0.73$ [$-1.39$, $-0.06$] \\
Gemma-3-27B & -- & -- & -- & -- & $-0.22$ [$-0.71$, $+0.25$] & $-0.73$ [$-1.37$, $-0.03$] & $-1.22$ [$-1.75$, $-0.67$] \\
InstructBLIP-13B & 274 & $-0.60$ & $-0.08$ & $-0.09$ & $-0.52$ [$-0.74$, $-0.30$] & $-0.70$ [$-1.17$, $-0.20$] & $-0.47$ [$-0.73$, $-0.18$] \\ \bottomrule
\end{tabular}\end{table}

\noindent\emph{Notes on \cref{tab:fivepath}.} \emph{Crossing.} Qwen2.5-VL crosses at 0.72 in the model-specific run of Supp.~\ref{app:transport}. LLaVA-1.5-13B crosses at 0.53 against 0.58 at 7B, Gemma-3-4B at 0.61 against 0.59 at 12B (Gemma-3-27B: 0.59, against 0.61 at 4B and 0.59 at 12B), and InstructBLIP-13B at 0.45 against 0.55 at 7B. On Gemma-3-4B the other prompt positions peak at layers 17--20, and at 12B they recover 0.34 [0.29, 0.40] at layers 24--29. \emph{Readout.} On Qwen2.5-VL the readout is 0.90 at layer 24 of 28 and 0.99 at the last layer. On InstructBLIP-13B the text-only control stays at 0.61 to 0.63 at every depth. \emph{Heads.} Qwen2.5-VL was compared with three random sets, LLaVA-1.5-13B with 20 (0.8\% of image-wins on average) and InstructBLIP-13B with three (0.0\% of prior-wins as well). On LLaVA-1.5-13B and InstructBLIP-13B the share of positive attribution is 0.47 and 0.49 and the correlation between halves 0.95 and 0.95; the random sets change the mean gap by $-0.03$ to $+0.08$ and by $-0.01$ to $+0.06$; and on LLaVA-1.5-13B the gap falls by $-0.50$ [$-0.66$, $-0.37$] on 71 prior-agreeing items. On Gemma-3-4B the random sets reverse 2.6--5.3\% of prior-wins and six of the 20 heads are in layer 0. The ablation there disrupts the model in both directions, so we do not count Gemma-3-4B as a model with decision-specific heads. On Gemma-3-27B the heads result fails as at 12B. \emph{Object (\cref{tab:fiveobj}).} Blurring the object lowers the margin by 1.91 logits on Qwen2.5-VL. The intervals of the gray edit on the object, on the rest and elsewhere are [$-0.64$, $-0.33$], [$-0.27$, $-0.03$] and [$-0.10$, $0.00$] on LLaVA-1.5-13B and [$-0.80$, $-0.40$], [$-0.22$, $+0.06$] and [$-0.15$, $-0.03$] on InstructBLIP-13B. On the Gemma-3 models the gray interval includes zero, as at 12B. InstructBLIP-13B was edited at 224 pixels; on the 221 items where the region elsewhere exists, the gray difference is $-0.52$ [$-0.80$, $-0.26$] and the blur and noise intervals include zero ($-0.33$ [$-0.89$, $+0.21$], $-0.28$ [$-0.63$, $+0.03$]). \emph{Paired-image map on LLaVA-1.5-13B} (ViLP, 233 image-wins; the prior-consistent image favors the expected answer on 0.98 of items; \cref{tab:paired}): the other prompt positions recover up to 0.38 [0.35, 0.41] and patching them alone reverses up to 58\% of image-wins, against 0.23 and 38\% at 7B; the crossing is 0.59, against 0.53 with the gray copy, and 0.63 on 361 prior-wins. \emph{Top-20 heads.} LLaVA-1.5-13B: 26.15, 36.27, 28.22, 24.6, 36.16, 39.32, 38.13, 39.35, 34.2, 20.31, 25.9, 32.19, 39.23, 23.33, 16.21, 26.30, 23.0, 26.13, 31.21, 24.2. InstructBLIP-13B: 33.5, 24.33, 27.35, 20.4, 25.18, 24.24, 21.23, 32.14, 22.8, 27.1, 26.24, 34.36, 23.0, 27.2, 25.10, 37.37, 21.35, 38.31, 38.23, 19.16.

\noindent\textbf{Handoff.} Crossings are 0.58, 0.55, 0.57, 0.66, 0.74, 0.68, 0.53 and 0.59 in the order of \cref{tab:archband}; \cref{tab:crossci} gives them with intervals. The prior-agreeing items cross in the same band as the conflicts on LLaVA-1.5, InstructBLIP, Idefics2, Fuyu, Emu3 and Gemma-3 and in the adjacent earlier band on LFM2-VL and Llama-3.2, where the two curves are nearly tied at that band. In LFM2-VL, whose 30 layers hold attention only at layers 2, 5, 9, 13, 17, 21, 24 and 27, a map with one layer per band shows that the steps fall at the attention layers (\cref{tab:lfmlayer}). Fuyu's readout is at chance through layer 19 of 36 and rises to 0.86 at layer 31. With one layer per band on Fuyu (199 conflicts), the image-site recovery stays at 0.91 or above through layer 22 and the curves cross between layer 28 (0.48 against 0.45) and layer 29 (0.42 against 0.51), at 0.78 of depth; the 43 prior-agreeing items cross between layers 26 and 27.

\begin{table}[h]\centering\footnotesize\setlength{\tabcolsep}{8pt}
\caption{\textbf{LFM2-VL with one layer per band.} Share of the gray edit's effect recovered at the image site and at the answer position, by the layer at which the patch enters. Attention layers: 2, 5, 9, 13, 17, 21, 24 and 27 of 30.}\label{tab:lfmlayer}
\begin{tabular}{lcc}\toprule
Entry layers & Image site & Answer position \\ \midrule
0--5 & 1.00 to 0.98 & 0.00 \\
6--9 & 0.85 & 0.01 \\
10--13 & 0.73 & 0.03 \\
14--17 & 0.59 & 0.06 to 0.08 \\
18--21 & 0.52 to 0.51 & 0.18 to 0.29 \\
22--24 & 0.29 to 0.30 & 0.56 to 0.64 \\
25--27 & 0.10 & 0.80 to 0.82 \\
28--29 & 0.00 & 0.97 to 0.98 \\ \bottomrule
\end{tabular}\end{table}

\begin{table}[h]\centering\footnotesize\setlength{\tabcolsep}{3pt}
\caption{\textbf{The additive model, per model} (common protocol; WHOOPS blur and noise doses). $a$, $b$ with 95\% intervals clustered by image; held-out deviance explained with the 95th percentile of a permutation null; $\Delta$: change in held-out deviance explained when a dose-by-prior interaction is added (interaction $p$); split columns: median (range) over 50 item splits and the share of splits on which both weights are positive. ``full'': dose applied to the full-size image; a pixel count: dose applied after resizing to that input size.}\label{tab:archfit}
\begin{tabular}{llcccccccc}\toprule
Model & grid & $a$ & $b$ & held out & null & $\Delta$ ($p$) & $a$, splits & $b$, splits & both $>0$ \\ \midrule
LLaVA-1.5 & 336 & 0.31 [0.23, 0.40] & 0.22 [0.10, 0.35] & 0.131 & 0.045 & $-$0.02 (0.61) & 0.30 (0.25--0.37) & 0.25 (0.19--0.36) & 1.00 \\
LLaVA-1.5 & full & 0.06 [$-$0.01, 0.13] & 0.22 [0.09, 0.36] & 0.115 & 0.050 & 0.00 (0.86) & 0.06 ($-$0.01--0.09) & 0.25 (0.17--0.39) & 0.98 \\
BakLLaVA & 336 & 0.31 [0.24, 0.39] & 0.15 [0.04, 0.26] & 0.107 & 0.044 & $-$0.10 (0.22) & 0.31 (0.27--0.36) & 0.16 (0.11--0.22) & 1.00 \\
BakLLaVA & full & 0.07 [0.00, 0.14] & 0.14 [0.03, 0.26] & 0.094 & 0.050 & $-$0.11 (0.33) & 0.07 (0.03--0.13) & 0.15 (0.09--0.21) & 1.00 \\
InstructBLIP & 224 & 0.33 [0.23, 0.42] & 0.18 [0.04, 0.31] & 0.120 & 0.050 & 0.00 (0.86) & 0.38 (0.32--0.45) & 0.18 (0.09--0.28) & 1.00 \\
InstructBLIP & full & $-$0.01 [$-$0.09, 0.07] & 0.16 [0.02, 0.30] & 0.077 & 0.048 & $-$0.05 (0.66) & 0.01 ($-$0.04--0.06) & 0.18 (0.07--0.28) & 0.64 \\
Idefics2 & full & 0.15 [0.06, 0.24] & 0.19 [0.02, 0.35] & 0.111 & 0.077 & $-$0.22 (0.35) & 0.15 (0.09--0.21) & 0.19 (0.08--0.33) & 1.00 \\
LFM2-VL & full & 0.17 [0.10, 0.23] & 0.19 [0.04, 0.33] & 0.071 & 0.043 & $-$0.23 (0.16) & 0.15 (0.11--0.21) & 0.22 (0.14--0.35) & 1.00 \\
Fuyu & full & 0.22 [0.13, 0.31] & 0.24 [0.13, 0.34] & 0.063 & 0.031 & $-$0.02 (0.76) & 0.22 (0.17--0.27) & 0.23 (0.16--0.32) & 1.00 \\
Emu3 & 512$^2$ & 0.31 [0.22, 0.40] & 0.23 [0.11, 0.35] & 0.145 & 0.051 & $-$0.03 (0.34) & 0.29 (0.23--0.34) & 0.20 (0.14--0.26) & 1.00 \\
Emu3 & full & 0.10 [0.03, 0.17] & 0.22 [0.09, 0.34] & 0.132 & 0.046 & $-$0.02 (0.44) & 0.12 (0.07--0.16) & 0.20 (0.14--0.26) & 1.00 \\
Llama-3.2 & 560 & 0.22 [0.14, 0.31] & 0.24 [0.11, 0.38] & 0.086 & 0.041 & $-$0.08 (0.24) & 0.23 (0.16--0.28) & 0.25 (0.15--0.32) & 1.00 \\
Llama-3.2 & full & 0.08 [$-$0.01, 0.17] & 0.23 [0.09, 0.37] & 0.088 & 0.051 & $-$0.05 (0.52) & 0.08 ($-$0.02--0.15) & 0.25 (0.14--0.34) & 0.98 \\
Gemma-3 & full & 0.16 [0.08, 0.25] & 0.09 [$-$0.02, 0.19] & 0.043 & 0.037 & $-$0.07 (0.48) & 0.19 (0.14--0.23) & 0.14 (0.09--0.23) & 1.00 \\ \bottomrule
\end{tabular}\end{table}

\begin{table}[h]\centering\footnotesize\setlength{\tabcolsep}{3.4pt}
\caption{\textbf{Dose curves} (vision-win rate on prior-valid WHOOPS conflicts). Grid as in \cref{tab:archfit}.}\label{tab:archdose}
\begin{tabular}{llccccccccccc}\toprule
 & & & \multicolumn{5}{c}{blur radius} & \multicolumn{5}{c}{noise $\sigma$} \\
Model & grid & clean & 1 & 2 & 3 & 5 & 8 & 5 & 10 & 20 & 40 & 60 \\ \midrule
LLaVA-1.5 & 336 & 0.734 & 0.727 & 0.705 & 0.669 & 0.549 & 0.357 & 0.736 & 0.722 & 0.710 & 0.688 & 0.635 \\
LLaVA-1.5 & full & 0.734 & 0.739 & 0.741 & 0.729 & 0.712 & 0.679 & 0.741 & 0.734 & 0.736 & 0.722 & 0.724 \\
BakLLaVA & 336 & 0.743 & 0.727 & 0.691 & 0.679 & 0.537 & 0.344 & 0.741 & 0.732 & 0.713 & 0.670 & 0.653 \\
BakLLaVA & full & 0.743 & 0.732 & 0.734 & 0.729 & 0.691 & 0.670 & 0.741 & 0.736 & 0.732 & 0.727 & 0.732 \\
InstructBLIP & 224 & 0.787 & 0.787 & 0.767 & 0.710 & 0.556 & 0.365 & 0.797 & 0.794 & 0.784 & 0.730 & 0.680 \\
InstructBLIP & full & 0.787 & 0.789 & 0.784 & 0.797 & 0.779 & 0.769 & 0.792 & 0.789 & 0.792 & 0.792 & 0.792 \\
Idefics2 & full & 0.816 & 0.816 & 0.804 & 0.785 & 0.749 & 0.677 & 0.818 & 0.818 & 0.797 & 0.787 & 0.744 \\
LFM2-VL & full & 0.771 & 0.776 & 0.759 & 0.747 & 0.693 & 0.640 & 0.771 & 0.771 & 0.766 & 0.759 & 0.766 \\
Fuyu & full & 0.675 & 0.680 & 0.685 & 0.642 & 0.570 & 0.458 & 0.680 & 0.662 & 0.662 & 0.616 & 0.514 \\
Emu3 & 512$^2$ & 0.641 & 0.622 & 0.587 & 0.553 & 0.437 & 0.314 & 0.649 & 0.644 & 0.617 & 0.570 & 0.491 \\
Emu3 & full & 0.641 & 0.641 & 0.639 & 0.622 & 0.568 & 0.511 & 0.649 & 0.654 & 0.636 & 0.627 & 0.602 \\
Llama-3.2 & 560 & 0.745 & 0.757 & 0.723 & 0.697 & 0.622 & 0.489 & 0.740 & 0.738 & 0.712 & 0.662 & 0.629 \\
Llama-3.2 & full & 0.745 & 0.780 & 0.780 & 0.764 & 0.723 & 0.657 & 0.738 & 0.749 & 0.742 & 0.695 & 0.683 \\
Gemma-3 & full & 0.806 & 0.813 & 0.804 & 0.794 & 0.759 & 0.674 & 0.797 & 0.799 & 0.785 & 0.768 & 0.728 \\ \bottomrule
\end{tabular}\end{table}

\subsection{Dose at the input size}
\noindent\textbf{Dose at the input size.} LLaVA-1.5, BakLLaVA and InstructBLIP resize every image to 336 or 224 pixels. With blur and noise applied to the full-size image the vision-win rate is nearly flat (\cref{tab:archdose}) and $a$ is 0.06, 0.07 and $-0.01$. With the same levels applied after resizing to the input size, vision-win falls to 0.34--0.37 at the strongest blur and $a$ is 0.31, 0.31 and 0.33, while $b$ changes by at most 0.02 (\cref{tab:archfit}). Each pair of runs uses the same items and code and differs only in the resize. Idefics2, LFM2-VL and Fuyu receive the image at or near full size. Emu3 encodes every image at a budget of 512 $\times$ 512 pixels with the aspect ratio kept. Llama-3.2 splits the image into up to four 560-pixel tiles, so the full-size grid is already in model-input pixels, but it is a weak dose there. On both models the dose has to be applied at a single input size before the clarity weight appears (\cref{tab:inputdose}). Main-paper Table~1 reports the input-size grid.

\begin{table}[h]\centering\footnotesize\setlength{\tabcolsep}{6pt}
\caption{\textbf{Emu3 and Llama-3.2: dose on the full-size image and at the input size.} Vision-win at the strongest blur and the weights of the additive model with 95\% intervals. Dash: not reported.}\label{tab:inputdose}
\begin{tabular}{llccc}\toprule
Model & Dose applied to & Vision-win, strongest blur & $a$ & $b$ \\ \midrule
Emu3 & full-size image & 0.511 & 0.10 [0.03, 0.17] & -- \\
 & image resized to the 512 $\times$ 512 budget & 0.314 & 0.31 [0.22, 0.40] & 0.23 [0.11, 0.35] \\
Llama-3.2 & full-size image & 0.657 & 0.08 [$-$0.01, 0.17] & -- \\
 & image resized to one 560-pixel tile & 0.489 & 0.22 [0.14, 0.31] & 0.24 [0.11, 0.38] \\ \bottomrule
\end{tabular}\end{table}

\noindent\emph{Notes on \cref{tab:inputdose}.} At the input size both weights are positive on all 50 item splits on both models. On Emu3 the held-out deviance explained is 0.145 against a null 95th percentile of 0.051. On Llama-3.2 the clean vision-win rate is unchanged by the resize (0.745).

\subsection{Heads: selection, ablation and scaling}
\noindent\textbf{Heads.} WHOOPS prior-valid conflicts (up to 400) are split alternately into a selection and an evaluation half. On the selection half each head's output at the answer position is projected, through the final norm, on the difference of the two answers' unembedding rows; the top 20 heads by mean attribution are then mean-ablated at the answer position on the evaluation half, with means from the selection half, against three layer-matched random sets of 20. Per-head attributions sum to each layer's projected output, and ablating every head with an item's own activations returns its gap exactly. Self-attention heads are scored on every model. On Llama-3.2 the cross-attention heads are scored as well, each attribution scaled by its layer's gate $\tanh(g)$, and the ablation is repeated within each kind of head.

\begin{table}[h]\centering\footnotesize\setlength{\tabcolsep}{4pt}
\caption{\textbf{Head ablation} (common protocol). Median layer and mean relative depth of the top 20 heads, their share of the positive attribution, the correlation of the top heads' attribution between halves, image-wins and prior-wins reversed and mean gap change under mean-ablation of the top 20, and the range over three layer-matched random sets. Main-paper Table~1 gives the later runs with 20 random sets (Emu3: 25.2\% of 139 image-wins), described in the paragraph on 20 random sets below.}\label{tab:archheads}
\begin{tabular}{lcccccccc}\toprule
Model & median layer & depth & share & corr. & image-wins, top / random & gap, top / random & prior-wins, top / random \\ \midrule
LLaVA-1.5 & 24 of 32 & 0.79 & 0.54 & 0.96 & 10.5\% / 0.0--1.4\% & $-0.88$ / $-0.03$ to $+0.03$ & 5.3\% / 0.0--1.8\% \\
InstructBLIP & 23 of 32 & 0.75 & 0.50 & 0.92 & 8.7\% / 0.0--0.6\% & $-1.04$ / $-0.01$ to $+0.07$ & 5.0\% / 0.0--2.5\% \\
Idefics2 & 26 of 32 & 0.83 & 0.52 & 0.97 & 4.1\% / 0.6--1.2\% & $-1.28$ / $+0.05$ to $+0.21$ & 3.2\% / 3.2--6.5\% \\
LFM2-VL & 26 of 30 & 0.87 & 0.84 & 0.97 & 21.1\% / 0.0--0.7\% & $-1.96$ / $-0.07$ to $+0.07$ & 6.2\% / 0.0--4.2\% \\
Fuyu & 31 of 36 & 0.88 & 0.49 & 0.90 & 11.5\% / 0.0--0.8\% & $-0.96$ / $-0.03$ to $+0.01$ & 5.5\% / 2.7--4.1\% \\
Emu3 & 31 of 32 & 0.95 & 0.67 & 0.95 & 27.0\% / 2.2--5.1\% & $-2.01$ / $-0.14$ to $+0.10$ & 4.8\% / 6.3--11.1\% \\
Llama-3.2 & 33 of 40 & 0.88 & 0.37 & 0.92 & 6.5\% / 0.6--1.3\% & $-0.71$ / $-0.10$ to $+0.10$ & 2.2\% / 0.0--4.3\% \\
Llama-3.2, self-attention only & 36 of 40 & 0.89 & 0.63 & 0.93 & 1.9\% / 0.0--0.6\% & $-0.06$ / $+0.01$ to $+0.04$ & 8.7\% / 0.0--4.3\% \\
Gemma-3 & 0 of 48 & 0.29 & 0.71 & 0.57 & 1.8\% / 0.0\% & $-0.77$ / $+0.05$ to $+0.15$ & 5.4\% / 0.0--5.4\% \\ \bottomrule
\end{tabular}\end{table}

\begin{table}[h]\centering\footnotesize\setlength{\tabcolsep}{5pt}
\caption{\textbf{Top-20 heads per model} (layer.head, in order of direct logit attribution; first row ranks 1 to 10, second row ranks 11 to 20). Llama-3.2 (all heads) ranks self-attention and cross-attention heads together; its cross-attention layers are 3, 8, \ldots, 38. LFM2-VL has 256 heads in 8 attention layers and Fuyu 2,304 heads.}\label{tab:topheads}
\begin{tabular}{lcccccccccc}\toprule
Model & \multicolumn{10}{c}{Heads, by rank} \\ \midrule
LLaVA-1.5 & 24.24 & 19.6 & 29.10 & 24.17 & 26.25 & 28.7 & 26.14 & 21.16 & 21.9 & 22.16 \\
 & 29.9 & 30.12 & 24.14 & 25.0 & 22.30 & 31.27 & 20.5 & 20.12 & 24.15 & 23.20 \\[2pt]
InstructBLIP & 23.24 & 20.27 & 23.20 & 28.17 & 29.22 & 19.26 & 24.7 & 25.25 & 21.23 & 25.14 \\
 & 30.4 & 20.22 & 24.5 & 18.28 & 24.8 & 21.4 & 21.15 & 23.17 & 28.16 & 16.28 \\[2pt]
Llama-3.2 (self-attention) & 35.0 & 36.8 & 39.7 & 36.11 & 37.24 & 37.2 & 39.0 & 37.26 & 34.16 & 26.26 \\
 & 30.3 & 39.14 & 37.27 & 35.18 & 26.24 & 31.18 & 36.9 & 26.14 & 39.1 & 39.6 \\[2pt]
Idefics2 & 29.12 & 24.21 & 31.22 & 29.2 & 28.25 & 29.14 & 31.25 & 31.18 & 28.18 & 24.11 \\
 & 18.12 & 24.22 & 25.10 & 27.29 & 26.6 & 21.2 & 28.26 & 19.5 & 21.11 & 19.9 \\[2pt]
Emu3 & 30.15 & 31.20 & 31.8 & 29.10 & 31.19 & 31.14 & 29.12 & 30.16 & 30.7 & 31.24 \\
 & 15.22 & 31.13 & 31.25 & 22.11 & 30.28 & 31.22 & 31.5 & 29.20 & 31.2 & 31.31 \\[2pt]
Gemma-3 & 0.10 & 0.2 & 0.15 & 0.12 & 0.13 & 0.1 & 0.9 & 0.3 & 46.4 & 44.10 \\
 & 0.8 & 45.1 & 0.14 & 38.15 & 38.8 & 2.5 & 4.9 & 1.9 & 36.5 & 4.6 \\[2pt]
Llama-3.2 (all heads) & 35.0 & 33.11 & 33.3 & 36.8 & 33.5 & 39.7 & 28.3 & 28.28 & 36.11 & 28.31 \\
 & 37.24 & 33.25 & 38.3 & 33.13 & 33.15 & 33.24 & 37.2 & 39.0 & 38.25 & 33.17 \\[2pt]
LFM2-VL & 27.4 & 27.23 & 24.17 & 24.7 & 27.6 & 24.19 & 24.6 & 27.27 & 27.24 & 24.30 \\
 & 27.25 & 27.30 & 27.21 & 24.20 & 21.4 & 27.20 & 27.8 & 21.11 & 24.24 & 21.21 \\[2pt]
Fuyu & 32.11 & 31.53 & 31.5 & 31.49 & 32.28 & 32.24 & 31.34 & 27.27 & 32.46 & 29.29 \\
 & 29.34 & 33.36 & 32.16 & 30.63 & 29.40 & 26.1 & 31.35 & 32.56 & 30.60 & 31.40 \\ \bottomrule
\end{tabular}\end{table}

\noindent\emph{Notes on \cref{tab:topheads}.} One head (23.20) is shared between LLaVA-1.5 and InstructBLIP. On Llama-3.2 with all heads ranked together, 13 of the 20 are cross-attention heads, which hold 0.65 of the positive attribution, with gates $\tanh(g)$ between $-0.0085$ and $0.0067$. The 20 best cross-attention heads reverse 7.8\% of 154 image-wins and no prior-wins (gap $-0.87$), against 1.3, 1.3 and 3.2\% for random cross-attention sets; the 20 best self-attention heads reverse 1.9\% (gap $-0.06$), against 0.6, 0.0 and 0.0\%. On Gemma-3 the selection fails: 10 of the top 20 heads are in layer 0 and 14 in layers 0 to 4, before the handoff, and ablating the 20 reverses 1.8\% of image-wins against 0.0\% for each random set. On Emu3, 18 of the top 20 heads are in layers 29 to 31, where the image tokens still carry 0.38 of the gray edit's effect (\cref{tab:archband}). On LFM2-VL all 20 are in attention layers 21, 24 and 27, after the crossing at 0.66 of depth, and the evaluation half has 152 image-wins. On Fuyu the evaluation half has 122 image-wins.

\begin{table}[h]\centering\footnotesize\setlength{\tabcolsep}{3pt}
\caption{\textbf{Object against rest} (common protocol; prior-valid WHOOPS conflicts with a SAM3 mask). Shift of the gap against the clean image, in logits, with bootstrap 95\% intervals, when the image is grayed inside the object mask, outside it, or inside the mask moved elsewhere in the frame with under 10\% overlap (available on 80 to 82\% of the items), and when the whole image is grayed. Edits are applied at full size.}\label{tab:archobj}
\begin{tabular}{lccccc}\toprule
Model & $n$ & gray object & gray rest & gray region elsewhere & gray whole image \\ \midrule
LLaVA-1.5 & 361 & $-0.47$ [$-0.63$, $-0.31$] & $-0.09$ [$-0.23$, $+0.03$] & $-0.03$ [$-0.08$, $+0.02$] & $-0.80$ [$-1.04$, $-0.56$] \\
InstructBLIP & 344 & $-0.57$ [$-0.73$, $-0.40$] & $-0.04$ [$-0.16$, $+0.07$] & $-0.06$ [$-0.11$, $-0.01$] & $-0.80$ [$-1.04$, $-0.57$] \\
Idefics2 & 356 & $-0.76$ [$-1.00$, $-0.56$] & $-0.36$ [$-0.58$, $-0.15$] & $-0.12$ [$-0.20$, $-0.05$] & $-1.19$ [$-1.54$, $-0.86$] \\
LFM2-VL & 352 & $-0.56$ [$-0.74$, $-0.41$] & $-0.29$ [$-0.43$, $-0.16$] & $-0.06$ [$-0.11$, $-0.00$] & $-0.91$ [$-1.16$, $-0.68$] \\
Fuyu & 334 & $-0.59$ [$-0.77$, $-0.42$] & $-0.02$ [$-0.16$, $+0.14$] & $-0.06$ [$-0.11$, $-0.01$] & $-0.99$ [$-1.25$, $-0.74$] \\
Emu3 & 349 & $-0.67$ [$-0.90$, $-0.45$] & $-0.33$ [$-0.51$, $-0.15$] & $-0.06$ [$-0.13$, $+0.01$] & $-1.22$ [$-1.53$, $-0.91$] \\
Llama-3.2 & 362 & $-0.61$ [$-0.82$, $-0.43$] & $-0.19$ [$-0.34$, $-0.04$] & $+0.00$ [$-0.10$, $+0.12$] & $-1.05$ [$-1.33$, $-0.79$] \\
Gemma-3-12B & 363 & $-0.93$ [$-1.35$, $-0.52$] & $-0.73$ [$-1.06$, $-0.39$] & $+0.10$ [$-0.06$, $+0.28$] & $-1.95$ [$-2.48$, $-1.41$] \\ \bottomrule
\end{tabular}\end{table}

\begin{table}[h]\centering\footnotesize\setlength{\tabcolsep}{3pt}
\caption{\textbf{Object against rest, paired contrasts and other edits} (same items as \cref{tab:archobj}). Blur radius 8 and noise $\sigma$ = 60 inside the object mask / outside it; paired object-minus-rest differences with bootstrap 95\% intervals for the three edits; paired difference between graying the object and graying the region elsewhere; share of image-wins whose gap changes sign when the object / the rest / the region elsewhere is grayed.}\label{tab:archobj2}
\resizebox{\textwidth}{!}{%
\begin{tabular}{lccccccc}\toprule
Model & blur obj.\ / rest & noise obj.\ / rest & obj.\ $-$ rest, gray & obj.\ $-$ rest, blur & obj.\ $-$ rest, noise & obj.\ $-$ elsewhere & wins reversed (\%) \\ \midrule
LLaVA-1.5 & $-0.59$ / $+0.03$ & $-0.30$ / $+0.06$ & $-0.38$ [$-0.57$, $-0.19$] & $-0.62$ [$-0.84$, $-0.41$] & $-0.37$ [$-0.53$, $-0.22$] & $-0.43$ [$-0.61$, $-0.25$] & 10.8 / 4.0 / 0.9 \\
InstructBLIP & $-0.71$ / $-0.12$ & $-0.27$ / $-0.04$ & $-0.52$ [$-0.71$, $-0.32$] & $-0.59$ [$-0.82$, $-0.38$] & $-0.23$ [$-0.35$, $-0.11$] & $-0.48$ [$-0.68$, $-0.31$] & 6.6 / 3.4 / 1.3 \\
Idefics2 & $-1.59$ / $-0.94$ & $-1.08$ / $-0.70$ & $-0.40$ [$-0.73$, $-0.11$] & $-0.65$ [$-0.99$, $-0.27$] & $-0.38$ [$-0.64$, $-0.14$] & $-0.68$ [$-0.92$, $-0.44$] & 5.9 / 3.9 / 0.8 \\
LFM2-VL & $-0.88$ / $-0.56$ & $-0.33$ / $-0.10$ & $-0.27$ [$-0.50$, $-0.08$] & $-0.32$ [$-0.58$, $-0.08$] & $-0.24$ [$-0.36$, $-0.12$] & $-0.52$ [$-0.74$, $-0.32$] & 5.9 / 3.5 / 0.0 \\
Fuyu & $-1.44$ / $-0.11$ & $-1.15$ / $-0.37$ & $-0.57$ [$-0.81$, $-0.36$] & $-1.33$ [$-1.76$, $-0.95$] & $-0.78$ [$-1.14$, $-0.44$] & $-0.52$ [$-0.73$, $-0.33$] & 7.2 / 4.8 / 2.4 \\
Emu3 & $-1.18$ / $-0.38$ & $-0.74$ / $-0.36$ & $-0.34$ [$-0.63$, $-0.05$] & $-0.80$ [$-1.13$, $-0.46$] & $-0.38$ [$-0.63$, $-0.14$] & $-0.61$ [$-0.87$, $-0.36$] & 10.4 / 5.8 / 1.6 \\
Llama-3.2 & $-1.04$ / $-0.40$ & $-0.76$ / $-0.34$ & $-0.43$ [$-0.67$, $-0.18$] & $-0.64$ [$-0.98$, $-0.32$] & $-0.42$ [$-0.62$, $-0.22$] & $-0.64$ [$-0.89$, $-0.39$] & 5.6 / 4.5 / 0.9 \\
Gemma-3-12B & $-2.12$ / $-0.90$ & $-1.45$ / $-0.34$ & $-0.20$ [$-0.67$, $+0.27$] & $-1.22$ [$-1.93$, $-0.59$] & $-1.11$ [$-1.56$, $-0.68$] & $-0.97$ [$-1.45$, $-0.53$] & 6.1 / 4.5 / 1.2 \\ \bottomrule
\end{tabular}}\end{table}

\noindent\textbf{Heads, 20 random sets and prior-agreeing items.} On LLaVA-1.5, Idefics2, Gemma-3-12B, InstructBLIP-7B, Fuyu-8B, Llama-3.2, LFM2-VL and Emu3 we repeated the ablation with 20 layer-matched random sets and with the same sets on prior-agreeing items, where the text-only answer is already the image's answer (\cref{tab:heads20}). On all eight models the top set is more extreme than all 20 random sets. On prior-agreeing items the ablation changes the margin about as much as on conflicts, so the heads write the image's answer whether or not it conflicts with the prior.

\begin{table}[h]\centering\footnotesize\setlength{\tabcolsep}{3.5pt}
\caption{\textbf{Top-20 heads against 20 layer-matched random sets, and on prior-agreeing items.} Image-wins reversed by mean-ablating the top 20 heads, with the mean and maximum over the random sets; mean change in the gap in logits, with the range over the random sets; on prior-agreeing items, the change in the margin and the share of decisions reversed. 95\% intervals in brackets.}\label{tab:heads20}
\begin{tabular}{lccccccc}\toprule
 & \multicolumn{4}{c}{Conflicts} & \multicolumn{3}{c}{Prior-agreeing items} \\ \cmidrule(lr){2-5}\cmidrule(lr){6-8}
Model & Reversed (\%) & Random & Gap change & Random & $n$ & Margin change & Rev.\ (\%) \\ \midrule
LLaVA-1.5 & 10.5 [5.6, 15.4] & 0.5 / 1.4 & $-0.88$ [$-1.04$, $-0.71$] & $-0.04$ to $+0.13$ & 75 & $-0.65$ [$-0.83$, $-0.45$] & 4.0 \\
InstructBLIP-7B & 8.7 [5.0, 13.8] & 0.5 / 1.9 & $-1.04$ [$-1.19$, $-0.89$] & $-0.11$ to $+0.12$ & 87 & $-0.82$ [$-1.04$, $-0.62$] & 4.7 \\
Idefics2 & 4.1 [1.2, 7.7] & 0.9 / 2.4 & $-1.28$ [$-1.52$, $-1.05$] & $-0.04$ to $+0.28$ & 74 & $-1.51$ [$-1.90$, $-1.10$] & 5.5 \\
LFM2-VL & 21.1 [15.1, 28.3] & 0.3 / 1.3 & $-1.96$ [$-2.24$, $-1.69$] & $-0.07$ to $+0.44$ & 79 & $-1.58$ [$-1.97$, $-1.20$] & 1.3 \\
Fuyu-8B & 11.5 [6.6, 17.2] & 0.2 / 0.8 & $-0.96$ [$-1.15$, $-0.76$] & $-0.06$ to $+0.05$ & 95 & $-0.90$ [$-1.12$, $-0.68$] & 2.6 \\
Llama-3.2 & 6.5 [3.2, 10.4] & 0.8 / 1.9 & $-0.71$ [$-0.85$, $-0.57$] & $-0.16$ to $+0.10$ & 74 & $-0.42$ [$-0.59$, $-0.26$] & 2.8 \\
Gemma-3-12B & 1.8 [0.0, 4.3] & 0.0 / 0.0 & $-0.77$ [$-0.98$, $-0.53$] & $-0.12$ to $+0.21$ & 68 & $-0.94$ [$-1.34$, $-0.54$] & 3.0 \\
Emu3 & 25.2 [18.0, 32.4] & 4.0 / 6.5 & $-2.01$ [$-2.32$, $-1.70$] & $-0.22$ to $+0.14$ & 86 & $-1.07$ [$-1.39$, $-0.74$] & 4.7 \\ \bottomrule
\end{tabular}\end{table}

\noindent\emph{Notes on \cref{tab:heads20}.} On Llama-3.2 the 20 best cross-attention heads reverse 7.8\% of image-wins against 0.6 to 2.6\% for 20 random cross-attention sets, and the 20 best self-attention heads 1.9\% against at most 0.6\%. On Emu3 the random sets reverse between 0.7 and 6.5\%; \cref{tab:archheads} gives 27.0\% for the top set, from a run with three random sets.

\noindent\textbf{Heads selected by their estimated ablation effect.} Direct logit attribution scores what a head writes straight into the answer logits and not what later layers do with its output. On Gemma-3 it selects heads in layer 0 whose ablation changes nothing. We therefore also rank every head by a first-order estimate of the change in the gap when its output at the answer position is replaced by its mean over the selection items: the gradient of the gap with respect to that output, taken through the whole forward pass, times (mean $-$ activation). The 20 heads with the most negative estimate on the selection half are mean-ablated on the held-out half, with 20 layer-matched random sets, as before. On 30 selection items the estimate is compared with the exact effect of ablating each selected head alone (\cref{tab:headgrad}).

\noindent\emph{Against the attribution set.} On LLaVA-1.5 the selected set shares 7 heads with the attribution set, sits earlier (median layer 18 against 24) and reverses more image-wins (14.7\% against 10.5\%). On Gemma-3-4B it reverses 14.1\% of image-wins against at most 2.5\% for a random set, and 8.1\% of prior-wins where the random sets reverse 8.1 to 21.6\%: the effect is directional, where the attribution set reversed both outcomes at similar rates. On Gemma-3-12B the selected heads lie in layers 27 to 47 of 48, three are shared with the attribution set, and ablating them reverses 7.4\% [3.7, 11.7] of 162 image-wins against at most 2.5\% for a random set, with the gap lower by 2.50 logits [2.12, 2.89] (random sets $-0.09$ to $+0.45$); the attribution set reversed 1.8\%. Prior-wins reverse at 10.5\% (4 of 38), inside the range of the random sets (0.0 to 10.5\%). On Idefics2 the selected set reverses 4.1\% [1.8, 7.1], the same rate as the attribution set, against at most 2.4\%. On InstructBLIP-7B it reverses 15.0\% [9.4, 21.2] against at most 1.3\% (8.7\% by attribution), with a median layer of 18 against 23.

\noindent\emph{Depth against the crossing.} \Cref{tab:headdepth} counts the selected heads that lie after the crossing of main Table~1, where a head in layer $l$ of $L$ is after the crossing when $(l{+}1)/L$ exceeds it. Qwen2.5-VL was not ranked this way.

\begin{table}[h]\centering\footnotesize\setlength{\tabcolsep}{5pt}
\caption{\textbf{Heads after the crossing}, of 20, under the two rankings. Llama-3.2: cross- and self-attention heads.}\label{tab:headdepth}
\begin{tabular}{lcccccccc}\toprule
Ranked by & Emu3 & Fuyu & Gemma-3-12B & Idefics2 & LFM2-VL-3B & Llama-3.2 & LLaVA-1.5 & InstructBLIP-7B \\ \midrule
Estimated effect & 18 & 17 & 19 & 14 & 17 & 16 & 12 & 11 \\
Attribution & 19 & 20 & 6 & 20 & 20 & 20 & 20 & 19 \\ \bottomrule
\end{tabular}\end{table}

\noindent\emph{Where the two rankings differ.} The ranking is not uniformly stronger than attribution: on Fuyu and LFM2-VL the selected set reverses 7.4\% [3.3, 12.3] and 13.8\% [8.6, 19.7], where attribution gave 11.5\% and 21.1\%; on both it shares more than half its heads with the attribution set (13 and 11 of 20) and exceeds all 20 random sets. On Llama-3.2 the selected set holds 10 cross-attention and 10 self-attention heads and reverses 7.7\% [3.8, 12.2] against at most 3.8\% (6.5\% by attribution). Ranked within kind, its 20 best cross-attention heads reverse 7.7\% against 0.6 to 3.2\% for random cross-attention sets, and its 20 best self-attention heads 9.0\% against 0.0 to 2.6\%, where the 20 best self-attention heads by attribution reversed 1.9\%: once what later layers do with a head's output is counted, self-attention heads carry as much of the decision as cross-attention heads.

\noindent\emph{Gemma-3-27B and Emu3.} On Gemma-3-27B the selected heads all lie in layers 31 to 61 and none is shared with the attribution set, whose median layer is 0. Ablating them lowers the gap by 1.60 logits [1.37, 1.84], against $-0.04$ to $+0.20$ for the random sets, and reverses 1.9\% of image-wins (3 of 160) against at most 0.6\%; on 67 prior-agreeing items the gap falls by 1.26 [0.91, 1.59]. Gemma-3-27B therefore has late heads that write the image's answer, and their removal moves the margin without changing many decisions. On Emu3 the selected set shares 15 heads with the attribution set, sits at the same depth (median layer 31 of 32) and reverses 21.6\% [15.1, 28.8] of 139 image-wins against at most 6.5\% for a random set (25.2\% by attribution).

\noindent\textbf{Scaling the heads up.} Ablation shows that the heads contribute. We also scaled them up. At the answer position each of the 20 selected heads' output is set to its mean plus $\gamma$ times its deviation from the mean ($\gamma=0$ is the mean ablation above, $\gamma=1$ the clean run), for $\gamma$ = 0, 0.5, 1.5, 2, 3 and 5, against 5 layer-matched random sets. The heads were chosen on one half of the WHOOPS conflicts; the test items are the other half, 300 prior-valid ViLP conflicts, and the prior-agreeing WHOOPS items. We also score gated scaling: the heads are scaled up only on items whose untrained margin at the first layer at or past the crossing depth favors the prior, so the trigger uses only the model's mid-depth margin. The criteria were pre-specified. The heads count as a graded control if (1) at $\gamma$ = 2 or 3 the selected set reverses more prior-wins than every random set on both sources, (2) the mean margin change rises with $\gamma$, and (3) at the smallest $\gamma$ that meets (1) at most 2\% of image-wins and of prior-agreeing answers are lost. The closed loop counts as a correction if its change in vision-win has a paired 95\% interval that excludes zero on both sources.

\Cref{tab:dial} gives the result on nine models from the eight architectures. On the held-out WHOOPS conflicts the mean margin is above its unscaled value at every $\gamma$ above 1 on every model (it need not rise at each step: on Idefics2 the change is $+$0.34, $+$0.46 and $+$0.39 at $\gamma$ = 2, 3 and 5), and at $\gamma$ = 3 the selected heads reverse 12.5 to 32.8\% of prior-wins, more than every random set on eight models; on Idefics2, with 31 prior-wins, the selected set ties the best random set at 12.9\%. The heads were chosen on WHOOPS, and on ViLP they transfer on five of the nine. There the selected set exceeds every random set at $\gamma$ = 3 on LLaVA-1.5, Emu3, Idefics2, Fuyu and Llama-3.2 (6.8 to 18.2\% against at most 12.8\%). It does not on four models. On InstructBLIP-7B (7.8\% against 2.4 to 8.4\%) and Qwen2.5-VL (9.3\% against 1.6 to 10.1\%) the ViLP margin does not move. On LFM2-VL-3B (9.8\% against 9.8 to 26.3\%) turning the heads up moves the ViLP margin toward the prior ($-$0.16 logits at $\gamma$ = 3, $-$0.78 on prior-wins). On Gemma-3-12B, where the heads are ranked by estimated ablation effect, it moves the margin on ViLP prior-wins by $-$12 logits at $\gamma$ = 3 and the selected set reverses 2.5\% against 1.2 to 30.0\% for random sets. So criterion (1) holds on both sources on LLaVA-1.5, Emu3, Fuyu and Llama-3.2 (cross-attention heads), and on InstructBLIP-7B at $\gamma$ = 2 only (6.0\% against at most 5.4\% on ViLP); (2) holds on WHOOPS on all nine. Criterion (3) fails on every model, because turning the heads up also costs image-wins: at $\gamma$ = 3, 0.6 to 4.3\% of the WHOOPS image-wins and 3.0 to 12.3\% of the ViLP image-wins are lost, and at the smallest $\gamma$ that meets (1) the loss on ViLP is 2.7\% on Fuyu, 3.0\% on Llama-3.2, 5.8\% on LLaVA-1.5, 6.0\% on InstructBLIP-7B and 12.3\% on Emu3, above the 2\% bound.

At $\gamma$ = 2 and 3 the closed loop changes vision-win by 0 to $+$7.0 points on WHOOPS and by $-$2.0 to $+$6.0 on ViLP. It is below zero in three of the thirty-six cells, all on ViLP: Llama-3.2 at $\gamma$ = 3 ($-$0.3 [$-$1.7, 1.0]) and Gemma-3-12B at $\gamma$ = 2 and 3 ($-$1.3 [$-$3.0, 0.0] and $-$2.0 [$-$3.7, $-$0.3]); on Gemma-3-12B the gated scaling lowers vision-win on the source the heads were not chosen on. On Fuyu it is a correction by the test fixed in advance: at $\gamma$ = 3 the interval excludes zero on both sources ($+$3.1 [1.0, 5.6] on WHOOPS and $+$6.0 [2.7, 9.3] on ViLP), and on ViLP the gain grows to $+$11.0 [7.3, 15.3] at $\gamma$ = 5. On Emu3 the interval excludes zero on WHOOPS only and on Idefics2 on ViLP at $\gamma$ = 3 only; on LLaVA-1.5 and LFM2-VL-3B the lower bound is 0.5 points or less on the one source where the gain is largest, and on InstructBLIP-7B, Llama-3.2 and Qwen2.5-VL the interval includes zero on both sources. On the prior-agreeing items the closed loop costs 1.3 and 2.7 points on LLaVA-1.5 (intervals reach zero) and nothing on the other eight at $\gamma$ = 2 and 3. Bootstrap bounds vary between repetitions by at most 0.5 points, so bounds within 0.5 points of zero are not read as excluding it.

The WHOOPS effect tends to be larger where the heads matter more under ablation (\cref{tab:dialrank}; rank correlation 0.83 on nine models). The effect of scaling is graded on WHOOPS on eight of nine models. Gated by the model's own mid-depth readout the heads raise vision-win reliably on one model of nine on both sources; by the test fixed in advance they are not a general correction, and on four models they do not transfer to ViLP.

\begin{table}[h]\centering\footnotesize\setlength{\tabcolsep}{6pt}
\caption{\textbf{Ablation against scaling, per model.} Image-wins reversed by mean-ablating the heads; closed-loop gain in vision-win at $\gamma$ = 3 on WHOOPS, in points; change in the mean margin on prior-wins when the heads' output is tripled, in logits. Llama-3.2: cross-attention heads.}\label{tab:dialrank}
\begin{tabular}{lccc}\toprule
Model & Image-wins reversed by ablation (\%) & Closed-loop gain at $\gamma$ = 3 & Margin on prior-wins at $\gamma$ = 3 \\ \midrule
Emu3 & 25.2 & $+$7.0 & $+$1.30 \\
LFM2-VL-3B & 21.1 & $+$2.0 & $+$0.70 \\
Fuyu & 11.5 & $+$3.1 & $+$0.91 \\
LLaVA-1.5 & 10.5 & $+$2.5 & $+$0.10 \\
InstructBLIP-7B & 8.7 & $+$1.5 & $+$0.20 \\
Llama-3.2 & 7.8 & $+$2.0 & $+$0.54 \\
Gemma-3-12B & 7.4 & 0.0 & $-$0.38 \\
Qwen2.5-VL & 7.3 & $+$1.5 & $+$0.45 \\
Idefics2 & 4.1 & $+$1.0 & $+$1.10 \\ \bottomrule
\end{tabular}\end{table}

\noindent\emph{Notes on \cref{tab:dialrank}.} On LLaVA-1.5 ablating the heads removes about 0.3 logits from the margin on prior-wins, where tripling them adds 0.10. On Gemma-3-12B the mean margin on prior-wins falls while 15.8\% of them reverse.

\begin{table*}[t]\centering\footnotesize\setlength{\tabcolsep}{3.5pt}
\caption{\textbf{Scaling the heads up.} Selected top-20 heads scaled by $\gamma$ at the answer position. Prior-wins reversed: selected set, with the range over 5 layer-matched random sets in parentheses; the 95\% interval at $\gamma$ = 3 is a bootstrap over items. Closed loop: change in vision-win, in points, when the heads are turned up only where the mid-depth untrained margin favors the prior; paired bootstrap 95\% intervals. $n$: image-wins / prior-wins among the test items. Image-wins lost among the prior-agreeing items at $\gamma$ = 2 / 3: LLaVA-1.5 1.3 / 4.0\% (75 items), Emu3 0.0 / 1.2\% (86), Idefics2 0.0 / 0.0\% (74), InstructBLIP-7B 1.2 / 1.2\% (87), Fuyu 1.3 / 3.9\% (95, of which 18 are prior-wins under the first-token margin), LFM2-VL-3B 0.0 / 0.0\% (79), Llama-3.2 0.0 / 0.0\% (74), Gemma-3-12B 3.0 / 4.5\% (68), Qwen2.5-VL 0.0 / 0.0\% (80). Applied to every test item, the scaling at $\gamma$ = 3 changes vision-win by $+$0.5 to $+$7.0 points on WHOOPS and by $-$3.0 to $+$7.3 on ViLP (computed from the two rates and $n$ of each row). Llama-3.2: cross-attention heads. Gemma-3-12B: heads ranked by estimated ablation effect.}\label{tab:dial}
\begin{tabular}{@{}llcccccc@{}}\toprule
 & & & \multicolumn{2}{c}{Prior-wins reversed (random sets)} & Image-wins lost & \multicolumn{2}{c}{Closed loop, vision-win change} \\
Model & Source & $n$ & $\gamma$ = 2 & $\gamma$ = 3 & $\gamma$ = 2 / 3 & $\gamma$ = 2 & $\gamma$ = 3 \\ \midrule
LLaVA-1.5 & WHOOPS & 143 / 57 & 5.3\% (0.0--3.5) & 14.0\% [5.3, 22.8] (0.0--3.5) & 0.0 / 2.1\% & $+$1.5 [0.0, 3.5] & $+$2.5 [$-$0.5, 5.5] \\
 & ViLP & 138 / 162 & 5.6\% (0.0--3.1) & 8.0\% [4.3, 12.3] (0.6--4.9) & 5.8 / 8.0\% & $+$2.0 [0.3, 4.0] & $+$2.3 [0.0, 4.7] \\
Emu3 & WHOOPS & 139 / 61 & 19.7\% (0.0--6.6) & 32.8\% [21.3, 44.3] (1.6--6.6) & 1.4 / 4.3\% & $+$4.0 [1.0, 7.0] & $+$7.0 [3.5, 11.0] \\
 & ViLP & 65 / 235 & 3.8\% (0.9--3.8) & 6.8\% [3.8, 10.2] (3.0--6.0) & 6.2 / 12.3\% & $+$1.3 [$-$0.7, 3.7] & $+$1.7 [$-$1.0, 4.3] \\
Idefics2 & WHOOPS & 169 / 31 & 6.5\% (0.0--6.5) & 12.9\% [3.2, 25.8] (3.2--12.9) & 1.2 / 0.6\% & 0.0 [$-$1.5, 1.5] & $+$1.0 [$-$1.0, 3.0] \\
 & ViLP & 159 / 141 & 9.9\% (1.4--6.4) & 14.2\% [9.2, 20.6] (3.5--12.8) & 2.5 / 5.0\% & $+$1.7 [0.0, 3.3] & $+$2.7 [0.7, 4.7] \\
InstructBLIP-7B & WHOOPS & 160 / 40 & 5.0\% (0.0--2.5) & 12.5\% [2.5, 22.5] (0.0--2.5) & 0.0 / 0.6\% & $+$0.5 [0.0, 1.5] & $+$1.5 [$-$0.5, 4.0] \\
 & ViLP & 134 / 166 & 6.0\% (2.4--5.4) & 7.8\% [4.2, 12.0] (2.4--8.4) & 6.0 / 9.0\% & $+$1.0 [$-$0.7, 3.0] & $+$0.7 [$-$1.7, 3.0] \\
Fuyu & WHOOPS & 123 / 72 & 12.5\% (1.4--2.8) & 18.1\% [9.7, 26.4] (1.4--4.2) & 1.6 / 2.4\% & $+$2.6 [0.5, 5.1] & $+$3.1 [1.0, 5.6] \\
 & ViLP & 146 / 154 & 9.1\% (1.3--3.9) & 18.2\% [12.3, 24.7] (1.3--6.5) & 2.7 / 4.1\% & $+$3.0 [0.7, 5.3] & $+$6.0 [2.7, 9.3] \\
LFM2-VL-3B & WHOOPS & 152 / 48 & 10.4\% (0.0--2.1) & 14.6\% [6.2, 25.0] (2.1--6.2) & 1.3 / 1.3\% & $+$1.0 [0.0, 2.5] & $+$2.0 [0.5, 4.0] \\
 & ViLP & 167 / 133 & 6.0\% (5.3--14.3) & 9.8\% [5.3, 15.0] (9.8--26.3) & 5.4 / 10.2\% & $+$0.3 [$-$1.3, 2.0] & $+$0.7 [$-$2.0, 3.3] \\
Llama-3.2 & WHOOPS & 156 / 44 & 15.9\% (2.3--9.1) & 20.5\% [9.1, 31.8] (4.5--13.6) & 1.9 / 2.6\% & $+$1.5 [$-$1.5, 4.5] & $+$2.0 [$-$1.0, 5.0] \\
 & ViLP & 237 / 63 & 6.3\% (3.2--6.3) & 9.5\% [3.2, 17.5] (4.8--7.9) & 1.3 / 3.0\% & $+$0.7 [0.0, 1.7] & $-$0.3 [$-$1.7, 1.0] \\
Gemma-3-12B & WHOOPS & 162 / 38 & 10.5\% (2.6--5.3) & 15.8\% [5.3, 28.9] (2.6--7.9) & 2.5 / 3.1\% & 0.0 [$-$2.0, 2.0] & 0.0 [$-$2.0, 2.5] \\
 & ViLP & 220 / 80 & 2.5\% (1.2--13.8) & 2.5\% [0.0, 6.2] (1.2--30.0) & 3.6 / 5.0\% & $-$1.3 [$-$3.0, 0.0] & $-$2.0 [$-$3.7, $-$0.3] \\
Qwen2.5-VL & WHOOPS & 165 / 35 & 8.6\% (0.0--2.9) & 14.3\% [2.9, 25.8] (2.9--2.9) & 0.0 / 1.8\% & $+$1.0 [0.0, 2.5] & $+$1.5 [$-$0.5, 4.0] \\
 & ViLP & 171 / 129 & 3.9\% (0.0--3.9) & 9.3\% [4.7, 14.7] (1.6--10.1) & 3.5 / 5.3\% & 0.0 & $+$1.3 [$-$0.7, 3.7] \\ \bottomrule
\end{tabular}\end{table*}

\begin{table}[h]\centering\footnotesize\setlength{\tabcolsep}{4pt}
\caption{\textbf{Heads selected by estimated ablation effect} (200 selection and 200 held-out WHOOPS conflicts; 196 and 195 on Fuyu). Median layer of the selected 20; heads shared with the attribution top 20; Pearson correlation of the estimate with exact single-head ablation over 600 head-item pairs; image-wins reversed by ablating the selected set, with 95\% interval, against the mean and maximum over 20 layer-matched random sets; prior-wins reversed (selected / random mean); mean change in the gap (selected / range of the random sets).}\label{tab:headgrad}
\begin{tabular}{lccccccc}\toprule
Model & Median layer & Shared & Pearson & Image-wins reversed & Random mean / max & Prior-wins & Gap change \\ \midrule
LLaVA-1.5 & 18 of 32 & 7 & 0.99 & 14.7\% [9.1, 20.3] & 0.8\% / 2.1\% & 0.0\% / 2.4\% & $-1.08$ / $-0.02$ to $+0.35$ \\
InstructBLIP-7B & 18 of 32 & 8 & 0.97 & 15.0\% [9.4, 21.2] & 0.6\% / 1.3\% & 2.5\% / 1.3\% & $-1.39$ / $-0.09$ to $+0.27$ \\
Idefics2 & 20 of 32 & 7 & 0.96 & 4.1\% [1.8, 7.1] & 0.9\% / 2.4\% & 0.0\% / 5.2\% & $-1.19$ / $-0.09$ to $+0.26$ \\
LFM2-VL & 22 of 30 & 11 & 0.95 & 13.8\% [8.6, 19.7] & 0.6\% / 2.6\% & 8.3\% / 4.2\% & $-1.47$ / $-0.20$ to $+0.29$ \\
Fuyu & 31 of 36 & 13 & 0.98 & 7.4\% [3.3, 12.3] & 0.7\% / 1.6\% & 1.4\% / 2.5\% & $-1.01$ / $-0.06$ to $+0.07$ \\
Llama-3.2 & 23 of 40 & 5 & 0.98 & 7.7\% [3.8, 12.2] & 2.4\% / 3.8\% & 4.5\% / 3.4\% & $-0.99$ / $-0.17$ to $+0.29$ \\
Gemma-3-4B & 20 of 34 & 5 & 0.91 & 14.1\% [9.2, 19.6] & 1.2\% / 2.5\% & 8.1\% / 13.9\% & $-4.11$ / $+0.91$ to $+1.62$ \\
Gemma-3-12B & 36 of 48 & 3 & 0.96 & 7.4\% [3.7, 11.7] & 0.8\% / 2.5\% & 10.5\% / 5.8\% & $-2.50$ / $-0.09$ to $+0.45$ \\
Gemma-3-27B & 46 of 62 & 0 & 0.97 & 1.9\% [0.0, 4.4] & 0.0\% / 0.6\% & 0.0\% / 2.1\% & $-1.60$ / $-0.04$ to $+0.20$ \\
Emu3 & 31 of 32 & 15 & 0.98 & 21.6\% [15.1, 28.8] & 4.0\% / 6.5\% & 3.3\% / 4.7\% & $-1.79$ / $-0.19$ to $+0.13$ \\ \bottomrule
\end{tabular}\end{table}

\subsection{The image-token subspace on other architectures}
\noindent\textbf{The image-token subspace on other architectures.} The subspace analysis of the main text is on three LLaVA models. We applied the same analysis to every architecture that places image tokens in the decoder's sequence, with one change: the basis is fit on each item's residual averaged over all image positions, where the basis of Supp.~\ref{app:necessity} averages over the attended image tokens. One definition then serves every architecture; the calibration at the end of this section gives what the change does on LLaVA-1.5. The blocks are those at relative depth 0.5, 0.625 and 0.75 (blocks 16, 20 and 24 of 32); every capture and intervention is on the block output at the image positions. The subspace is the top 256 principal components of the per-item mean image-position residual over the prior-valid ViLP and Visual CounterFact conflicts, and it is evaluated on the WHOOPS conflicts. The controls are those of Supp.~\ref{app:necessity}: the next 256 components, a random 256-D subspace inside the top 1,024, an off-block set from the per-token spectrum outside its top 1,024 block (with 20 weighted draws), and a random 256-D subspace of the whole space; the per-token covariance and mean use at least 64 clean forwards and 36,864 image positions. The criteria were pre-specified. At the image positions with 256 dimensions: (A) projecting the subspace out reverses more image-wins than either control, with the lower end of its interval above both controls' rates; (B) that lower end is above the share of prior-wins it reverses; (C) setting the subspace to its mean produces at least half of the change in the gap that replacing every image position by the mean produces (lower end of the ratio at or above 0.5), where the two controls produce under 0.2. The result holds on an architecture when A, B and C hold; sufficiency against the two span controls (D) is reported with it. \Cref{tab:archsub,tab:archsubmean} give the six architectures: InstructBLIP-7B, LFM2-VL-3B, Fuyu-8B and Emu3 meet A to D, Gemma-3-12B fails A, and Idefics2 fails C by the lower end of its interval. On all six the subspace set to its mean reverses 18 to 62\% of image-wins, against at most 8.5\% for a control set to its mean, and produces 0.51 to 0.94 of the change in the gap from replacing every image position. Seven further runs test scale, a second checkpoint or a second encoder within an architecture (middle and lower blocks of the two tables). Gemma-3 fails A at every size and also C at 27B, where the share falls to 0.28; InstructBLIP-13B fails B; LFM2-VL-1.6B fails C because a control gives 0.23 of the full change, above the 0.2 bound; Idefics2-8B-chatty, LLaVA-1.5-13B and Qwen2.5-VL meet A to D.

\begin{table}[h]\centering\footnotesize\setlength{\tabcolsep}{4.5pt}
\caption{\textbf{The image-token subspace on other architectures: setup of each run.} Image positions, the three blocks, the hidden size, the number of items the basis is fit on, the WHOOPS conflicts evaluated with their split into image-wins and prior-wins, and the share of the per-token variance the fitted subspace holds (one value, or one per block). Dash: not reported. The two calibration runs use the bases' fitting set of Supp.~\ref{app:necessity}.}\label{tab:subsetup}
\begin{tabular}{@{}llcccccc@{}}\toprule
Model & Image positions & Blocks & Hidden size & Fit items & Conflicts & Image- / prior-wins & Variance held \\ \midrule
InstructBLIP-7B & 32 Q-Former query tokens & 16, 20, 24 of 32 & 4,096 & 1,017 & 402 & 320 / 82 & -- \\
LFM2-VL-3B & image tokens & 15, 19, 23 of 30 & 2,048 & 1,037 & 410 & 318 / 92 & -- \\
Gemma-3-12B & pooled SigLIP tokens & 24, 30, 36 of 48 & 3,840 & 1,013 & 423 & 340 / 83 & -- \\
Idefics2-8B & 64 latent tokens & 16, 20, 24 of 32 & 4,096 & 1,018 & 417 & 341 / 76 & 0.43 \\
Fuyu-8B & raw-patch embeddings & 18, 23, 27 of 36 & 4,096 & 1,029 & 391 & 247 / 144 & -- \\
Emu3 & about 4,160 VQ tokens & 16, 20, 24 of 32 & 4,096 & 1,024 & 406 & 263 / 143 & 0.36 \\ \midrule
Gemma-3-4B & pooled SigLIP tokens & 17, 21, 26 of 34 & 2,560 & 1,033 & 421 & 341 / 80 & -- \\
Gemma-3-27B & pooled SigLIP tokens & 31, 39, 47 of 62 & 5,376 & 1,016 & 428 & 347 / 81 & 0.98, 0.92, 0.81 \\
InstructBLIP-13B & 32 Q-Former query tokens & 20, 25, 30 of 40 & 5,120 & 704 & 322 & 252 / 70 & -- \\
Idefics2-8B-chatty & 64 latent tokens & 16, 20, 24 of 32 & 4,096 & 1,022 & 423 & 338 / 85 & 0.46 \\ \midrule
LLaVA-1.5-7B (float32) & image tokens & 16, 20, 24 of 32 & 4,096 & -- & 416 & 307 / 109 & -- \\
BakLLaVA (bfloat16) & image tokens & 16, 20, 24 of 32 & 4,096 & -- & 420 & 314 / 106 & 0.92 to 0.96 \\
LLaVA-1.5-13B (bfloat16) & image tokens & 20, 25, 30 of 40 & 5,120 & 1,070 & 422 & 315 / 107 & 0.69, 0.53, 0.44 \\
Qwen2.5-VL-7B & image tokens & 14, 18, 21 of 28 & 3,584 & 1,057 & 410 & 330 / 80 & 0.50, 0.56, 0.51 \\ \bottomrule
\end{tabular}\end{table}

\begin{table}[h]\centering\footnotesize\setlength{\tabcolsep}{3.5pt}
\caption{\textbf{The image-token subspace on other architectures: sufficiency and removal} (256 dimensions; blocks at relative depth 0.5, 0.625, 0.75; WHOOPS conflicts; cluster-bootstrap 95\% intervals over images). Sufficiency: share of the gray shift recovered. Removal is at the image positions: share of image-wins reversed, and for the fitted subspace the share of prior-wins. Middle block: other sizes and a second checkpoint within an architecture. Lower block: the LLaVA architecture through the same code, LLaVA-1.5-7B and BakLLaVA as calibrations, LLaVA-1.5-13B, and Qwen2.5-VL-7B, which has a different encoder. The mean arms and the outcome are in \cref{tab:archsubmean}.}\label{tab:archsub}
\begin{tabular}{@{}lccccccc@{}}\toprule
 & \multicolumn{3}{c}{Sufficiency} & \multicolumn{3}{c}{Projected out, image-wins} & Prior-wins \\ \cmidrule(lr){2-4}\cmidrule(lr){5-7}\cmidrule(lr){8-8}
Model & Whole & Fitted & In-span / next & Fitted & Random & Off-block & Fitted \\ \midrule
InstructBLIP-7B & 0.67 & 0.53 [0.48, 0.58] & 0.20 / 0.06 & 29.4\% [23.7, 35.0] & 2.8\% & 0.0\% & 4.9\% [1.1, 9.9] \\
LFM2-VL-3B & 0.84 & 0.84 [0.76, 0.92] & 0.36 / 0.10 & 60.7\% [55.8, 66.1] & 5.3\% & 2.2\% & 10.9\% [4.9, 17.9] \\
Gemma-3-12B & 0.77 & 0.50 [0.41, 0.57] & 0.30 / $-$0.04 & 30.0\% [25.1, 34.8] & 47.6\% & 0.9\% & 7.2\% [2.5, 13.5] \\
Idefics2-8B & 0.74 & 0.53 [0.43, 0.64] & 0.13 / 0.02 & 14.4\% [11.1, 18.1] & 0.9\% & 0.3\% & 7.9\% [2.6, 14.5] \\
Fuyu-8B & 0.97 & 0.77 [0.69, 0.83] & 0.20 / 0.11 & 61.9\% [55.7, 67.9] & 8.5\% & 2.4\% & 12.5\% [7.4, 17.7] \\
Emu3 & 0.66 & 0.43 [0.37, 0.50] & 0.19 / 0.05 & 27.4\% [22.1, 33.0] & 1.5\% & 0.8\% & 4.9\% [1.9, 8.8] \\
\midrule
Gemma-3-4B & 0.66 & 0.36 [0.24, 0.46] & 0.19 / $-$0.02 & 36.7\% [31.6, 41.9] & 41.9\% & 2.6\% & 5.0\% [1.2, 10.2] \\
Gemma-3-27B & 0.67 & 0.35 [0.27, 0.42] & 0.08 / 0.04 & 15.3\% [11.7, 19.2] & 16.1\% & 0.6\% & 9.9\% [3.8, 17.4] \\
InstructBLIP-13B & 0.45 & 0.37 [0.32, 0.42] & 0.19 / 0.04 & 18.7\% [14.0, 23.5] & 1.2\% & 0.0\% & 18.6\% [9.5, 27.3] \\
Idefics2-8B-chatty & 0.71 & 0.55 [0.48, 0.62] & 0.18 / 0.02 & 24.3\% [19.6, 28.8] & 1.8\% & 0.3\% & 1.2\% [0.0, 4.3] \\
LFM2-VL-1.6B & 0.77 & 0.73 [0.65, 0.80] & 0.31 / 0.11 & 49.7\% [43.8, 55.4] & 2.9\% & 1.0\% & 10.6\% [4.8, 17.0] \\
\midrule
LLaVA-1.5-7B (float32) & 0.77 & 0.61 [0.55, 0.67] & 0.22 / 0.13 & 33.9\% [28.8, 39.2] & 1.6\% & 0.3\% & 2.8\% [0.0, 6.2] \\
BakLLaVA (bfloat16) & 0.74 & 0.66 [0.59, 0.73] & 0.16 / 0.05 & 34.7\% [29.2, 39.9] & 1.3\% & 0.3\% & 7.5\% [2.9, 13.5] \\
LLaVA-1.5-13B & 0.54 & 0.46 [0.41, 0.53] & 0.17 / 0.09 & 11.1\% [7.8, 14.6] & 1.0\% & 0.6\% & 6.5\% [2.6, 11.8] \\
Qwen2.5-VL-7B & 0.86 & 0.78 [0.72, 0.84] & 0.32 / 0.06 & 43.9\% [38.3, 49.1] & 1.5\% & 0.9\% & 2.5\% [0.0, 6.3] \\ \bottomrule
\end{tabular}\end{table}

\begin{table}[h]\centering\footnotesize\setlength{\tabcolsep}{5pt}
\caption{\textbf{The image-token subspace on other architectures: mean arms and outcome} (models and blocks as in \cref{tab:archsub}). Share of image-wins reversed when every image position, or only the fitted subspace, is set to its mean, and when only the subspace is kept. Share: change in the gap with the fitted subspace set to its mean, as a fraction of the change with every image position set to the mean. Outcome: the pre-specified readings A to C (\cref{tab:pre-specified}).}\label{tab:archsubmean}
\begin{tabular}{@{}lccccc@{}}\toprule
 & \multicolumn{4}{c}{Mean arms, image-wins} & \\ \cmidrule(lr){2-5}
Model & All positions & Subspace & Share & Only subspace kept & Outcome \\ \midrule
InstructBLIP-7B & 39.1\% & 25.6\% & 0.67 [0.62, 0.72] & 16.9\% & holds \\
LFM2-VL-3B & 66.0\% & 61.9\% & 0.94 [0.92, 0.96] & 8.8\% & holds \\
Gemma-3-12B & 49.4\% & 27.4\% & 0.70 [0.68, 0.73] & 14.7\% & fails A \\
Idefics2-8B & 32.0\% & 18.5\% & 0.51 [0.47, 0.57] & 5.9\% & fails C \\
Fuyu-8B & 61.9\% & 53.4\% & 0.92 [0.86, 0.98] & 28.3\% & holds \\
Emu3 & 56.7\% & 43.0\% & 0.75 [0.71, 0.78] & 32.3\% & holds \\
\midrule
Gemma-3-4B & 33.4\% & 22.9\% & 0.71 [0.67, 0.75] & 6.5\% & fails A \\
Gemma-3-27B & 26.5\% & 8.9\% & 0.28 [0.24, 0.33] & 11.5\% & fails A, C \\
InstructBLIP-13B & 24.2\% & 20.2\% & 0.63 [0.53, 0.73] & 5.2\% & fails B \\
Idefics2-8B-chatty & 34.6\% & 25.7\% & 0.70 [0.68, 0.73] & 6.8\% & holds \\
LFM2-VL-1.6B & 63.1\% & 59.2\% & 0.90 [0.89, 0.92] & 12.7\% & fails C \\
\midrule
LLaVA-1.5-7B (float32) & 54.7\% & 48.2\% & 0.87 [0.85, 0.88] & 12.4\% & holds \\
BakLLaVA (bfloat16) & 40.8\% & 36.3\% & 0.87 [0.84, 0.89] & 6.1\% & holds \\
LLaVA-1.5-13B & 23.2\% & 16.2\% & 0.84 [0.80, 0.88] & 8.6\% & holds \\
Qwen2.5-VL-7B & 73.9\% & 57.0\% & 0.76 [0.74, 0.78] & 9.4\% & holds \\ \bottomrule
\end{tabular}\end{table}

\begin{table}[h]\centering\footnotesize\setlength{\tabcolsep}{3.5pt}
\caption{\textbf{The image-token subspace on other architectures: controls under the mean arms, and the answer position.} Image-wins reversed (\%), with 95\% intervals, when every image position or only the fitted subspace is set to its mean. Controls (random / off-block): image-wins reversed with the control set to its mean, the share of the all-positions change in the gap it produces, and image-wins reversed when only the control is kept. Answer position: image-wins / prior-wins reversed by projecting the subspace out there. One value where the two controls agree or one is reported. Dash: not reported.}\label{tab:subctrl}
\begin{tabular}{@{}lcccccc@{}}\toprule
 & \multicolumn{2}{c}{Set to the mean} & \multicolumn{3}{c}{Controls (random / off-block)} & Answer \\ \cmidrule(lr){2-3}\cmidrule(lr){4-6}
Model & All positions & Subspace & At mean & Share & Only control kept & position \\ \midrule
InstructBLIP-7B & 39.1 [33.6, 44.4] & 25.6 & 1.6 / 0.3 & $-0.02$ / $-0.03$ & 36.2 / 37.8 & 6.9 / 14.6 \\
LFM2-VL-3B & 66.0 [60.8, 71.2] & 61.9 & 3.8 / 0.6 & 0.08 / 0.01 & 69.2 / 64.8 & 6.0 / 19.6 \\
Gemma-3-12B & 49.4 [44.1, 54.9] & 27.4 [22.7, 32.3] & 5.3 / 1.2 & 0.12 / 0.00 & 50.0 / 48.2 & -- \\
Idefics2-8B & 32.0 [27.3, 36.9] & 18.5 [14.7, 22.6] & 0.6 / 0.0 & 0.01 / 0.00 & 29.6 & 2.6 / 7.9 \\
Fuyu-8B & 61.9 [56.0, 68.0] & 53.4 [47.1, 59.8] & 8.5 / 1.2 & 0.14 / 0.02 & 61.9 / 58.7 & 8.1 / 6.2 \\
Emu3 & 56.7 [50.7, 62.7] & 43.0 [37.2, 49.6] & 2.7 / 0.4 & 0.06 / 0.02 & 56.3 & 5.3 / 7.7 \\ \midrule
Gemma-3-4B & 33.4 [28.8, 38.3] & 22.9 [18.7, 27.2] & 3.2 / 1.2 & -- & 37.5 / 33.1 & -- \\
Gemma-3-27B & 26.5 [21.9, 31.3] & 8.9 [6.1, 11.9] & 2.0 / 1.2 & 0.10 / $-0.02$ & 27.1 / 26.8 & -- \\
InstructBLIP-13B & 24.2 [19.3, 30.2] & 20.2 [15.3, 25.6] & 1.6 / 0.0 & $-0.02$ / 0.00 & 23.0 / 23.8 & -- \\
Idefics2-8B-chatty & 34.6 & 25.7 [20.9, 30.4] & 1.2 / 0.0 & -- & -- & -- \\ \midrule
LLaVA-1.5-13B & 23.2 [18.9, 27.5] & 16.2 [12.5, 20.3] & 1.6 / 0.3 & 0.07 / 0.01 & 23.2 & -- \\
Qwen2.5-VL-7B & 73.9 [69.3, 78.6] & 57.0 [51.5, 61.9] & 1.2 / 0.3 & 0.04 / 0.00 & 73.3 & -- \\ \bottomrule
\end{tabular}\end{table}

\begin{table}[h]\centering\footnotesize\setlength{\tabcolsep}{6pt}
\caption{\textbf{Removal against the number of dimensions.} Image-wins reversed (\%) when the top $k$ components of the fitted subspace are projected out at the image positions. The 128-dimension row of LFM2-VL-3B uses the basis cut at 128. The last two rows are the calibration on LLaVA-1.5-7B in float32: the all-positions basis of this section against the basis of Supp.~\ref{app:necessity}.}\label{tab:subdims}
\begin{tabular}{@{}lccccccccc@{}}\toprule
Model & $k$ = 1 & 2 & 4 & 8 & 16 & 32 & 64 & 128 & 256 \\ \midrule
InstructBLIP-7B & 0.9 & 0.6 & 1.6 & 3.1 & 4.4 & 9.1 & 15.3 & 24.1 & 29.4 \\
LFM2-VL-3B & 4.1 & 7.9 & 10.1 & 13.2 & 14.5 & 24.5 & 37.1 & 48.4 & 60.7 \\
LFM2-VL-3B, 128 dimensions & 3.8 & 8.2 & 10.4 & 12.9 & 14.5 & 23.0 & 37.4 & 48.7 & -- \\
Idefics2-8B & 0.9 & 1.2 & 1.2 & 1.5 & 1.5 & 2.6 & 4.7 & 8.8 & 14.4 \\
Fuyu-8B & 21.5 & 25.1 & 25.9 & 29.6 & 36.4 & 39.7 & 51.0 & 57.9 & 61.9 \\
Emu3 & 0.4 & 1.5 & 1.1 & 1.5 & 4.2 & 8.4 & 8.4 & 17.5 & 27.4 \\
InstructBLIP-13B & 0.0 & 9.5 & 7.5 & 9.9 & 11.5 & 9.5 & 14.7 & 18.3 & 18.7 \\
LLaVA-1.5-13B & 1.9 & 1.9 & 1.9 & 2.2 & 2.9 & 3.2 & 7.0 & 9.5 & 11.1 \\
Qwen2.5-VL-7B & 4.2 & 14.5 & 14.5 & 11.5 & 10.6 & 13.0 & 15.8 & 26.4 & 43.9 \\ \midrule
LLaVA-1.5-7B, all-positions basis & 1.6 & 1.6 & 2.6 & 3.3 & 4.6 & 8.1 & 11.1 & 20.8 & 33.9 \\
LLaVA-1.5-7B, basis of Supp.~\ref{app:necessity} & 1.6 & 3.6 & 4.9 & 7.8 & 11.4 & 15.6 & 24.8 & 32.6 & 42.3 \\ \bottomrule
\end{tabular}\end{table}

\noindent\textbf{Notes by model.} \Cref{tab:subsetup} gives the setup of each run, \cref{tab:archsub,tab:archsubmean} the main arms, \cref{tab:subctrl} the controls under the mean arms and the removal at the answer position, and \cref{tab:subdims} the removal against the number of dimensions. The notes below give what is particular to each model.

\noindent\emph{InstructBLIP-7B.} The image positions are the Q-Former query tokens. Keeping only the subspace still reverses 16.9\% of image-wins, about half as many as keeping only a control. At the answer position the same removal reverses more prior-wins than image-wins, the pattern of the LLaVA models. In the sufficiency test the off-block control recovers 0.01 of the gray shift. Removing the subspace from each block's own update reverses no image-win.

\noindent\emph{LFM2-VL-3B.} The subspace is nearly all of what the image tokens carry: it recovers as much of the gray shift as the whole residual (0.84), and set to its mean it produces 0.94 of the all-positions change. At 256 dimensions the subspace is an eighth of this model's space. With 128 dimensions, the same sixteenth as on the 4,096-wide models, A to D still hold: projecting the subspace out reverses 48.7\% [43.4, 53.7] of image-wins and 4.3\% of prior-wins, against 2.2\% and 0.9\% of image-wins for the two controls; set to its mean it reverses 47.8\% [42.1, 52.8] and produces 0.77 [0.74, 0.79] of the all-positions change (controls 0.04 and 0.00); keeping only the subspace reverses 15.1\%, against 67\% when a control is kept; and it recovers 0.83 [0.76, 0.92] of the gray shift against 0.19 and 0.18 for the random-in-span and next-128 controls.

\noindent\emph{Gemma-3-12B.} The test fails at A. Projecting the fitted subspace out reverses 30.0\% [25.1, 34.8] of image-wins and 7.2\% of prior-wins, and the off-block control 0.9\%, but projecting a random 256-D subspace out reverses 47.6\% [42.0, 53.3] of image-wins and 20.5\% of prior-wins. On this architecture setting any 256 directions of the image tokens to zero is destructive unless they come from the low-variance tail: the random projection lowers the gap of image-wins by 9.41 logits, as much as replacing every image token by the mean (9.34 logits; 49.4\% [44.1, 54.9] reversed), and zeroing the single top component already reverses 28.2\%. The arms that leave the component's average in place separate the subspace from the controls. Set to its mean, the fitted subspace reverses 27.4\% [22.7, 32.3] of image-wins and 6.0\% of prior-wins and produces 0.70 [0.68, 0.73] of the change in the gap from replacing every image token; the random control reverses 5.3\% [2.9, 7.8] (0.12) and the off-block control 1.2\% (0.00). Keeping only the subspace reverses 14.7\%, where keeping only a control reverses 50.0\% and 48.2\%. Sufficiency is 0.50 [0.41, 0.57] against 0.30 [0.22, 0.38] for the random-in-span control and $-0.04$ for the next 256 components (whole residual 0.77). B, C and D therefore hold and A does not. The comparison of the mean arms was not among the pre-specified criteria; we added it as a further criterion (A$'$: the subspace set to its mean reverses more image-wins than either control set to its mean) before any later result was known, and it is post hoc for the three architectures above, all of which meet it (InstructBLIP-7B 25.6\% against 1.6\% and 0.3\%; LFM2-VL-3B 61.9\% against 3.8\% and 0.6\%).

\noindent\emph{Idefics2-8B.} The image positions are the latent tokens of the perceiver resampler. The subspace holds 0.43 of the per-token variance, the least of any architecture here. A, B and D hold and C fails: set to its mean the subspace produces 0.51 [0.47, 0.57] of the all-positions change in the gap, so the lower end of the ratio is under the 0.5 that C requires. On this architecture the subspace is about half of what the image tokens carry for the decision. It meets A$'$ (18.5\% against 0.6\% and 0.0\%), which was specified before this model's results were known. At the answer position the removal reverses as many prior-wins as the random control does. With 384 dimensions (an additional dimension; the 256-dimension outcome is the one counted) A to D hold: set to its mean the subspace reverses 19.6\% [15.8, 23.6] of image-wins, against 0.9\% and 0.3\% for the controls, and produces 0.59 [0.55, 0.64] of the all-positions change; projected out it reverses 17.9\% [14.1, 21.7] of image-wins and 6.6\% of prior-wins, against 0.6\% and 0.3\% of image-wins for the controls; keeping only the subspace reverses 5.0\%; sufficiency is 0.57 [0.47, 0.68] against 0.23 for the random-in-span control. A second checkpoint of the same architecture, Idefics2-8B-chatty (the same architecture with a different fine-tune; pre-specified as an additional model; the Idefics2-8B outcome is the one counted), meets A to D at 256 dimensions. Its subspace holds 0.46 of the per-token variance at each block, as on Idefics2-8B (0.43), and produces 0.70 of the all-positions change. The share near one half on Idefics2-8B is therefore a property of that checkpoint and not of the perceiver-resampler architecture.

\noindent\emph{Fuyu-8B.} Fuyu has no vision encoder: the image positions are linear embeddings of raw patches. A to D hold, and A$'$, specified before this model's results were known, holds. The top component alone reverses 21.5\% of image-wins, and the random control projected out reverses 8.5\% [5.5, 12.1]. The whole residual recovers 0.97 of the gray shift, the highest of any architecture, in line with the late crossing of the band map. Keeping only the subspace still reverses 28.3\% of image-wins: the subspace carries nearly all of the removable effect and the remaining dimensions are still needed to hold the decision.

\noindent\emph{Emu3.} The image positions are discrete VQ tokens from the vocabulary the model also generates. The subspace holds 0.36 of the per-token variance. A to D and A$'$ hold.

\noindent\emph{Gemma-3-4B} repeats the 12B pattern: A fails and B, C, D and A$'$ hold. A random 256-D subspace projected out reverses 41.9\% [36.9, 47.1] of image-wins, more than the fitted subspace (36.7\%), and zeroing the single top component 34.0\%, as many as replacing every image token by the mean (33.4\%). The failure of A is a property of this architecture at both sizes: on Gemma-3, zeroing a random 256-D subspace or the single top component removes about as much as replacing every image token by the mean.

\noindent\emph{Gemma-3-27B} fails A and C; B, D and A$'$ hold. Here 256 dimensions are 4.8\% of the space, against 10\% at 4B. A random 256-D subspace projected out reverses 16.1\% [12.2, 20.1] of image-wins, as many as the fitted subspace (15.3\%), and zeroing the top two components 23.1\%, close to replacing every image token by the mean (26.5\%). Set to its mean, the fitted subspace produces 0.28 [0.24, 0.33] of the all-positions change in the gap, where the share is 0.71 at 4B and 0.70 at 12B. Keeping only the subspace reverses 11.5\% [8.3, 15.5] and produces 0.37 of the all-positions change. So with the subspace at its mean 0.72 of the image tokens' effect remains, and with only the subspace kept 0.63 remains: at this size either part alone carries most of it, although the subspace holds 0.98, 0.92 and 0.81 of the per-token variance at the three blocks. With 384 dimensions (7.1\% of the space, the most the 1,016 fit items allow with a next-384 control) A and C still fail: set to its mean the subspace reverses 12.1\% [8.9, 15.8] of image-wins, against 1.7\% and 1.2\% for the controls, and produces 0.39 [0.35, 0.44] of the all-positions change; projected out it reverses 16.4\% [12.8, 20.5], against 20.2\% for a random 384-D subspace; sufficiency is 0.41 [0.34, 0.48] against 0.20 for the random-in-span control.

\noindent\emph{InstructBLIP-13B} fails B; A, C, D and A$'$ hold. Projected out, the subspace reverses 18.6\% [9.5, 27.3] of prior-wins, where it reversed 4.9\% at 7B. Prior-wins are unstable on this model under every arm: the random control projected out reverses 11.4\% of them, the off-block control 5.7\%, and replacing every query token by the mean 22.9\%. The whole residual recovers 0.45 of a gray shift that is small on this model (0.96 logits).

\noindent\emph{LLaVA-1.5-13B} meets A to D and A$'$, with smaller effects than the 7B model: replacing every image token by the mean reverses 23.2\% of image-wins, against 54.7\% at 7B.

\noindent\emph{Qwen2.5-VL-7B} meets A to D. It has the LLaVA architecture, a ViT encoder with an MLP projector, with a different encoder and dynamic resolution.

\noindent\textbf{Numerical precision and calibration.} The projections are computed in float32 and written back in the model's precision; on four models the float32 matrix products used a reduced-precision mode, which moved the margin by at most 0.016 of its value in the most sensitive check (a full-rank identity basis); on the models listed next the mode is off and that check deviates by at most $6\times10^{-5}$ (Fuyu, Emu3, Gemma-3-4B, Gemma-3-27B, InstructBLIP-13B). No gap was non-finite on any of the six models of the upper block, on Gemma-3 at 4B and 27B, on InstructBLIP-13B or on LLaVA-1.5-13B.

\noindent\emph{Calibration on LLaVA-1.5-7B.} LLaVA-1.5-7B was analyzed with the all-positions basis in float32 (416 conflicts: 307 image-wins, 109 prior-wins; no non-finite gap; A to D hold; first row of the lower block of \cref{tab:archsub}). The arms that leave the component's average in place reproduce Supp.~\ref{app:necessity}, and the zeroing arm and sufficiency come out lower (\cref{tab:subcalib}; \cref{tab:subdims} gives the two curves over dimensions). The difference is the basis and not the precision. Repeated in float32 with the bases of Supp.~\ref{app:necessity}, the removal of Supp.~\ref{app:necessity} is unchanged: projecting the subspace out reverses 42.3\% [37.1, 48.2] of image-wins and 3.7\% of prior-wins and lowers the gap of image-wins by 3.28 logits, as in float16 (42.3\%, 3.7\%, 3.28), with 0.72 [0.68, 0.75] of the all-positions change; the subspace at its mean reverses 47.2\% (47.9\% in float16; share 0.86 in both), every image position at the mean 55.4\% (54.7\%), and keeping only the subspace 13.4\% (13.4\%). The two bases differ in what is averaged: the attended image tokens in Supp.~\ref{app:necessity}, all image positions here. On LLaVA-1.5 the two 256-D subspaces are not the same: their mean squared cosine is 0.53, 0.58 and 0.59 at the three blocks (0.06 by chance), the leading directions agree more (the top 32 of the Supp.~\ref{app:necessity} basis lie 0.83 to 0.86 inside the all-positions span), and the all-positions basis holds more of the mean image-token residual (0.74, 0.70 and 0.80 of its squared norm, against 0.50, 0.50 and 0.68). Zeroed, the all-positions basis removes less (33.9\% against 42.3\%); set to its mean, it removes as much (48.2\% against 47.2\%, share 0.87 against 0.86).

\noindent\emph{Calibration on BakLLaVA.} BakLLaVA with the all-positions basis (bfloat16; 420 conflicts: 314 image-wins, 106 prior-wins; A to D hold; second row of the lower block) agrees with Supp.~\ref{app:necessity} on the image-wins of every arm (\cref{tab:subcalib}, against the float32 values of Supp.~\ref{app:necessity}); prior-wins reversed by the removal are higher. On BakLLaVA, where the fitted subspace holds 0.92 to 0.96 of the per-token variance, which tokens are averaged does not move the removal of image-wins.

\noindent\emph{float16.} In float16 the zeroing arm with the all-positions basis is not usable on LLaVA-1.5: with 128 or more dimensions projected out the gap is non-finite on many items (152 of 416 at 128 dimensions, 260 at 256). The float16 curve of Supp.~\ref{app:necessity} on LLaVA-1.5 has non-finite gaps at 1 and 2 dimensions at the image tokens, counted as reversals (7.2\% and 13.0\%); main-paper Fig.~5b shows the float32 curve, which has none (Supp.~\ref{app:necessity}, Precision). Llama-3.2 has no image positions in the sequence and is not run.

\begin{table}[h]\centering\footnotesize\setlength{\tabcolsep}{6pt}
\caption{\textbf{Calibration of the all-positions basis} against the attended-token basis of Supp.~\ref{app:necessity} on the two LLaVA-architecture models. Image-wins reversed (\%) unless noted; 95\% intervals in brackets.}\label{tab:subcalib}
\begin{tabular}{@{}lcccc@{}}\toprule
 & \multicolumn{2}{c}{LLaVA-1.5-7B} & \multicolumn{2}{c}{BakLLaVA} \\ \cmidrule(lr){2-3}\cmidrule(lr){4-5}
Arm & All positions & Supp.~\ref{app:necessity} & All positions & Supp.~\ref{app:necessity} \\ \midrule
Every image position at the mean & 54.7 & 54.7 & 40.8 & 40.0 \\
Subspace at its mean & 48.2 & 47.9 & 36.3 & 35.9 \\
Share of the all-positions change & 0.87 [0.85, 0.88] & 0.86 & 0.87 [0.84, 0.89] & 0.89 \\
Only the subspace kept & 12.4 & 13.4 & 6.1 & 6.7 \\
Subspace projected out & 33.9 [28.8, 39.2] & 42.3 [37.1, 48.2] & 34.7 [29.2, 39.9] & 34.3 \\
Subspace projected out, prior-wins & 2.8 & 3.7 & 7.5 [2.9, 13.5] & 2.9 \\
Share of the gray shift recovered & 0.61 [0.55, 0.67] & 0.69 & 0.66 [0.59, 0.73] & 0.67 \\ \bottomrule
\end{tabular}\end{table}

\subsection{Further checkpoints: training stage and reasoning mode}
\noindent\textbf{Eight further checkpoints.} Outcomes, the dose grid and the additive-model fit were also measured on Qwen3-VL-8B (instruct and thinking), MiMo-VL-7B (SFT and RL) and InternVL3.5-8B (the pretrained, instruct, MPO and final stages; tiles of 448 pixels), all with dynamic resolution and the dose on the full-size image (\cref{tab:stage}). Both weights are positive on 50 of 50 item splits on each. With reasoning switched on or off, 100 sampled items per model give the same answer on 0.95 (Qwen3-VL thinking), 0.93 (MiMo-VL SFT) and 0.96 (MiMo-VL RL) of the pairs where both replies parse. On MiMo-VL RL the held-out deviance explained is negative on the main split and 0.098 at the median of the 50 splits; the SFT and RL stages of the same model give the same precision weight (0.24 and 0.25) and vision-win rate (0.774 and 0.779). In the boundary fit the dose-by-prior term is significant on both MiMo-VL stages ($p$ = 0.01 at SFT and $p$ $<$ 0.005 at RL; adding it lowers held-out deviance explained by 0.30 and 0.86) and on none of the other six ($p$ $\geq$ 0.23). These eight grids are not among the 28 refit on every cell in Supp.~\ref{app:rulecmp}. Across the four InternVL3.5 stages the precision weight is 0.14, 0.13, 0.17 and 0.16 and the vision-win rate 0.760, 0.739, 0.746 and 0.752; the pretrained stage has a negative held-out value on the main split and 0.061 at the median split. With reasoning on or off the final InternVL3.5 checkpoint gives the same answer on 0.92 of 59 parsed pairs (19\% of its reasoning replies hit the token limit).

\begin{table}[h]\centering\footnotesize\setlength{\tabcolsep}{4pt}
\caption{\textbf{The additive model on eight further checkpoints} (WHOOPS; dose on the full-size image). Vision-win on the clean image and at blur radius 8; $a$, $b$, held-out deviance explained and the 95th percentile of the permutation null as in \cref{tab:archfit}.}\label{tab:stage}
\begin{tabular}{lcccccc}\toprule
Model & Vision-win & $a$ & $b$ & Held out & Null 95th & Vision-win at blur 8 \\ \midrule
Qwen3-VL-8B instruct & 0.815 & 0.14 [0.05, 0.24] & 0.23 [0.10, 0.36] & 0.082 & 0.043 & 0.709 \\
Qwen3-VL-8B thinking & 0.786 & 0.14 [0.06, 0.22] & 0.17 [0.05, 0.30] & 0.083 & 0.039 & 0.650 \\
MiMo-VL-7B SFT & 0.774 & 0.24 [0.16, 0.32] & 0.29 [0.14, 0.44] & 0.024 & 0.024 & 0.541 \\
MiMo-VL-7B RL & 0.779 & 0.25 [0.16, 0.35] & 0.38 [0.22, 0.54] & $-$0.077 & 0.018 & 0.550 \\
InternVL3.5-8B pretrained & 0.760 & 0.14 [0.06, 0.22] & 0.29 [0.15, 0.44] & $-$0.042 & 0.022 & 0.622 \\
InternVL3.5-8B instruct & 0.739 & 0.13 [0.07, 0.20] & 0.19 [0.04, 0.34] & 0.166 & 0.079 & 0.595 \\
InternVL3.5-8B MPO & 0.746 & 0.17 [0.10, 0.24] & 0.17 [0.04, 0.30] & 0.081 & 0.044 & 0.634 \\
InternVL3.5-8B final & 0.752 & 0.16 [0.09, 0.23] & 0.23 [0.09, 0.37] & 0.041 & 0.034 & 0.636 \\ \bottomrule
\end{tabular}\end{table}

\subsection{Object edits, enhancement and the positive control}
\noindent\textbf{Object edits.} The object-minus-rest difference excludes zero for all three edits on the first seven models. On Gemma-3-12B it excludes zero for blur and for noise and not for gray ($-0.20$ [$-0.67$, $+0.27$]): graying everything except the object costs $-0.73$ logits, more than on any other model, and graying the whole image $-1.95$. Graying the object still lowers the margin by 0.97 logits [0.53, 1.45] more than graying a same-area region elsewhere. The gray interval also includes zero at 4B and 27B, so on this architecture graying the surroundings costs about as much as graying the object, while blur and noise separate the two. The next statements are for the first seven models. On the subset where the region elsewhere exists, every difference keeps its sign, and three of the 21 intervals include zero (blur on LFM2-VL, noise on Idefics2 and Emu3); on the near-square images, where the 336-pixel masks map onto the image without distortion, every interval again excludes zero. The LLaVA-1.5 row reproduces the gray-object and gray-rest values of the model-specific analysis ($-0.47$, $-0.09$). The rest of the image carries more on Idefics2, LFM2-VL, Emu3 and Llama-3.2 than on LLaVA-1.5, InstructBLIP and Fuyu.

\noindent\textbf{Enhancing the object (nine models, eight architectures).} The additive model predicts that degrading the object lowers vision-win. We pre-specified the converse as a separate hypothesis: that enhancing the object alone raises the vision-win rate above the clean image, where whole-image enhancement did not in an earlier study \cite{vankadaru2026trust}. On prior-valid WHOOPS conflicts with a mask we applied an unsharp mask at 150, 300 and 500\% and a contrast gain of 1.25, 1.5 and 2 to the object, and we cropped the image to the object's box or to a box of the same size elsewhere. The hypothesis fails on all nine models: no enhancement of the object raises vision-win (\cref{tab:enhance}), and a crop that keeps the object keeps most of the decision where a crop that loses it does not (\cref{tab:crop}).

\begin{table}[h]\centering\footnotesize\setlength{\tabcolsep}{4.5pt}
\caption{\textbf{Enhancing the object.} Change in vision-win, in points, against the clean image when only the object is sharpened or its contrast raised. $^\ast$: the 95\% interval excludes zero (always below). Edits are applied at the size given (full: the original image; Emu3: its input budget of 262,144 pixels). Fuyu: one item failed.}\label{tab:enhance}
\begin{tabular}{llcccccccc}\toprule
 & & & & \multicolumn{3}{c}{Sharpening (\%)} & \multicolumn{3}{c}{Contrast gain} \\ \cmidrule(lr){5-7}\cmidrule(lr){8-10}
Model & Edits at & Conflicts & Clean vision-win & 150 & 300 & 500 & 1.25 & 1.5 & 2 \\ \midrule
LLaVA-1.5 & 336 & 361 & 0.778 & $-0.3$ & $-0.3$ & $-1.4$ & $+0.3$ & $-0.3$ & $-0.6$ \\
InstructBLIP-7B & 224 & 344 & 0.846 & $+0.6$ & $-0.9$ & $-2.3^\ast$ & $+0.3$ & $+0.3$ & $-0.6$ \\
Idefics2 & full & 356 & 0.860 & $+0.3$ & $0.0$ & $-1.4$ & $+0.6$ & $+0.8$ & $-0.6$ \\
LFM2-VL & full & 352 & 0.827 & $-1.7^\ast$ & $-0.9$ & $-1.4$ & $-0.3$ & $-0.3$ & $-0.6$ \\
Llama-3.2 & 560 & 362 & 0.807 & $-0.8$ & $-1.7$ & $-2.2$ & $0.0$ & $-0.6$ & $0.0$ \\
Emu3 & budget & 349 & 0.699 & $0.0$ & $+0.6$ & $-0.9$ & $-0.9$ & $-1.1$ & $+0.3$ \\
Gemma-3-12B & full & 363 & 0.851 & $+0.6$ & $-0.3$ & $-1.4$ & $0.0$ & $-0.3$ & $-0.3$ \\
Qwen2.5-VL & full & 351 & 0.840 & $0.0$ & $-0.3$ & $-1.1$ & $-0.6$ & $-0.6$ & $-2.0^\ast$ \\
Fuyu-8B & full & 333 & 0.733 & $-0.9$ & $-1.8$ & $-4.5^\ast$ & $-1.2$ & $-1.8$ & $-2.7$ \\ \bottomrule
\end{tabular}\end{table}

\noindent\emph{Notes on \cref{tab:enhance}.} The intervals of the starred cells are [$-4.4$, $-0.3$] (InstructBLIP-7B), [$-3.1$, $-0.6$] (LFM2-VL), [$-3.7$, $-0.3$] (Qwen2.5-VL) and [$-7.5$, $-1.5$] (Fuyu-8B); Fuyu-8B at contrast gain 2 has [$-5.4$, 0.0] and InstructBLIP-7B at gain 1.25 has [0.0, 0.9]. LLaVA-1.5 has 281 image-wins and 80 prior-wins. On it, sharpening the object moves the gap by $-0.02$ [$-0.08$, $+0.04$], $-0.08$ [$-0.16$, $-0.00$] and $-0.14$ [$-0.25$, $-0.05$] logits, and raising its contrast by $+0.00$, $-0.01$ and $-0.07$. Sharpening the whole image costs more ($-0.08$, $-0.16$ and $-0.25$ [$-0.35$, $-0.17$] logits), and the object-minus-whole difference at 500\% is $+0.11$ [$+0.00$, $+0.23$], so enhancing only the object is less harmful than enhancing everything. The same edits were also applied to everything except the object and to a same-area region elsewhere (292 items). Idefics2 has 306 image-wins and 50 prior-wins. On it the gap moves by $-0.01$ to $-0.32$ logits, object-only enhancement reverses 6 to 12\% of the 50 prior-wins and loses 0.3 to 3.6\% of the 306 image-wins, and sharpening the whole image at 500\% costs 3.1 points [1.1, 5.3].

\begin{table}[h]\centering\footnotesize\setlength{\tabcolsep}{5pt}
\caption{\textbf{Cropping to the object.} Change in vision-win, in points, when the image is cropped to the object's box, and when it is cropped to a box of the same size elsewhere (with the resulting vision-win). Last column: margin under the object crop minus margin under the crop elsewhere, in logits. Intervals are 95\%. Dash: not reported.}\label{tab:crop}
\begin{tabular}{lcccc}\toprule
Model & Object crop & Crop elsewhere & Vision-win after crop elsewhere & Object crop minus elsewhere \\ \midrule
LLaVA-1.5 & $-$4.5 [$-$8.1, $-$0.8] & -- & 0.405 & $+2.98$ [$+2.51$, $+3.40$] \\
InstructBLIP-7B & $-$2.1 [$-$5.1, $+$1.2] & $-$33.3 & 0.496 & $+2.81$ [$+2.30$, $+3.33$] \\
Idefics2 & $-$6.2 [$-$10.1, $-$2.5] & -- & 0.402 & $+4.21$ [$+3.56$, $+4.84$] \\
LFM2-VL & $-$13.4 [$-$17.3, $-$9.4] & $-$44.9 & 0.379 & $+2.63$ [$+2.08$, $+3.18$] \\
Llama-3.2 & $-$6.9 [$-$10.5, $-$3.3] & $-$42.3 & 0.396 & $+3.87$ [$+3.28$, $+4.49$] \\
Emu3 & $-$7.8 [$-$11.8, $-$4.0] & $-$37.3 & 0.306 & $+3.17$ [$+2.72$, $+3.66$] \\
Gemma-3-12B & $-$6.9 [$-$11.0, $-$3.6] & $-$44.7 & 0.398 & $+5.97$ [$+4.97$, $+6.97$] \\
Qwen2.5-VL & $-$9.7 [$-$13.4, $-$6.0] & $-$46.9 & 0.372 & $+4.21$ [$+3.61$, $+4.91$] \\
Fuyu-8B & $+0.3$ [$-4.2$, $+4.2$] & $-$32.9 & 0.387 & $+3.44$ [$+2.90$, $+4.01$] \\ \bottomrule
\end{tabular}\end{table}

\noindent\emph{Notes on \cref{tab:crop}.} On LLaVA-1.5 the object crop lowers the gap by $-0.51$ [$-0.76$, $-0.27$], reversing 17.7\% of prior-wins and 10.8\% of image-wins, and the crop elsewhere lowers it by $-3.69$ [$-4.21$, $-3.18$]. On Idefics2 the two crops lower the gap by $-1.05$ [$-1.43$, $-0.71$] and $-5.69$ [$-6.36$, $-4.99$]. On Fuyu-8B the object crop leaves the rate unchanged (gap $+0.14$ [$-0.16$, $+0.41$]), reversing 29.2\% of prior-wins and 10.2\% of image-wins. The object crop costs more on LFM2-VL than on any other architecture. Split by the clean outcome, the object crop reverses 12.5 to 29.2\% of prior-wins on the nine models (a crop of the same size elsewhere 2.3 to 15.9\%) and raises the mean gap on prior-wins on eight (LFM2-VL: $-0.20$ logits). It loses 6.7 to 19.2\% of the image-wins (the crop elsewhere 42 to 57\%), and since image-wins outnumber prior-wins by 2.3 to 6.1 to one, the rate falls on eight models. The prior-wins that the object crop reverses have smaller clean gaps than prior-wins overall (median absolute gap 0.50 to 1.75 logits against 1.82 to 4.19), 14 to 64\% of them lie within 1 logit of zero on the clean image, and after the crop their median gap is 1.0 to 2.6 logits toward the image.

\noindent\textbf{Positive control for the object edits.} No enhancement or crop above raises vision-win, so we tested whether the measurement registers a rise when the object's evidence improves. The baseline is the whole image under the strongest blur (radius 8 at the size of the model's dose grid); each arm puts the clean pixels back in one region. The reading was fixed before the run: the control passes on a model if restoring the object raises vision-win above the blurred baseline with a 95\% interval above zero, and by more than restoring a region of the same area elsewhere (paired interval above zero). It passes on all nine models (\cref{tab:poscontrol}). Restoring the object, a median 10 to 11\% of the pixels, raises vision-win by 9.8 to 35.2 points; restoring the same area elsewhere changes it by $-2.1$ to $+5.5$ points, with intervals that include zero on seven models. The object's gain exceeds the gain of the region elsewhere on every model, and its gap shift exceeds the shift from restoring everything except the object on every model (by 0.32 to 1.33 logits, every interval above zero). Restoring the object reverses 32 to 59\% of the baseline's prior-wins, and the region elsewhere 9 to 15\%. The gain is larger where the blur lowers vision-win more: 21 to 35 points on the five models whose blurred baseline is at or below 0.52, and 10 to 15 on the four whose baseline is above 0.64. LLaVA-1.5 ran this control in bfloat16, where its enhancement run used float16 (clean image 0.781 against 0.778).

\begin{table}[h]\centering\footnotesize\setlength{\tabcolsep}{3.5pt}
\caption{\textbf{Positive control.} Vision-win on the image under blur 8 (baseline) and on the clean image, and the change in points [95\% interval] from restoring the clean pixels in the object, in a same-area region elsewhere, and in everything except the object. Last column: the object's gain minus the gain of the region elsewhere, paired over the items where the region exists. The first four models take the blur at their input size, the rest on the full-size image.}\label{tab:poscontrol}
\begin{tabular}{lrcccccc}
\toprule
Model & Items & Baseline & Object & Elsewhere & Rest & Clean & Object $-$ elsewhere \\ \midrule
LLaVA-1.5 & 361 & 0.377 & $+35.2$ [30.2, 40.4] & $+5.5$ [1.4, 9.2] & $+23.8$ [18.6, 29.1] & 0.781 & 30 [24, 36] \\
InstructBLIP-7B & 344 & 0.375 & $+34.9$ [29.6, 40.4] & $-0.4$ [$-5.4$, 4.6] & $+31.4$ [25.6, 36.9] & 0.846 & 35 [29, 41] \\
Emu3 & 349 & 0.327 & $+24.4$ [19.8, 28.9] & $+1.1$ [$-2.8$, 4.9] & $+21.8$ [16.6, 26.9] & 0.699 & 20 [15, 26] \\
Llama-3.2 & 362 & 0.519 & $+21.3$ [16.9, 26.0] & $-2.1$ [$-5.8$, 1.7] & $+15.2$ [10.5, 20.2] & 0.798 & 23 [17, 28] \\
Qwen2.5-VL & 351 & 0.641 & $+14.8$ [10.5, 19.1] & $-0.3$ [$-3.5$, 2.8] & $+9.4$ [4.6, 13.7] & 0.835 & 15 [10, 20] \\
Fuyu & 334 & 0.488 & $+21.9$ [16.8, 26.9] & $+0.0$ [$-3.3$, 3.7] & $+7.8$ [2.7, 12.9] & 0.725 & 20 [14, 25] \\
Idefics2 & 356 & 0.716 & $+9.8$ [6.2, 13.8] & $-0.3$ [$-3.5$, 2.8] & $+6.2$ [2.5, 10.4] & 0.860 & 9 [5, 13] \\
LFM2-VL & 352 & 0.685 & $+9.9$ [6.8, 13.4] & $+4.2$ [1.8, 6.7] & $+8.8$ [5.4, 12.5] & 0.827 & 7 [3, 10] \\
Gemma-3-12B & 363 & 0.708 & $+12.1$ [8.3, 16.3] & $-1.7$ [$-5.4$, 1.7] & $+4.4$ [0.8, 8.3] & 0.857 & 13 [9, 18] \\
\bottomrule
\end{tabular}
\end{table}

\subsection{LLaVA-NeXT, BakLLaVA and second checkpoints under the common protocol}
\noindent\textbf{LLaVA-NeXT and BakLLaVA under the common protocol.} The LLaVA-NeXT and BakLLaVA rows of main Table~1 are from this protocol (\cref{tab:commonllava}). On LLaVA-NeXT (425 prior-valid WHOOPS conflicts, vision-win 0.736; the image enters as 2,928 tokens on the median item) the dose grid on the full-size image reproduces the model-specific grid of Supp.~\ref{app:equation} at every level (blur: 0.736, 0.729, 0.722, 0.673, 0.645; noise: 0.736, 0.729, 0.734, 0.725, 0.706), and the additive-model fit gives $a$ = 0.11 [0.03, 0.18] and $b$ = 0.19 [0.06, 0.32] (medians over 50 splits 0.09 and 0.18, both positive on every split); the model-specific fit gave 0.13 [0.05, 0.21] and 0.16 [0.03, 0.30].

\begin{table}[h]\centering\footnotesize\setlength{\tabcolsep}{6pt}
\caption{\textbf{LLaVA-NeXT and BakLLaVA under the common protocol.} Crossing on the conflicts that the gray edit moves by at least 0.5 logits, with the 95\% interval over items; shares of the gray edit's effect in the bands that start at 0.50 / 0.625 of depth; top-20 heads by attribution against 20 layer-matched random sets; object edits in logits. 95\% intervals in brackets. Dash: not reported.}\label{tab:commonllava}
\begin{tabular}{lcc}\toprule
 & LLaVA-NeXT & BakLLaVA \\ \midrule
Conflicts moved by at least 0.5 logits & 162 & 200 \\
Crossing & 0.57 [0.55, 0.60] & 0.59 [0.57, 0.61] \\
Crossing, model-specific run & 0.58 & -- \\
Image-site share, bands at 0.50 / 0.625 & 0.68 [0.62, 0.74] / 0.41 [0.35, 0.46] & 0.79 [0.72, 0.87] / 0.43 [0.37, 0.50] \\
Answer-site share, bands at 0.50 / 0.625 & 0.37 [0.31, 0.45] / 0.63 [0.56, 0.71] & 0.30 [0.23, 0.35] / 0.61 [0.54, 0.67] \\
Prior-agreeing items: $n$, crossing (same band) & 31, 0.58 & 43, 0.56 \\ \midrule
Median layer of the top 20 heads & 28 of 32 & 26 of 32 \\
Held-out image-wins & 153 & 150 \\
Image-wins reversed (\%) & 11.8 [7.2, 16.3] & 14.7 [9.3, 20.0] \\
\quad random sets & 0.0 to 0.7 & 0.7 to 3.3 \\
Prior-wins reversed (\%) & none & 8.0 [2.0, 16.0] \\
\quad random sets & -- & 0.0 to 6.0 \\
Gap lowered by (logits) & 1.30 [1.10, 1.52] & 1.86 [1.53, 2.20] \\
\quad on 68 prior-agreeing items & 0.66 [0.46, 0.88] & 1.53 [1.14, 1.94] \\
Model-specific run: image-wins reversed / random (\%) & 7.0 / 0.3 & 7.7 / 2.2 \\ \midrule
Graying the object & $-0.83$ [$-1.05$, $-0.64$] & $-0.65$ [$-0.91$, $-0.43$] \\
Graying the rest & $-0.28$ [$-0.42$, $-0.14$] & $-0.34$ [$-0.53$, $-0.15$] \\
Difference (items) & $-0.55$ [$-0.80$, $-0.32$] (364) & $-0.31$ [$-0.60$, $-0.04$] (360) \\
Difference, model-specific run & $-0.56$ [$-0.80$, $-0.33$] & -- \\ \bottomrule
\end{tabular}\end{table}

\noindent\emph{Notes on \cref{tab:commonllava}.} The model-specific head ablation on BakLLaVA, reported above, is on its own item sample.

\noindent\textbf{Crossing and rule on further checkpoints.} Two of the five architectures that had one model each have a second public checkpoint that fits on one GPU: Idefics2-8B-chatty, which is Idefics2-8B fine-tuned for dialogue, and LFM2-VL-1.6B, a smaller model of the LFM2-VL architecture (16 decoder layers against 30). The reading was fixed before the runs: a second checkpoint agrees with the first if its crossing is within 0.10 of depth of the first and both weights of the additive model are positive with intervals that exclude zero. Both agree (\cref{tab:secondckpt}).

\begin{table}[h]\centering\footnotesize\setlength{\tabcolsep}{8pt}
\caption{\textbf{Second checkpoints of two architectures.} Dose on the full-size image; crossing on the conflicts that the gray edit moves by at least 0.5 logits. Both weights are positive on every one of 50 splits on both models.}\label{tab:secondckpt}
\begin{tabular}{lcc}\toprule
 & Idefics2-8B-chatty & LFM2-VL-1.6B \\ \midrule
Prior-valid WHOOPS conflicts & 424 & 401 \\
Vision-win & 0.788 & 0.761 \\
$a$ & 0.20 [0.10, 0.30] & 0.13 [0.05, 0.21] \\
$b$ & 0.15 [0.02, 0.28] & 0.21 [0.06, 0.36] \\
Dose-by-prior term, $p$ & 0.05 & 0.38 \\
Crossing (conflicts) & 0.57 (145) & 0.72 (180) \\
Crossing of the first checkpoint & 0.57 (Idefics2-8B) & 0.66 (LFM2-VL-3B) \\
Prior-agreeing items & 19, same band & 42, same band, at 0.65 \\ \bottomrule
\end{tabular}\end{table}

\noindent On LFM2-VL-1.6B a second scoring pass over 40 items changes the gap by up to 0.15 nats and changes no decision. Fuyu, Emu3 and Llama-3.2 remain one model each.

We also mapped the crossing on checkpoints that had the additive model only (the eight further checkpoints above), scoring WHOOPS again under the prompt of the common protocol. MiMo-VL-7B has the Qwen2.5-VL architecture with 36 decoder layers. At its SFT stage (408 conflicts, vision-win 0.772; 0.774 in the stage sweep) the crossing is at 0.67 on 196 conflicts, and at its RL stage (409 conflicts, vision-win 0.787; 0.779 in the stage sweep) at 0.67 on 204 conflicts; the prior-agreeing items (52 and 46) cross in the same band on both. Qwen2.5-VL-7B crosses at 0.73. Qwen3-VL-8B (instruct; 36 decoder layers) adds vision features to the image positions after decoder layers 0 to 2, and the first band (layers 0 to 3) covers them. On its 422 conflicts (vision-win 0.815, as in the stage sweep) the crossing is at 0.68 on 208 conflicts, and the 35 prior-agreeing items cross in the same band. InternVL3.5-8B at its final stage, loaded from the HF-format weights (36 decoder layers; 413 conflicts, vision-win 0.760; 0.752 in the stage sweep), crosses at 0.66 on 222 conflicts, and the 44 prior-agreeing items cross in the same band.

\section{The additive model against alternatives}\label{app:rulecmp}
\noindent\textit{Supports main-paper Sec.~4. The additive model against seven alternatives, and refit on every cell of each grid.}\par\smallskip
\noindent\textbf{Held-out comparison.} For every dose grid of the common protocol we refit the additive model and seven alternatives on 50 random item splits (70\% of the items for fitting, 30\% held out) and score held-out deviance explained. All models share an intercept and a noise-family term. \Cref{tab:rulecmp} gives medians at each model's grid of Table~1 of the main paper, except Emu3, shown on the full-size grid (0.102 at its 512$^2$ budget, below); LLaVA-1.5-13B is added, and LLaVA-NeXT and Qwen2.5-VL (the run that fixes the library version) are given in the text below. The additive model exceeds the dose-only model on 84 to 100\% of splits and the prior-only model on 94 to 100\%. The interaction term changes the median by at most 0.001. A dose term linear in blur radius or noise amplitude fits better than log precision by up to 0.003, and one free term per dose level by up to 0.013 on the 224- and 336-pixel grids and by at most 0.004 elsewhere. On an extrapolation split (fit on clean rows and the middle prior tertile, test on dosed rows of the outer tertiles) the additive model beats the prior-only model on 86 to 100\% of splits on seven models, and on 64\%, 28\% and 10\% on Emu3, LFM2-VL and Gemma-3.

\noindent\textbf{A positive prior weight is not automatic.} If the image added a fixed amount to the text-only margin, the margin with the image would have slope 1 on the text-only margin and the shift would be uncorrelated with it. We regress the margin with the image on the text-only margin, log precision and the noise-family term (ordinary least squares, errors clustered by item). The slope is 0.57 to 0.88 and its interval lies below 1 on seven of ten models. On clean images the shift (margin with the image minus text-only margin) is negatively correlated with the text-only margin on every model. The text-only margin explains 0.09 to 0.29 of the variance of the margin with the image and the dose adds up to 0.05.

\begin{table}[h]\centering\footnotesize\setlength{\tabcolsep}{3.2pt}
\caption{\textbf{The additive model against alternatives} (WHOOPS dose grids; medians over 50 item splits of held-out deviance explained). Lin.: dose linear in blur radius or noise amplitude; Free: one term per dose level. $>$dose, $>$prior: share of splits on which the additive model beats the dose-only and the prior-only model. Slope: of the margin with the image on the text-only margin, with 95\% interval; $r$: correlation of the clean-image shift with the text-only margin. The prior-only values of this table and the prior-alone values of the every-cell refits in the text below come from separate fits and differ by at most 0.009 (the additive-model value of LLaVA-1.5-13B by 0.007).}\label{tab:rulecmp}
\begin{tabular}{llcccccccccc}\toprule
Model & Grid & Dose only & Prior only & Both & $+$ interaction & Lin. & Free & $>$dose & $>$prior & Slope & $r$ \\ \midrule
LLaVA-1.5-7B & 336 & 0.027 & 0.068 & 0.093 & 0.092 & 0.096 & 0.105 & 0.96 & 1.00 & 0.66 [0.47, 0.84] & $-$0.28 \\
LLaVA-1.5-13B & 336 & 0.031 & 0.052 & 0.079 & 0.078 & 0.081 & 0.092 & 0.92 & 1.00 & 0.70 [0.47, 0.93] & $-$0.22 \\
BakLLaVA & 336 & 0.031 & 0.047 & 0.073 & 0.073 & 0.076 & 0.082 & 0.84 & 1.00 & 0.76 [0.52, 1.00] & $-$0.19 \\
InstructBLIP & 224 & 0.047 & 0.045 & 0.087 & 0.087 & 0.090 & 0.101 & 0.90 & 1.00 & 0.57 [0.34, 0.79] & $-$0.34 \\
Idefics2 & full & 0.007 & 0.052 & 0.059 & 0.058 & 0.060 & 0.060 & 0.90 & 0.98 & 0.86 [0.51, 1.20] & $-$0.07 \\
LFM2-VL & full & 0.005 & 0.051 & 0.054 & 0.054 & 0.054 & 0.056 & 0.90 & 0.98 & 0.64 [0.43, 0.85] & $-$0.25 \\
Fuyu & full & 0.013 & 0.067 & 0.082 & 0.082 & 0.084 & 0.085 & 0.98 & 1.00 & 0.64 [0.43, 0.84] & $-$0.30 \\
Emu3 & full & 0.004 & 0.081 & 0.084 & 0.084 & 0.085 & 0.085 & 1.00 & 0.96 & 0.88 [0.66, 1.10] & $-$0.09 \\
Llama-3.2 & 560 & 0.016 & 0.054 & 0.067 & 0.067 & 0.069 & 0.072 & 0.98 & 0.98 & 0.71 [0.51, 0.90] & $-$0.20 \\
Gemma-3 & full & 0.004 & 0.024 & 0.029 & 0.029 & 0.030 & 0.030 & 0.88 & 0.94 & 0.65 [0.45, 0.86] & $-$0.17 \\ \bottomrule
\end{tabular}\end{table}

\noindent\textbf{The additive model on every cell: what it explains and the interaction.} The fits above train on boundary cells and score interior cells. Here the additive model is refit on every cell of each grid with item-clustered intervals, and scored over 50 random 70/30 item splits with all cells on both sides. \Cref{tab:everycell} gives the result on WHOOPS and \cref{tab:everycellvilp,tab:vilpboundary} on ViLP. Three things hold across the tables. The dose adds to the prior wherever it is applied at a fixed input size, and adds almost nothing on the full-size image. The interaction between dose and prior is inside the pre-specified bound on most grids. Both weights are positive on ViLP on all ten models.

\begin{table}[h]\centering\footnotesize\setlength{\tabcolsep}{5pt}
\caption{\textbf{The additive model refit on every cell, WHOOPS.} Median held-out deviance explained over 50 item splits by both terms and by the prior alone, and the increment from adding the dose. Interaction: 95\% cluster-robust interval of the change in the precision weight per standard deviation of the prior, as a share of the precision weight ($^\ddagger$: outside the $\pm$0.5 bound; $^\S$: excludes zero; unstable: the precision weight is near zero; dash: not reported). $^\ast$: on the full-size grids and items of Supp.~\ref{app:equation} (425 and 407 prior-valid conflicts; blur and noise at five levels each; the Qwen2.5-VL run that fixes the library version).}\label{tab:everycell}
\begin{tabular}{llcccc}\toprule
Model & Dose grid & Both terms & Prior alone & Dose increment & Interaction share \\ \midrule
\multicolumn{6}{l}{\emph{Dose at a fixed input size}} \\
LLaVA-1.5 & 336 & 0.093 & 0.062 & $+$0.030 & [$-$0.20, 0.15] \\
BakLLaVA & 336 & 0.073 & 0.045 & $+$0.033 & [$-$0.14, 0.16] \\
InstructBLIP-7B & 224 & 0.087 & 0.036 & $+$0.050 & [$-$0.22, 0.05] \\
Emu3 & 512$^2$ & 0.102 & 0.075 & $+$0.028 & [$-$0.07, 0.32] \\
Llama-3.2 & 560 (one tile) & 0.067 & 0.053 & $+$0.017 & [$-$0.17, 0.20] \\
LLaVA-1.5-13B & 336 & 0.086 & 0.049 & $+$0.032 & [$-$0.19, 0.14] \\
InstructBLIP-13B & 224 & 0.048 & 0.012 & $+$0.035 & [$-$0.16, 0.15] \\ \midrule
\multicolumn{6}{l}{\emph{The same models, dose on the full-size image}} \\
LLaVA-1.5 & full & 0.072 & 0.071 & $+$0.001 & unstable \\
BakLLaVA & full & 0.046 & 0.045 & $+$0.001 & unstable \\
InstructBLIP-7B & full & 0.054 & 0.054 & 0.000 & -- \\
Emu3 & full & 0.084 & 0.081 & $+$0.004 & -- \\
Llama-3.2 & full & 0.065 & 0.063 & $+$0.004 & $-$0.31 [$-$0.60, $-$0.01]$^\S$ \\ \midrule
\multicolumn{6}{l}{\emph{Models that take the image at its own resolution}} \\
Fuyu & full & 0.082 & 0.067 & $+$0.014 & [$-$0.38, 0.10] \\
Idefics2 & full & 0.059 & 0.051 & $+$0.008 & [$-$0.41, 0.28] \\
LFM2-VL-3B & full & 0.054 & 0.049 & $+$0.005 & [$-$0.02, 0.75]$^\ddagger$ \\
Gemma-3-4B & full & 0.052 & 0.046 & $+$0.004 & [$-$0.58, 0.12]$^\ddagger$ \\
Gemma-3-12B & full & 0.029 & 0.024 & $+$0.004 & [$-$0.39, 0.22] \\
Gemma-3-27B & full & 0.051 & 0.044 & $+$0.007 & [$-$0.29, 0.23] \\ \midrule
LLaVA-NeXT$^\ast$ & full & 0.064 & 0.062 & $+$0.002 & -- \\
Qwen2.5-VL$^\ast$ & full & 0.062 & 0.052 & $+$0.008 & -- \\ \bottomrule
\end{tabular}\end{table}

\noindent\emph{Notes on \cref{tab:everycell}.} The dose increment is positive on 98 to 100\% of splits on the seven fixed-input fits, and on 96\% of splits on LLaVA-NeXT and on Qwen2.5-VL. On those two grids the additive model exceeds the dose-only model (0.002 and 0.008) on 96\% and 98\% of splits, the interaction term changes the median by less than 0.001, and the interaction coefficient is 0.009 [$-$0.007, 0.026] and 0.007 [$-$0.009, 0.024] with intervals not clustered by item; the two grids are not among the 28 counted below. On LLaVA-NeXT, which splits the image into crops of 336 pixels, the full-size dose adds little, as on the fixed-input models with the dose on the full-size image. On Llama-3.2 the interaction interval excludes zero on the full-size grid and does not at 560 pixels. For the intervals inside the bound, cluster-bootstrap intervals over items (500 resamples) agree to within 0.04 at each end; on LFM2-VL-3B the bootstrap interval is [$-$0.04, 0.89].

\begin{table}[h]\centering\footnotesize\setlength{\tabcolsep}{3.4pt}
\caption{\textbf{The additive model refit on every cell, ViLP.} Columns as in \cref{tab:everycell}, with the weights of the every-cell refit. $^\S$: the interval excludes zero.}\label{tab:everycellvilp}
\begin{tabular}{llcccccc}\toprule
Model & Dose grid & Both terms & Prior alone & Dose increment & $a$ & $b$ & Interaction share \\ \midrule
LLaVA-1.5 & 336 & 0.078 & 0.056 & $+$0.020 & 0.21 [0.18, 0.24] & 0.17 [0.13, 0.22] & [$-$0.13, 0.18] \\
BakLLaVA & 336 & 0.083 & 0.056 & $+$0.027 & 0.25 [0.22, 0.29] & 0.13 [0.09, 0.16] & [$-$0.22, 0.08] \\
InstructBLIP-7B & 224 & 0.063 & 0.040 & $+$0.028 & 0.25 [0.22, 0.29] & 0.16 [0.11, 0.21] & [$-$0.03, 0.28] \\
Emu3 & 512$^2$ & 0.151 & 0.126 & $+$0.021 & 0.26 [0.21, 0.31] & 0.24 [0.19, 0.29] & [$-$0.02, 0.42] \\
Llama-3.2 & 560 & 0.087 & 0.053 & $+$0.030 & 0.28 [0.24, 0.32] & 0.17 [0.13, 0.22] & [$-$0.18, 0.14] \\
LFM2-VL-3B & full & 0.064 & 0.061 & $+$0.003 & 0.07 [0.05, 0.10] & 0.18 [0.13, 0.23] & [$-$0.28, 0.39] \\
Idefics2 & full & 0.095 & 0.087 & $+$0.007 & 0.14 [0.11, 0.17] & 0.21 [0.16, 0.25] & $-$0.06 [$-$0.32, 0.20] \\
Qwen2.5-VL & full & 0.085 & 0.071 & $+$0.012 & 0.18 [0.14, 0.21] & 0.20 [0.15, 0.25] & [$-$0.29, 0.12] \\
Fuyu & full & 0.047 & 0.036 & $+$0.010 & 0.16 [0.13, 0.19] & 0.10 [0.07, 0.14] & [$-$0.12, 0.27] \\
Gemma-3-12B & full & 0.070 & 0.066 & $+$0.005 & 0.12 [0.09, 0.16] & 0.14 [0.11, 0.18] & $-$0.31 [$-$0.59, $-$0.04]$^\S$ \\ \bottomrule
\end{tabular}\end{table}

\noindent\emph{Notes on \cref{tab:everycellvilp}.} The dose increment is positive on 98 to 100\% of splits on the first six models with a ViLP fit (LLaVA-1.5, BakLLaVA, InstructBLIP-7B, Emu3, LFM2-VL-3B and Idefics2) and on 94\% of splits on Gemma-3-12B. On the ten ViLP grids $a$ is 0.07 to 0.28 and $b$ 0.10 to 0.24; the dose adds 0.02 to 0.03 on the five fits at a fixed input size and 0.003 to 0.012 on the five at native resolution. All interaction intervals are inside the $\pm$0.5 bound except that of Gemma-3-12B, where the interaction is present by the pre-specified criterion: the precision weight falls by about a third per standard deviation of the prior. Over the 28 grids refit on every cell in this section the interaction excludes zero on two, this one and the full-size WHOOPS grid of Llama-3.2, both negative.

\begin{table}[h]\centering\footnotesize\setlength{\tabcolsep}{4pt}
\caption{\textbf{ViLP by the boundary fit of the main protocol.} Prior-valid items, weights with 95\% intervals, held-out deviance explained on the interior cells of the main split against the 95th percentile of its permutation null, and the median weights over 50 item splits (both weights are positive on every split on every model listed with medians). $^\dagger$: below the null. Dash: not reported.}\label{tab:vilpboundary}
\begin{tabular}{llcccccc}\toprule
Model & Dose grid & Items & $a$ & $b$ & Held out & Null 95th & Median $a$ / $b$ \\ \midrule
LLaVA-1.5 & 336 & -- & 0.23 [0.16, 0.30] & 0.13 [0.04, 0.22] & 0.115 & 0.037 & -- \\
BakLLaVA & 336 & -- & 0.31 [0.24, 0.38] & 0.11 [0.06, 0.17] & 0.117 & 0.039 & -- \\
InstructBLIP-7B & 224 & 567 & 0.28 [0.21, 0.35] & 0.15 [0.05, 0.25] & 0.057 & 0.023 & 0.28 / 0.14 \\
Emu3 & 512$^2$ & 590 & 0.27 [0.15, 0.40] & 0.24 [0.15, 0.33] & 0.178 & 0.055 & 0.32 / 0.24 \\
Llama-3.2 & 560 & 570 & 0.26 [0.18, 0.33] & 0.13 [0.03, 0.23] & 0.106 & 0.041 & 0.28 / 0.14 \\
LFM2-VL-3B & full & 572 & 0.09 [0.04, 0.15] & 0.18 [0.08, 0.28] & 0.124 & 0.050 & 0.09 / 0.17 \\
Idefics2 & full & 573 & 0.17 [0.10, 0.23] & 0.36 [0.24, 0.48] & $-$0.070$^\dagger$ & 0.034 & 0.14 / 0.31 \\
Qwen2.5-VL & full & 582 & 0.17 [0.11, 0.23] & 0.31 [0.19, 0.43] & 0.092 & 0.039 & 0.17 / 0.27 \\
Fuyu & full & 582 & 0.15 [0.09, 0.20] & 0.11 [0.04, 0.18] & 0.052 & 0.029 & 0.16 / 0.10 \\
Gemma-3-12B & full & 550 & 0.16 [0.09, 0.23] & 0.14 [0.06, 0.23] & 0.141 & 0.054 & 0.16 / 0.18 \\ \bottomrule
\end{tabular}\end{table}

\noindent\emph{Notes on \cref{tab:vilpboundary}.} The pre-specified criterion is that both weights are positive with intervals that exclude zero and that the held-out deviance is above the 95th percentile of the permutation null. It holds on nine of the ten models. On Idefics2 it does not hold in full: on the main split the boundary fit predicts the interior cells worse than the null, with a dose-by-prior term in the boundary cells ($p$ = 0.01) that the all-cell refit of \cref{tab:everycellvilp} does not show. Over 50 item splits its weights are positive on every split and the held-out median is 0.080 ($-$0.08 to 0.22). In total, both weights are positive on ViLP on ten models from eight architectures, and nine on which the held-out test passes. How vision-win moves with the dose on ViLP differs by model. On InstructBLIP-7B blur 8 lowers vision-win from 0.40 to 0.10. On Llama-3.2 it falls from 0.76 to 0.41 at blur 8 and to 0.66 at the strongest noise. On LFM2-VL-3B the dose effect comes from blur alone (vision-win 0.55 clean, 0.42 at blur 8, 0.55 at the strongest noise). On Idefics2 neither blur below radius 3 nor noise below level 40 lowers vision-win (0.50 clean, 0.32 at blur 8, 0.47 at the strongest noise).

\noindent\textbf{Summary across grids.} AUROC of the additive model on held-out cells is 0.60 to 0.71, against 0.58 to 0.68 for the prior alone. At the strongest blur vision-win falls by 20 to 48 points in each prior tertile on the seven fixed-input fits. Adding a dose-by-prior interaction changes the median held-out deviance by at most 0.0003 in either direction on the WHOOPS grids and 0.0007 on the ViLP grids. The pre-specified criterion for the interaction: it is absent on a model if the 95\% interval of the change in the precision weight per standard deviation of the prior, as a share of the precision weight, lies inside $\pm$0.5, and present if it excludes zero.

\section{Frontier API models}\label{app:frontier}
\noindent\textit{Supports main-paper Sec.~4. The protocol and the results for the three API models.}\par\smallskip
\noindent\textbf{Protocol.} The models are GPT-6.1 Sol \cite{openai2026gpt61sol} (\texttt{gpt-6.1-sol}), Claude Sonnet 5.5 \cite{anthropic2026sonnet55} (\texttt{claude-sonnet-5-5}) and Gemini 4 Argon \cite{google2026gemini4argon} (\texttt{gemini-4-argon}), called through the providers' APIs in November 2026. Images are sent at up to 1,024 pixels on the long side, and blur and noise are applied at that size. Free completion on the WHOOPS statements (``complete this statement with the single most fitting word''), with the settings each API accepted, recorded on every call: GPT-6.1 Sol with reasoning effort \texttt{low}, its lowest accepted setting (2,557 calls), and Claude Sonnet 5.5 with thinking at effort \texttt{low}, since thinking could not be switched off (2,431 calls), both at the provider's default sampling because no temperature setting was accepted; Gemini 4 Argon at temperature 0 with a thinking budget of 128 tokens and the top-20 token log-probabilities (2,535 calls). The image is given or withheld, blurred at radius 2, 5 and 8, noised at 20 and 60, and cropped to the object's box or to a box of equal size elsewhere. A reply is labeled by the earliest whole-word match to the item's expected-answer set or image-answer set; replies that match neither are dropped. Prior-valid items are those whose text-only reply is in the expected set. The prior of an item is the mean text-only margin of the open models, since the APIs return no usable log-odds for it; Gemini also returns token log-probabilities, and with its own log-odds as the prior its weights are $a$ = 0.18 [0.08, 0.28] and $b$ = $-0.05$ [$-0.17$, 0.08]. Crop items are prior-valid items with a SAM3 mask and a near-square image.

\begin{table}[h]\centering\footnotesize\setlength{\tabcolsep}{3.5pt}
\caption{\textbf{Three API models} (WHOOPS). Calls, with none failed. Vision-win on prior-valid items with 95\% intervals; at blur 8 and noise 60; $a$ and $b$ of the additive model with the prior from the open models; share of crop items that keep the image's answer under the object crop and the random crop, and their difference.}\label{tab:frontier}
\resizebox{\linewidth}{!}{%
\begin{tabular}{lccccccccc}\toprule
Model & Calls & Prior-valid & Vision-win & Blur 8 & Noise 60 & $a$ & $b$ & Object / random crop & Difference \\ \midrule
GPT-6.1 Sol & 2,557 & 318 & 0.901 [0.862, 0.933] & 0.761 & 0.793 & 0.27 [0.19, 0.36] & 0.13 [0.03, 0.23] & 0.83 / 0.31 ($n$=58) & 0.52 [0.38, 0.66] \\
Claude Sonnet 5.5 & 2,431 & 297 & 0.920 [0.885, 0.950] & 0.842 & 0.879 & 0.18 [0.08, 0.27] & 0.07 [$-$0.07, 0.20] & 0.91 / 0.30 ($n$=53) & 0.60 [0.47, 0.74] \\
Gemini 4 Argon & 2,535 & 314 & 0.951 [0.922, 0.975] & 0.908 & 0.924 & 0.18 [0.09, 0.28] & 0.11 [$-$0.06, 0.26] & 0.90 / 0.51 ($n$=41) & 0.39 [0.22, 0.56] \\ \bottomrule
\end{tabular}}\end{table}

\noindent Between 126 and 141 of 473 text-only replies and 30 to 34 clean-image replies per model match neither set. On the prior-valid items, 10 to 12\% of clean-image replies and 12 to 13\% of dosed replies match neither set, and the fits drop them. \Cref{tab:apiunmatched} gives the weights when they are counted for the prior or for the image instead.

\begin{table}[h]\centering\footnotesize\setlength{\tabcolsep}{7pt}
\caption{\textbf{Replies that match neither answer set, counted for one side.} Weights of the additive model with 95\% intervals.}\label{tab:apiunmatched}
\begin{tabular}{lcccc}\toprule
 & \multicolumn{2}{c}{Counted for the prior} & \multicolumn{2}{c}{Counted for the image} \\ \cmidrule(lr){2-3}\cmidrule(lr){4-5}
Model & $a$ & $b$ & $a$ & $b$ \\ \midrule
GPT-6.1 Sol & 0.20 [0.14, 0.26] & 0.09 [0.02, 0.16] & 0.25 [0.17, 0.33] & 0.07 [$-$0.01, 0.15] \\
Claude Sonnet 5.5 & 0.13 [0.07, 0.19] & 0.07 [$-$0.00, 0.14] & 0.16 [0.07, 0.25] & 0.03 [$-$0.06, 0.12] \\
Gemini 4 Argon & 0.12 [0.06, 0.18] & 0.07 [$-$0.01, 0.14] & 0.17 [0.07, 0.27] & 0.06 [$-$0.04, 0.15] \\ \bottomrule
\end{tabular}\end{table}

\noindent The precision weight excludes zero either way; the prior weight stays positive, and its interval excludes zero only on GPT-6.1 Sol with the replies dropped or counted for the prior. The prior overrides the image on 2 to 4 crop items per model, too few to test whether the object crop restores the image's answer.

\section{Further analyses of the main runs}\label{app:saved}
\noindent\textit{Supports main-paper Secs.~6 and 8. Further analyses of the main runs.}\par\smallskip
\noindent\textbf{Readout against the prior.} A mid-depth margin could predict the decision only because it has already absorbed the prior. On prior-valid WHOOPS conflicts we compare three predictors of the decision: the text-only margin (the prior), the lens margin of the text-only run at 0.625 of depth, and the lens margin of the image run at 0.625 of depth, and we fit cross-validated logistic predictors on the prior alone and on the prior plus each lens margin (\cref{tab:savedread}). The text-run lens adds at most 0.01 to the prior (Idefics2) and nothing on the other eleven models. The image-run lens adds 0.18 to 0.26 AUROC on eight models, 0.08 to 0.09 on Emu3 and Gemma-3-4B, and nothing on LFM2-VL and Fuyu.

\begin{table}[h]\centering\footnotesize\setlength{\tabcolsep}{4pt}
\caption{\textbf{Readout against the prior} (prior-valid WHOOPS conflicts; AUROC for the decision). Increment: cross-validated prior plus image-run lens minus prior alone, with bootstrap 95\% interval.}\label{tab:savedread}
\begin{tabular}{lcccc}\toprule
Model & Prior alone & Text-run lens & Image-run lens & Increment over the prior \\ \midrule
LLaVA-1.5-7B & 0.67 & 0.59 & 0.83 [0.78, 0.87] & $+$0.18 [0.12, 0.25] \\
LLaVA-1.5-13B & 0.64 & 0.57 & 0.84 [0.80, 0.88] & $+$0.21 [0.15, 0.28] \\
BakLLaVA & 0.61 & 0.57 & 0.80 [0.75, 0.84] & $+$0.20 [0.13, 0.27] \\
InstructBLIP & 0.64 & 0.60 & 0.89 [0.85, 0.92] & $+$0.26 [0.20, 0.33] \\
Idefics2 & 0.62 & 0.58 & 0.83 [0.79, 0.88] & $+$0.21 [0.13, 0.29] \\
LFM2-VL & 0.61 & 0.54 & 0.63 [0.57, 0.69] & $+$0.04 [$-$0.01, 0.09] \\
Fuyu & 0.67 & 0.53 & 0.54 [0.48, 0.61] & $-$0.01 [$-$0.02, 0.00] \\
Emu3 & 0.68 & 0.57 & 0.74 [0.69, 0.79] & $+$0.09 [0.04, 0.14] \\
Llama-3.2 & 0.65 & 0.55 & 0.82 [0.77, 0.86] & $+$0.18 [0.12, 0.23] \\
Gemma-3-4B & 0.63 & 0.50 & 0.69 [0.63, 0.75] & $+$0.08 [0.02, 0.14] \\
Gemma-3-12B & 0.56 & 0.55 & 0.82 [0.78, 0.86] & $+$0.26 [0.18, 0.33] \\
Gemma-3-27B & 0.63 & 0.54 & 0.89 [0.85, 0.92] & $+$0.26 [0.19, 0.33] \\ \bottomrule
\end{tabular}\end{table}

\noindent\textbf{Object edits as decisions.} \Cref{tab:savedobj} counts the image-wins whose gap changes sign under each edit. Graying the object reverses 5.9 to 11.2\% of image-wins and graying a same-area region elsewhere 0.0 to 2.2\%; the paired difference excludes zero on all nine models. Graying everything else, about nine tenths of the image, reverses 2.8 to 6.1\%; the object-minus-rest difference in reversals excludes zero for gray on LLaVA-1.5-7B and 13B and for blur on LLaVA-1.5-7B, InstructBLIP, Llama-3.2 and Gemma-3-4B, and includes zero elsewhere. The margin-level differences of \cref{tab:archobj2} are the more sensitive measure.

\begin{table}[h]\centering\footnotesize\setlength{\tabcolsep}{4pt}
\caption{\textbf{Image-wins reversed by gray edits} (prior-valid WHOOPS conflicts with a mask). Paired differences with bootstrap 95\% intervals, in percentage points.}\label{tab:savedobj}
\begin{tabular}{lcccccc}\toprule
Model & Image-wins & Object & Rest & Elsewhere & Object $-$ rest & Object $-$ elsewhere \\ \midrule
LLaVA-1.5-7B & 278 & 11.2\% & 4.3\% & 0.9\% & $+$6.8 [3.2, 10.8] & $+$10.6 [6.4, 14.7] \\
LLaVA-1.5-13B & 290 & 7.2\% & 2.8\% & 0.9\% & $+$4.5 [1.0, 7.9] & $+$6.5 [3.5, 10.4] \\
InstructBLIP & 291 & 6.5\% & 3.8\% & 1.7\% & $+$2.7 [$-$1.0, 6.2] & $+$4.3 [1.3, 7.8] \\
Idefics2 & 306 & 5.9\% & 3.9\% & 0.8\% & $+$2.0 [$-$1.3, 5.2] & $+$6.2 [3.3, 9.9] \\
LFM2-VL & 290 & 5.9\% & 3.4\% & 0.0\% & $+$2.4 [$-$0.7, 5.5] & $+$6.0 [3.0, 9.4] \\
Fuyu & 224 & 7.1\% & 4.9\% & 2.2\% & $+$2.2 [$-$1.3, 5.8] & $+$5.0 [1.7, 8.3] \\
Emu3 & 245 & 10.6\% & 6.1\% & 2.1\% & $+$4.5 [$-$0.4, 9.4] & $+$9.8 [5.2, 14.4] \\
Llama-3.2 & 290 & 5.9\% & 4.5\% & 1.3\% & $+$1.4 [$-$1.7, 4.8] & $+$3.9 [0.9, 7.3] \\
Gemma-3-4B & 307 & 5.9\% & 5.5\% & 1.6\% & $+$0.3 [$-$3.3, 3.6] & $+$3.7 [0.8, 6.6] \\ \bottomrule
\end{tabular}\end{table}

\noindent\textbf{Band map by outcome.} Splitting the conflicts of the band map by which answer won, the crossing of the prior-wins minus that of the image-wins is $+0.01$ on LLaVA-1.5-7B, LLaVA-1.5-13B, LFM2-VL, Llama-3.2 and Gemma-3-27B, $+0.02$ on InstructBLIP, $0.00$ on Idefics2 and Fuyu, $-0.01$ on Emu3, $+0.06$ on Gemma-3-4B and $-0.08$ on Gemma-3-12B. Every bootstrap interval includes zero; the widest is [$-0.21$, $+0.13$] on Emu3. No difference is detected; the widest interval is wider than the $\pm$0.10 margin used above, so equivalence is not shown.

\section{Extended related work}\label{app:related}
\noindent\textit{Extends main-paper Sec.~2.}\par\smallskip
\noindent\textbf{Knowledge conflicts in language models.} When a context contradicts a fact a language model holds, the model often answers from memory \cite{longpre2021entity}, and follows the context when it is coherent and convincing \cite{xie2023chameleon} or the prior is weak \cite{wu2024clasheval,yu2023characterizing,du2024persuasion}. The preference is carried by identifiable heads \cite{jin2024cutting} that compete across depth \cite{ortu2024competition}, and it is detectable in the residual stream before the answer \cite{zhao2024conflictdetect}.

\noindent\textbf{Image versus prior in VLMs.} Benchmarks and studies of language priors \cite{agrawal2018vqacp,niu2021cfvqa,lee2024vlind,goyal2017vqav2,bittonguetta2023whoops,luo2025vilp,li2026seeingorknowing}, of text-image conflicts \cite{deng2025blindfaith,hua2025conflicting,paraselli2026entities,ghosh2026whichsource,liu2025insight,jia2026mmkc} and of conflicts between a model's visual and textual answers \cite{zhu2024crossmodal} establish that VLMs often answer from the prior and that the preference shifts with task and legibility. \citet{barua2026bayesian} show that VLM preferences between conflicting prompt text and images follow a biased Bayesian rule with per-model weights, and name the parametric-prior case as future work. The main paper fits that case on eleven models, with the prior measured per item. \citet{zhang2025modalitiesconflict} tie modality following to unimodal uncertainty, and \citet{ma2025bayesbench} find Bayes-consistent, though not uniformly efficient, cue combination in multimodal LLMs on magnitude-estimation tasks.

\noindent\textbf{Mechanisms of the decision.} \citet{ortu2026seeing} identify counterfactual heads on WHOOPS-AHA by logit inspection and steer by reweighting their attention. \citet{lietzow2026visiondefault} locate prior-override heads in the second half of the network (52 to 100\% of depth) on three VLM families, all of which insert image tokens into the stream, whose ablation under a prompt contrast flips most prior-grounded predictions. We study the complementary heads, which write the image's answer, on seven of eight architectures, and find them at a similar depth without a conflict. Information-flow studies find that image information moves to the question tokens and, in the middle layers, from there to the last position \cite{zhang2025crossmodal,kaduri2025whats}, or directly from the image tokens to the last position, depending on the task \cite{salazar2026pathways}, and through a single gate token in natively multimodal models \cite{serra2025narrowgate}; we measure that flow by patching, with the same image edited as the counterfactual, under conflict and without it, on eight architectures. Other work reports that visual information stays decodable while larger models follow priors more \cite{grob2026bigger}, a mid-network commitment in LLaVA-NeXT against a late one in Qwen2.5-VL \cite{he2026modalityconflict}, attention on the evidence even when the answer is wrong \cite{liu2025seeingnotbelieving}, late alignment of visual with textual representations \cite{nikankin2025sametask}, steerable modality preference \cite{zhang2025modalitypreferences}, and a prior-to-visual shift in the mid-to-late layers \cite{golovanevsky2025pixels}. \citet{nooralahzadeh2026arbitration} show the visual evidence is encoded even when the prior wins, and that last-token patching flips 0--1\% of decisions. Object information sits at the object's token positions and reaches the answer through the middle layers \cite{neo2024towards,jiang2024devils}, head subsets control what the model reports \cite{wang2025vseam,gandikota2026gaze,rudman2026pih}, and models can represent visual evidence without using it \cite{zhou2026ignorance,liao2026decodable}.

\noindent\textbf{Hallucination detection and training-free intervention.} Hallucination and adversarial benchmarks score answers \cite{rohrbach2018chair,li2023pope,guan2023hallusionbench,li2024naturalbench}, and POPE can be answered with three quarters of the image tokens removed \cite{lan2026seeingwithoutlooking}; the readout we report scores the decision before the answer. Detectors read the logit lens \cite{nostalgebraist2020logitlens} of image representations \cite{jiang2024interpretingedit} or probe hidden states \cite{kogilathota2026halp}. Training-free interventions raise the weight of the image by contrastive decoding, an over-trust penalty, attention reweighting or route gating within heads \cite{leng2023vcd,huang2023opera,liu2024pai,kang2025var,cheng2026crg}.

\noindent\textbf{Methods.} Activation patching and its metrics \cite{zhang2024patching,heimersheim2024patching}, subspace interventions and their illusions \cite{geiger2024das,makelov2024illusion}, and linear concept erasure \cite{belrose2023leace} supply the controls we adopt.

\noindent\textbf{Further related work.} In text models a one-dimensional subspace sets how much the context is followed \cite{minder2025knob}. Causal tracing and head attribution have been applied to VLMs \cite{palit2023causaltracing,basu2024information,golovanevsky2025notice,kim2025headattribution,luo2025sink,liu2026dualpathway,park2026retrievalheads}; a layer at which the image begins to change the last-token state is reported per model by \citet{long2026chainofembedding}, localization heads sit at different depths in different models \cite{schaumloffel2026localization}, and conflict detection and resolution occur at different depths \cite{nguyen2025challenges}. Visual information that is represented and not used is reported by \citet{fu2025hidden} and \citet{liu2025ignorance}; intermediate layers can hold the visual answer while the output follows the prior \cite{wang2025deco}, and first-token logits predict the outcome before generation \cite{zhao2024firsttoknow}. Interventions on heads, subspaces and priors include \citet{yang2025modular}, \citet{yang2025nullu}, \citet{zhou2025causalmm}, \citet{favero2024m3id} and \citet{liang2026enar}, and layer-matched random head sets are the control recommended by \citet{kadir2026audit}. Models with discrete image tokens are studied by \citet{hegde2026routing}. Cropping or masking to the evidence helps on other tasks \cite{zhang2025mllmsknow,liu2025seeingnotbelieving}, and on counterfactual images crops had weak or negative effects on several open models \cite{sun2026prive}. Object-against-background edits are a standard test in object recognition \cite{xiao2021noise}, blur and noise are standard corruptions \cite{hendrycks2019benchmarking}, and \citet{schild2026waldo} name the loss of detail when a processor downsamples the image. The yes-bias on binary questions is studied by \citet{chan2026yes}.

\noindent\textbf{Components named in this supplement.} CLIP \cite{radford2021clip}, SigLIP \cite{zhai2023siglip}, the Q-Former \cite{li2023blip2}, Vicuna \cite{zheng2023vicuna}, CLIPSeg \cite{luddecke2022clipseg}, the transformers library \cite{wolf2020transformers}, MME \cite{fu2023mme} and COCO \cite{lin2014coco}.

\end{document}